\documentclass[b5paper,10pt]{book}

\usepackage{phdthesis}
\usepackage[
  paperwidth=17cm
  ,paperheight=24cm
  ,inner=2.6cm
  ,outer=2.3cm
  ,top=2.3cm
  ,bottom=2.3cm
]{geometry}

\usepackage{amsmath}
\usepackage{amsfonts}
\usepackage{bbm}
\usepackage{booktabs}
\usepackage[bottom]{footmisc}
\usepackage{placeins}
\usepackage{graphicx}
\usepackage{rotating}
\usepackage{hyperref}
\hypersetup{
    hidelinks,
    colorlinks=false
}
\usepackage[utf8]{inputenc}
\usepackage[T1]{fontenc}
\usepackage{textcomp}
\usepackage{pdfpages}
\usepackage{multirow}
\usepackage[round]{natbib}
\usepackage{soul}
\usepackage{subcaption}
\usepackage{textcomp}
\usepackage[table]{xcolor}
\usepackage{xcolor}
\usepackage{arydshln}
\usepackage{enumitem}
\usepackage{xparse}
\usepackage{float}

\newenvironment{myquote}
  {\list{}{\leftmargin=0.3in\rightmargin=0.0in}\item[]}%
  {\endlist}

\DeclareUnicodeCharacter{0308}{\"}

\newcommand{\RQ}[1]{
    \par\vspace{0.5em}
    \noindent
    \ifcase#1
        \or \textbf{Research Question 1}: \textit{How can we measure and enhance cross-lingual knowledge transfer in multilingual neural machine translation through the lens of representational similarities?}
        \or \textbf{Research Question 2}: \textit{What role does cross-lingual knowledge transfer play when extending \textnormal{k}-nearest neighbor machine translation to multilingual approaches with the goal of improving low-resource translation?}
        \or \textbf{Research Question 3}: \textit{How can we mitigate negative transfer and preserve beneficial capabilities when fine-tuning large language models for machine translation?}
        \or \textbf{Research Question 4}: \textit{What are the effects of scaling language diversity during fine-tuning of large language models for machine translation?}
        \fi
    \vspace{0.5em}
    \noindent
}

\newcommand{\RQsub}[2]{
    \par
    \noindent
    \ifcase#1
    \or
        \ifcase#2
            \or \textbf{RQ1.1} \quad \textit{To what extent can knowledge transfer in multilingual neural machine translation be explained and predicted by representational similarities between languages?}            
            \or \textbf{RQ1.2} \quad \textit{Which dataset and linguistic characteristics are most predictive of representational similarities and therefore knowledge transfer between languages?}
            \or \textbf{RQ1.3} \quad \textit{How can we leverage insights about representational similarities to improve translation quality for low-resource languages?}
        \fi
    \or
        \ifcase#2
            \or \textbf{RQ2.1} \quad \textit{To what extent can cross-lingual datastores improve translation quality for low-resource languages compared to traditional bilingual datastores?}
            \or \textbf{RQ2.2} \quad \textit{How can we effectively combine information from multiple languages into a single datastore to improve translation quality across both low and high-resource languages?}
            \or \textbf{RQ2.3} \quad \textit{How can we optimize multilingual datastores to maintain translation quality while improving computational efficiency?}
        \fi
    \or
        \ifcase#2
            \or \textbf{RQ3.1} \quad \textit{What is the impact of fine-tuning on the qualitative advantages that large language models possess over traditional neural machine translation systems?}
            \or \textbf{RQ3.2} \quad \textit{How does the size of fine-tuning data and model scale influence the preservation or degradation of LLM translation capabilities?}
            \or \textbf{RQ3.3} \quad \textit{How can we develop fine-tuning strategies that improve translation quality while maintaining the beneficial properties of LLMs?}
        \fi
    \or
        \ifcase#2
            \or \textbf{RQ4.1} \quad \textit{How does incrementally scaling the number of languages during fine-tuning affect translation performance across seen and unseen language pairs?}
            \or \textbf{RQ4.2} \quad \textit{To what extent do the effects of multilingual scaling vary across different model architectures and fine-tuning approaches?}
            \or \textbf{RQ4.3} \quad \textit{What representational changes occur within LLMs as a result of exposure to varying degrees of language diversity during fine-tuning?}
        \fi    
    \fi
}

\begin{document}

% \includepdf[pages=1]{cover.pdf}

\frontmatter
% !TEX root = thesis-main.tex
% This is the first page, consisting of your title and name.

% Title goes here
{\pagestyle{empty}
\newcommand{\printtitle}{%
{\Huge\bf Analyzing and Improving Cross-lingual\\ Knowledge Transfer\\ for Machine Translation \\[0.8cm]
}}

% Some markup followed by your name
\begin{titlepage}
\par\vskip 2cm
\begin{center}
\printtitle
\vfill
{\LARGE\bf David Stap}
\vskip 2cm
\end{center}
\end{titlepage}

% Skip a page to start on a right page again.
\mbox{}\newpage
\setcounter{page}{1}

% This is the title page (titelblad)
% You also need to send it to the Bureau Pedel
% Make sure you change:
%   the date of your defense
%   your name
%   your place of birth (including country, but ONLY if you were born outside the Netherlands)
\clearpage
\par\vskip 2cm
\begin{center}
\printtitle
\par\vspace {4cm}
{\large \sc Academisch Proefschrift}
\par\vspace {1cm}
{ ter verkrijging van de graad van doctor aan de \\
Universiteit van Amsterdam\\
op gezag van de Rector Magnificus\\
% prof.\ dr.\ ir.\ K.I.J. Maex\\
prof.\ dr.\ ir.\ P.P.C.C. Verbeek\\
ten overstaan van een door het College voor Promoties ingestelde commissie,\\
in het openbaar te verdedigen in de Aula der Universiteit\\
op woensdag 26 november 2025, te 11.00 uur \\ }
\par\vspace {1cm} {\large door}
\par \vspace {1cm}
{\Large David Stap}
\par\vspace {1cm}
{\large geboren te Amsterdam} %MAKE VERY SURE THIS MATCHES YOUR PASSPORT, THE BEADLE WILL CHECK THIS! 
\end{center}

% The page following the titelblad. This usiallu contains:
%   committee members
%   SIKS logo + text
%   sponsors/ project numbers
%   ISBN
%   copyrights, cover design, printer
\clearpage
\noindent%
\textbf{\textit{Promotiecommissie}} \\\\
\begin{tabular}{@{}l l l}
\textit{Promotor}: \\
& prof. dr. C. Monz & Universiteit van Amsterdam \\  %promotor
\textit{Co-promotor}: \\
& dr. V. Niculae & Universiteit van Amsterdam \\
\textit{Overige leden}: \\
& dr. A. Bisazza & Rijksuniversiteit Groningen \\
& prof. dr. W. Byrne & University of Cambridge \\
& prof. dr. R. Fernández Rovira & Universiteit van Amsterdam \\
& prof. dr. E. Kanoulas & Universiteit van Amsterdam \\
& dr. A. Lucic & Universiteit van Amsterdam \\
\end{tabular}

\bigskip\noindent%
Faculteit der Natuurwetenschappen, Wiskunde en Informatica\\

\vfill

% Sponsors and projects
\noindent
The work described in this thesis has been carried out at the Language Technology Lab of the University of Amsterdam and in part during an internship at Amazon AGI, Berlin.
The research carried out at the University of Amsterdam was funded in part by the Association of Collaborating Dutch Universities (VSNU) under project Onderzoek Digitale Samenleving (DiSa) and in part by the Netherlands Organization for Scientific Research (NWO) under project numbers VI.C.192.080 and 2023.017.
We thank SURF (www.surf.nl) for the support in using the National Supercomputer Snellius.

\bigskip

% Copyrights
\noindent
Copyright \copyright~2025 David Stap, Haarlem, The Netherlands\\
Cover by Christel Staal \\
ISBN: 978-94-93483-28-6
\clearpage
}

\chapter*{Acknowledgements}
\addcontentsline{toc}{chapter}{Acknowledgements} % add to table of contents

PhDone!
Even though I did not fully understand what I signed myself up for when starting this journey, it was rewarding and I would do it again.
My PhD spanned a period of explosive growth in the field, a transformation that saw my methods evolve from training 93-million-parameter models from scratch to fine-tuning pre-trained 65-billion-parameter models.
Yet, for all the rapid advancements in technology, the truly indispensable support came from people.
Thankfully, I got a lot of help along the way for which I am very grateful.

First, Christof.
Thank you for taking a chance on me all those years ago.
You patiently taught me how to become an independent researcher, and I am especially grateful that you never made me feel stupid in the process. 
I will miss your humor and our discussions, especially the ones where we disagreed, as those forced me to articulate half-formed thoughts and connect them into a coherent and (hopefully) convincing argument.

Vlad, thank you for being my co-supervisor.
You are a great teacher and researcher, and I am grateful that I could learn from you.

I am honored to have Ana, Arianna, Bill, Evangelos and Raquel as part of my PhD committee.
Thank you for your valuable time to read and critically evaluate my thesis.

The Language Technology Lab (LTL) steadily grew in size as I progressed in my PhD journey.
Ali, Amir, Baohao, Di, Evgeniia, Kata, Maya, Sara, Sergey, Seth, Shaomu, Wafaa, Yan, thank you for sharing the highs and lows of our existence during many lunches, dinners, coffee breaks, walks, random chats and conference trips.
I'll cherish these memories!
Also thanks to my university friends outside of LTL: Joris and Maurits.
Thank you for all the discussions about life, the PhD, and everything in between.

I am grateful that I got to do two internships at the great multilingual NLP team of Amazon AGI in Berlin.
It was a very welcome break from my PhD, and a great introduction to research outside a university setting.
Arda, Bill, Eva, Felix, Hagen, Ke, Luke, Rexhina, Sony, Tobias, Zach, thank you for the many interesting discussions, lunches and table tennis breaks.

To Heleen, thank you for being the warm and wise grandmother figure I have been so fortunate to know.

To my parents.
Thank you for your unconditional support and pride.
While the academic world is very different from yours, your constant encouragement and interest made all the difference.

Finally, my deepest thanks to Christel.
Your presence made the journey better.
You were my essential connection to the real world, and I am incredibly grateful for it.
I’m happy to have started our next chapter.

\begin{flushright}
David\\
Haarlem, October 2025
\end{flushright}

\tableofcontents

\mainmatter

\chapter{Introduction}
Machine translation can be considered as one of the most challenging and impactful applications of artificial intelligence, with the potential to break down language barriers and connect people across the globe.
The field has seen remarkable progress over the past decade, transitioning from statistical approaches to neural networks and, most recently, to large language models.
Despite these advances, a fundamental challenge remains: how can we create systems that effectively translate between thousands of languages when parallel data is abundant for only a small fraction of them?

Cross-lingual knowledge transfer---the ability of multilingual models to apply what they have learned from one language to another---offers a promising solution to this challenge.
When successful, this transfer allows models to leverage information from high-resource languages to improve translation for low-resource ones, potentially extending high-quality translation to many more of the world's languages.
However, understanding when, why, and how this transfer occurs remains an open research question with important practical implications.

The complexity of cross-lingual transfer stems from the intricate interplay between languages at various linguistic dimensions, from phonology and morphology to syntax and semantics, combined with the challenges of representation alignment in neural networks, parameter sharing across dissimilar languages, and the dynamic optimization landscapes that unfold when multiple languages compete for limited model capacity.

Languages share universal properties while differing in specific features: they may use similar grammatical structures but different vocabularies, share vocabulary items but employ different word orders, or have entirely different writing systems despite syntactic similarities.
A model's ability to abstract away from language-specific features while preserving meaning is central to effective cross-lingual transfer.

Traditional neural machine translation (NMT) systems are typically trained on parallel data for specific language pairs, limiting their ability to generalize to new languages or language combinations.
Multilingual approaches attempt to overcome this limitation by training on multiple language pairs simultaneously, enabling the model to learn shared representations across languages.
However, the effectiveness of these approaches varies widely across language pairs, with some languages benefiting significantly while others show minimal improvement or even degradation.

Large language models (LLMs) have introduced a significant shift in machine translation approaches, demonstrating impressive zero-shot translation capabilities despite not being explicitly trained for the task.
These models develop translation abilities through incidental exposure to multilingual and parallel text during pre-training, offering a different mechanism for cross-lingual knowledge transfer.
However, fine-tuning these models on specific translation tasks introduces new challenges, as the process of optimization for one objective may interfere with beneficial capabilities acquired during pre-training.

% Throughout this thesis, we approach cross-lingual knowledge transfer from multiple complementary perspectives.
% We begin by developing a theoretical framework for measuring and analyzing transfer through representational similarities between languages.
% Building on these insights, we implement practical methods for enhancing transfer in both neural machine translation and retrieval-augmented approaches.
% We then address the challenge of preserving beneficial capabilities when fine-tuning large language models, exploring strategies to balance task-specific optimization with capability retention.
% Finally, we investigate how language diversity during fine-tuning implicitly shapes cross-lingual generalization patterns in large language models.

Understanding the mechanisms of cross-lingual knowledge transfer is not merely an academic pursuit.
It has practical implications for extending high-quality translation to the thousands of languages that currently lack sufficient resources for traditional training approaches.
By developing more effective methods for cross-lingual transfer, we work toward the goal of making machine translation accessible to speakers of all languages, regardless of their resource status.

\section{Research outline and questions}
This thesis explores the advancement of cross-lingual knowledge transfer in multilingual machine translation through three interconnected perspectives: representational analysis, practical implementation using nearest neighbors, and the preservation of capabilities when fine-tuning large language models. We develop novel methods and analysis frameworks to examine how cross-lingual knowledge can be effectively transferred and preserved in various multilingual translation settings.

The overarching question that guides this thesis is: How can we better understand and improve cross-lingual knowledge transfer to improve multilingual machine translation quality while preserving beneficial model capabilities?

Before investigating practical implementations, we first examine the fundamental question of how knowledge transfer occurs between languages at the representational level. We study how similarities between languages in the representational space relate to improvements in translation quality and what factors influence these similarities.

Next, in more detail, we focus on implementing cross-lingual knowledge transfer in a practical setting using \textit{k}-nearest neighbor approaches. While recent advances in \textit{k}-nearest-neighbor machine translation have shown impressive results for high-resource languages, their effectiveness for low-resource languages remains limited. We investigate how combining representations from multiple languages into a single datastore can benefit both low and high-resource translation directions.

Further, we address the challenge of maintaining beneficial cross-lingual capabilities when adapting large language models for machine translation. We examine how fine-tuning affects various qualitative advantages that LLMs hold over traditional NMT models and propose methods to preserve these capabilities while improving general translation quality.

Lastly, we investigate how language diversity during fine-tuning affects cross-lingual generalization in large language models, examining how diverse language exposure implicitly shapes internal model representations and resolving conflicting evidence from prior work on multilingual fine-tuning strategies.

Concretely, we set out to answer the following research questions in this thesis:

\RQ{1}
\begin{myquote}
In this research question, we investigate how representational similarities between languages are related to cross-lingual knowledge transfer in multilingual machine translation.
We examine whether the improvements in translation quality observed in multilingual models stem from genuine knowledge transfer or merely from increased exposure to target language data.
Additionally, we explore which factors influence these representational similarities and develop methods to enhance them for improving low-resource translation.
We divide this research question into three sub-questions and address them in Chapter~\ref{ch3}:
\end{myquote}

\RQsub{1}{1}
\begin{myquote}
To answer this question, we introduce a metric called Representational Transfer Potential (RTP) that quantifies the similarity between cross-attention context vectors of different languages.
We demonstrate that RTP scores correlate strongly with improvements in translation quality, confirming that knowledge transfer occurs at the representational level.
\end{myquote}

\RQsub{1}{2}
\begin{myquote}
We investigate dataset features (such as vocabulary occupancy and source subword overlap) and linguistic features (such as genetic distance) to identify which characteristics best predict representational similarities between languages.
Our analysis reveals that multi-parallel overlap is the most important yet under-explored feature.
\end{myquote}

\RQsub{1}{3}
\begin{myquote}
Based on our findings, we develop a novel training approach that incorporates an auxiliary similarity loss to exploit multi-parallel data.
This method increases the degree of language invariance across source representations, leading to substantial improvements in translation quality for low-resource languages.
\end{myquote}
Having established a theoretical framework for understanding and measuring cross-lingual knowledge transfer through representational similarities, we next explore further practical applications of these insights in retrieval-augmented machine translation.
Specifically, we ask:

\RQ{2}
\begin{myquote}
In this research question, we explore how cross-lingual knowledge transfer can be leveraged in the context of \textit{k}-nearest neighbor machine translation to address its limitations for low-resource languages.
We examine how datastores built from high-resource languages can benefit low-resource translation and how multiple languages can be effectively combined into a single datastore.
We divide this question into three sub-questions and address them in Chapter~\ref{ch4}:
\end{myquote}

\RQsub{2}{1}
\begin{myquote}
We examine whether cross-lingual datastores (using examples from high-resource languages to improve low-resource translation) can overcome the limitations of small bilingual datastores for low-resource languages.
Our experiments across various language pairs demonstrate that cross-lingual datastores consistently outperform bilingual datastores for low-resource languages.
\end{myquote}

\RQsub{2}{2}
\begin{myquote}
We investigate how to effectively combine examples from multiple source languages into a single datastore to benefit both low-resource and high-resource translation.
Our results show that multilingual datastores consistently outperform both bilingual and cross-lingual datastores, with particular benefits for low-resource languages.
\end{myquote}

\RQsub{2}{3}
\begin{myquote}
Given that larger datastores lead to slower inference, we explore strategies to optimize multilingual datastores for efficiency while maintaining translation quality.
We find that language-group-specific multilingual datastores achieve comparable performance to comprehensive datastores while being significantly smaller, resulting in up to 5.3× faster inference speeds.
\end{myquote}
While our first two research questions focus on understanding and enhancing positive cross-lingual transfer in multilingual neural machine translation, we next shift our attention to large language models where fine-tuning on translation data can lead to negative transfer or interference.
Therefore, we investigate:

\RQ{3}
\begin{myquote}
In this research question, we investigate the trade-offs involved when fine-tuning large language models (LLMs) for machine translation.
While fine-tuning improves general translation quality, it may compromise beneficial capabilities that emerge from pre-training.
We examine which capabilities are affected, how these effects vary across model scales, and develop strategies to maintain these capabilities.
We divide this question into three sub-questions and address them in Chapter~\ref{ch5}:
\end{myquote}

\RQsub{3}{1}
\begin{myquote}
We identify and evaluate several capabilities that are relevant for translation, including formality steerability, few-shot adaptation to specialized domains, document-level translation, and non-literal translation of idiomatic expressions.
Our findings reveal that most of these capabilities significantly degrade after fine-tuning on parallel data, even as general translation quality improves.
\end{myquote}

\RQsub{3}{2}
\begin{myquote}
Through comprehensive experiments across multiple model scales (7B to 65B parameters) and data quantities, we find that capability degradation begins with as few as 18K fine-tuning examples and becomes more pronounced with larger datasets.
These patterns are consistent across all model sizes we investigate.
\end{myquote}

\RQsub{3}{3}
\begin{myquote}
Based on our findings, we develop a mixed-data fine-tuning approach that combines parallel translation data with monolingual data.
This method not only better preserves the model's qualitative advantages compared to parallel-only fine-tuning but also achieves better overall translation quality.
\end{myquote}
Having examined both explicit methods for enhancing cross-lingual transfer (RQ1 and RQ2) and strategies to mitigate negative transfer during fine-tuning (RQ3), we finally explore how language diversity during fine-tuning implicitly affects cross-lingual generalization.
Specifically, we ask:

\RQ{4}
\begin{myquote}
In this research question, we investigate how language diversity during fine-tuning affects both seen and unseen translation directions.
Previous research presents conflicting evidence regarding the optimal multilingual fine-tuning strategy, with some studies showing benefits from minimal language exposure while others suggest advantages from scaling to many languages.
We resolve these disparities through controlled experiments and analyze the underlying representational changes.
We divide this question into three sub-questions and address them in Chapter~\ref{ch6}:
\end{myquote}

\RQsub{4}{1}
\begin{myquote}
Through controlled fine-tuning experiments across 132 translation directions, we systematically increase language diversity to assess its impact on translation quality and cross-lingual transfer.
Our results demonstrate that models fine-tuned on the most diverse set of languages consistently outperform those trained on limited language combinations, even for language pairs explicitly included in more specialized training.
\end{myquote}

\RQsub{4}{2}
\begin{myquote}
We examine whether these patterns are consistent across different model architectures and fine-tuning conditions.
Our experiments with various model sizes, regularization techniques, and data configurations confirm that the benefits of language diversity are robust across these variations, though the effects may plateau beyond a certain diversity threshold.
\end{myquote}

\RQsub{4}{3}
\begin{myquote}
To understand the mechanisms behind these improvements, we analyze how internal model representations change when exposed to different levels of language diversity.
Our findings reveal that middle layers adapt most significantly during fine-tuning, with increased specialization for target-specific languages and enhanced cross-lingual overlap for linguistically related languages, facilitating improved transfer.
\end{myquote}

\section{Main contributions}
This thesis advances the understanding and application of cross-lingual knowledge transfer in machine translation through several approaches.
The contributions span theoretical frameworks, practical implementations, and extensive empirical analyses across diverse language settings and model architectures.

\subsection{Algorithmic contributions}
We develop novel methods and analytical frameworks for measuring, enhancing, and preserving cross-lingual knowledge transfer in multilingual translation systems:

\begin{enumerate}
    \item We formalize a representational approach to measuring cross-lingual transfer by introducing the Representational Transfer Potential (RTP) metric, which quantifies knowledge transfer between languages based on similarities in their cross-attention context vectors (Chapter~\ref{ch3}).

    \item We design an auxiliary training objective that exploits multi-parallel data to increase representational alignment across languages, enhancing knowledge transfer from high-resource to low-resource languages (Chapter~\ref{ch3}).

    \item We extend \textit{k}-nearest neighbor machine translation to the multilingual setting by developing cross-lingual and multilingual datastore architectures that enable effective retrieval of translation examples across language boundaries (Chapter~\ref{ch4}).

    \item We implement cross-lingual representation alignment through linear mapping techniques that enhance retrieval effectiveness in multilingual \textit{k}NN-MT by bringing representations from different languages into a more unified space (Chapter~\ref{ch4}).

    \item We create a mixed-data fine-tuning strategy for large language models that balances task-specific optimization with capability preservation, combining parallel translation data with monolingual text (Chapter~\ref{ch5}).
\end{enumerate}

\subsection{Empirical contributions}
Through systematic experimentation and analysis, we provide empirical insights into cross-lingual knowledge transfer across various translation paradigms:

\begin{enumerate}
    \item We establish the predictive relationship between representational similarities and translation quality improvements in multilingual models, demonstrating that RTP scores correlate significantly with BLEU gains (Spearman's $\rho=.77$, $p<0.001$).
    We identify key predictive factors for cross-lingual transfer, revealing that dataset characteristics like multi-parallel overlap and source subword overlap often outweigh traditional linguistic similarity measures (Chapter~\ref{ch3}).

    \item We demonstrate that datastores constructed from linguistically similar high-resource languages consistently improve translation quality for low-resource languages beyond what can be achieved with bilingual datastores alone.
    Our multilingual datastores organized by language groups achieve performance comparable to comprehensive datastores while reducing size by up to 75\% and increasing inference speed by up to 5.3× (Chapter~\ref{ch4}).

    \item We provide the first comprehensive analysis of how fine-tuning affects various qualitative advantages of LLMs for machine translation across multiple model scales and data regimes.
    We demonstrate that capabilities such as formality control, few-shot domain adaptation, and document-level translation degrade during fine-tuning, even as general translation quality improves (Chapter~\ref{ch5}).

    \item We resolve conflicting evidence regarding optimal multilingual fine-tuning strategies through controlled experiments across 132 translation directions.
    Our findings show that scaling language diversity during fine-tuning consistently improves translation performance for both seen and unseen language pairs (Chapter~\ref{ch6}).

    \item We analyze representational changes during multilingual fine-tuning, revealing that increased language diversity creates greater representational overlap between linguistically related languages in middle model layers while simultaneously increasing specialization for language-specific characteristics in other layers (Chapter~\ref{ch6}).
\end{enumerate}

\subsection{Resource contributions}
To facilitate reproducibility and future research, we release the following resources:

\begin{enumerate}
    \item Chapter~\ref{ch3}: Code for computing representational similarities between languages using the RTP metric and implementing the auxiliary similarity loss for enhanced cross-lingual knowledge transfer in multilingual NMT systems.

    \item Chapter~\ref{ch4}: Implementation of multilingual \textit{k}NN-MT with cross-lingual and language-group-specific datastores, including tools for datastore construction, optimization, and cross-lingual mapping.

    \item Chapter~\ref{ch5}: The IdiomsInCtx-MT dataset, containing high-quality human translations of idiomatic expressions in context across German, Russian, and English language pairs, designed for evaluating non-literal translation capabilities of machine translation systems.
    
    \item Chapter~\ref{ch6}: Code and model checkpoints for conducting controlled fine-tuning experiments with varying levels of language diversity.
\end{enumerate}

\section{Thesis overview}
After this introductory chapter, the remainder of this thesis consists of a background chapter (Chapter~\ref{ch2}), four research chapters (Chapters~\ref{ch3}--\ref{ch6}), and a concluding chapter (Chapter~\ref{ch7}).
Below we present a high-level overview of the main content of each of these chapters.

\paragraph{Chapter~\ref{ch2}: Background}
provides the necessary background for understanding the research presented in this thesis.
We begin by discussing the data foundations of machine translation, covering parallel and monolingual corpora.
We then introduce the two machine translation paradigms used throughout this thesis: neural machine translation (NMT) and large language model-based machine translation (LLM-based MT).
We explain the foundations of NMT and transformer architectures, basic transfer learning approaches, and multilingual neural machine translation.
The chapter also examines the recent paradigm shift towards large language models, their architectures, and zero-shot translation capabilities.
Finally, we discuss methods for evaluating translation quality.

\paragraph{Chapter~\ref{ch3}: Representational Knowledge Transfer in Multilingual MT}
introduces the concept of representational knowledge transfer in multilingual machine translation.
We formalize the relationship between cross-lingual transfer and cross-attention similarities between languages by introducing a new metric: Representational Transfer Potential (RTP).
Through extensive experiments, we investigate whether RTP can quantify positive and negative transfer, predict the factors relevant for transfer, and optimize translation performance through auxiliary losses.
Our findings demonstrate that representational similarities between languages are strongly correlated with effective knowledge transfer, providing insights into why certain languages benefit more from multilingual training than others.
Our results in this chapter provide answers to \textbf{RQ1}.

\paragraph{Chapter~\ref{ch4}: Multilingual \textit{k}-Nearest-Neighbor Machine Translation}
explores how cross-lingual knowledge transfer can be leveraged in the context of \textit{k}NN-MT to overcome limitations for low-resource languages.
We investigate whether a low-resource language can benefit from a datastore built from a related high-resource language, and how to effectively combine multiple languages into datastores.
Our results show that cross-lingual datastores significantly improve translation quality for low-resource languages, and that multilingual datastores benefit both low-resource and high-resource languages while maintaining computational efficiency.
This chapter addresses \textbf{RQ2}.

\paragraph{Chapter~\ref{ch5}: Fine-Tuning and Capability Preservation in LLM Translation}
examines the trade-off between improving translation quality and preserving the beneficial capabilities of large language models during fine-tuning.
We investigate which specific LLM capabilities are affected by fine-tuning, how these effects vary across different model scales and data quantities, and how to maintain capabilities while still improving translation quality.
Our findings reveal that fine-tuning on parallel data improves general translation quality but compromises several capabilities that are relevant for translation, such as formality steerability and document-level translation.
We introduce a mixed-data fine-tuning approach that better preserves these capabilities while still improving translation performance. 
This chapter provides answers to \textbf{RQ3}.

\paragraph{Chapter~\ref{ch6}: The Effect of Language Diversity in LLM Translation}
investigates how language diversity during fine-tuning affects translation performance across both seen and unseen language pairs.
Through controlled experiments across 132 translation directions, we find that increasing language diversity consistently improves performance for both seen and unseen language combinations.
Our representational analysis reveals that middle layers undergo the most substantial adaptation, and that for linguistically related languages, our most diverse model shows enhanced representational overlap, indicating improved cross-lingual transfer.
These findings explain the superior performance of models fine-tuned on greater linguistic diversity and provide answers to \textbf{RQ4}.

\paragraph{Chapter~\ref{ch7}: Conclusions}
summarizes the main findings of this thesis in relation to our research questions and proposes directions for future work.

\section{Origins}
The research presented in Chapters \ref{ch3}--\ref{ch6} of this thesis is based on several peer-reviewed publications. We indicate the origins of each chapter below.

\begin{itemize}
    \item \textbf{Chapter 3}\\  
    \textbf{David Stap}, Vlad Niculae, and Christof Monz.
    2023.
    Viewing Knowledge Transfer in Multilingual Machine Translation Through a Representational Lens.
    In \textit{Findings of the Association for Computational Linguistics: EMNLP 2023}, pages 14973–14987.
    \nocite{stap_viewing_2023}

    \item \textbf{Chapter 4}\\
    \textbf{David Stap} and Christof Monz.
    2023.
    Multilingual \textit{k}-Nearest-Neighbor Machine Translation.
    In \textit{Proceedings of the 2023 Conference on Empirical Methods in Natural Language Processing}, pages 9200–9208.
    \nocite{stap_multilingual_2023}

    \item \textbf{Chapter 5}\\
    \textbf{David Stap}, Eva Hasler, Bill Byrne, Christof Monz, Ke Tran.
    2024.
    The Fine-Tuning Paradox: Boosting Translation Quality Without Sacrificing LLM Abilities.
    In \textit{Proceedings of the 62nd Annual Meeting of the Association for Computational Linguistics}, pages 6189–6206.
    \nocite{stap_finetuning_2024}

    \item \textbf{Chapter 6}\\
    \textbf{David Stap} and Christof Monz.
    2025.
    The Effect of Language Diversity When Fine-Tuning Large Language Models for Translation.
    Accepted for publication in the \textit{Findings of the Association for Computational Linguistics: EMNLP 2025}.
    \nocite{stap_effect_2025}
\end{itemize}
The writing of this thesis also benefited from work on the following publications:

\begin{itemize}
    \item \textbf{David Stap}, Maurits Bleeker, Sarah Ibrahimi, and Maartje ter Hoeve.
    2020.
    Conditional Image Generation and Manipulation for User-Specified Content.
    In \textit{Proceedings of the IEEE/CVF Conference on Computer Vision and Pattern Recognition (CVPR) Workshop}.
    \nocite{stap_conditional_2020}
    
    \item Yizhong Wang, Swaroop Mishra, Pegah Alipoormolabashi, [...], \textbf{David Stap} (10/35), [...], Sumanta Patro, Tanay Dixit, and Xudong Shen.
    2022.
    Super-NaturalInstructions: Generalization via Declarative Instructions on 1600+ NLP Tasks.
    In \textit{Proceedings of the 2022 Conference on Empirical Methods in Natural Language Processing}.
    \nocite{wang_super-naturalinstructions_2022}
    
    \item Just Zwennicker and \textbf{David Stap}.
    2022.
    Towards a general purpose machine translation system for Sranantongo. In \textit{Proceedings of the Sixth Widening NLP Workshop (WiNLP)}.
    \nocite{zwennicker_towards_2022}
    
    \item \textbf{David Stap} and Ali Araabi.
    2023.
    ChatGPT is not a good indigenous translator.
    In \textit{Proceedings of the Workshop on Natural Language Processing for Indigenous Languages of the Americas (AmericasNLP)}.
    \nocite{stap_chatgpt_2023}
    
    \item Di Wu, Shaomu Tan, \textbf{David Stap}, Ali Araabi, and Christof Monz.
    2023.
    UvA-MT’s Participation in the WMT 2023 General Translation Shared Task.
    In \textit{Proceedings of the Eighth Conference on Machine Translation}.
    \nocite{wu_uva-mts_2023}
    
    \item Danai Xezonaki, Talaat Khalil, \textbf{David Stap}, and Brandon Denis.
    2023.
    Improving Domain Robustness in Neural Machine Translation with Fused Topic Knowledge Embeddings.
    In \textit{Proceedings of Machine Translation Summit XIX, Vol. 1: Research Track}.
    \nocite{xezonaki-etal-2023-improving}
    
    \item Di Wu, Shaomu Tan, Yan Meng, \textbf{David Stap}, and Christof Monz.
    2024.
    How Far Can 100 Samples Go? Unlocking Overall Zero-Shot Multilingual Translation via Tiny Multi-Parallel Data.
    In \textit{Findings of the Association for Computational Linguistics: ACL 2024}.
    \nocite{wu-etal-2024-far}
    
    \item Wei-Lin Chen, Jenny Chim, Leshem Choshen, Luca D’Amico-Wong, Melissa Dell, Run-Ze Fan, Shahriar Golchin, Yucheng Li, Pengfei Liu, Bhavish Pahwa, Ameya Prabhu, Suryansh Sharma, Emily Silcock, Kateryna Solonko, \textbf{David Stap}, Mihai Surdeanu, Yu-Min Tseng, Vishaal Udandarao, Zengzhi Wang, Ruijie Xu, and Jinglin Yang.
    2024.
    Data Contamination Report from the 2024 CONDA Shared Task.
    In \textit{Proceedings of the 1st Workshop on Data Contamination (CONDA)}.
    \nocite{sainz-etal-2024-data}

    \item Shaomu Tan, Di Wu, \textbf{David Stap}, Seth Aycock, and Christof Monz.
    2024.
    UvA-MT’s Participation in the WMT 2024 General Translation Shared Task.
    In \textit{Proceedings of the Ninth Conference on Machine Translation}.
    \nocite{tan-etal-2024-uva}

    \item David Ifeoluwa Adelani, Duygu Ataman, Mammad Hajili, Raghav Mantri, \textbf{David Stap}, Jonne Sälevä, and Abraham Owodunni.
    2024.
    \textit{Proceedings of the Fourth Workshop on Multilingual Representation Learning (MRL 2024)}.
    \nocite{mrl-2024-1}

    \item Kenneth Enevoldsen, Isaac Chung, Imene Kerboua, Márton Kardos, Ashwin Mathur, \textbf{David Stap} (6/81), [...],  Orion Weller, Siva Reddy, and Niklas Muennighoff.    
    2024.
    MMTEB: Massive multilingual text embedding benchmark.
    In \textit{Proceedings of the thirteenth international conference on learning representations (ICLR)}.
    \nocite{enevoldsen2025mmteb}

    \item Seth Aycock, \textbf{David Stap}, Di Wu, Christof Monz, and Khalil Sima{\textquotesingle}an.
    2024.
    Can LLMs Really Learn to Translate a Low-Resource Language from One Grammar Book?
    In \textit{Proceedings of the thirteenth international conference on learning representations (ICLR)}.
    \nocite{aycock2025can}

    \item Long Phan, Alice Gatti, Ziwen Han, [...], \textbf{David Stap} (456/732), [...], Summer Yue, Alexandr Wang, and Dan Hendrycks.
    2024.
    Humanity's Last Exam.
    In \textit{arXiv:2501.14249 [cs]}.
    \nocite{phan_humanitys_2025}
    
\end{itemize}

\chapter{Background}\label{ch2}
This chapter provides the necessary background for understanding the research presented in this thesis.
We begin by discussing the data foundations of machine translation in Section~\ref{ch2:sec:data}, covering parallel and monolingual corpora that are essential for training translation systems.
Next, we discuss the two machine translation paradigms used throughout this thesis: neural machine translation (NMT, Section~\ref{ch2:sec:nmt}) and large language model-based machine translation (LLM-based MT, Section~\ref{ch2:sec:llm}).
We first explain the foundations of NMT and transformer architectures (Section~\ref{ch2:sec:transformer}). We then introduce basic transfer learning in NMT (Section~\ref{ch2:sec:basic_transfer}), explaining how knowledge can be transferred from one language pair to another. This leads to our discussion of multilingual neural machine translation (Section~\ref{ch2:sec:mnmt}), which enables translation between multiple language pairs simultaneously. We then explore advanced mechanisms for cross-lingual knowledge transfer (Section~\ref{ch2:sec:adv_transfer}) that are central to this thesis.
The LLM section examines the recent paradigm shift towards large language models, their architectures and pre-training objectives (Section~\ref{ch2:sec:llm_arch}), zero-shot translation capabilities (Section~\ref{ch2:sec:llm_zero}), and approaches for adapting them to translation tasks (Section~\ref{ch2:sec:llm_ft}).
Finally, Section~\ref{ch2:sec:evaluation} discusses methods for evaluating translation quality.

\section{Data foundations for machine translation}\label{ch2:sec:data}
Translation systems rely heavily on data for learning cross-lingual mappings between languages. In this section, we discuss the types of data used for training translation models, their characteristics, and how they are employed in different machine translation paradigms.

\subsection{Parallel corpora}
Parallel corpora, also known as bitexts, consist of aligned texts in two or more languages, where each segment in one language is paired with its translation in another language.
These corpora are the primary training data for supervised machine translation systems, from traditional statistical models to modern neural approaches and fine-tuned large language models.
Typically, parallel corpora are aligned at the sentence level, with each source sentence paired with its corresponding target translation.
The size of parallel datasets varies significantly across language pairs.
High-resource language pairs like English-German or English-French often have millions of sentence pairs available, while low-resource language pairs may have only thousands or even fewer parallel examples \citep{arivazhagan_massively_2019}.

The quality and quantity of parallel data significantly impact translation performance.
High-quality parallel corpora often come from professionally translated sources such as parliamentary proceedings like Europarl \citep{koehn_europarl_2005} or legal documents such as JRC-Acquis \citep{steinberger-etal-2006-jrc} containing European Union law texts, news commentary from sources like WMT shared tasks \citep{barrault_findings_2019,barrault-etal-2020-findings,akhbardeh-etal-2021-findings,kocmi_findings_2022,kocmi_findings_2023,kocmi-etal-2024-findings}, and United Nations documents \citep{ziemski_united_2016}.

However, obtaining high-quality parallel data through professional translation is expensive and time-consuming.
This has led to the development of various methods for web mining to create larger parallel datasets. These methods automatically extract potential translation pairs from multilingual websites, resulting in datasets such as ParaCrawl \citep{banon_paracrawl_2020}, CCMatrix \citep{schwenk_ccmatrix_2021}, and NLLB \citep{fan_beyond_2021}.

A particularly valuable type of parallel resource are multi-parallel corpora, where the same content is available in \textit{more than two languages} simultaneously.
Notable examples include the FLORES-101 benchmark \citep{goyal_flores-101_2022}, which provides professional translations of the same 3,001 sentences across 101 languages, and NTREX-128 \citep{federmann_ntrex-128_2022}, which consists of 1,997 multi-parallel sentences in 128 languages.
Multi-parallel test sets enable consistent evaluation across diverse language pairs by providing semantically identical content in all languages, thus removing content variation as a confounding factor.
This makes them valuable for studying cross-lingual knowledge transfer (Chapters \ref{ch3}, \ref{ch4}, and \ref{ch6}) and for comparing translation performance between high and low-resource language pairs.

\subsection{Monolingual corpora}\label{ch2:sec:data:monolingual}
Monolingual corpora contain text in a single language and are vastly more abundant than parallel data.
Their role in machine translation has evolved significantly across different paradigms.

In NMT, monolingual data has traditionally been used for back-translation \citep{sennrich_improving_2016}, where synthetic parallel data is created by translating target-language monolingual text into the source language; forward translation or self-training \citep{zhang_exploiting_2016}, generating synthetic parallel data by translating source-language monolingual text; and building stronger language models for the target language \citep{domhan_using_2017,burlot_using_2018}.

In the LLM era, monolingual data serves a foundational role.
Pre-training of large language models occurs on massive text corpora that are heavily dominated by English and other high-resource languages.
Datasets like C4 \citep{raffel_exploring_2020}, The Pile \citep{gao_pile_2020}, RefinedWeb \citep{penedo_refinedweb_2023}, and Dolma \citep{soldaini_dolma_2024}, while containing some multilingual content, are predominantly English-centric.
Other efforts like mC4 \citep{xue_mt5_2021}, OSCAR \citep{abadji_towards_2022}, MADLAD-400 \citep{kudugunta2023madlad}, and HPLT \citep{de-gibert-etal-2024-new} have focused on creating more balanced multilingual pre-training corpora, though significant resource disparities across languages remain a challenge.
This pre-training enables LLMs to develop broad linguistic knowledge across languages, forming the basis for their zero-shot translation capabilities (see Section~\ref{ch2:sec:llm_zero}).
After pre-training, LLMs can be fine-tuned on parallel data to improve their translation performance, a topic explored in more detail in Section~\ref{ch2:sec:llm_ft}.

\section{Neural machine translation}\label{ch2:sec:nmt}
\subsection{Core principles and architecture}\label{ch2:sec:nmt_architecture}
NMT represents a paradigm shift from earlier statistical approaches, modeling the translation process as a single neural network that directly maps source language text to target language text.
Unlike phrase-based statistical models that decompose translation into multiple components, NMT systems learn to encode the source sentence into a continuous representation and decode it into the target language within an end-to-end framework.

At its core, NMT aims to model the conditional probability of a target sentence $\mathbf{y}=(y_1,y_2,...,y_m)$ given a source sentence $\mathbf{x} = (x_1, x_2, ..., x_n)$, where $x_i$ represents the $i$-th token (typically a word or subword unit) in the source sentence of length $n$, and $y_j$ represents the $j$-th token in the target sentence of length $m$:
\begin{equation}
    p(\mathbf{y}\mid\mathbf{x})=\prod_{t=1}^m p(y_t\mid\mathbf{y}_{<t},\mathbf{x}),
\end{equation}
where $\mathbf{y}_{<t}$ represents the target word sequence generated before position $t$.
This formulation captures the auto-regressive nature of translation, where each target word depends on both the source sentence and previously generated target words.

The training objective for NMT models is to maximize the log-likelihood of the parallel training data, or equivalently, to minimize the cross-entropy loss:
\begin{equation}\label{ch2:eq:nmt_loss}
    \mathcal{L}={-}\sum_{(\mathbf{x},\mathbf{y})\in\mathcal{D}} \log p(\mathbf{y}\mid\mathbf{x}),
\end{equation}
where $\mathcal{D}$ represents the parallel training corpus.
This optimization is typically performed using variants of stochastic gradient descent, with the gradient computed via backpropagation.

\paragraph{Inference}
During inference, a trained NMT model generates a target translation $\mathbf{y}$ given a source sentence $\mathbf{x}$.
While the model is trained to predict the next token $y_t$ given previous context $\mathbf{y}_{<t}$ and $\mathbf{x}$, generating a complete translation requires an efficient search strategy through the vast space of possible sequences.

The most widely used approach is \textit{beam search}, a heuristic search algorithm originating from early artificial intelligence research \citep{10.5555/907741} and adopted in various natural language processing tasks.
In NMT, beam search maintains a fixed number ($k$) of partial hypotheses at each decoding step.
The beam size $k$ represents the trade-off between translation quality and computational efficiency.
Formally, beam search aims to find:
\begin{equation}
\hat{\mathbf{y}} = \underset{\mathbf{y}}{\arg\max} \log p(\mathbf{y}|\mathbf{x}).
\end{equation}
At each step, the algorithm expands each hypothesis by considering all possible next tokens from the vocabulary, scores the resulting sequences, and keeps only the top-$k$ highest scoring candidates.

Recent research has identified several limitations of standard beam search.
One significant challenge is the length bias problem, where beam search tends to favor shorter translations \citep{murray-chiang-2018-correcting}. 
Various normalization techniques have been proposed to address this, including length normalization \citep{wu_googles_2016} and coverage-based penalties \citep{tu-etal-2016-modeling}.
Another issue is the ``beam search curse''—the paradoxical observation that increasing beam size beyond a certain point often degrades translation quality despite exploring more of the search space \citep{koehn_six_2017,stahlberg-byrne-2019-nmt}.
This phenomenon has been attributed to exposure bias and model calibration issues.

To address these limitations, researchers have developed various decoding alternatives.
Some approaches modify beam search directly, such as diverse beam search \citep{vijayakumar_diverse_2018}, which promotes diversity among hypotheses, and minimum Bayes risk decoding \citep{eikema-aziz-2022-sampling}, which selects translations based on expected utility rather than probability alone.
Other methods take more radical departures from traditional decoding, including non-autoregressive translation \citep{gu_non-autoregressive_2018}, which generates multiple target tokens simultaneously, and neural noisy channel approaches \citep{yee-etal-2019-simple}, which incorporate reverse translation probabilities to improve robustness.

Despite these innovations, standard beam search with moderate beam sizes (typically 4-10) remains predominant in production MT systems due to its reasonable balance between quality, computational efficiency, and implementation simplicity.

\paragraph{Architectures}
Early NMT systems relied on recurrent neural networks (RNNs), with \citet{kalchbrenner_recurrent_2013} introducing recurrent continuous translation models.
These approaches evolved to incorporate LSTMs (Long Short-Term Memory; \citealp{hochreiter_long_1997}) or GRUs (Gated Recurrent Unit    s; \citealp{cho_learning_2014}), structured as encoder-decoder architectures \citep{sutskever_sequence_2014, cho_learning_2014}.
Subsequently, attention mechanisms \citep{bahdanau_neural_2015, luong_effective_2015} enhanced these architectures, allowing the decoder to focus on relevant parts of the source sentence when generating each target word.
While effective, these recurrent architectures struggled with long-range dependencies and were limited in their parallelization capabilities during training.

The transformer architecture \citep{vaswani_attention_2017} addressed these limitations and has since become the dominant approach not only in NMT but throughout the entire field of natural language processing, forming the foundation for the models used throughout this thesis.
Unlike recurrent models that process text sequentially, transformers rely entirely on attention mechanisms, enabling highly parallel processing of the entire sequence.

\subsection{Transformer models}\label{ch2:sec:transformer}
The transformer architecture \citep{vaswani_attention_2017} revolutionized natural language processing by introducing a model based entirely on attention mechanisms, eliminating recurrence.

\subsubsection{Architecture overview}
At a high level, transformers follow the encoder-decoder architecture common in sequence-to-sequence models but replace recurrent connections with multi-head attention mechanisms.
The architecture consists of stacked encoder and decoder layers, with each layer containing two main components: a multi-head attention mechanism and a position-wise feed-forward network.

The encoder processes the source sentence $\mathbf{x} = (x_1, x_2, ..., x_n)$ and transforms it into a sequence of continuous representations $\mathbf{z} = (z_1, z_2, ..., z_n)$.
Each encoder layer processes its input through a multi-head self-attention mechanism where each position $i$ attends to all positions $j$ in the previous layer with attention weights $\alpha_{ij}$.
This allows the model to incorporate context from the entire sentence when encoding each word.
The self-attention mechanism is followed by a position-wise feed-forward network $\text{FFN}(\mathbf{z}) = \max(0, \mathbf{z}\mathbf{W}_1 + \mathbf{b}_1)\mathbf{W}_2 + \mathbf{b}_2$ applied identically to each position.

The decoder generates the target sentence $\mathbf{y} = (y_1, y_2, ..., y_m)$ one token at a time in an auto-regressive manner, modeling $p(y_t\mid\mathbf{y}_{<t}, \mathbf{z})$.
Each decoder layer contains two multi-head attention mechanisms: a masked self-attention layer that prevents positions from attending to future positions $j>i$ (ensuring auto-regressive generation), and a cross-attention layer that attends to the encoder's output $\mathbf{z}$ with attention weights $\beta_{ij}$ between decoder position $i$ and encoder position $j$.
This cross-attention mechanism allows each position in the decoder to focus on relevant parts of the source sentence.

\subsubsection{Attention mechanism}
The attention mechanism in transformers is a generalization of attention mechanisms proposed in earlier neural machine translation models \citep{bahdanau_neural_2015,luong_effective_2015}, where it was primarily used to allow the decoder to focus on relevant parts of the source sequence.
The transformer expands this concept by applying attention more broadly within the architecture.
In particular, the transformer employs two types of attention: \textit{cross-attention} (between encoder and decoder) and \textit{self-attention} (within encoder or decoder).
Both share the same underlying computation.

Formally, the attention mechanism computes attention scores as:
\begin{equation}
    \text{Attention}(\mathbf{Q}, \mathbf{K}, \mathbf{V}) = \text{softmax}\left(\frac{\mathbf{Q}\mathbf{K}^T}{\sqrt{d_k}}\right)\mathbf{V},
\end{equation}
where $\mathbf{Q}$, $\mathbf{K}$, and $\mathbf{V}$ represent query, key, and value matrices derived from the input representations, and $d_k$ is the dimension of the key vectors, serving as a scaling factor.
In self-attention, these matrices are all derived from the same sequence, while in cross-attention, the queries come from one sequence (decoder) and the keys and values from another (encoder).

In multi-head attention, this computation is performed in parallel across several `heads', allowing the model to jointly attend to information from different representation subspaces:
\begin{equation}
    \text{MultiHead}(\mathbf{Q}, \mathbf{K}, \mathbf{V}) = \text{Concat}(\text{head}_1, ..., \text{head}_h)\mathbf{W}^O,
\end{equation}
where each head is computed as:
\begin{equation}
    \text{head}_i = \text{Attention}(\mathbf{Q}\mathbf{W}_i^Q, \mathbf{K}\mathbf{W}_i^K, \mathbf{V}\mathbf{W}_i^V).
\end{equation}
While attention effectively captures dependencies regardless of distance, it comes with computational challenges.
The attention mechanism requires computing pairwise interactions between all tokens, resulting in quadratic $O(n^2)$ time and memory complexity with respect to sequence length.
This limitation becomes particularly problematic for long documents or when processing large batches.

Numerous approaches have been proposed to address this efficiency bottleneck.
Sparse attention mechanisms like Sparse Transformer \citep{child_generating_2019} reduce complexity by attending only to a subset of positions.
Linear attention methods such as Linformer \citep{wang_linformer_2020} and Performer \citep{choromanski2021rethinking} approximate the attention computation to achieve linear complexity.
Other approaches include Reformer \citep{kitaev_reformer_2020}, which uses locality-sensitive hashing to attend only to similar representations, and Longformer \citep{beltagy_longformer_2020}, which combines local windowed attention with task-specific global attention.
These efficient attention variants have enabled transformers to process significantly longer sequences while maintaining strong performance.

\subsubsection{Positional encoding}
Since the transformer lacks recurrence, positional information must be explicitly provided.
This is accomplished through positional encodings added to the input embeddings.
The standard approach uses sine and cosine functions of different frequencies:
\begin{equation}
\begin{split}
\text{PE}_{(pos, 2i)} &= \sin\left(\frac{pos}{10000^{2i/d{model}}}\right) \\
\text{PE}_{(pos, 2i+1)} &= \cos\left(\frac{pos}{10000^{2i/d{model}}}\right),
\end{split}
\end{equation}
where $pos$ is the position and $i$ is the dimension.
These encodings have the useful property that the relative positions of tokens can be computed as linear functions of their positional encodings.
The constant 10000 was chosen to create a spectrum of wavelengths from very short to very long, allowing the model to attend to both local patterns and long-range dependencies.
This creates a unique encoding for each position while ensuring that positions with similar semantic relationships have similar relative encodings across different sequence lengths.

The original transformer paper \citep{vaswani_attention_2017} also experimented with \textit{learnable} positional embeddings, which were first introduced by \citet{gehring_convolutional_2017} for convolutional sequence-to-sequence models, where distinct vectors for each position are directly optimized during training.
While slightly outperformed by sinusoidal encodings \citep{vaswani_attention_2017}, learnable positional embeddings have become common in many modern transformer implementations like BERT \citep{devlin_bert_2019} and GPT \citep{radford_language_2019}.

Subsequent research has introduced more sophisticated positional encoding methods.
Rotary Position Embedding (RoPE; \citealp{su_roformer_2024}) directly encodes relative position information into the attention computation by applying a rotation matrix to the query and key vectors, preserving the ability to generalize to longer sequences.
Relative positional encoding \citep{shaw_self-attention_2018} explicitly models the relationships between pairs of positions rather than absolute positions.
T5 \citep{raffel_exploring_2020} introduced relative attention bias, adding learned scalar biases to attention scores based on relative positions.
These advancements in positional encoding have helped transformers better model sequential information across various natural language tasks while maintaining their parallelization advantages.

\subsubsection{Model variants}
The original transformer architecture features both an encoder and decoder component, making it well-suited for sequence-to-sequence tasks like machine translation.
However, subsequent research has explored various architectural adaptations, including encoder-only models (like BERT; \citealp{devlin_bert_2019}) for tasks such as text classification and natural language understanding, and decoder-only models for text generation tasks.

Decoder-only transformers, which use only the self-attention component without cross-attention to an encoder, have become particularly important in the development of LLMs.
These models, trained auto-regressively on vast amounts of text data, have demonstrated remarkable capabilities across diverse language tasks.
The specifics of decoder-only architectures are discussed further in Section~\ref{ch2:sec:llm} on LLMs.

\subsubsection{Wider impact on natural language processing and beyond}
The transformer architecture has proven highly effective not only for machine translation but across the entire spectrum of natural language processing tasks.
Its success has led to the development of transformer-based pre-trained models that have redefined the state of the art in areas such as text classification \citep{devlin_bert_2019}, question answering \citep{NEURIPS2020_6b493230}, summarization \citep{zhang_pegasus_2020}, and language generation \citep{brown_language_2020}.
The transformer's influence extends across diverse applications in NLP, from biomedical text analysis \citep{beltagy_scibert_2019} to code understanding \citep{feng_codebert_2020} and multimodal learning \citep{10.5555/3454287.3454289}.

Beyond language processing, the transformer architecture has driven remarkable advances across multiple AI domains.
In computer vision, Vision Transformers \citep{dosovitskiy2021an} challenged the dominance of convolutional neural networks, delivering exceptional performance on image classification by treating image patches as token sequences.
This powerful approach has been successfully extended to video understanding\citep{9710415}, 3D point cloud processing \citep{Zhao_2021_ICCV}, and medical image analysis \citep{CHEN2024103280}, often surpassing previous state-of-the-art methods. 
Additionally, transformers have become essential building blocks for groundbreaking multimodal systems bridging text and visual information, including CLIP \citep{pmlr-v139-radford21a}, which learns visual concepts from natural language supervision at unprecedented scale, and DALL-E \citep{ramesh2021zeroshot}, which generates remarkably detailed images from text descriptions.
In scientific applications, transformers have enabled major breakthroughs such as highly accurate protein structure prediction with AlphaFold \citep{jumper2021highly} and sophisticated chemical reaction prediction \citep{doi:10.1021/acscentsci.9b00576}.
Importantly, the architecture's scalability has been fundamental to the development of large language models (see Section~\ref{ch2:sec:llm}), enabling powerful capabilities in understanding and generating human language across diverse and increasingly complex applications \citep{openai_gpt-4_2023}.

Reflecting this architectural evolution, Chapters \ref{ch3} and \ref{ch4} of this thesis make use of transformer-based encoder-decoder architectures, while Chapters \ref{ch5} and \ref{ch6} investigate translation capabilities in transformer-based decoder-only LLMs.

\section{Basic transfer learning in neural machine translation}\label{ch2:sec:basic_transfer}
Transfer learning is a fundamental approach in machine learning where knowledge gained from solving one problem is applied to a different but related problem.
In machine learning broadly, and NMT specifically, transfer learning addresses the data scarcity problem that plagues many applications and languages, allowing models to benefit from data-rich settings when working with low-resource scenarios.

\subsection{Parent-child transfer learning}
The basic principle of transfer learning in NMT involves training a model on a high-resource language pair (parent) and transferring knowledge to a low-resource language pair (child).
This transfer learning process can be formalized as:
\begin{enumerate}
    \item Train parent model $\theta_p$ on high-resource data $D_p$: $\theta_p = \arg\min_\theta \mathcal{L}(\theta, D_p)$
    \item Initialize child model with parent parameters and fine-tune on low-resource data $D_c$: $\theta_c = \arg\min_\theta \mathcal{L}(\theta, D_c)$, where $\theta$ is initialized with $\theta_p$
\end{enumerate}
This approach follows the same training objective as discussed in Section~\ref{ch2:sec:nmt} (Equation \ref{ch2:eq:nmt_loss}), applied in two stages.
\citet{zoph_transfer_2016} demonstrated that this approach could achieve significant improvements compared to training on the low-resource data alone.
The effectiveness depends on several factors, including the language relatedness, corpus sizes, and specific transfer methodology.

\subsection{Impact of language relatedness}
The relationship between the parent and child languages influences the effectiveness of transfer learning.
Empirical studies have yielded somewhat contradictory results: \citet{zoph_transfer_2016} found that using a related language as the parent model produced better results, while \citet{kocmi_trivial_2018} observed that the size of the parent model's training data could be more important than linguistic similarity.
\citet{dabre-etal-2017-empirical} showed that transfer between related languages tends to be more effective for morphologically rich languages.
These findings suggest that while language relatedness is beneficial, it may be overshadowed by other factors such as data availability and quality.

Recent studies have investigated which linguistic factors contribute most significantly to cross-lingual transfer.
Syntactic similarity—particularly word order—has been identified as a key factor across multiple studies \citep{pires_how_2019,lauscher_zero_2020,dolicki_analysing_2021,ahuja-etal-2022-multi,de-vries-etal-2022-make}.
Additional linguistic features that positively impact transfer include low genetic distance \citep{lin_choosing_2019,eronen_transfer_2022}, geographical proximity \citep{lauscher_zero_2020,ahuja-etal-2022-multi}, and phonological similarity \citep{lin_choosing_2019,ahuja-etal-2022-multi}, though their relative importance varies by task type and language pair combination.

\subsection{Limitations for multiple languages}
While the parent-child transfer approach is effective for a single language pair, it faces significant limitations when scaling to multiple languages:
\begin{enumerate}
    \item \textbf{Sequential nature}: The traditional parent-child paradigm requires training models sequentially, which becomes impractical as the number of languages increases.
    \item \textbf{Model proliferation}: Each language pair requires a separate model, leading to a quadratic growth in the number of models needed as languages are added ($n^2$ models for $n$ languages).
    \item \textbf{Inefficient parameter usage}: Each model maintains its own parameters, even though many language processing capabilities could potentially be shared.
    \item \textbf{Inconsistent knowledge transfer}: When transferring knowledge to multiple child languages independently, there is no mechanism to ensure consistent cross-lingual representations across all languages.
\end{enumerate}
These limitations motivated the development of multilingual neural machine translation (MNMT) approaches that enable simultaneous training on multiple language pairs within a single model, which we discuss in the next section.

\section{Multilingual neural machine translation}\label{ch2:sec:mnmt}
Multilingual neural machine translation (MNMT) extends the standard NMT framework to accommodate multiple language pairs within a single model.
This approach offers several advantages over maintaining separate bilingual models, including more efficient parameter usage and improved translation quality through parameter sharing \citep{johnson_googles_2017,ha_toward_2016}.
MNMT systems are typically categorized based on their directionality: one-to-many (O2M), many-to-one (M2O), or many-to-many (M2M).

\subsection{Translation directionality}
\subsubsection{One-to-many (O2M) translation}
In O2M translation, a single source language is translated into multiple target languages.
The encoder processes input from a specific source language, while the decoder generates outputs in various target languages.

This setup is particularly useful for scenarios where content in a resource-dominant language (often English) needs to be disseminated in multiple target languages.
\citet{dong_multi-task_2015} demonstrated that an O2M NMT system with a shared encoder and language-specific decoders could outperform individual bilingual models, suggesting that the shared encoder learns more robust source representations when trained on multiple translation tasks simultaneously.

\subsubsection{Many-to-one (M2O) translation}
M2O translation involves multiple source languages being translated into a single target language.
In this setup, the model may employ language-specific encoders or a shared encoder with language indicators, while the decoder focuses solely on generating text in the target language.

\citet{firat_multi-way_2016} showed that M2O systems benefit from the diverse inputs across source languages, which helps the model develop more generalizable representations and improves target-language fluency through the shared decoder.
M2O systems are particularly effective for aggregating information from multiple languages into a high-resource language like English.

\subsubsection{Many-to-many (M2M) translation}
M2M translation represents the most comprehensive and challenging MNMT configuration, supporting translation between any pair within a set of languages.

Early approaches to M2M translation, such as the one proposed by \citet{firat_multi-way_2016}, employed multiple encoders and decoders with a shared attention mechanism.
However, this architecture resulted in models with a large number of parameters that scaled poorly with the addition of new languages \citep{johnson_googles_2017}.

\subsection{Parameter sharing in MNMT}
Parameter sharing across languages is essential in MNMT to enable efficient scaling to many languages and facilitate cross-lingual knowledge transfer.
By sharing representations and parameters, MNMT systems can learn from high-resource languages to improve performance on low-resource ones, while maintaining reasonable model sizes.
The extent of parameter sharing can range from minimal to complete:

\subsubsection{Minimal sharing}
In the minimal sharing approach, separate encoders and decoders are maintained for each language pair, with only the attention mechanism shared across all pairs \citep{firat_multi-way_2016}.
This approach allows for language-specific processing but results in models whose size grows linearly with the number of languages:
\begin{equation}
    \text{Parameter count} \approx |L_s| \times E + |L_t| \times D + A
\end{equation}
where $|L_s|$ is the number of source languages, $|L_t|$ is the number of target languages, $E$ is the encoder size, $D$ is the decoder size, and $A$ is the attention mechanism size.

\subsubsection{Complete sharing}
\citet{johnson_googles_2017} introduced a more elegant solution to M2M translation using a completely shared model for all language pairs.
This approach uses a single encoder, attention mechanism, and decoder for all languages, with a special token prepended to the source sentence to indicate the desired target language:
\begin{equation}
    \text{Input} = [\langle\text{2$l_t$}\rangle, x_1^{l_s}, x_2^{l_s}, \ldots, x_n^{l_s}]
\end{equation}
where $\langle\text{2$l_t$}\rangle$ indicates the target language, e.g., $\langle\text{2fr}\rangle$ indicating that French is the desired target language), and $(x_1^{l_s}, x_2^{l_s}, ..., x_n^{l_s})$ is the source sentence.
Despite its simplicity, this method has proven highly effective, especially for low-resource languages that benefit from cross-lingual transfer.
The parameter efficiency of this approach has made it the foundation for massively multilingual NMT systems covering 100+ languages \citep{aharoni_massively_2019,arivazhagan_massively_2019,fan_beyond_2021}.

\subsubsection{Intermediate sharing strategies}
Between these extremes, various intermediate parameter sharing strategies exist.
\citet{ha_toward_2016} proposed an approach that maintains separate vocabularies for each language while sharing other model parameters.
\citet{sachan_parameter_2018} explored different parameter sharing configurations, and found that sharing encoder self-attention and decoder cross-attention mechanisms is particularly beneficial for linguistically dissimilar languages.

\subsection{Main challenge in multilingual neural machine translation}
A central challenge in MNMT is balancing language-specific and language-agnostic representations.
While sharing parameters across languages enables cross-lingual transfer, it can also introduce a representational bottleneck that limits performance, particularly for high-resource languages \citep{arivazhagan_massively_2019}.
This phenomenon, known as the ``curse of multilinguality'' \citep{conneau_unsupervised_2020}, describes the trade-off between the number of languages a model can effectively handle and its per-language performance given fixed model capacity \citep{ahuja-etal-2022-multi,eronen_transfer_2022}.

Several approaches aim to address this through techniques like language-specific adapter layers \citep{bapna_non-parametric_2019,philip_monolingual_2020,pfeiffer_mad-x_2020,ustun_hyper-x_2022}, which can be added to a pre-trained multilingual model to enhance performance for specific language pairs without significant parameter growth:
\begin{equation}
    h_l = \text{Adapter}_l(h)
\end{equation}
where $h$ is the shared multilingual representation from the main model and $h_l$ is the language-specific adapted representation after applying the transformation $\text{Adapter}_l$ for language $l$.

\section{Advanced cross-lingual knowledge transfer techniques}\label{ch2:sec:adv_transfer}
Building on the basic transfer learning principles and multilingual NMT approaches described in previous sections, researchers have developed more sophisticated techniques for cross-lingual knowledge transfer. These advanced methods aim to better align representations across languages and optimize the transfer of knowledge from high-resource to low-resource languages.

\subsection{Representation alignment techniques}
Several approaches have been developed to enhance cross-lingual transfer by aligning representations across languages.

\subsubsection{Lexical transfer}
For effective cross-lingual knowledge transfer, lexical alignment between languages is crucial.
Techniques include mapping pre-trained monolingual word embeddings to a common vector space using bilingual lexicons \citep{gu_universal_2018}, aligning child source embeddings to parent source embeddings before fine-tuning \citep{kim_effective_2019}, and using shared subword vocabularies through BPE or character-level modeling \citep{kunchukuttan_leveraging_2018}.
These techniques can be represented by a mapping function: $E_c = f(E_p)$ where $E_p$ and $E_c$ are the embedding spaces of the parent and child languages, and $f$ is the alignment function.

The role of lexical overlap in cross-lingual transfer has been extensively studied with mixed findings.
While some studies show a positive correlation between lexical overlap and transfer performance \citep{wu_beto_2019,patil_overlap-based_2022}, others suggest its impact is dependent on other factors \citep{k_cross-lingual_2020,deshpande-etal-2022-bert}.
The relationship appears more nuanced: lexical overlap becomes particularly important when languages have dissimilar word orders \citep{deshpande-etal-2022-bert} or when working with low-resource languages \citep{patil_overlap-based_2022}.
Different metrics for measuring lexical overlap, such as subword overlap, normalized Levenshtein distance, and EzGlot similarity \citep{kovacevic_ezglot_2022}, have been proposed to better quantify this factor.

\subsubsection{Syntactic transfer}
The challenge of syntactic divergence between languages has also been addressed in cross-lingual transfer.
Methods include pre-ordering (reordering parent sentences to match child word order) \citep{murthy_addressing_2019}, noise injection (training parent encoders with artificially noisy input through probabilistic word insertion, deletion, and permutation) \citep{kim_effective_2019}, and mixture of experts (using language-specific expert modules to handle syntactic differences) \citep{gu_universal_2018}.
For languages with different scripts, techniques such as transliteration to a common script \citep{maimaiti_multi-round_2019} help facilitate cross-lingual transfer by leveraging lexical similarities that might otherwise be obscured by different writing systems.

While cross-script transfer is challenging, it remains possible even between languages with completely different writing systems \citep{pires_how_2019,de-vries-etal-2022-make}.
This suggests that multilingual models can develop language-neutral representations that capture deeper linguistic features beyond surface-level lexical similarities.
The development of specific techniques to bridge script differences, such as transliteration and script-agnostic encoders, has further improved cross-script transfer capabilities.

\subsection{Model architecture factors}
Architectural considerations significantly impact a model's cross-lingual transfer abilities.
Studies have shown that network depth can be more important than the number of attention heads for cross-lingual performance \citep{k_cross-lingual_2020}.
The sharing strategy for model parameters also matters—sharing more layers generally improves transfer capabilities \citep{conneau_emerging_2020}.
There appears to be a delicate balance in model capacity: overparameterized models may create language-specific subspaces that hinder cross-lingual transfer, while models with fewer parameters are forced to develop more efficient, language-neutral representations \citep{conneau_emerging_2020,dufter_identifying_2020}.
The embedding layer plays a particularly critical role, with studies showing that cross-lingual alignment of token embeddings significantly impacts transfer performance \citep{wu-etal-2023-oolong,deshpande-etal-2022-bert}.

\subsection{Pre-training considerations}
Pre-training settings and data characteristics substantially influence cross-lingual transfer capabilities.
Larger pre-training corpora generally improve cross-lingual performance, with target language corpus size being particularly important for higher-level tasks \citep{lauscher_zero_2020,srinivasan_predicting_2021,ahuja-etal-2022-multi}.
The domain similarity between pre-training corpora across languages also affects transfer quality, with greater similarity leading to better performance \citep{dufter_identifying_2020,deshpande-etal-2022-bert}.

Regarding pre-training objectives, eliminating the next sentence prediction task has been shown to improve cross-lingual transfer \citep{k_cross-lingual_2020}.
Training on subwords rather than words or characters and using longer sequence lengths during pre-training have also demonstrated positive effects on cross-lingual capabilities \citep{k_cross-lingual_2020,liu_study_2020}.
The quality of tokenizers, measured by metrics such as fertility and proportion of continued words, significantly impacts token-level tasks like Named Entity Recognition and Part-of-Speech tagging \citep{ahuja-etal-2022-multi,rust-etal-2021-good}.

\subsection{Advanced transfer learning approaches}
\subsubsection{Meta-learning for cross-lingual transfer}
The transfer learning paradigm has evolved with the emergence of meta-learning approaches.
\citet{gu_meta-learning_2018} applied Model-Agnostic Meta-Learning (MAML) to learn parameter initializations that facilitate rapid adaptation to new language pairs:
\begin{equation}
    \theta^* = \arg\min_\theta \sum_{(L_s, L_t) \in \mathcal{T}} \mathcal{L}_{(L_s, L_t)}(\theta - \alpha \nabla_\theta \mathcal{L}_{(L_s, L_t)}(\theta)),
\end{equation}
where $\mathcal{T}$ is the set of high-resource language pairs, and $\alpha$ is the adaptation learning rate.
This approach explicitly optimizes the parent model parameters to be amenable to fast adaptation for child tasks, rather than merely optimizing for performance on the parent language pair.

\subsubsection{Target-side transfer challenges}
Transfer learning on the target side has proven more challenging than source-side transfer, likely because target languages require more language-specific representations for generation, similar to the challenges faced in multilingual decoder design.
\citet{dabre_exploiting_2019} addressed this challenge by proposing a multi-stage fine-tuning process for models with multiple target languages, showing the importance of carefully designing the training curriculum when transferring to new target languages.

\subsubsection{Zero-shot approaches}
Zero-shot learning has emerged as a significant area in cross-lingual transfer, allowing models to generalize to unseen language pairs or tasks without explicit training data \citep{artetxe_cross-lingual_2020,wu-etal-2023-oolong}.
This approach is particularly valuable for low-resource languages where parallel data is scarce.
Zero-shot capabilities have been demonstrated across various tasks including text classification \citep{pushp_train_2017}, sentiment analysis \citep{pelicon_zero-shot_2020}, question answering \citep{ma_knowledge-driven_2021}, and named entity recognition \citep{wu-etal-2020-single}.
The effectiveness of zero-shot transfer varies by task type, with syntactic tasks generally transferring more readily than semantic ones \citep{pires_how_2019,lin_choosing_2019}.

\subsubsection{Automated language selection}
Recent research has explored the automation of source language selection for cross-lingual transfer.
Several studies have developed prediction frameworks to determine the most suitable source language for a given target language and task \citep{lin_choosing_2019, lauscher_zero_2020, srinivasan_predicting_2021, dolicki_analysing_2021, eronen_transfer_2022}.
These frameworks typically incorporate multiple factors including linguistic similarity metrics, pre-training corpus sizes, and lexical overlap.
Such approaches provide valuable guidance for efficient deployment of cross-lingual transfer, particularly in scenarios with limited computational resources or when working with very low-resource languages.

\subsubsection{Massively multilingual transfer}
In massively multilingual settings, transfer learning becomes more nuanced.
Benefits of cross-lingual transfer increase with the number of languages incorporated in the model, particularly for low-resource languages \citep{aharoni_massively_2019}.
Performance on high-resource languages may degrade as more languages are added, suggesting a trade-off in representational capacity \citep{arivazhagan_massively_2019}.
Language-family-specific models trained on clusters of related languages often show better performance than models trained on mixed language families \citep{wang_compact_2019}.

This finding has led to the development of language-specific pre-trained models that focus on a single language to achieve better performance than multilingual counterparts for that language.
Examples include CamemBERT for French \citep{martin-etal-2020-camembert}, BERTje \citep{vries_bertje_2019} and RobBERT \citep{delobelle-etal-2020-robbert} for Dutch, AraBERT for Arabic \citep{antoun-etal-2020-arabert}, and PhoBERT for Vietnamese \citep{nguyen-tuan-nguyen-2020-phobert}.
The success of these language-specific models suggests that while multilingual models facilitate cross-lingual transfer, they may not always optimize performance for individual languages \citep{de-vries-etal-2022-make}.

These observations highlight the complex dynamics of cross-lingual transfer in large-scale multilingual settings and point to the need for sophisticated approaches that can balance the benefits of transfer against the constraints of model capacity.

\section{Large language models}\label{ch2:sec:llm}
Large language models (LLMs) represent a notable evolution in natural language processing, evolving from encoder-decoder and encoder-only transformer architectures toward predominantly decoder-only models trained on vast amounts of text data.
These models have demonstrated effectiveness across diverse language tasks, including machine translation, a capability achieved without explicit task-specific supervision but rather through incidental bilingualism in the pre-training data \citep{briakou_searching_2023}.

Unlike traditional NMT systems that are trained specifically on parallel corpora to optimize translation performance, LLMs develop from a more general pre-training approach that develops broad linguistic and world knowledge through self-supervised learning on text data in multiple languages, though the distribution is often imbalanced (as discussed in Section~\ref{ch2:sec:data:monolingual}).
This pre-training followed by adaptation methodology has significantly influenced how machine translation tasks are approached.

Advancing from dedicated multilingual NMT systems to LLM-based translation represents a shift in how cross-lingual capabilities are acquired.
While multilingual NMT models are explicitly designed to translate between specific language pairs through supervised training on parallel corpora, LLMs develop translation abilities more implicitly through exposure to diverse multilingual content during self-supervised pre-training.
After pre-training, LLMs might be further optimized for translation, which will be discussed in Section~\ref{ch2:sec:llm_ft}.
Despite these different training approaches, cross-lingual knowledge transfer in both paradigms relies fundamentally on parameter sharing across languages.
The primary distinction lies in the training objective: multilingual NMT models are optimized specifically for translation tasks from the beginning, whereas LLMs acquire translation as one of many emergent capabilities from their general language modeling objective.

\subsection{Architecture and pre-training}\label{ch2:sec:llm_arch}
While the original transformer decoder was designed to work with an encoder in sequence-to-sequence tasks, LLMs typically employ standalone decoder-only architectures that rely exclusively on self-attention, eliminating the encoder-decoder cross-attention mechanism.

The core of most LLMs is a stack of transformer decoder blocks.
Each block consists of a masked multi-head self-attention layer that prevents positions from attending to future positions, ensuring the autoregressive property required for next-token prediction, followed by a position-wise feed-forward network.
Each component is complemented by residual connections and layer normalization \citep{ba_layer_2016}.

This decoder-only design has become the foundation for widely-used LLMs, including GPT models \citep{radford_language_2019, brown_language_2020, openai_gpt-4_2023}, PaLM \citep{chowdhery_palm_2023,anil_palm_2023}, LLaMA \citep{touvron_llama_2023,touvron_llama2_2023,llama_team_llama_nodate}, and Gemini \citep{gemini_team_gemini_nodate}.

At their core, LLMs aim to model the probability distribution over text by predicting the next token given the preceding context:
\begin{equation}
p(\mathbf{x}) = \prod_{t=1}^n p(x_t|\mathbf{x}_{<t}),
\end{equation}
where $\mathbf{x} = (x_1, x_2, ..., x_n)$ represents a sequence of tokens, and $\mathbf{x}_{<t}$ denotes all tokens before position $t$.
This autoregressive formulation enables LLMs to generate coherent text by repeatedly sampling from the predicted distribution of the next token.

The training objective for LLMs is to maximize the log-likelihood of the training corpus, or equivalently, to minimize the cross-entropy loss:
\begin{equation}
\mathcal{L} = -\sum_{\mathbf{x} \in \mathcal{D}} \sum_{t=1}^{n} \log p(x_t|\mathbf{x}_{<t}),
\end{equation}
where $\mathcal{D}$ represents the pre-training corpus.
In contrast to the NMT objective (Equation~\ref{ch2:eq:nmt_loss}), which models the conditional probability of a target sequence given a source sequence, the LLM objective models the probability of each token given only its preceding tokens in the same sequence.
This objective encourages the model to accurately predict the next token in a sequence based on all preceding tokens.

What distinguishes LLMs from their predecessors is not just their architectural choices but also their scale in three dimensions: 
(1) model size, with parameter counts ranging from billions to trillions; 
(2) training data size, often comprising hundreds of billions to trillions of tokens; and 
(3) computational resources devoted to training, typically measured in thousands to millions of GPU-hours.
This scaling has led to capabilities not observed in smaller models \citep{brown_language_2020,wei2022emergent}, including zero-shot and few-shot learning across tasks and improved reasoning abilities.

\subsection{Large language models for machine translation}
This section discusses the translation capabilities and adaptation methods of large language models, which are built upon the decoder-only transformer architecture described in the previous section.

\subsubsection{Zero-shot translation capabilities}\label{ch2:sec:llm_zero}
Translation abilities in base LLMs (before any fine-tuning) can be traced back to incidental exposure to multilingual content during pre-training \citep{briakou_searching_2023}.
This incidental bilingualism enables translations without explicit parallel data, although performance varies across language pairs.
This multilingual content typically follow specific templates such as:
\begin{align*}
    \texttt{[src\_lang]: [src\_sentence]}\\
    \texttt{[tgt\_lang]: [tgt\_sentence]}
\end{align*}
For example, a template for English to German translation might look like:
\begin{align*}
    \text{English}&: \text{How are you today?}\\
    \text{German}&: \text{Wie geht es dir heute?}
\end{align*}
LLMs learn to associate this template pattern with the translation task through repeated exposure to such bilingual snippets during pre-training, despite not being explicitly trained for translation \citep{briakou_searching_2023,jiao_is_2023}.
While convenient, zero-shot performance typically lags behind dedicated NMT systems, especially for low-resource languages \citep{bang-etal-2023-multitask}.

Several strategies enhance zero-shot translation: pivot prompting (routing translation through English) improves results for distant language pairs \citep{jiao_is_2023}, while including domain-specific keywords or context benefits specialized translations \citep{aycock-bawden-2024-topic}.

\subsubsection{In-context learning for translation}
In-context learning (ICL) significantly improves LLM translation by providing exemplar translations within the prompt, allowing pattern inference without parameter updates \citep{garcia_unreasonable_2023}.
Three key factors affect ICL effectiveness:
\begin{enumerate}
    \item \textbf{Example quantity}: Five examples typically represents the optimal balance between performance and efficiency, with diminishing returns beyond this point \citep{vilar_prompting_2023,agrawal_-context_2023}.
    \item \textbf{Example selection}: Methods range from similarity-based retrieval to sophisticated approaches considering coherence and diversity \citep{m-etal-2023-ctqscorer,sia-duh-2023-context}.
    Submodular functions optimize selection by balancing coverage and minimizing redundancy \citep{ji-etal-2024-submodular}.
    \item \textbf{Example ordering}: Placing the most similar examples closest to the input generally yields better results \citep{fu-etal-2024-gptscore}.
\end{enumerate}
High-quality in-context examples can achieve performance comparable to fine-tuned models \citep{zhang_prompting_2023}, particularly valuable for low-resource languages.

\subsubsection{Adaptation approaches}\label{ch2:sec:llm_ft}
Parameter adaptation further enhances LLMs' translation capabilities through several approaches: 

\paragraph{Supervised Fine-Tuning (SFT)} on parallel corpora significantly improves translation quality across multiple language directions.
Even with limited examples (as few as 32), SFT produces meaningful improvements \citep{zhu_finetuning_2024}.
The effectiveness of SFT scales with data quality and quantity, with best results achieved through curated high-quality parallel datasets \citep{xu_paradigm_2024}.
This approach typically combines translation-specific instructions with source-target pairs, allowing LLMs to adapt their generative capabilities to translation tasks while preserving their broader abilities.
However, SFT requires substantial computational resources and can lead to catastrophic forgetting of beneficial LLM capabilities, such as formality control and contextual disambiguation, which we discuss in Chapter~\ref{ch5}.
Techniques like continued pre-training on a mix of monolingual and parallel data help mitigate these trade-offs while maintaining translation quality improvements.

\paragraph{Parameter-efficient fine-tuning (PEFT)} methods like LoRA \citep{hu2022lora} insert trainable low-rank matrices into frozen LLMs, achieving performance comparable to full fine-tuning while updating only 0.1-2\% of parameters \citep{alves_steering_2023, zhang_machine_2023} .

\paragraph{Advanced adaptation approaches} like X-ALMA \citep{xu_paradigm_2024} and Tower \citep{alves_tower_2024} employ two-stage fine-tuning: first on non-English monolingual data to strengthen multilingual proficiency, then on high-quality parallel data.
This strategy yields substantial improvements across diverse language pairs, establishing new state-of-the-art results in recent machine translation benchmarks \citep{kocmi_findings_2023,kocmi-etal-2024-findings}.

\paragraph{Human preference-based Methods} including Contrastive Preference Optimization (CPO) \citep{xu_contrastive_2025} and Reinforcement Learning with Human Feedback (RLHF) \citep{zhu-etal-2024-preference,xu_advancing_2024}, further refine translations by aligning outputs with human preferences, enabling models to distinguish between adequate and exceptional translations.

\section{Translation evaluation}\label{ch2:sec:evaluation}
Reliable evaluation methods are essential for comparing different machine translation systems and measuring progress in the field; this section outlines the primary evaluation metrics used throughout this thesis to assess and contrast translation quality.

\subsection{BLEU}
The Bilingual Evaluation Understudy (BLEU) score, introduced by \citet{papineni_bleu_2002}, has been the standard metric to evaluate machine translation quality for over two decades.
BLEU measures the overlap between a candidate translation and one or more reference translations using n-gram precision.
The metric calculates how many n-grams in the candidate translation appear in the reference translation, applying a brevity penalty to penalize overly short translations.
Formally, BLEU is defined as
\begin{equation}
\text{BLEU}_N = \text{BP} \cdot \exp\left(\sum_{n=1}^{N} w_n \log p_n\right),
\end{equation}
where $\text{BP}$ is the brevity penalty, $w_n$ are weights for different n-gram precisions (typically uniform weights such that $\sum_{n=1}^{N} w_n = 1$ with $N=4$), and $p_n$ represents the modified n-gram precision.

The modified precision $p_n$ for n-grams of size $n$ is computed as:

\begin{equation}
p_n = \frac{\sum_{c \in \{\text{candidates}\}} \sum_{n\text{-gram} \in c} \text{count}_{\text{clip}}(n\text{-gram})}{\sum_{c' \in \{\text{candidates}\}} \sum_{n\text{-gram}' \in c'} \text{count}(n\text{-gram}')},
\end{equation}
where $\text{count}_{\text{clip}}$ is defined as:

\begin{equation}
\text{count}_{\text{clip}}(x) = \min(\text{count}(x), \text{max\_ref\_count}(x)).
\end{equation}
This clipping mechanism prevents the inflation of BLEU scores by repetitive n-grams in the candidate translation.
Without clipping, a translation system could artificially boost its score by repeating high-confidence phrases multiple times beyond their occurrence in the reference translations.
In these equations, \textit{candidates} refers to the set of translation candidates, and \textit{max\_ref\_count} is the maximum number of times an n-gram appears in any single reference translation.

BLEU's widespread adoption can be attributed to several key advantages: it is language-agnostic, easy to compute, and provides researchers with a consistent benchmark for comparing systems across publications.
Its ubiquity in the field has made it an essential metric to situate new research within the existing literature.
Until recently, virtually all machine translation papers reported BLEU scores as their primary evaluation metric, creating a valuable historical record for tracking progress in the field.

An important contribution to the continued relevance of BLEU was made by \citet{post_call_2018}, who addressed inconsistencies in the implementation of BLEU by introducing a standardized version called SacreBLEU.
This tool ensures reproducibility across research papers by providing a canonical implementation and a signature that specifies tokenization and other pre-processing details, helping to mitigate one of the major criticisms of BLEU: inconsistent reporting practices.
In the research presented in Chapters~\ref{ch3} and \ref{ch4}, unless otherwise mentioned, we use SacreBLEU for our BLEU-based evaluations to ensure reproducibility and comparability with other work in the field.

Although more recent research \citep{mathur_tangled_2020,freitag_bleu_2020,freitag-etal-2022-results} has identified limitations in BLEU's correlation with human judgments, particularly for high-quality translations and certain language pairs, the metric continues to provide useful signals about translation quality. BLEU is particularly effective at identifying poor translations and can reliably detect substantial improvements in translation systems.

The evolution of evaluation practices is evident in newer metrics such as COMET \citep{rei_comet_2020} (introduced in the next section), which uses contextualized embeddings to better align with human judgments. Recent WMT shared tasks \citep{freitag-etal-2022-results} have begun to recommend these newer metrics alongside BLEU.

\subsection{COMET}\label{ch2:sec:comet}
Recent years have seen the emergence of neural metrics that better correlate with human judgments of translation quality.
COMET (Cross-lingual Optimized Metric for Evaluation of Translation), introduced by \citet{rei_comet_2020}, represents a significant advancement in translation evaluation by leveraging contextualized embeddings to predict human assessments.

Unlike traditional metrics that rely on surface-level n-gram matching, COMET operates by encoding the source text, machine translation output, and reference translation into a shared embedding space using a pre-trained multilingual encoder.
These embeddings capture deeper semantic relationships between texts rather than simply measuring lexical overlap.

The core architecture of COMET can be formalized as:

\begin{equation}
\text{COMET}(s, h, r) = f_{\theta}(\text{enc}(s), \text{enc}(h), \text{enc}(r)),
\end{equation}
where $s$ is the source text, $h$ is the machine translation hypothesis, $r$ is the reference translation, $\text{enc}(\cdot)$ is the encoder function that maps text to a contextualized embedding, and $f_{\theta}$ is a feed-forward estimator that predicts a quality score.

COMET is trained on human judgments from past WMT shared tasks, learning to predict Direct Assessment (DA,  \citealt{GRAHAM_BALDWIN_MOFFAT_ZOBEL_2017}) scores or Multidimensional Quality Metrics (MQM, \citealt{burchardt-2013-multidimensional}) annotations.
DA is a human evaluation method where annotators rate translation quality on a continuous scale, while MQM provides a more detailed error-based evaluation framework that categorizes and weights different types of translation errors.
The model is optimized to minimize the mean squared error between predicted scores and human assessments:

\begin{equation}
\mathcal{L}(\theta) = \frac{1}{N}\sum_{i=1}^{N}(f_{\theta}(\text{enc}(s_i), \text{enc}(h_i), \text{enc}(r_i)) - y_i)^2,
\end{equation}
where $y_i$ represents the human judgment score for the $i$-th translation triple.

A key advantage of COMET is its ability to capture more nuanced aspects of translation quality, including semantic equivalence beyond surface form, fluency and grammaticality, preservation of meaning, and adequacy in conveying the source content.

The correlation between COMET scores and human judgments has been consistently higher than traditional metrics like BLEU across multiple language pairs and domains \citep{freitag-etal-2022-results}.

In Chapters~\ref{ch5} and \ref{ch6} of this thesis, we transition from BLEU to COMET as our primary evaluation metric, following the recommendation of \citet{kocmi-etal-2021-ship}.

\chapter{Representational Knowledge Transfer in Multilingual MT}
\label{ch3}

\renewcommand{\thefootnote}{}
\footnotetext{This chapter was published as: 
David Stap, Vlad Niculae, and Christof Monz.
Viewing Knowledge Transfer in Multilingual Machine Translation Through a Representational Lens.
In \textit{Findings of the Association for Computational Linguistics: EMNLP 2023}, pages 14973–14987, Singapore. Association for Computational Linguistics, 2023.}
\renewcommand{\thefootnote}{\arabic{footnote}}

\section{Introduction and research questions}
As discussed in Section \ref{ch2:sec:mnmt}, multilingual neural machine translation (mNMT) \citep{ha_toward_2016,johnson_googles_2017} can support multiple translation directions in a single model, with low-resource languages benefiting most and high-resource languages degrading in quality \citep{arivazhagan_massively_2019}. 
However, there is a large discrepancy in quality among low-resource languages, with some languages benefiting a lot, while others see relatively little improvement. 
Conflicting findings have emerged in cross-lingual knowledge transfer research, leaving the underlying causes for this discrepancy unclear. 
For example, some studies have found that token overlap can be leveraged to increase translation performance \citep{patil_overlap-based_2022,wu_beyond_2023}, while others have found that token overlap is unimportant for cross-lingual transfer \citep{k_cross-lingual_2020,conneau_emerging_2020}.

In the context of transferring knowledge from a parent translation model to a child model, prior research has shown contradictory results, as discussed in Section~\ref{ch2:sec:basic_transfer}: some studies demonstrated that quality improvements are larger when using a closely related parent \citep{zoph_transfer_2016}, while others found that unrelated language pairs can work even better \citep{kocmi_trivial_2018}.
Another finding is that an English-centric model benefits most from positive transfer for directions \textit{into} English, while improvement in the other directions is modest \citep{arivazhagan_massively_2019}. 

One of the most striking observations in the literature is that the improvements of many-to-one mNMT can be explained to a large extent by the increased amount of target data \citep{fan_beyond_2021}, rather than by cross-lingual knowledge transfer.  

Understanding cross-lingual knowledge transfer in the context of mNMT is an under-explored research direction \citep{hupkes_taxonomy_2023}.
Despite some existing studies that have examined mNMT representations, none have yet connected these representations to knowledge transfer.
For instance, when translating ``voiture'' in French and ``Auto'' in German to ``car'' in English, one would expect that the cross-attention context vectors for French-English and German-English would be similar.
However, \citet{johnson_googles_2017} show that clustering occurs on the \textit{sentence level} rather than the \textit{word level}.
Even identical sentences in various languages do not occupy the same position in the representation space \citep{escolano_multilingual_2022}, and encoder representations are dependent on the target language \citep{kudugunta_investigating_2019} instead of source meaning.

In this chapter, we investigate the relationship between cross-lingual transfer and representational similarities between languages.
To this end, we ask the following question:
\RQ{1}
\begin{myquote}
We formalize the relationship between cross-lingual transfer and cross-attention similarities between languages, by introducing a new metric: Representational Transfer Potential (RTP). 
This allows us to reason about knowledge transfer in a way translation quality (BLEU) is unable to capture.
We investigate cross-attention because it acts as \textit{bottleneck} between the encoder (mostly responsible for representing the source sentence) and the decoder.\\\\
This overarching question is addressed through three interconnected steps. First, we develop and validate our representational metric to quantify knowledge transfer (RQ1.1). Then, we investigate which factors predict successful transfer between languages (RQ1.2). Finally, we leverage these insights to develop a novel training approach that enhances cross-lingual knowledge transfer (RQ1.3).
\end{myquote}

\RQsub{1}{1}
\begin{myquote}
Through extensive experiments we investigate whether RTP can be used to quantify positive and negative transfer (also known as interference).
Furthermore, we investigate to what extent these representational similarities correlate with improvements in translation quality, with the goal of finding out whether knowledge transfer occurs or the improved translation quality is only due to the increased data on the target side \citep{fan_beyond_2021}.\\\\
Having established RTP as a metric for quantifying cross-lingual transfer, we next explore:
\end{myquote}

\RQsub{1}{2}
\begin{myquote}
We investigate how to predict data and language characteristics that are relevant for transfer.
Our objective is to use dataset and linguistic features to predict the representational similarities, which we treat as a regression problem.
\end{myquote}

\RQsub{1}{3}
\begin{myquote}
Based on our answers to RQ1.1 and RQ1.2, we explore optimization methods to improve transfer. 
We propose a method for training a multilingual translation model using an auxiliary similarity loss that exploits multi-parallel data, thereby increasing the degree of language invariance across source representations.
Contrary to common perception, a significant amount of multi-parallel data exists within parallel datasets such as WMT, making it more abundant than commonly assumed \citep{freitag_bleu_2020}.
Our method works by alternately feeding parallel and multi-parallel batches to a model.
For multi-parallel batches, we minimize an auxiliary similarity loss that encourages context vectors, resulting from cross-attention, to be similar.
\end{myquote}

\paragraph{Organization.} This chapter is organized as follows: After reviewing the previous work in Section \ref{ch3:sec:related_work}, we present our methodology for analyzing transfer in multilingual MT models in Section \ref{ch3:sec:transfer_m2m_models}. In Section \ref{ch3:sec:predicting_transfer}, we investigate the factors that predict cross-lingual transfer, followed by Section \ref{ch3:sec:multiparallel_training} where we propose a novel training approach to optimize for representational invariance. Finally, we conclude and summarize our findings in Section \ref{ch3:sec:conclusion}.

\section{Related work}
\label{ch3:sec:related_work}
In this section, we discuss previous work that investigates representations in multilingual NMT.

\subsection{Analyzing multilingual neural machine translation}
Using Singular Value Canonical Correlation Analysis \citep[SVCCA,][]{raghu_svcca_2017}, \citet{kudugunta_investigating_2019} demonstrated that encoder representations cluster based on linguistic similarity and are dependent on the target language.
Additionally, the set of most important attention heads are similar across language pairs, which enables language clustering \citep{kim_multilingual_2021}. 
Furthermore, representations of different languages cluster together when they are semantically related \citep{johnson_googles_2017,escolano_bilingual_2019}. 
In particular, visualising cross-attention per decoding time-step shows that meaning equivalent sentences generally cluster together \citep{johnson_googles_2017}.

However, the extent of these phenomena has not been quantified per language.
Moreover, these studies have primarily focused on representations in isolation, or its relation with linguistic similarity, with less focus on the role of representations in knowledge transfer. 
In contrast, we explicitly connect representations to transfer, which allows for a deeper understanding of the impact of transfer on translation quality.

\section{Analyzing transfer in many-to-many models}\label{ch3:sec:transfer_m2m_models}
In this section, we aim to delve deeper into the understanding of knowledge transfer across languages in mNMT models, moving beyond the commonly used metric of translation quality as a proxy for transfer.
By exploring the relationship between transfer and hidden representations in a multilingual model, we aim to gain insight into why certain languages benefit more from multilingual training (as discussed in Section~\ref{ch3:sec:predicting_transfer}).
Furthermore, we aim to develop training strategies that can increase representational similarity and thus enhance knowledge transfer (as outlined in Section~\ref{ch3:sec:multiparallel_training}).

\subsection{Experimental setup} \label{ch3:sec:baseline_setup}
\subsubsection*{Data} To investigate the relationship between transfer and representation in multilingual machine translation, we conduct our experiments on the TED Talks corpus \citep{qi_when_2018}.
The corpus comprises parallel data from 59 languages and is chosen over other large parallel corpora such as OPUS-100 \citep{zhang_improving_2020} due to its high translation quality and inclusion of relatively large portions of \textit{explicit} multi-parallel data, which is an important characteristic for our analysis.
We train a many-to-many model on all language pairs that contain English in the source or target, resulting in 116 translation directions. 
To ensure comparable results, we apply joint subword segmentation \citep{sennrich_neural_2016} and use a vocabulary size of 32K.
We also train and evaluate bilingual baselines using the same setup.

Additionally we evaluate on the out-of-domain FLORES-101 evaluation benchmark \citep{goyal_flores-101_2022}. Out-of-domain data helps to assess robustness and generalization capabilities, and provides a more realistic measure of how well the system can handle diverse and unexpected inputs.
This dataset is completely multi-parallel, which is a necessary property for our analysis. 
It consists of a dev (997 sentences) and devtest (1012 sentences) split, both of which we combine to enhance the robustness of our findings.
Sentences are extracted from English Wikipedia, and translated to 101 languages by professional translators.
\subsubsection*{Evaluation} We calculate BLEU scores \citep{papineni_bleu_2002} using SacreBLEU \citep{post_call_2018}.
\footnote{\scriptsize{nrefs:1$|$case:mixed$|$eff:no$|$tok:13a$|$smooth:exp$|$version:2.3.1}}

\subsubsection*{Experimental setup}
\begin{table}[htbp]
    \begin{center}
    \small
        \begin{tabular}{l|r@{\hspace{0.5em}}|r@{\hspace{0.5em}}r@{\hspace{0.5em}}|r@{\hspace{0.5em}}r}
        \toprule
         & & Neubig$^{\text{a}}$ & Aharoni$^{\text{b}}$ & Ours &  \\
        Language pair & Data size & M2O & M2M & M2O & M2M \\
        \midrule
        Belarusian-English (be-en) & $4.5$K & $18.3$ & $21.7$ & $23.8$ & $\mathbf{24.9}$ \\
        Azerbaijani-English (az-en) & $5.9$K & $11.7$ & $12.8$ & $14.3$ & $\mathbf{15.2}$ \\
        Galician-English (gl-en) & $10$K & $29.1$ & $30.7$ & $34.9$ & $\mathbf{36.0}$ \\
        Slovak-English (sk-en) & $61$K & $28.3$ & $29.5$ & $33.4$ & $\mathbf{34.2}$ \\
        German-English (de-en) & $167$K & -- & $33.0$ & $36.3$ & $\mathbf{37.5}$ \\
        Italian-English (it-en) & $203$K & -- & $35.1$ & $38.5$ & $\mathbf{39.8}$ \\
        Hebrew-English (he-en) & $211$K & -- & $33.2$ & $36.5$ & $\mathbf{37.3}$ \\
        Arabic-English (ar-en) & $213$K & -- & $28.3$ & $31.3$ & $\mathbf{32.6}$ \\
        \midrule
        Average & $109$K & $21.6$ & $28.04$ & $31.1$ & $\mathbf{32.2}$ \\
        \bottomrule
        \end{tabular}
    \end{center}
    \caption{X$\rightarrow$English test BLEU scores on TED Talks corpus. Results shown for both many-to-many (M2M) and many-to-one (M2O) systems. $^{\text{a}}$\citet{neubig_rapid_2018}; $^{\text{b}}$\citet{aharoni_massively_2019}.}
\label{ch3:tab:ted_bleu}
\end{table}

\subsubsection*{Models and optimization}
We train many-to-one and many-to-many transformer base models \citep{vaswani_attention_2017} with 6 layers, model dimension 512, hidden dimension 2048, and 8 attention heads.
All parameters are shared between language pairs \citep{ha_toward_2016,johnson_googles_2017}.
We use Adam \citep{kingma_adam_2015} ($\beta_1 = 0.9$, $\beta_2 = 0.98$ and $\epsilon = 10^{-9}$) to optimize a label smoothed \citep{szegedy_rethinking_2016} (smoothing=0.1) cross entropy loss function.
To be able to make use of multilingual data within a single system we use a target-language prefix tag to each source sentence \citep{johnson_googles_2017,ha_toward_2016}. 
We tie the weights of the decoder input embeddings and the decoder softmax layer \citep{press_using_2017} and apply a 0.2 dropout rate \citep{srivastava_dropout_2014} on the sum of the input embeddings and positional embeddings, on the output of each sublayer, on the output after the ReLU activation in each feedforward sublayer, and to the attention weights. 
The resulting model has ~93M trainable parameters. 
We use a batch size of 25k tokens.
Following \citet{neubig_rapid_2018} and \citet{aharoni_massively_2019}, we do not use temperature sampling.
Models are implemented in our open-source translation system. 
All models we train converge in approximately 2 days of training, using 4x NVIDIA TITAN V (12GB) GPUs.

\subsubsection*{Results} For evaluation on TED, we used tokenized BLEU to be comparable with \citet{neubig_rapid_2018} and \citet{aharoni_massively_2019}.
Table~\ref{ch3:tab:ted_bleu} shows that our many-to-one and many-to-many models obtains comparable or better BLEU scores for X$\rightarrow$English directions.

\subsection{(Dis-)advantages of multilingual training}
\begin{table}[!htb]
    \begin{center}
    \small
        \begin{tabular}{llll}
        \toprule
        & low ($<10$K) & mid ($10$K-$150$K) & high ($>150$K) \\
        & 12 languages & 23 languages & 17 languages \\
        \midrule
        bilingual       & $1.2$          & $12.3$           & $\mathbf{18.6}$ \\
        many-to-one     & $8.8^{*(12/0)}$ & $14.6^{*(20/3)}$ & $15.0^{*(0/17)}$ \\
        many-to-many    & $\mathbf{9.7}^{*(12/0)}$ & $\mathbf{16.8}^{*(21/0)}$ & $17.9^{*(0/15)}$ \\
        \bottomrule
        \end{tabular}
    \end{center}
    \caption{\small{X$\rightarrow$English} BLEU on FLORES-101 for bilingual, many-to-one and many-to-many models. Results are bucketed by number of training examples in TED. $*(n/m)$ denote the fraction of scores in a bucket the are significantly better ($n$) or worse ($m$) to the bilingual baseline, according to bootstrap resampling.}
    \label{ch3:tab:flores_bleu}
\end{table}

Having validated that our model meets strong baselines, we will use the FLORES-101 \citep{goyal_flores-101_2022} evaluation dataset for our subsequent analyses.
X$\rightarrow$English results are summarized in Table~\ref{ch3:tab:flores_bleu}.

In general, low-resource and mid-resource languages benefit ($+8.5$ and $+4.5$ BLEU), and high-resource language scores are weakened ($-0.7$ BLEU) compared to bilingual baselines.
Similar to previous findings \citep{johnson_googles_2017} we find that a many-to-many setup outperforms a many-to-one setup.
Low-resource BLEU scores have a large standard deviation ($\pm 6.9$), indicating that some languages benefit much more than others.

\subsection{Representational view on transfer}\label{ch3:subsec::representational_view}
To further investigate the differences between multilingual and bilingual models, we will now focus on understanding the underlying mechanics of knowledge transfer.
Using translation quality alone as a measure of knowledge transfer is inadequate, as differences in translation quality can have various causes, such as target data distribution \citep{fan_beyond_2021}. 
Therefore, in the following experiments, we aim to gain deeper insight into the mechanisms behind knowledge transfer in multilingual models, focusing on the many-to-many model, which produced the highest translation scores.

When translating two semantically equivalent sentences from different source languages to the same target language, if the context vectors produced by the cross-attention mechanism are (almost) identical for every decoding timestep, the resulting translations will be the same.
However, the reverse may not hold true; it is possible for distinct context vectors to produce the same output, and these variations may correspond to specific aspects of the target language.
The question of whether source language invariance is a desirable or even necessary trait for an mNMT model remains unresolved.

\subsubsection*{Quantifying cross-attention language invariance}
Our goal is to determine the degree of language invariance in the encoder representations of our multilingual model, and how this affects translation quality and transfer.
Unlike previous studies that have focused on the investigation of hidden encoder and decoder representations \citep{kudugunta_investigating_2019}, we concentrate on cross-attention, which connects the encoder and decoder.
To investigate the degree of language invariance, we sample semantically equivalent sentence triples $\{\mathbf{x}^{\ell}, \mathbf{x}^{\ell'}, \mathbf{y}^\tau\}$ from dataset $\mathcal{D}$. 
Here, $\mathbf{x}^{\ell}$ and $\mathbf{x}^{\ell'}$ are sentences that originate from two different non-English source languages $\ell$ and $\ell'$, while $\mathbf{y}^\tau$ is the target sentence in language $\ell^\tau$, which is always English.
We then measure the average cosine similarity of the cross-attention vectors of all sentences in $\ell$ and $\ell'$ at different decoding time steps $t$ when translating into $\ell^\tau$:
\begin{equation}\label{ch3:eq:xsim}
\textrm{xsim}_{(\ell,\ell',\ell^\tau)} = 
    \sum_{\{\mathbf{x}^{\ell},\mathbf{x}^{\ell'},\mathbf{y}^\tau\}\in \mathcal{D}}
    \sum_{t}
    \frac{1}{t}
    \cos(\times_t(\mathbf{x}^{\ell}, \mathbf{y}^\tau), \times_t(\mathbf{x}^{\ell'}, \mathbf{y}^\tau)),
\end{equation}
where $\cos$ is the cosine similarity, and $\times_t(\cdot, \cdot)$ is the context vector, i.e., the result of encoder-decoder cross-attention at decoding time step $t$.
We use FLORES-101, consisting of 2,009 multi-parallel sentences. 
As we need multiple source sentences and a single target sentence, our analysis focuses on many-to-one directions.
We only consider cross-attention within the final decoder layer in this analysis, and leave extensions to non-English target languages to future work.

\begin{figure}[t!]
\centering
\includegraphics[width=0.7\linewidth]{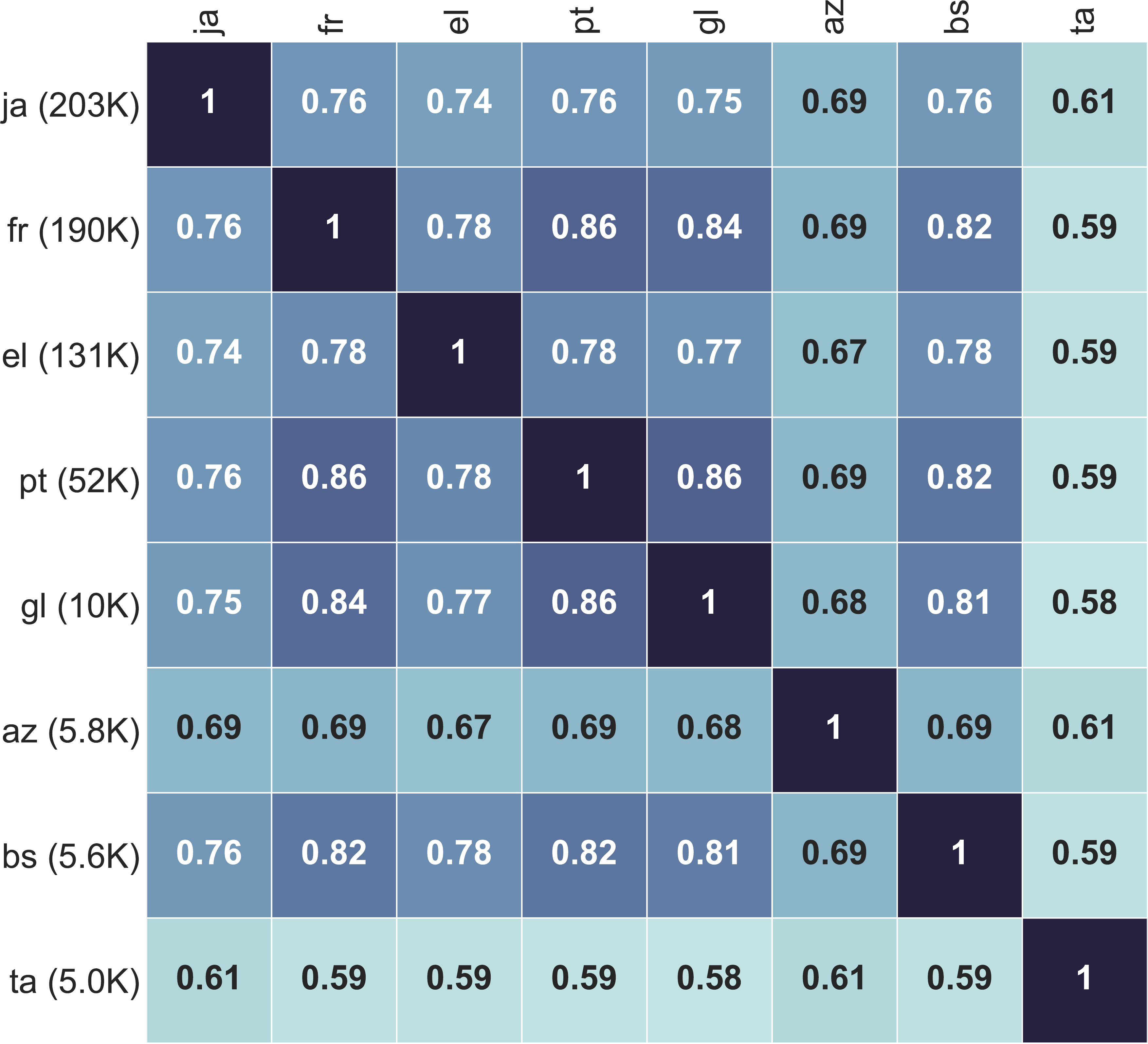}
\caption{Average cosine similarities between context vectors (see Equation~\ref{ch3:eq:xsim}) for different source language combinations into English.
Train data size is shown between brackets.
The higher the similarity, the higher the degree of language invariance.
}
\label{ch3:fig:partial_sim_matrix}
\end{figure}

\begin{figure*}[!htb]
    \centering
    \includegraphics[width=\linewidth]{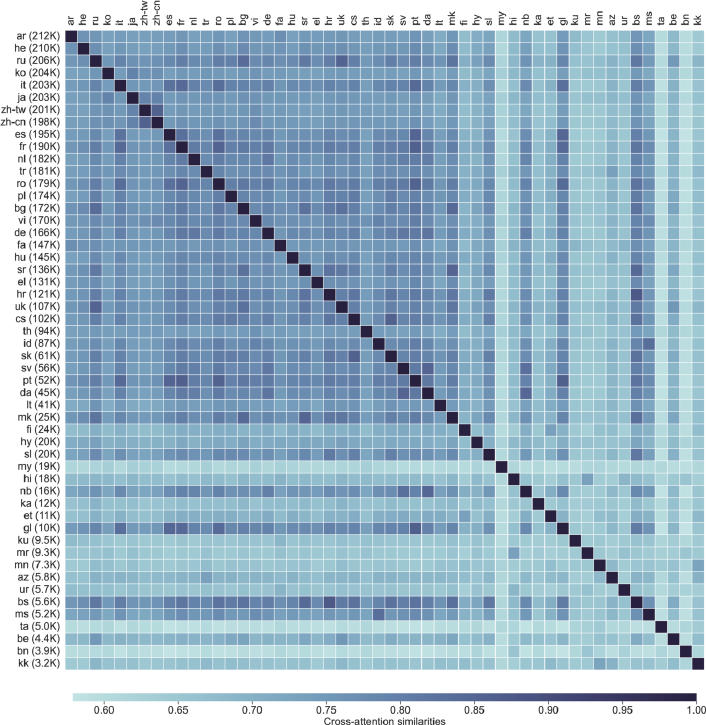}
    \caption{Cross-attention similarities for all language combinations. Training data size into English depicted between brackets.}
    \label{ch3:fig:full_sim_matrix}
\end{figure*}

\begin{figure}[!htb]
\centering
\includegraphics[width=0.7\linewidth]{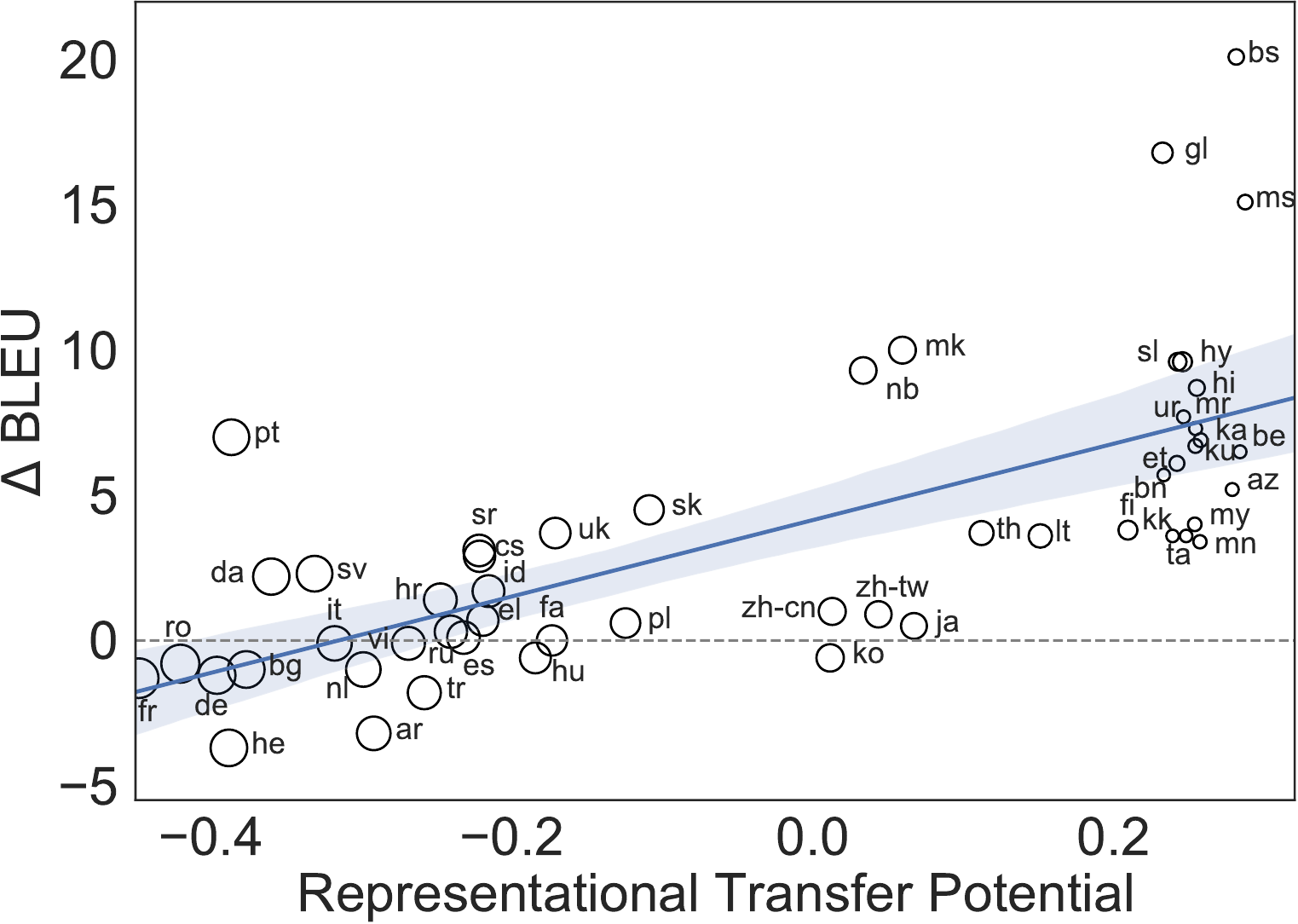}
\caption{The x-axis represents Representational Transfer Potentials (RTP), which measure the total transfer potential for a language (as detailed in Equation~\ref{ch3:eq:rtp}), on FLORES-101.
The y-axis illustrates the difference in BLEU scores (multilingual BLEU - bilingual BLEU) on FLORES-101.
The size of the dots indicates the bilingual BLEU score.
The correlation coefficient (Spearman's $\rho$) is $.77$ and it is statistically significant ($p<0.001$).
The trend illustrates that a higher RTP value is positively associated with changes in translation performance in a multilingual setting.}
\label{ch3:fig:delta_bleu_vs_rtp}
\end{figure}

The resulting similarity matrix is displayed in Figure~\ref{ch3:fig:partial_sim_matrix} for eight languages. 
An $\textrm{xsim}$ similarity value of 1 indicates that the encoder representations are identical for all decoding time steps, i.e., the representations are language invariant.
Conversely, a low similarity suggests that the representations are dissimilar on average, indicating that they are far from being language invariant.
From the matrix, we can observe several patterns:

\begin{itemize}
    \item High-resource languages tend to have relatively high similarity with other high-resource languages.
    For instance, the similarity between French (fr) and Portuguese (pt), $\textrm{xsim}_{(\textrm{fr, pt, en})}$, is $0.86$, and between Greek (el) and French (fr), $\textrm{xsim}_{(\textrm{el, fr, en})}$, is $0.78$.
    \item Furthermore, we find that some low-resource languages, such as Galician (gl) and Bosnian (bs), have high similarities with high-resource languages.
    These languages benefit greatly from multilingual modeling, as evidenced by an increase of $16.8$ and $20.1$ BLEU points, respectively, compared to their bilingual scores.
    \item Other low-resource languages, such as Tamil (ta), do not have high similarities with high-resource languages.
    These languages do not benefit as much from transfer, as demonstrated by a small increase of only $3.6$ BLEU points in the case of Tamil. 
\end{itemize}
For completeness, Figure~\ref{ch3:fig:full_sim_matrix} shows cross-attention similarities between \textit{all} language combinations in our dataset.
This comprehensive visualization reveals additional patterns beyond those observed in the partial matrix.
We observe distinct clustering along linguistic family lines, with Romance and Slavic languages forming visible blocks of higher similarity.
The matrix also illustrates a gradient effect where similarity generally decreases as we move toward lower-resource languages (bottom right), though with notable exceptions where linguistic relatedness appears to override resource limitations.

\subsubsection*{Connecting representations to translation quality}
We quantify the potential for knowledge transfer into language $\ell \in L$ from other languages $\ell' \in L \setminus \{\ell\}$, by connecting context vector similarity and translation quality.
To the best of our knowledge, this is the first approach that quantifies transfer at the representational level.
We define the \textit{Representational Transfer Potential} (RTP) as follows:
\begin{equation}\label{ch3:eq:rtp}
\mathrm{RTP}_{(\ell)} = \sum_{\ell' \in L \setminus \{\ell, \textrm{en}\}} \frac{\Delta\mathrm{BLEU}(\ell, \ell')}{\max |\Delta\mathrm{BLEU}(\ell, \ell')|} \mathrm{xsim}_{(\ell, \ell', \textrm{en})},
\end{equation}
where $\Delta\mathrm{BLEU}(\ell,\ell')$ is the difference in bilingual BLEU scores between the languages when translating into English, which can be thought of as an upper bound for the potential transfer between $\ell$ and $\ell'$. 
$\Delta\mathrm{BLEU}(\ell,\ell')$ is then weighted by the average representational similarity between $\ell$ and $\ell'$ when translating into English, $\textrm{xsim}_{(\ell,\ell', \textrm{en})}$ (see Equation~\ref{ch3:eq:xsim}). 
RTP thus shows to what extent languages act as donor, i.e., benefiting other languages, or recipient, i.e., benefiting from other languages. 
Positive transfer can occur when a language $\ell'$ has better translation performance than $\ell$, which increases the weighted $\textrm{RTP}_{(\ell)}$ score.
Negative transfer can occur when language $\ell'$ has worse translation performance than $\ell$, which decreases the score.
It is important to note that RTP is \textit{not} a score of a language in isolation, but rather a score of a language dataset in the context of other language datasets.
Thus, RTP depends on the languages involved and the available resources in a dataset.

\subsubsection*{Results}
In Figure~\ref{ch3:fig:delta_bleu_vs_rtp}, we plot the resulting RTP scores on the x-axis, and the changes in BLEU scores in a multilingual model versus a bilingual model on the y-axis.
We observe a strongly positive and significant correlation ($\rho=.77, p<0.001$), where a higher RTP score implies increased translation performance, and a lower RTP score implies lower translation performance.
Consider Hebrew (he), which has high similarities with lower performing languages, and smaller similarities with better performing languages. 
Therefore, RTP can correctly predict that Hebrew \textit{does not} benefit from the multilingual setup, which is evidenced by its negative RTP score ($-.39$) and decreased BLEU score ($-3.7$).
On the other hand, Bosnian (bs) has a relatively high RTP score of $.28$, meaning it is similar to languages with stronger translation quality.
Bosnian is the language that benefits most from the multilingual setup ($+20.1$ BLEU).
This means that the resulting differences in translation quality are due to knowledge transfer as captured by the RTP score, and not as a side effect of increased target data size.
However, it is worth mentioning that this trend is not perfect and can only explain part of the transfer.
For instance, Galician (gl) and Finnish (fi) have similar RTP scores ($.23$ and $.21$) but the increase in translation quality for Galician is far greater: $16.8$ vs $3.8$ for Finnish.
The discrepancies in RTP scores warrant further investigation (see Section~\ref{ch3:sec:predicting_transfer}).

\begin{figure}[t]
\centering
\includegraphics[width=0.7\linewidth]{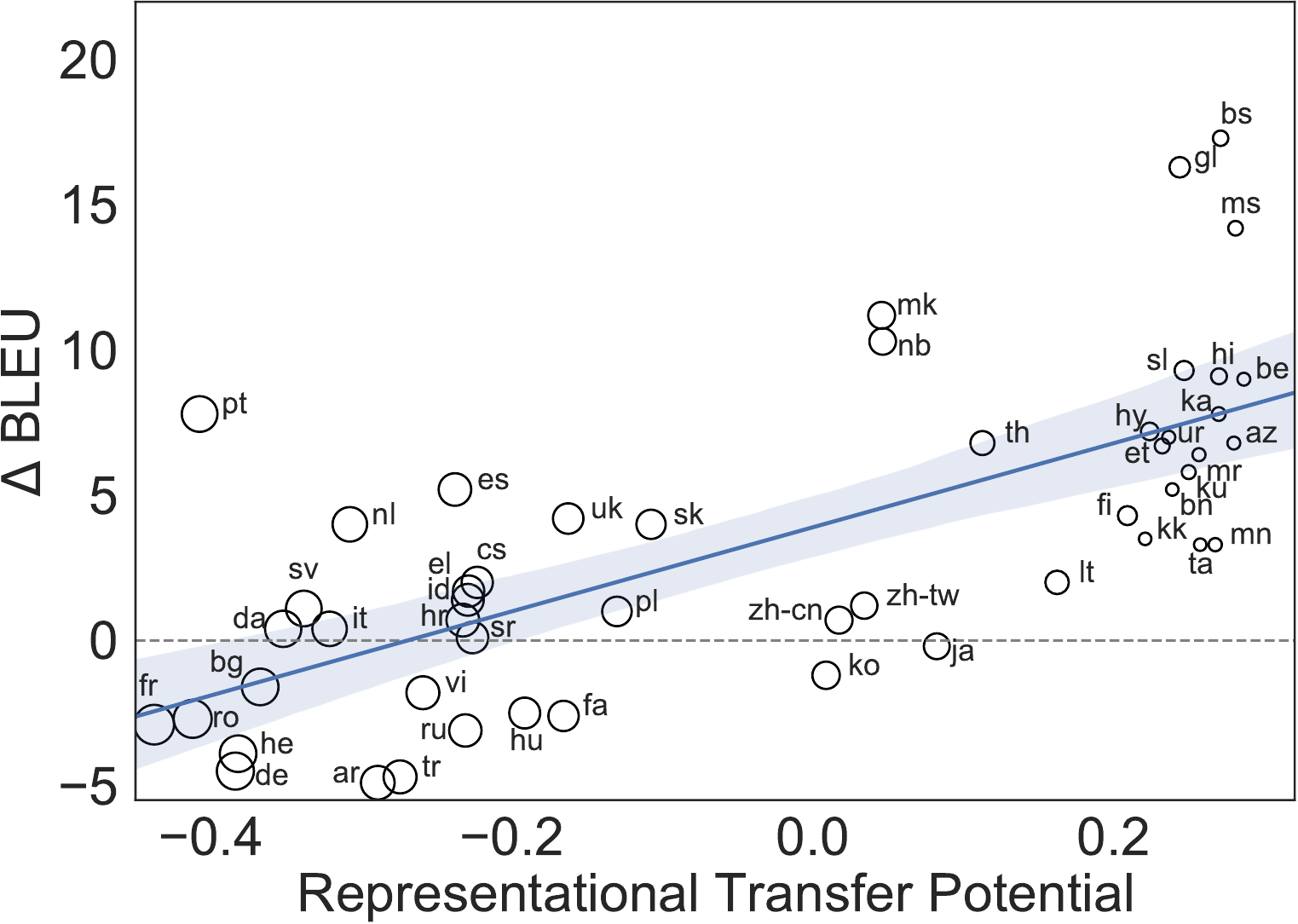}
\caption{The x-axis represents Representational Transfer Potentials (RTP), which measure the total transfer potential for a language (as detailed in Equation~\ref{ch3:eq:rtp}), on NTREX-128.
The y-axis illustrates the difference in BLEU scores (multilingual BLEU - bilingual BLEU) on NTREX-128.
The size of the dots indicates the bilingual BLEU score.}
\label{ch3:fig:delta_bleu_vs_rtp_ntrex}
\end{figure}

To ensure the validity and generalizability of our RTP analysis findings beyond a single test dataset (FLORES-101), we incorporate an additional test dataset, NTREX-128 \citep{federmann_ntrex-128_2022}.
It consists of 1,997 multi-parallel sentences in 128 languages. 
See Figure~\ref{ch3:fig:delta_bleu_vs_rtp_ntrex} for the corresponding plot.
For NTREX-128, we again observe a strongly positive correlation ($\rho=.73, p<0.001$) between RTP and translation quality, further establishing their relationship.
Figures \ref{ch3:fig:delta_bleu_vs_rtp_ntrex} and \ref{ch3:fig:delta_bleu_vs_rtp} (RTP vs delta BLEU on FLORES-101) are highly similar, indicating that RTP generalizes to different test sets.

\subsubsection*{Ablations}
Table~\ref{ch3:tab:rtp_ablation} shows ablations on RTP, and several linguistic distance baselines as described in \citet{lin_choosing_2019} (which we will introduce in Section \ref{ch3:subsec:linguistic_features}).
We calculate correlation coefficients (Spearman's $\rho$) on the metrics and the difference in BLEU scores (multilingual BLEU - bilingual BLEU) on FLORES-101.
Removing the xsim term in RTP gives $\rho=.56$, and removing $\Delta$~BLEU results in $\rho=.28$.
The linguistic distances (Section \ref{ch3:subsec:linguistic_features}) range from $\rho=-.14$ (inventory distance) to $\rho=.14$ (syntactic distance) and thus are not correlated with BLEU difference.
We conclude that RTP has far better correlation with translation quality than linguistic distances and ablations.

Additionally, the mean absolute RTP deviation per language on FLORES-101 and NTREX-128 is $0.008$, and the correlation is extremely robust ($\rho=.99, p<0.001$).
These results provide further evidence that RTP scores are consistent across different test sets, rather than being an artifact of a specific dataset.

\begin{table}[t]
    \begin{center}
        \small
        \begin{tabular}{lrr}
        metric                  & $\rho$    & $p$\\\toprule
        RTP                     & $.77$     & $<.001$ \\
        only $\Delta$BLEU       & $.56$     & $<.001$ \\
        only xsim               & $.28$    & $<.05$\\\midrule
        genetic distance        & $-.11$    & $>.30$ \\
        inventory distance      & $-.14$    & $>.30$ \\
        syntactic distance      & $.14$     & $>.30$ \\
        phonological distance   & $-.13$    & $>.30$ \\
        combined distances      & $-.01$    & $>.30$ \\
        \bottomrule
        \end{tabular}
    \end{center}
    \caption{RTP ablations and linguistic distance baselines, calculated on FLORES-101.}
    \label{ch3:tab:rtp_ablation}
\end{table}

\subsubsection{Picking suitable transfer languages}
Finally, we show that RTP can be used to pick suitable auxiliary transfer languages.
We find that training a language with its top 5 RTP contributors leads to substantially better results of up to $6.8+$ BLEU, compared to training with its bottom 5 contributors.

For low-resource languages Belarusian (be) and Bengali (bn), we identify the top 5 and bottom 5 contributors to their RTP scores, as shown in Table \ref{ch3:tab:RTP_contributors}.
We selected these two languages because they represent similar resource tiers (Belarusian with 4.5K training examples and Bengali with 3.9K) but come from distinct language families (Slavic and Indo-Aryan, respectively).
allowing us to evaluate the effectiveness of our RTP metric for picking suitable transfer languages.
This controlled comparison allows us to isolate the impact of linguistic factors on transfer effectiveness beyond mere data size considerations.

To ensure fair comparison between the top and bottom contributor groups, we create balanced training sets by controlling for data size. 
For each language pair (Belarusian or Bengali), we determine which dataset size to use for training by finding the minimum size within each group (top 5 or bottom 5).
This minimum size, indicated by underlined values in Table \ref{ch3:tab:RTP_contributors}, is then used to subsample the larger datasets when training the models.

\begin{table*}[th!]
    \centering
    \small
    \begin{tabular}{l|cr|cr}
    \toprule
    & \multicolumn{2}{c|}{$\mathrm{RTP}_{\textrm{top}}$ contributors} & \multicolumn{2}{c}{$\mathrm{RTP}_{\textrm{min}}$ contributors} \\
    \midrule
    \multirow{5}{*}{be} 
    & uk (0.75) & 107K & bn (0.59) & \underline{3.9K}* \\
    & ru (0.75) & 206K & ta (0.6)  & 5.1K \\
    & bg (0.73) & 172K & my (0.6)  & 19K  \\
    & mk (0.72) & 249K & mr (0.64) & 9.3K \\
    & sr (0.72) & 136K & ur (0.64) & 5.7K \\
    \midrule
    \multirow{5}{*}{bn} 
    & hi (0.65) & 178K & es (0.58) & 195K \\
    & mr (0.65) & 9.3K & gl (0.58) & 9.9K \\
    & ur (0.62) & \underline{5.7K}* & pt (0.58) & 52K  \\
    & mn (0.62) & 7.4K & it (0.59) & 203K \\
    & hy (0.62) & 203K & nb (0.59) & 16K \\
    \bottomrule
    \end{tabular}
    \caption{Top and bottom RTP contributing languages for Belarusian (be) and Bengali (bn). The values in parentheses show similarity scores. Dataset sizes (in sentences) are shown for each language. Underlined values with asterisks (*) indicate the reference sizes used for subsampling: 3.9K for Belarusian and 5.7K for Bengali experiments.}
    \label{ch3:tab:RTP_contributors}
\end{table*}

We then train a many-to-many system on the resulting datasets, after including Belarusian or Bengali and English.
Results can be found in Table \ref{ch3:tab:rtp_topn_bleu}.
We observe large discrepancies in scores for the top 5 and bottom 5 datasets, even though the dataset sizes are identical, for both in-domain (TED) and out-of-domain (FLORES-101) settings.
In all cases, the model trained on the top 5 RTP contributors outperforms the one trained on the bottom 5 contributors.
The difference is substantial: Belarusian-English on TED with the top RTP contributors scores 16.2 BLEU, whereas the system trained on the bottom contributors results in 9.4 BLEU.
These findings show that RTP can be used to identify suitable auxiliary transfer languages.

\begin{table*}[th!]
    \centering
    \small
    \begin{tabular}{lrrrrrr}
        \toprule
        & \multicolumn{2}{c}{FLORES-101}
        & \multicolumn{2}{c}{TED} \\ 
        \cmidrule{2-3} 
        \cmidrule{4-5}

        & \multicolumn{1}{l}{be-en}
        & \multicolumn{1}{l}{bn-en}
        & \multicolumn{1}{l}{be-en}
        & \multicolumn{1}{l}{bn-en} \\
        
        \midrule
        bilingual         
            & $0.5$
            & $0.4$
            & $6.1$
            & $6.8$ \\

        many-to-many (all)      
            & $\mathbf{7.0}$
            & $\mathbf{6.1}$
            & $\mathbf{24.9}$
            & $\mathbf{19.3}$ \\\midrule

        many-to-many ($\textrm{RTP}_{\textrm{top}}$)
            & $\mathbf{5.4}$
            & $\mathbf{4.1}$
            & $\mathbf{16.2}$
            & $\mathbf{12.9}$ \\

        many-to-many ($\textrm{RTP}_{\textrm{min}}$)
            & $2.3$
            & $1.6$
            & $9.4$
            & $6.2$ \\

        $\Delta$ BLEU
            & $+3.1$
            & $+2.5$
            & $+6.8$
            & $+6.7$ \\
        
        \bottomrule
        \end{tabular}
        \caption{Belarusian-English (be-en) and Bengali-English (bn-en) BLEU scores on FLORES-101 and TED. We compare systems trained on bilingual data with many-to-many systems trained on all data (all), the top 5 contributors to RTP ($\textrm{RTP}_{\textrm{top}}$), and the bottom 5 contributors to RTP ($\textrm{RTP}_{\textrm{min}}$).}
        \label{ch3:tab:rtp_topn_bleu}
\end{table*}

\section{Analyzing causes for transfer} \label{ch3:sec:predicting_transfer}
Next, we investigate characteristics that are relevant for transfer.
Our objective is to use dataset and linguistic features to predict the representational similarities $\textrm{xsim}_{(\ell, \ell', \ell^\tau)}$, as defined in Equation~\ref{ch3:eq:xsim}.

\subsection{Dataset features}\label{ch3:subsec:dataset_features}
\subsubsection*{Dataset size}
The difference in training data size for two languages may serve as a predictor for transfer. It is likely that a low-resource language would benefit from a high-resource language.
Let $S_{\ell}$ denote the set of parallel sentences to English for language $\ell$, and $S_{\ell'}$ be defined similarly for language $\ell'$. 
We then compute the ratio of the smaller size to the larger size as follows:
\begin{equation}
S_{(\ell, \ell')} = \frac{\min (|S_{\ell}|, |S_{\ell'}|)}{\max (|S_{\ell}|, |S_{\ell'}|)}. 
\end{equation}

Since $\textrm{xsim}$ is symmetric, we design features that are also symmetric, when applicable.\\\\

\subsubsection*{Vocabulary occupancy}
We calculate the difference in vocabulary occupancy for $\ell$ and $\ell'$. 
The fraction of the vocabulary that is used by a language captures information about how well the subwords are optimized for that language. 
Let $V_{\ell}$ be the set of unique subwords in vocabulary $V$ that are present in the training data $S_{\ell}$ of language $\ell$. 
The vocabulary occupancy is then computed as: $|V_{\ell}|/|V|$. $V_{\ell'}$ is defined similarly.
The vocabulary occupancy ratio between $\ell$ and $\ell'$ is defined as:
\begin{equation}
V_{(\ell,\ell')}=\frac{\min (|V_{l}|/|V|,|V_{\ell'}|/|V|)}{\max (|V_{l}|/|V|,|V_{\ell'}|/|V|)}.
\end{equation}

\subsubsection*{Source subword overlap}
We measure the similarity between the (subword) vocabularies of language $\ell$ and language $\ell'$.
This is calculated by taking the ratio of the number of subwords that are common to both languages ($|V_{\ell} \cap V_{\ell'}|$) and the total number of unique subwords in both languages ($|V_{\ell} \cup V_{\ell'}|$) according to the following equation:
\begin{equation}
\quad\quad O_{\textrm{src}(\ell,\ell')} = \frac{|V_{\ell} \cap V_{\ell'}|}{|V_{\ell} \cup V_{\ell'}|}.
\end{equation}
We also investigated the use of frequency-weighted subwords, which produced similar results.\\

\subsubsection*{Multi-parallel overlap}
 We are interested to see how generating identical target sentences (in English) affects transfer. 
To calculate this, we take the ratio of the number of multi-parallel sentences shared by the languages $\ell$ and $\ell'$, denoted as $|S_{\ell} \cap S_{\ell'}|$, to the total number of training sentences in both languages ($|S_{\ell} \cup S_{\ell'}|$):
% equation 3.5
\begin{equation}
S_{\textrm{shared}(\ell,\ell')} = \frac{|S_{\ell} \cap S_{\ell'}|}{|S_{\ell} \cup S_{\ell'}|}.
\end{equation}

\subsubsection*{Target n-gram overlap}
We also measure the similarity between the \textit{generated} target n-grams for the languages.
This is similar to the (weighted) source subword overlap but applied to the target side. 
Let $S_{(\ell,\ell^p)}$ be the set of aligned training sentence pairs of language $\ell$ with pivot language $\ell^p$ (English is taken as pivot here). The (weighted) target n-gram overlap is then defined as:
\begin{equation}
O_{\textrm{tgt}(\ell, \ell')} = \sum_{i}\sum_{n} \textrm{n-gram-count}(S_{(\ell',\ell^p)}[i]) \cdot \textrm{\textit{n}-gram-count}(S_{(\ell,\ell^p)}[i]) \cdot n,
\end{equation}
where $\textrm{n-gram-count}(\cdot)$ is the n-gram count in a sentence.

We have also experimented with higher-order n-grams, and found similar results as unigram, thus we only report the results for unigram.\\

\subsection{Linguistic features}\label{ch3:subsec:linguistic_features}
Following \citet{lin_choosing_2019}, we adopt five linguistic features:

\subsubsection{Geographic distance}
Geographic distance measures the physical separation between the traditional territories of languages $\ell$ and $\ell'$. This is calculated as the orthodromic (great-circle) distance between the approximate geographic centers of the languages' historical usage areas, normalized by dividing by the maximum possible distance on Earth's surface.
Languages that developed in proximity often share areal features due to contact effects, even when not closely related genetically, making this a potentially useful predictor for representational similarity.

\subsubsection{Genetic distance}
Genetic distance quantifies the phylogenetic relationship between languages $\ell$ and $\ell'$ based on their position in the language family tree.
This measure is derived from historical linguistics research on language descent and evolution paths through the Glottolog database \citep{hammarstrom_glottologglottolog_2021}.
Languages with smaller genetic distances typically share more grammatical structures, morphological patterns, and lexical properties due to their common ancestral language, potentially leading to better transfer learning capabilities between them.

\subsubsection{Inventory distance}
Inventory distance captures the similarity between the phonological inventories (the sets of sounds used) in languages $\ell$ and $\ell'$.
This is calculated as the cosine distance between feature vectors derived from the PHOIBLE database \citep{moran_cldf-datasetsphoible_2019}, which catalogs the distinctive sounds across the world's languages.
Languages with similar sound inventories often exhibit parallel phonological constraints and processes, which may influence subword tokenization patterns and subsequently affect cross-lingual transfer at the representation level.

\subsubsection{Syntactic distance}
Syntactic distance measures the structural differences between languages $\ell$ and $\ell'$ based on their grammatical properties.
This feature is computed as the cosine distance between typological feature vectors that encode fundamental syntactic characteristics such as word order patterns, morphological type, and the presence of features like grammatical gender or case marking.
Languages with lower syntactic distance tend to organize information in similar ways, potentially facilitating more effective transfer of structural knowledge between their representations.

\subsubsection{Phonological distance}
Phonological distance quantifies the differences in sound systems and phonological processes between languages $\ell$ and $\ell'$.
Unlike inventory distance, which focuses on the sets of individual phonemes, this measure incorporates phonological rules, constraints, and patterns (such as syllable structure, stress patterns, and phonotactic constraints) derived from WALS \citep{dryer_wals_2013} and Ethnologue \citep{lewis_ethnologue_2009} databases.
Languages with similar phonological systems might encode phonological information in comparable ways in their representations, potentially affecting cross-lingual transfer performance.

\subsection{Experimental setup}
We treat the prediction of the representational similarities $\textrm{xsim}_{(\ell, \ell', \ell^\tau)}$ (see Equation~\ref{ch3:eq:xsim}) when translating into target language $\ell^\tau$ (English) between source languages $\ell$ and $\ell'$ as a regression problem. 
We use the features described in the previous subsection as input variables. 
To account for variations in feature values across different language pairs, we scale the features between 0 and 1. 
We consider all 52 source languages.
Considering that representational similarities are symmetric, and discarding combinations where $\ell = \ell'$, the resulting number of to be predicted representational similarities is $\frac{(52 \cdot 52)-52}{2} = 1326$. 
We use a leave-one-out cross-validation approach, leaving out all similarities for a single language in each round. 
To evaluate the performance of the model, we use the average (over all language pairs) mean absolute error (MAE) as the evaluation metric. 
Since different machine learning algorithms have different inductive biases, we train and evaluate three regression models using the scikit-learn library \citep{pedregosa_scikit-learn_2011}: linear regression (LR), multilayer perceptron (MLP), and gradient boosting (GB). 
For MLP and GB, we report the average score over 3 random seeds. We do not report STD as it is negligible.
For MLP, we used hidden layer dimensionality 80 using 3 layers.
We use the ReLU activation function, Adam \citep{kingma_adam_2015} ($\beta_1 = 0.9$, $\beta_2 = 0.98$ and $\epsilon = 10^{-9}$).
For GB, we used squared error, 0.1 learning rate, and 100 estimators.

\subsection{Prediction results}
The results for predicting representational similarities are shown in Table~\ref{ch3:tab:mae}. 
First, combined features lead to better MAE scores than single features. 
Using all dataset features results in better predictions than using all linguistic features, and combining dataset and linguistic features results in best results for all algorithms.
Furthermore, all single features have the potential to improve over a na\"{i}ve baseline (random input), indicating that they have at least some predictive power.

\begin{table}[!htb]
    \begin{center}
    \small
        \begin{tabular}{llrrr}
        \toprule
        \multicolumn{2}{c}{Feature} & \multicolumn{3}{c}{Regressor} \\
        \cmidrule(lr){3-5}
         &  & \multicolumn{1}{c}{LR} & \multicolumn{1}{c}{MLP} & \multicolumn{1}{c}{GB}\\
        \midrule
        \parbox[t]{2mm}{\multirow{1}{*}{\rotatebox[origin=c]{90}{}}}
        & baseline (noise)          & $0.061$ & $0.061$ & $0.061$ \\
        \midrule
        \parbox[t]{2mm}{\multirow{5}{*}{\rotatebox[origin=c]{90}{dataset}}}
        & dataset size              & $0.052$ & $0.052$ & $0.046$ \\
        & vocabulary occupancy      & $0.041$ & $0.041$ & $0.035$ \\
        & multi-parallel overlap    & $0.047$ & $0.042$ & $0.034$ \\
        & source subword overlap     & $\underline{0.040}$ & $\underline{0.036}$ & $\underline{0.031}$ \\
        & target n-gram overlap     & $0.050$ & $0.046$ & $0.042$ \\
        \midrule
        \parbox[t]{2mm}{\multirow{5}{*}{\rotatebox[origin=c]{90}{linguistic}}}
        & geographic distance       & $0.062$ & $0.053$ & $\underline{0.049}$  \\
        & genetic distance          & $0.054$ & $0.053$ & $\underline{0.049}$ \\
        & inventory distance        & $0.062$ & $0.061$ & $0.055$ \\
        & syntactic distance        & $\underline{0.051}$ & $\underline{0.050}$ & $0.050$ \\
        & phonological distance     & $0.061$ & $0.061$ & $0.052$ \\
        \midrule
        \parbox[t]{2mm}{\multirow{3}{*}{\rotatebox[origin=c]{90}{}}}
        & all data                  & $0.031$ & $0.029$ & $0.021$ \\
        & all linguistic            & $0.049$ & $0.043$ & $0.034$ \\
        & all data + all linguistic & $\mathbf{0.028}$ & $\mathbf{0.025}$ & $\mathbf{0.016}$ \\
        \bottomrule
        \end{tabular}
    \end{center}
    \caption{Mean absolute error (MAE) scores averaged over language pairs for transfer prediction, i.e., predicting $\textrm{xsim}_{(\ell, \ell', \ell^\tau)}$ (similarity scores between languages $\ell$ and $\ell'$ when translating into English, see Equation~\ref{ch3:eq:xsim}) using dataset features (Section~\ref{ch3:subsec:dataset_features}) and linguistic features (Section~\ref{ch3:subsec:linguistic_features}).
    Best scores per regressor in \textbf{bold} and per feature class \underline{underlined}.}
    \label{ch3:tab:mae}
\end{table}

\subsection{Feature importance}
We investigate the importance of features to gain a better understanding of their role in transfer.\\
\subsubsection{Linear regression coefficients} 
\begin{figure}[!htb]
\centering
\includegraphics[width=0.7\linewidth]{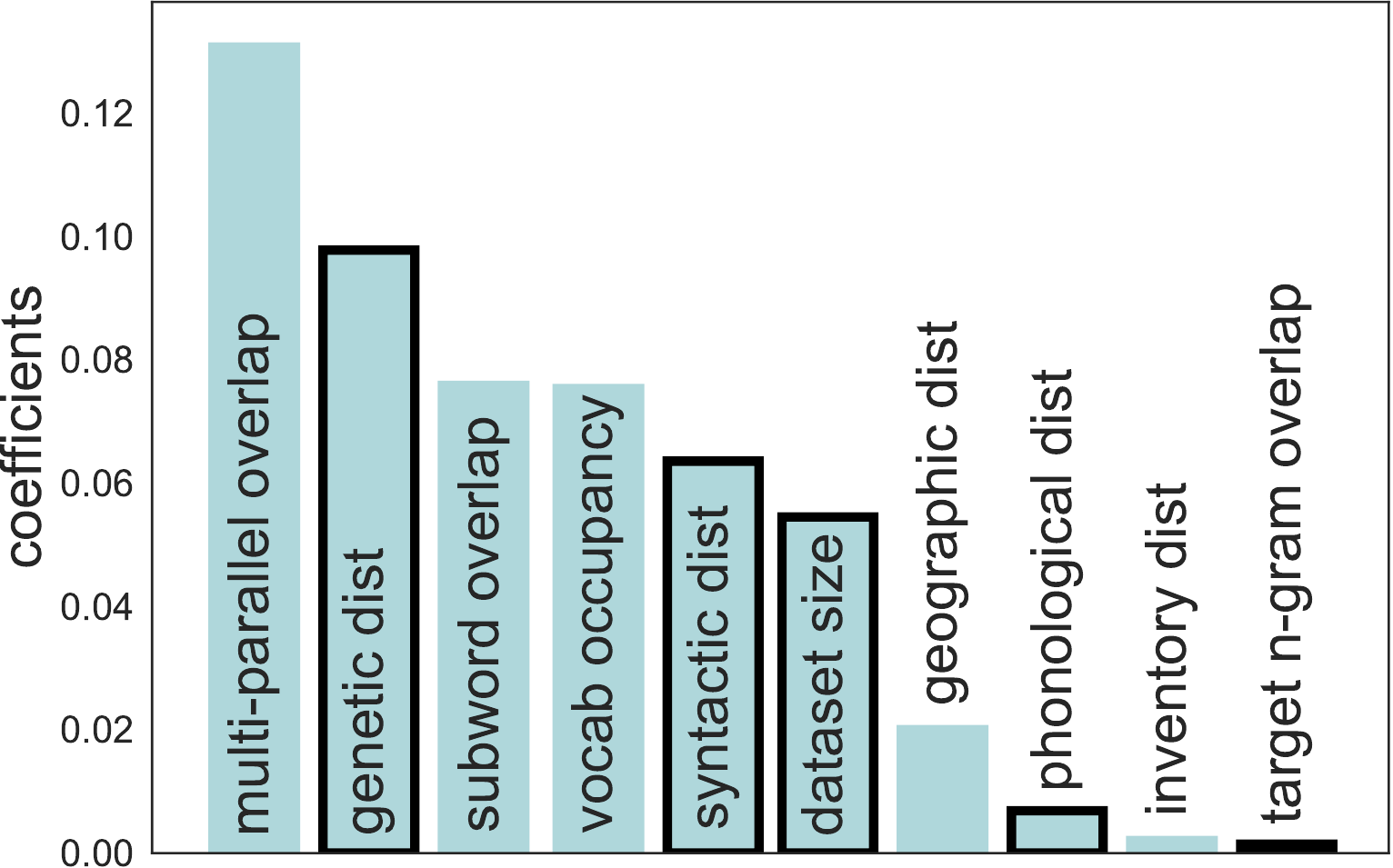}
\caption{Feature importance for transfer prediction: linear regression sign of coefficients.
Absolute values are plotted. 
Black line indicates negative coefficient (e.g., genetic distance is negative).}
\label{ch3:fig:coeff_importance}
\end{figure}

Weight coefficients are used as a crude measure of feature importance. 
These coefficients quantify the conditional association between the target $\textrm{xsim}$ and a given feature, while holding other features constant.
The sign of the coefficients shows the direction of the association, and the magnitude is an indication of the strength of the association. 
In Figure~\ref{ch3:fig:coeff_importance}, we can see that multi-parallel overlap, source subword overlap, and vocabulary occupancy have the largest positive weights among the data features, which implies that these features are positively associated with the target variable and have a strong influence on the prediction. 
Furthermore, genetic and syntactic distance have the highest importance among the linguistic features.

\subsubsection{Permutation importance}
\begin{figure}[!htb]
    \centering
    \includegraphics[width=0.7\linewidth]{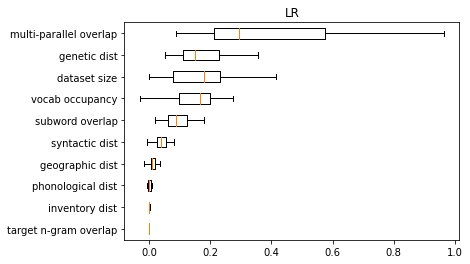}
    \includegraphics[width=0.7\linewidth]{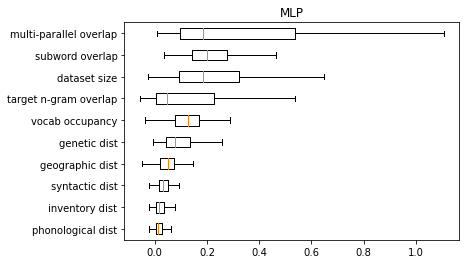}
    \includegraphics[width=0.7\linewidth]{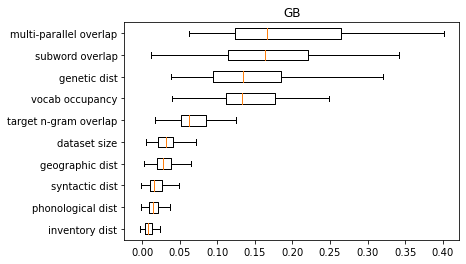}
    \caption{Sorted permutation feature importance scores for linear regression (LR, top), multilayer perceptron (MLP, middle), and gradient boosting (GB, bottom).}
    \label{ch3:fig:permut_importance}
\end{figure}

To further understand the importance of each feature, we additionally calculate permutation feature importance scores \citep{breiman_random_2001,fisher_all_2019}. 
This method evaluates the decrease in model score when a single feature value is randomly shuffled. 
This model-agnostic procedure breaks the relationship between the feature and the target, thus the drop in the model score is indicative of how much the model depends on the feature.
The results using permutation feature importance are consistent with the results obtained using linear regression coefficients.
Specifically, we find that multi-parallel overlap is the most important feature for all three regression models. 
Source subword overlap is also important for MLP and GB, and slightly less for LR. 
Vocabulary occupancy and dataset size also score relatively high on importance. 
Genetic distance is consistently the most important linguistic feature among all models.
See Figure~\ref{ch3:fig:permut_importance} for the feature importance box plots.

\section{Optimizing for representational invariance} \label{ch3:sec:multiparallel_training}
Some features that we have shown to be predictive of transfer have been used in previous work. 
Higher vocab overlap leads to more positive transfer \citep{chung_improving_2020,patil_overlap-based_2022,sun_alternative_2022}.
Temperature sampling addresses dataset size (imbalance) \citep{arivazhagan_massively_2019}.
Backtranslated data can be used for similar effect \citep{liao_back-translation_2021}. 
Grouping languages by their linguistic similarity outperforms English-centric models \citep{oncevay_bridging_2020,fan_beyond_2021}.

In contrast, there are no such methods for multi-parallel data.
Parallel datasets contain a large number of hidden multi-parallel sentences that remain unused, and resurfacing these improves multilingual translation quality \citep{freitag_complete_2020,xu_eag_2022}.
However, these approaches merely increase the available multi-parallel data without explicitly incorporating its structural properties into the learning objective.
Our method, by comparison, directly exploits the characteristics of multi-parallel data.

We introduce an auxiliary similarity loss that encourages context vectors to be more similar when generating the same target token.
When sampling a parallel batch, consisting of a source sentence $x$ and the corresponding target sentence $y$, we optimize the cross-entropy loss as usual. 
When sampling a multi-parallel batch, consisting of meaning equivalent triples $\{x^1,x^2,y\}$ (as defined in Section \ref{ch3:subsec::representational_view}), such that $y \neq x^1$ and $x^1 \neq x^2$, we optimize a similarity loss function: 
\begin{equation}
\mathcal{L}_{\textrm{xsim}(x^1, x^2, y)} = \sum_{t=1}^n \textrm{sim}(\times_t(x^1,y), \times_t(x^2,y)),
\end{equation}
where $\textrm{sim}(\cdot,\cdot)$ is a similarity function and $\times_t(\cdot,\cdot)$ is the context vector resulting from the cross-attention at decoding timestep $t$. 
The goal of minimizing $\mathcal{L}_{\textrm{xsim}}$ is to encourage representations that are invariant across languages. 
The final learning objective for multi-parallel batches ($x^1,x^2,y$) combines minimizing $\mathcal{L}_{\textrm{xsim}}$ and cross-entropy ($\mathcal{L}_{CE}$): 
\begin{equation}
\mathcal{L}_{(x^1,x^2,y)} = \lambda\mathcal{L}_{\textrm{xsim}(x^1,x^2,y)} + \displaystyle\sum_{i=1} ^{2} \mathcal{L}_{\textrm{CE}(x^i,y)}.
\end{equation}

\subsection{Experimental setup}
We follow the setup as described in Section~\ref{ch3:sec:baseline_setup} and make the following modifications: 
1) we sample parallel and multi-parallel batches in a 1:1 ratio, 2) for the multi-parallel batches, we optimize an auxiliary cosine similarity loss and set weight to $\lambda = 1$. 
To reduce the impact of a small number of dimensions that can dominate similarity metrics, known as rogue dimensions \citep{timkey_all_2021}, we subtract the mean context vector $\bar{\mathrm{c}}$ from each context vector in the batch before calculating similarities. 
If we sample a batch where English is \textit{not} the target, we do not calculate a similarity loss, i.e., $\lambda=0$. 
Note, that our method does not require a dataset that is fully multi-parallel. 
The parallel dataset consists of all X$\rightarrow$En and En$\rightarrow$X pairs. 
Its size is 10M pairs. 
The multi-parallel dataset consists of all $(x^1,x^2)$ source combinations, with target $y$ fixed to English. 
The size is 5.9M triples.

Additionally, we perform an ablation experiment where we set the similarity loss to 0, to investigate the role of the loss versus the modified data sampling strategy. Note that we cannot ablate the multi-parallel batching, since the similarity loss requires multi-parallel batches.

\subsection{Results}
\begin{table*}[!htb]
    \centering
    \small
    \begin{tabular}{l|llllll}
        \toprule
        \multirow{2}{*}{Dataset} & \multirow{2}{*}{Model} & \multicolumn{3}{c}{Resource level} \\
        & & low ($<$10K) & mid (10K-150K) & high ($>$150K) \\
        & & 12 languages & 23 languages & 17 languages \\
        \midrule
        \multirow{6}{*}{FLORES} 
        & many-to-many         
            & $9.7$
            & $16.8$
            & $\mathbf{17.9}$ \\
        & + multi-parallel
            & $9.9^{*(0/0)}$
            & $16.9^{*(0/0)}$
            & $17.7^{*(0/0)}$ \\
        & + xsim
            & $\mathbf{11.5}^{*(12/0)}$
            & $\mathbf{17.8}^{*(18/0)}$
            & $17.4^{*{(0/13)}}$ \\
        \cmidrule{2-5}
        & many-to-one
            & $8.8$
            & $14.6$
            & $\mathbf{15.0}$ \\
        & + multi-parallel
            & $8.6^{*(0/0)}$
            & $14.7^{*(0/0)}$
            & $14.9^{*(0/0)}$ \\
        & + xsim
            & $\mathbf{10.8}^{*(12/0)}$
            & $\mathbf{15.7}^{*{(16/2)}}$
            & $14.5^{*(0/10)}$ \\
        \midrule
        \multirow{6}{*}{TED} 
        & many-to-many         
            & $20.5$
            & $30.2$
            & $\mathbf{31.2}$ \\
        & + multi-parallel
            & $20.8^{*(0/0)}$
            & $30.1^{*(0/0)}$
            & $31.0^{*(0/0)}$ \\
        & + xsim
            & $\mathbf{21.8}^{*(12/0)}$
            & $\mathbf{30.7}^{*(14/1)}$
            & $30.4^{*{(0/14)}}$ \\
        \cmidrule{2-5}
        & many-to-one
            & $19.9$
            & $27.9$
            & $\mathbf{27.2}$ \\
        & + multi-parallel
            & $19.7^{*(0/0)}$
            & $27.6^{*(0/0)}$
            & $27.0^{*(0/0)}$ \\
        & + xsim
            & $\mathbf{22.0}^{*(12/0)}$
            & $\mathbf{28.8}^{*(14/2)}$
            & $26.6^{*(0/11)}$ \\
        \bottomrule
    \end{tabular}
    \caption{X$\rightarrow$En BLEU scores on FLORES-101 and TED test datasets for multilingual many-to-many and many-to-one models. We compare baseline models against models with multi-parallel batches during training (+ multi-parallel) and models with both multi-parallel batches and auxiliary similarity loss (+ xsim). 
    Results are grouped by resource level, with $*(n/m)$ denoting the fraction of scores in a bucket that are significantly better ($n$) or worse ($m$) than the bilingual baseline, according to bootstrap resampling.}
    \label{ch3:tab:flores_bleu_sim}
\end{table*}

We include results for our method on both in-domain (TED) and out-of-domain test sets (FLORES-101), for both many-to-many as well as many-to-one models. 
Table~\ref{ch3:tab:flores_bleu_sim} shows BLEU scores and a comparison to the baselines. 
Adding multi-parallel batches and our similarity loss yields improvements for low- and mid-resource languages, in both many-to-many and many-to-one models.
Including multi-parallel batches without applying a similarity loss leads to scores that are not statistically significantly different from the baseline.
Furthermore, many-to-many models have the best performance on all aggregated test score buckets. 
Lowest resource languages benefit most from this approach, with an average BLEU increase of $+1.8$ and $+1.3$ (many-to-many). 
This makes sense, since $\mathcal{L}_\textrm{xsim}$ encourages the representations of these languages to be more similar to other languages, most of which have better performance. 
Mid-resource languages also benefit from adding $\mathcal{L}_{\textrm{xsim}}$: $+1.0$ and $+0.5$ average increase for FLORES-101 and TED. 
Higher resource languages suffer from adding the auxiliary loss ($-0.5$ for FLORES-101, $-0.8$ for TED). 
These results demonstrate that lower- and mid-resource languages improve when explicitly optimizing for language invariance using multi-parallel data.
Higher-resource languages pay a small price in performance. 
This trend holds for in- and out-of-domain test sets, and different types of multilingual models.

\section{Conclusion}\label{ch3:sec:conclusion}
In this chapter, we have investigated the relationship between cross-lingual transfer and representational similarities between languages.
We asked:\\
\RQsub{1}{1}
\begin{myquote}
We argue that translation quality alone is not a sufficient metric for measuring knowledge transfer in multilingual neural machine translation. 
To support this claim, we introduce Representational Transfer Potential (RTP), which measures representational similarities between languages.
As demonstrated in Figure~\ref{ch3:fig:delta_bleu_vs_rtp}, RTP effectively quantifies both positive and negative transfer.
Furthermore, we show that these similarities correlate with improvements in translation quality, indicating that there \textit{is} genuine knowledge transfer occurring, and the improved translation quality is \textit{not} only due to the increased data on the target side.
Having established RTP as a reliable metric for measuring cross-lingual transfer, we next addressed:
\end{myquote}

\RQsub{1}{2}
\begin{myquote}
Through our feature importance analysis in Section~\ref{ch3:sec:predicting_transfer}, we identify the dataset and language characteristics that are most relevant for transfer.
As shown in Figure~\ref{ch3:fig:permut_importance}, multi-parallel overlap emerges as the most important yet under-explored feature, followed by source subword overlap and vocabulary occupancy.
Among linguistic features, genetic distance consistently shows the highest importance across all regression models.
\end{myquote}

\RQsub{1}{3}
\begin{myquote}
Drawing on our findings from RQ1.1 and RQ1.2, we develop a novel training scheme that incorporates an auxiliary similarity loss.
This approach encourages greater cross-lingual invariance in representations by leveraging the properties of multi-parallel data.
Through extensive experimentation, we show that our method increases the degree of language invariance across source representations.
Our experimental results in Table~\ref{ch3:tab:flores_bleu_sim} clearly demonstrate that this approach leads to increased performance for low- and mid-resource languages across multiple data and model setups.
\end{myquote}
Taken together, these steps answered our main research question:

\RQ{1}
\begin{myquote}
We formalize the relationship between cross-lingual transfer and cross-attention similarities between languages, by introducing a new metric: Representational Transfer Potential (RTP). 
This allows us to reason about knowledge transfer in a way translation quality (BLEU) is unable to capture.
By incorporating an auxiliary similarity loss that encourages representations to be invariant across languages, we obtain a higher degree of invariance, which yields substantial improvements in translation quality in low- and mid-resource languages.
\end{myquote}

\subsubsection{Limitations}
While our focus is on English-centric many-to-many and many-to-English models, it is important to note that there has been prior work that has explored non-English-centric setups, such as the studies by \citet{fan_beyond_2021} and \citet{freitag_complete_2020}. 
This may present limitations in the generalizability of our results to other multilingual settings. 

While our analysis already uses 53 languages, we did not measure to what extent our findings hold when using even more languages. 

Furthermore, our training data size is relatively small which may affect model performance.

We use TED instead of the larger OPUS-100 dataset, because TED has higher translation quality and consists of partly multi-parallel data.

\chapter{Multilingual \textit{k}-Nearest-Neighbor Machine Translation}
\label{ch4}

\renewcommand{\thefootnote}{}
\footnotetext{This chapter was published as: 
David Stap and Christof Monz.
Multilingual \textit{k}-Nearest-Neighbor Machine Translation.
In \textit{Proceedings of the 2023 Conference on Empirical Methods in Natural Language Processing}, pages 9200–9208, Singapore. Association for Computational Linguistics, 2023.}
\renewcommand{\thefootnote}{\arabic{footnote}}

\section{Introduction and research questions}
Building upon our findings in the previous chapter, where we established that representational similarities between languages facilitate cross-lingual knowledge transfer in multilingual NMT, we now shift our focus to \textit{k}-nearest neighbor machine translation (\textit{k}NN-MT), investigating how multilingual representations can be leveraged in datastores to improve translation quality, particularly for low-resource languages.

Recently, semi-parametric approaches such as \textit{k}NN-MT \citep{khandelwal_nearest_2021} have attracted interest due to a series of impressive results in language modeling and machine translation \citep{guu_generating_2018,bapna_non-parametric_2019,khandelwal_generalization_2020}.
These techniques capitalize on information retrieved from an extensive repository of translation examples cached in a datastore.
One of the most important limitations of \textit{k}NN-MT is that the extent of quality improvements strongly depends on the size of the datastore \citep{khandelwal_generalization_2020,khandelwal_nearest_2021,zhu_knn-box_2024}.
This dependence on datastore size is problematic for low-resource languages and the improvements that \textit{k}NN-MT can offer in low-resource settings are modest at best \citep{Vardhan2022}.
On the other hand, there are general methods that can improve low-resource performance, such as transfer learning \citep{zoph_transfer_2016,kocmi_trivial_2018} and multilingual NMT (mNMT) \citep{johnson_googles_2017,arivazhagan_massively_2019,stap_viewing_2023}.

As discussed in Section \ref{ch3:sec:transfer_m2m_models}, multilingual NMT demonstrates varying degrees of knowledge transfer between languages, with low-resource languages typically benefiting most.
Preliminary findings on combining \textit{k}NN-MT with mNMT suggest that mNMT representations generalize sufficiently well across languages to make cross-lingual retrieval effective \citep{khandelwal_nearest_2021}.
However, its effectiveness for low-resource languages remains an open question.

In this chapter, we investigate the following research question:

\RQ{2}
\begin{myquote}
While \textit{k}NN-MT has demonstrated impressive results for high-resource languages, its reliance on large datastores presents a significant challenge for low-resource languages.
We explore how cross-lingual knowledge transfer can be leveraged in the context of \textit{k}NN-MT to overcome this limitation.
We investigate methods for effectively combining information from multiple languages into datastores that can benefit both low-resource and high-resource languages while maintaining computational efficiency.\\\\
This overarching question is addressed through three interconnected steps.
First, we explore cross-lingual datastores for improving low-resource translation (RQ2.1).
Then, we examine the creation of multilingual datastores that benefit both low and high-resource languages (RQ2.2).
Finally, we investigate optimization strategies to balance performance and computational efficiency (RQ2.3).
\end{myquote}

\RQsub{2}{1}
\begin{myquote}
We investigate whether a low-resource language can benefit from a datastore built from a related high-resource language, thus overcoming the limitation of small datastore size.
We examine various language pairs to understand what factors influence the effectiveness of cross-lingual retrieval.
\end{myquote}

\RQsub{2}{2}
\begin{myquote}
Having examined cross-lingual datastores, we next investigate how to combine multiple languages into a datastore.
We explore the creation of multilingual datastores that integrate translation examples from various languages, potentially leading to more robust retrieval and better translation quality.
We examine whether such datastores can benefit both low-resource and high-resource language translation.
\end{myquote}

\RQsub{2}{3}
\begin{myquote}
Based on our findings from RQ2.1 and RQ2.2, we then explore how can we optimize multilingual datastores to maintain translation quality while improving computational efficiency.
Since larger datastores lead to slower inference, we investigate strategies to reduce datastore size without sacrificing translation quality.
We explore whether linguistic similarities can be leveraged to create more efficient datastores and whether cross-lingual representation alignment can further enhance retrieval effectiveness.
\end{myquote}

\paragraph{Organization.}
This chapter is organized as follows: Section \ref{ch4:sec:related_work} provides background on \textit{k}NN-MT and reviews related work. In Section \ref{ch4:sec:multilingual_knn_mt}, we introduce our approach to multilingual \textit{k}NN-MT, including the construction of cross-lingual and multilingual datastores. We then present our experimental setup and results. Finally, we analyze which languages contribute most to the multilingual datastores and evaluate the trade-offs between performance and inference speed. In Section \ref{ch4:sec:conclusion}, we summarize our findings in relation to our research questions.

\section{Related work}\label{ch4:sec:related_work}
In this section, we first describe the \textit{k}NN-MT approach in detail, then review subsequent research focused on improving its efficiency and retrieval quality.
Finally, we examine previous efforts to extend \textit{k}NN-MT beyond bilingual settings and address low-resource scenarios, highlighting the research gap our work addresses in leveraging cross-lingual knowledge transfer for languages with limited data.

\begin{figure}[!htb]
    \centering
    \includegraphics[width=\linewidth]{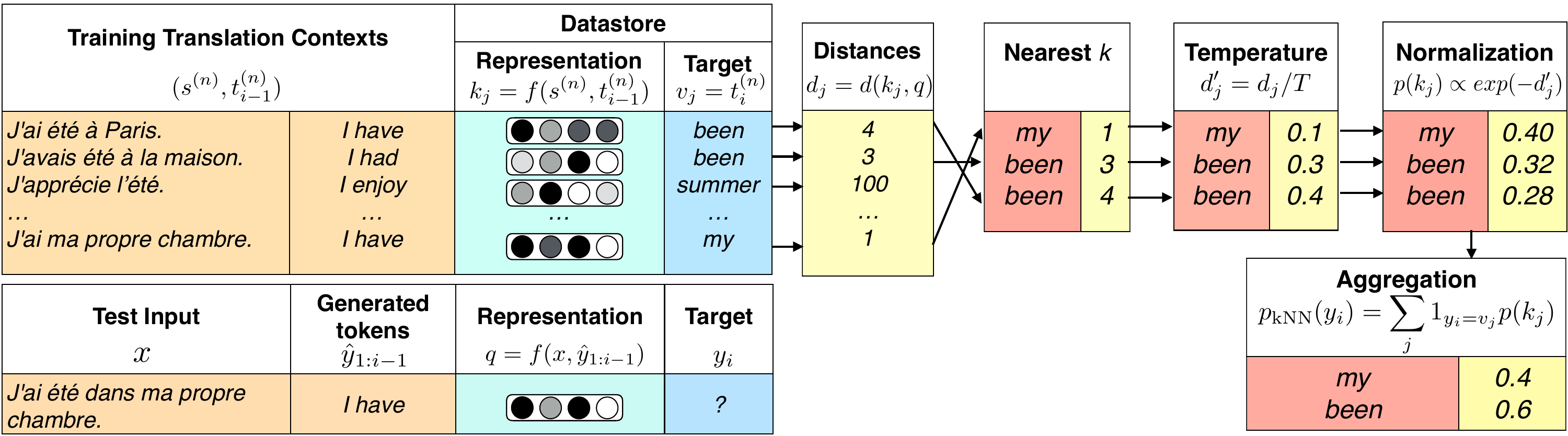}
    \caption{Schematic overview of \textit{k}NN-MT distribution computation.
    The datastore contains representation vectors from training data translation contexts paired with their corresponding target tokens.
    During inference, a query representation derived from the test input sentence and previously generated tokens retrieves the \textit{k} most similar neighbors from the datastore.
    The system calculates distances between the query and retrieved items, applies a softmax temperature function, and aggregates the results to produce the final \textit{k}NN probability distribution over candidate translations.
    Figure reproduced from \citet{khandelwal_generalization_2020}.}
    \label{ch4:fig:knnmt}
\end{figure}

\subsection{\textit{k}-nearest neighbor machine translation}
$k$NN-MT is a semi-parametric method that combines a parametric component with a nearest neighbor retrieval mechanism that allows direct access to a datastore of cached examples \citep{khandelwal_nearest_2021}.
This approach augments a standard NMT model by retrieving similar translation contexts from a datastore during inference, enabling the model to directly reference and leverage specific examples from the training data.
As illustrated in Figure \ref{ch4:fig:knnmt}, the system retrieves the most similar contexts based on representation similarity, applies a temperature-scaled distribution over the corresponding target tokens, and interpolates this with the NMT model's predictions to produce the final translation output.

\paragraph{Datastore}
More precisely, for a translation direction from source language $\ell$ to target language $\ell^\prime$, the datastore $\mathcal{D}_{(\ell, \ell^\prime)}$ consists of key-value pairs.
Each key is a \textit{translation context}—the decoder output representation $f(\mathbf{x},\mathbf{y}_{<t})$ captured from the final decoder layer before the softmax—and each value is the corresponding target token $y_t$ that was generated in that context.
For a parallel text collection $\mathcal{B}_{(\ell, \ell^\prime)}$, the complete datastore is defined as:
\begin{equation}\label{ch4:eq:datastore}
    \mathcal{D}_{(\ell, \ell^\prime)} =
    \{ (f(\mathbf{x},\mathbf{y}_{<t}), y_t),
    \forall y_t \in \mathbf{y}
    \mid
    (\mathbf{x}, \mathbf{y} \in \mathcal{B}_{(\ell, \ell^\prime)})\}.
\end{equation}

The datastore is constructed offline through a single forward pass over each training example.
This process is computationally efficient as it requires significantly less resources than training on these examples.
The key representations capture information from both the source sentence and the target prefix, effectively encoding the full translation context.

\paragraph{Inference}
At inference time, given a source sentence $\mathbf{x}$, the NMT model outputs a distribution over the vocabulary $p_{\mathrm{NMT}}(y_t|\mathbf{y}_{<t}, \mathbf{x})$ for each target position.
Simultaneously, the model computes the representation $f(\mathbf{x},\mathbf{y}_{<t})$, which is used to query the datastore for the $k$ nearest neighbors $\mathcal{N}$ according to squared-$L^2$ distance $d$.
In practice, the search over potentially billions of key-value pairs is carried out using efficient approximate nearest neighbor search algorithms.

The retrieved set of nearest neighbors is converted into a distribution over the vocabulary by applying a softmax with temperature parameter $T$ to the negative distances and aggregating over multiple occurrences of the same vocabulary item:
\begin{equation}
p_{k\mathrm{NN}}(y_t|\mathbf{y}_{<t}, \mathbf{x}) \propto \sum_{(k_j,v_j) \in \mathcal{N}} \mathbbm{1}_{y_t=v_j} \exp\left(\frac{-d(k_j, f(\mathbf{x},\mathbf{y}_{<t}))}{T}\right),
\end{equation}
where $T$ controls the sharpness of the distribution.
Values greater than 1 flatten the distribution, preventing the model from overfitting to the most similar retrievals and allowing for greater diversity in the predictions.
This is particularly important when multiple relevant examples with different translations are retrieved.

The final probability distribution is obtained by interpolating the NMT model distribution and the $k$NN distribution using a tuned parameter $\lambda$:
\begin{equation}
p(y_t|\mathbf{y}_{<t}, \mathbf{x}) = \lambda \cdot p_{k\mathrm{NN}}(y_t|\mathbf{y}_{<t}, \mathbf{x}) + (1 - \lambda) \cdot p_{\mathrm{NMT}}(y_t|\mathbf{y}_{<t}, \mathbf{x}).
\end{equation}
This interpolation combines the two probability distributions with fixed weights.
$p_{k\mathrm{NN}}(y_t|\mathbf{y}_{<t}, \mathbf{x})$ will naturally assign higher probabilities to tokens found in similar contexts in the datastore, while having a more uniform distribution when no good matches exist.
$p_{\mathrm{NMT}}(y_t|\mathbf{y}_{<t}, \mathbf{x})$ contributes predictions with consistent weighting regardless of the quality of \textit{k}NN matches.
The final combined distribution is then used to generate the translation through beam search.

\paragraph{Advantages}
An advantage of \textit{k}NN-MT is it improves the translation quality of NMT models without retraining.
By providing the decoder direct access to examples at test time, it creates a more expressive model.
Moreover, it enables domain adaptation through domain-specific datastores without fine-tuning---particularly valuable when handling diverse texts with varying terminology and style conventions.

\subsubsection{Reducing computational and memory costs}
Methods like \textit{k}NN-MT lead to a steep increase in decoding time, because each decoding timestep requires a computationally expensive nearest neighbor search.
In light of this, subsequent research has predominantly focused on reducing computational and memory expenses \citep{zheng_adaptive_2021,martins_chunk-based_2022,martins_efficient_2022,yang_nearest_2022,meng_fast_2022,jiang-etal-2022-towards}.
Some notable examples are moving the \textit{k}NN search forward to the preprocessing phase \citep{yang_nearest_2022}, limiting the search space \citep{meng_fast_2022}, or datastore compression and dimension reduction \citep{he_efficient_2021,martins_efficient_2022}.
These optimizations enable \textit{k}NN-MT to be more practically applicable in real-world translation scenarios where efficiency is critical, while preserving most of the quality improvements that the approach offers.

\subsubsection{Reducing retrieval noise}
Another challenge in \textit{k}NN-MT is retrieval noise, which occurs when irrelevant or misleading examples are retrieved from the datastore.
This noise can negatively impact translation quality by introducing incorrect tokens into the final output.
Research has addressed this issue through two main approaches: dynamically estimating \textit{k}NN-MT hyperparameters \citep{jiang_learning_2021,wang2022non,jiang-etal-2022-towards,drozdov-etal-2022-cant} and explicitly aligning pre-trained representations with model representations \citep{jin-etal-2022-plug,li_better_2022}.

Our work differs from both the computational efficiency research and the retrieval noise reduction efforts.
We focus on improving performance through cross-lingual knowledge transfer, particularly for low-resource languages where the benefits of \textit{k}NN-MT have been previously limited due to small datastore sizes.

\subsubsection{Semi-parametric approaches for multilingual and low-resource settings}
While \textit{k}NN-MT has shown significant improvements in high-resource settings, its effectiveness for low-resource languages remains limited.
Several alternative approaches have been developed to address this challenge.
\citet{cai_neural_2021} introduced neural machine translation with monolingual translation memory, which improves low-resource translation through monolingual data.
Unlike \textit{k}NN-MT, however, their approach requires additional parameters and a more complex setup rather than directly leveraging model representations.
For low-resource scenarios, \citet{Vardhan2022} proposed an additional filtering layer to reduce noise from small datastores.
However, quality gains remain modest, with only 0.2 BLEU points improvement on average.

Despite the potential benefits, the integration of multilingual NMT with \textit{k}NN-MT has only been minimally explored.
\citet{khandelwal_nearest_2021} found that using a cross-lingual datastore with English on the source side can improve multilingual performance, but it remains unclear to what extent this works for low-resource languages where representation quality is inherently limited.
\citet{li_better_2022} obtain improvements in multilingual settings, but their method relies on pre-trained models and aligning low-resource representations is challenging due to data scarcity.
Notably, these studies do not include any low-resource languages, nor consider linguistic similarities as a factor in datastore construction.

Our work addresses this research gap by investigating how linguistically informed cross-lingual and multilingual datastores can improve translation quality for low-resource languages.

\section{Multilingual \textit{k}-nearest-neighbor machine translation}\label{ch4:sec:multilingual_knn_mt}
In this section, we present our approach to multilingual \textit{k}-nearest neighbor machine translation.

Our goal is to improve performance for low-resource languages by constructing cross-lingual and multilingual datastores.
These datastores consist of keys generated from mNMT representations, allowing semantically related sentences from different languages to cluster together \citep{johnson_googles_2017,escolano_bilingual_2019}, a phenomenon we explored in depth in Section \ref{ch3:sec:transfer_m2m_models}.

\subsection{Bilingual and cross-lingual datastores}
We use bilingual datastores as defined in Section~\ref{ch4:sec:related_work}, Equation \ref{ch4:eq:datastore}. To recap, a \textit{bilingual datastore} $\mathcal{D}_{(\ell, \ell^\prime)}$ contains key-value pairs derived from bi-text data $\mathcal{B}_{(\ell, \ell^\prime)}$ originating from a single source language $\ell$ into target language $\ell^\prime$.

When we use a bilingual datastore $\mathcal{D}_{(\ell, \ell^\prime)}$ to augment the translation direction of \textit{another} source language $\ell^* \neq \ell$ into the same target language $\ell^\prime$, we call the datastore \textit{cross-lingual}.
For instance, a Russian-English datastore $\mathcal{D}_{(\textrm{ru}, \textrm{en})}$ may be used to enhance Belarusian-English translation.

An important advantage of cross-lingual datastores is that they can be significantly larger than their bilingual counterparts, potentially resulting in better translation quality compared to using a smaller language-specific datastore.

\subsection{Multilingual datastores}\label{sec:mlds}
Earlier work is limited to monolingual or bilingual datastores \citep{khandelwal_nearest_2021,cai_neural_2021,li_better_2022}.
In contrast, we create multilingual datastores consisting of multiple source languages, resulting in larger datastores.\\

We construct a \textit{multilingual datastore} $\mathcal{D}_{(\text{L}_{\text{ML}}, \ell^\prime)}$ by considering a set of source languages $\text{L}_{\text{ML}}$ that map to a target language $\ell^\prime$:
\begin{equation}
    \mathcal{D}_{(\text{L}_{\text{ML}}, \ell^\prime)}
    =
    \{(f(\mathbf{x},\mathbf{y}_{<t}), y_t),
    \forall y_t \in \mathbf{y}
    \mid
    (\mathbf{x}, \mathbf{y}) \in \mathcal{B}_{(\text{L}_{\text{ML}}, \ell^\prime)} \},
\end{equation}
where $\mathcal{B}_{(\text{L}_{\text{ML}}, \ell^\prime)} = \bigcup_{\ell\in\text{L}_{\text{ML}}} \mathcal{B}_{(\ell, \ell^\prime)}$ is the combined data from all $\text{L}_{\text{ML}}$ source languages into target $\ell^\prime$.\\

\subsubsection{Linear cross-lingual representation alignment}
To further align multilingual representations, we learn a linear mapping between two languages.
Our goal is to let language $\ell^1$ more effectively query from a $\ell^2$ datastore, thereby improving cross-lingual transfer.
This approach is complementary to the auxiliary similarity loss method we introduced in Section \ref{ch3:sec:multiparallel_training}, both aiming to increase cross-lingual invariance in representations.

For our training data $\mathbb{T}$, we include translation contexts from the $\ell^1$ datastore $\mathcal{D}_{(\ell^1, \ell^\prime)}$ and the $\ell^2$ datastore $\mathcal{D}_{(\ell^2, \ell^\prime)}$ that correspond to the \textit{same} target sentence $\mathbf{y}$ and target token $y_t$:
\begin{equation}
    \mathbb{T} =
    \{
    (\mathcal{D}_{(\ell^1, \ell^\prime)}^i, \mathcal{D}_{(\ell^2, \ell^\prime)}^j) 
    \mid
    i,j \textrm{ map to } y_t \in (\mathbf{y} \in \mathcal{B}_{(L_{12}, \ell^\prime)}
    )\},
\end{equation}
where $\mathcal{B}_{(L_{12}, \ell^\prime)} = \{\mathcal{B}_{(\ell^1, \ell^\prime)}, \mathcal{B}_{(\ell^2, \ell^\prime)}\}$ is the combined data from $\ell^1$ and $\ell^2$ into $\ell^\prime$.

Given these paired representation vectors, we learn a linear transformation matrix $A$ that maps representations from language $\ell^1$ to language $\ell^2$.
We formulate this as a least squares problem, minimizing the L2 norm of the differences:
\begin{equation}
    \min_{A}\sum^n_{i=1}||\mathbb{T}_{\ell^2}^i - A\mathbb{T}_{\ell^1}^i||_2^2,
\end{equation}
where $\mathbb{T}_{\ell^1}^i$ and $\mathbb{T}_{\ell^2}^i$ are the representation vectors originating from tuples of $\mathbb{T}$, and $||v||_2^2$ represents the squared Euclidean norm of vector $v$.
This optimization problem can be solved efficiently using the normal equation approach, yielding a closed-form solution.

We then use $A$ to map a translation context from $\ell^1$ to $\ell^2$.
This allows queries from language $\ell^1$ to more effectively search through a datastore built from language $\ell^2$.
We also learn the inverse relation, i.e., $\ell^2$ to $\ell^1$, and create an optimized multilingual datastore for $\ell^1$ by applying the $\ell^2$ to $\ell^1$ mapping prior to storing the datastore.

\subsection{Experimental setup}
\subsubsection{Model}
The 418M parameter version of the M2M100 multilingual translation model \citep{fan_beyond_2021} is used for all experiments.
It is a transformer \citep{vaswani_attention_2017} with 24 layers, 16 heads and hidden dimensionality of 1024 supporting 100 languages.

\subsubsection{Data}
We conduct our experiments on the widely used TED Talks corpus \citep{qi_when_2018}.
We use the train set to create datastores, the development set for tuning \textit{k}NN-MT hyperparameters, and the test set to report results. We use 51 languages into English, from 23 different language families, that are supported by both TED and M2M100.

We follow the language groupings defined in the M2M100 model \citep{fan_beyond_2021}, which clusters languages based on a combination of linguistic similarity and shared script.
These groupings are designed to facilitate better parameter sharing during multilingual model training. We list information about all 51 bilingual datastores in Table~\ref{ch4:tabknn_all}.
Note that often, language family and language grouping are consistent, i.e., all languages in a language grouping are from the same language family. However, there are some exceptions such as the language grouping Greek, which includes languages from the Hellenic, Kartvelian, Armenian, and Albanian families.
This discrepancy between language groupings and language families illustrates how M2M100's grouping criteria sometimes prioritize practical considerations such as script similarity over strict phylogenetic relationships.

\begin{table*}[!htb]
\scriptsize
    \begin{center}
    \begin{tabular}{llllrrrr}
    \toprule
    $\mathcal{D}$ & $|\mathcal{D}|$ & \textbf{family/grouping} & \textbf{script} & \textbf{b} & \textbf{M2M} & \textbf{+\textit{k}NN} & $\Delta$\\\midrule
    \textbf{kk-en}  & $84\mathrm{K}$    & Turkic/Turkic & Cyrillic &          & $2.1$   &   $2.6$   & $0.5$\\
    \textbf{be-en}  & $116\mathrm{K}$   & Slavic/Slavic & Cyrillic &          & $19.2$  &   $20.9$  & $1.7$\\
    \textbf{bn-en}  & $127\mathrm{K}$   & Indo-Aryan/Indo & Eastern-Nagari & \checkmark   & $9.3$   &   $13.9$  &	$4.6$\\
    \textbf{ms-en}  & $132\mathrm{K}$   & Malayo-Polyn./Malayo & Latin &             & $28.8$  &   $30.6$  & $1.8$\\
    \textbf{bs-en}  & $146\mathrm{K}$   & Slavic/Slavic & Latin &             & $31.5$  &   $33.1$  &   $1.6$\\
    \textbf{az-en}  & $153\mathrm{K}$   & Turkic/Turkic & Cyrillic &          & $8.8$   &   $10.0$  &   $1.2$\\
    \textbf{ta-en}  & $156\mathrm{K}$   & Dravidian/Indo & Tamil & \checkmark  & $0.4$   &   $0.9$   &   $0.5$\\
    \textbf{ur-en}  & $158\mathrm{K}$   & Indo-Aryan/Indo & Arabic &            & $14.4$  &   $16.9$  &	$2.5$\\
    \textbf{mn-en}  & $181\mathrm{K}$   & Mongolic/Mongolic & Cyrillic &          & $5.1$   &   $7.1$   &	$2.0$\\
    \textbf{mr-en}  & $241\mathrm{K}$   & Indo-Aryan/Indo & Devanagari &        & $3.9$   &   $6.2$   &	$2.3$\\
    \textbf{gl-en}  & $254\mathrm{K}$   & Romance/Romance & Latin &             & $32.4$  &   $34.0$    &	$1.6$\\
    \textbf{et-en}  & $280\mathrm{K}$   & Uralic/Uralic & Latin &             & $23.5$  &   $25.3$  &	$1.8$\\
    \textbf{ka-en}  & $332\mathrm{K}$   & Kartvelian/Greek & Georgian &          & $10.8$  &   $14.7$  &	$3.9$\\
    \textbf{no-en}  & $411\mathrm{K}$   & Germanic/Germanic & Latin &             & $42.8$  &   $45.6$	&   $2.8$\\
    \textbf{hi-en}  & $481\mathrm{K}$   & Indo-Aryan/Indo & Devanagari & \checkmark       & $17.9$  &   $23.3$  &	$5.4$\\
    \textbf{sl-en}  & $520\mathrm{K}$   & Slavic/Slavic & Latin &             & $24.9$  &   $27.3$  &	$2.4$\\
    \textbf{hy-en}  & $544\mathrm{K}$   & Armeian/Greek & Armenian &          & $16.8$  &   $20.1$  &	$3.3$\\
    \textbf{my-en}  & $558\mathrm{K}$   & Sino-Tibetan/Mongolic & Burmese &           & $0.4$   &   $1.5$   &	$1.1$\\
    \textbf{fi-en}  & $623\mathrm{K}$   & Uralic/Uralic & Latin & \checkmark  & $21.0$    &   $22.8$  &	$1.8$\\
    \textbf{mk-en}  & $683\mathrm{K}$   & Slavic/Slavic & Cyrillic &          & $29.3$  &   $32.8$  &	$3.5$\\
    \textbf{lt-en}  & $1.1\mathrm{M}$   & Baltic/Uralic & Latin & \checkmark & $24.7$  &   $28.2$  &	$3.5$\\
    \textbf{sq-en}  & $1.2\mathrm{M}$   & Albanian/Greek & Latin &             & $31.9$  &   $35.8$  &	$3.9$\\
    \textbf{da-en}  & $1.2\mathrm{M}$   & Germanic/Germanic & Latin &             & $40.0$    &   $44.5$  &	$4.5$\\
    \textbf{pt-en}  & $1.2\mathrm{M}$   & Romance/Romance & Latin &  \checkmark & $38.9$  &   $42.0$	&   $3.1$\\
    \textbf{sv-en}  & $1.4\mathrm{M}$   & Germanic/Germanic & Latin & \checkmark  & $37.3$  &   $41.0$	&   $3.7$\\
    \textbf{sk-en}  & $1.6\mathrm{M}$   & Slavic/Slavic & Latin &             & $28.4$  &   $32.6$	&   $4.2$\\
    \textbf{id-en}  & $2.3\mathrm{M}$   & Malayo-Polyn./Malayo & Latin & \checkmark            & $29.1$  &   $32.5$	&   $3.4$\\
    \textbf{th-en}  & $2.6\mathrm{M}$   & Kra-Dai/Mongolic & Thai &              & $2.3$   &   $8.5$	&   $6.2$\\
    \textbf{cs-en}  & $2.7\mathrm{M}$   & Slavic/Slavic & Latin &             & $27.5$  &   $31.4$	&   $3.9$\\
    \textbf{uk-en}  & $2.9\mathrm{M}$   & Slavic/Slavic & Cyrillic &          & $24.7$  &   $29.1$	&   $4.4$\\
    \textbf{hr-en}  & $3.3\mathrm{M}$   & Slavic/Slavic & Latin &             & $32.2$  &   $37.0$	&   $4.8$\\
    \textbf{el-en}  & $3.5\mathrm{M}$   & Hellenic/Greek & Greek & \checkmark            & $32.6$  &   $38.3$	&   $5.7$\\
    \textbf{sr-en}  & $3.6\mathrm{M}$   & Slavic/Slavic & Cyrillic &          & $30.7$  &   $35.9$	&   $5.2$\\
    \textbf{hu-en}  & $3.9\mathrm{M}$   & Uralic/Uralic & Latin & \checkmark & $23.3$  &   $26.8$	&   $3.5$\\
    \textbf{fa-en}  & $4.0\mathrm{M}$   & Iranian/Arabic & Arabic &  \checkmark          & $22.7$  &   $27.6$	&   $4.9$\\
    \textbf{de-en}  & $4.5\mathrm{M}$   & Germanic/Germanic & Latin & \checkmark             & $31.7$  &   $36.9$	&   $5.2$\\
    \textbf{vi-en}  & $4.6\mathrm{M}$   & Vietic/Chinese & Latin &  \checkmark           & $23.7$  &   $27.2$	&   $3.5$\\
    \textbf{bg-en}  & $4.7\mathrm{M}$   & Slavic/Slavic & Cyrillic &          & $34.4$  &   $39.5$	&   $5.1$\\
    \textbf{pl-en}  & $4.7\mathrm{M}$   & Slavic/Slavic & Latin & \checkmark  & $21.1$  &   $25.0$	&   $3.9$\\
    \textbf{ro-en}  & $4.8\mathrm{M}$   & Romance/Romance & Latin &             & $30.6$  &   $35.4$	&   $4.8$\\
    \textbf{nl-en}  & $4.9\mathrm{M}$   & Germanic/Germanic & Latin &  \checkmark  & $31.9$  &   $36.2$	&   $4.3$\\
    \textbf{tr-en}  & $4.9\mathrm{M}$   & Turkic/Turkic & Latin & \checkmark          & $22.4$  &   $26.5$	&   $4.1$\\
    \textbf{fr-en}  & $5.1\mathrm{M}$   & Romance/Romance & Latin & \checkmark & $35.1$  &   $40.3$	&   $5.2$\\
    \textbf{es-en}  & $5.2\mathrm{M}$   & Romance/Romance & Latin &   \checkmark   & $36.6$  &   $41.9$	&   $5.3$\\
    \textbf{zh-en}  & $5.4\mathrm{M}$   & Chinese/Chinese & Chinese & \checkmark           & $16.3$  &   $20.2$	&   $3.9$\\
    \textbf{ja-en}  & $5.5\mathrm{M}$   & Japonic/Chinese & Kanji &  \checkmark  & $10.6$  &   $13.5$	&   $2.9$\\
    \textbf{it-en}  & $5.5\mathrm{M}$   & Romance/Romance & Latin &             & $33.4$  &   $38.4$	&   $5.0$\\
    \textbf{ko-en}  & $5.5\mathrm{M}$   & Koreanic/Chinese & Hangul & \checkmark & $16.2$  &   $19.5$	&   $3.3$\\
    \textbf{ru-en}  & $5.6\mathrm{M}$   & Slavic/Slavic & Cyrillic & \checkmark & $21.6$  &   $25.8$	&   $4.2$\\
    \textbf{he-en}  & $5.7\mathrm{M}$   & Semitic/Arabic & Hebrew  & \checkmark         & $31.2$  &   $36.8$	&   $5.6$\\
    \textbf{ar-en}  & $5.8\mathrm{M}$   & Arabic/Arabic & Arabic &   \checkmark         & $26.2$  &   $31.4$	&   $5.2$\\\midrule
    \textbf{avg}& $2.5\mathrm{M}$   & & & & $23.4$ & $27.0$ & $3.6$\\
    \textbf{total}  & $125\mathrm{M}$ & \\\bottomrule
        \end{tabular}
        \end{center}
    \caption{Datastore information for all 51 languages into English that we use. Column family/grouping shows language family and grouping from M2M100. Column bridge (b) indicates whether the language is a bridge language in M2M100. The three final columns show BLEU scores for the base model (M2M), augmented with \textit{k}NN-MT (+\textit{k}NN), and their difference ($\Delta$).}
    \label{ch4:tabknn_all}
\end{table*}

\subsubsection{Evaluation and \textit{k}NN settings}
We use \textit{k}NN-BOX \citep{zhu_knn-box_2024} for our experiments, and FAISS \citep{douze_faiss_2025,johnson_billion-scale_2019} for efficient approximate similarity search.
Following \citet{khandelwal_nearest_2021}, we tune the number of neighbors $k\in\{16,32,64\}$, interpolation $\lambda\in\{0.2, 0.3, ..., 0.7\}$ and softmax temperature $T\in\{10,100\}$ hyperparameters on the development set.
We use a beam size of 5. We evaluate our models using sacreBLEU \citep{post_call_2018,papineni_bleu_2002}.\footnote{\scriptsize{nrefs:1$|$case:mixed$|$eff:no$|$tok:13a$|$smooth:exp$|$version:2.3.1}}

\subsubsection{Cross-lingual and multilingual datastores}
We construct datastores for languages from three M2M100 language groupings into English: Slavic (12 languages), Germanic (5 languages), and Greek (4 languages).
We then generate translations for all possible combinations of source language and datastore, with the goal of investigating the potential of cross-lingual datastores to improve low-resource performance.

Additionally, we construct several multilingual datastores:

\begin{itemize}
    \item $\mathcal{D}_{(\textrm{ALL}, \textrm{en})}$: We created a comprehensive datastore, integrating 51 languages that occur in both TED and M2M100 into English, resulting in $125\textrm{M}$ entries of key-value pairs.

    \item $\mathcal{D}_{(\textrm{BR}, \textrm{en})}$: mNMT is typically English-centric, i.e., English occurs on the source or target side in the training data.
    M2M100 instead uses a set of \textit{bridge languages}, which leads to a greater coverage of direct translation directions.
    To align with these languages, we additionally create a smaller datastore of size $86\textrm{M}$, consisting of 24 bridge languages.

    \item $\mathcal{D}_{(\textrm{LG}, \textrm{en})}$: We investigate to what extent datastore size can be further decreased.
    We hypothesize that more similar multilingual representations result in better cross-lingual retrieval, and that representations within the same language grouping are more similar.
    In line with this, we create three multilingual datastores consisting of all languages within a language grouping: Slavic (12 source languages, datastore size $31\textrm{M}$), Germanic (5 source languages, datastore size $20\textrm{M}$), and Greek (4 source languages, datastore size $5.6\textrm{M}$).\\
\end{itemize}

\begin{table}[!htb]
    \centering
    \setlength\dashlinedash{2.0pt}
    \setlength\dashlinegap{1.5pt}
    \setlength\arrayrulewidth{0.3pt}
    \definecolor{gray}{RGB}{217,217,217}
    \sethlcolor{gray}
    \rotatebox{90}{
        \footnotesize
        \begin{tabular}{
            >{\centering\arraybackslash}m{0.0294\textwidth}|
            >{\centering\arraybackslash}m{0.0294\textwidth}|
            >{\centering\arraybackslash}m{0.0294\textwidth}
            >{\centering\arraybackslash}m{0.0294\textwidth}
            >{\centering\arraybackslash}m{0.0294\textwidth}
            >{\centering\arraybackslash}m{0.0294\textwidth}
            >{\centering\arraybackslash}m{0.0294\textwidth}
            >{\centering\arraybackslash}m{0.0294\textwidth}
            >{\centering\arraybackslash}m{0.0294\textwidth}
            >{\centering\arraybackslash}m{0.0294\textwidth}
            >{\centering\arraybackslash}m{0.0294\textwidth}
            >{\centering\arraybackslash}m{0.0294\textwidth}
            >{\centering\arraybackslash}m{0.0294\textwidth}
            >{\centering\arraybackslash}m{0.0294\textwidth}|
            >{\centering\arraybackslash}m{0.0294\textwidth}
            >{\centering\arraybackslash}m{0.0294\textwidth}
            >{\centering\arraybackslash}m{0.0294\textwidth}}
                
            $\mathcal{D}$ &
            base &
            $\mathcal{D}_{\textrm{be}}$ &
            $\mathcal{D}_{\textrm{bs}}$ &
            $\mathcal{D}_{\textrm{sl}}$ &
            $\mathcal{D}_{\textrm{mk}}$ &
            $\mathcal{D}_{\textrm{sk}}$ &
            $\mathcal{D}_{\textrm{cs}}$ &
            $\mathcal{D}_{\textrm{uk}}$ &
            $\mathcal{D}_{\textrm{hr}}$ &
            $\mathcal{D}_{\textrm{sr}}$ &
            $\mathcal{D}_{\textrm{bg}}$ &
            $\mathcal{D}_{\textrm{pl}}$ &
            $\mathcal{D}_{\textrm{ru}}$ &
            $\mathcal{D}_{\textrm{LG}}$ &
            $\mathcal{D}_{\textrm{BR}}$ &
            $\mathcal{D}_{\textrm{ALL}}$ \\
                    
            $|\mathcal{D}|$ &
            $0$ &
            $116\mathrm{K}$ &
            $146\mathrm{K}$ &
            $520\mathrm{K}$ &
            $683\mathrm{K}$ &
            $1.6\mathrm{M}$ &
            $2.7\mathrm{M}$ &
            $2.9\mathrm{M}$ &
            $3.3\mathrm{M}$ &
            $3.6\mathrm{M}$ &
            $4.7\mathrm{M}$ &
            $4.7\mathrm{M}$ &
            $5.6\mathrm{M}$ &
            $31\mathrm{M}$ &
            $86\mathrm{M}$ &
            $125\mathrm{M}$\\\hline 
            
            \textbf{be} &
            $19.2$ &
            \cellcolor[rgb]{0.85, 0.85, 0.85} $20.9$ &
            $20.5$ &
            $20.9$ &
            $20.8$ &
            $21.0$ &
            $21.5$ &
            $22.4$ &
            $21.4$ &
            $21.3$ &
            $21.3$ &
            $21.3$ &
            $\underline{21.7}$ &
            $\mathbf{23.1}$ &
            $22.2$ &
            $22.5$ \\
            
            \textbf{bs}&
            $31.5$ &
            $33.2$ &
            \cellcolor[rgb]{0.85, 0.85, 0.85} $33.1$ &
            $34.0$ &
            $34.0$ &
            $34.4$ &
            $34.6$ &
            $34.7$ &
            $\underline{36.0}$ &
            $35.8$ &
            $35.2$ &
            $34.6$ &
            $35.0$ &
            $\mathbf{36.7}$ &
            $36.0$ &
            $36.2$ \\
            
            \textbf{sl}&
            $24.9$ &
            $26.4$ &
            $26.4$ &
            \cellcolor[rgb]{0.85, 0.85, 0.85} $27.3$ &
            $27.0$ &
            $27.6$ &
            $27.9$ &
            $27.9$ &
            $28.3$ &
            $\underline{28.4}$ &
            $27.9$ &
            $28.1$ &
            $28.2$ &
            $29.2$ &
            $29.2$ &
            $\mathbf{29.5}$ \\
            
            \textbf{mk} &
            $29.3$ &
            $32.0$ &
            $32.1$ &
            $32.5$ &
            \cellcolor[rgb]{0.85, 0.85, 0.85} $32.8$ &
            $33.2$ &
            $33.8$ &
            $33.3$ &
            $\underline{34.8}$ &
            $33.9$ &
            $34.1$ &
            $33.4$ &
            $33.4$ &
            $35.5$ &
            $35.5$ &
            $\mathbf{35.6}$ \\\hdashline
            
            \textbf{sk}&
            $28.4$ &
            $30.3$ &
            $30.5$ &
            $31.5$ &
            $31.4$ &
            \cellcolor[rgb]{0.85, 0.85, 0.85} $32.6$ &
            $\underline{33.0}$ &
            $32.1$ &
            $32.2$ &
            $32.4    $ &
            $32.8$ &
            $32.4$ &
            $32.5$ &
            $\mathbf{34.1}$ &
            $33.6$ &
            $\mathbf{34.1}$ \\
            
            \textbf{cs}&
            $27.5$ &
            $29.3$ &
            $29.4$ &
            $30.0$ &
            $30.0$ &
            $30.9$ &
            \cellcolor[rgb]{0.85, 0.85, 0.85} $\underline{31.4}$ &
            $30.7$ &
            $30.8$ &
            $30.9$ &
            $31.2$ &
            $30.8$ &
            $31.0$ &
            $32.0$ &
            $31.9$ &
            $\mathbf{32.1}$ \\
            
            \textbf{uk}&
            $24.7$ &
            $26.6$ &
            $27.0$ &
            $27.4$ &
            $27.6$ &
            $27.9$ &
            $28.3$ &
            \cellcolor[rgb]{0.85, 0.85, 0.85} $\underline{29.1}$ &
            $28.5$ &
            $28.3$ &
            $28.8$ &
            $28.3$ &
            $28.9$ &
            $\mathbf{29.9}$ &
            $29.6$ &
            $29.7$ \\
            
            \textbf{hr}&
            $32.2$ &
            $33.8$ &
            $34.4$ &
            $34.8$ &
            $34.9$ &
            $35.3$ &
            $35.6$ &
            $35.5$ &
            \cellcolor[rgb]{0.85, 0.85, 0.85} $\underline{37.0}$ &
            $36.6$ &
            $36.0$ &
            $35.5$ &
            $35.7$ &
            $37.5$ &
            $37.1$ &
            $\mathbf{37.8}$ \\
            
            \textbf{sr}&
            $30.7$ &
            $32.2$ &
            $32.7$ &
            $33.3$ &
            $33.6$ &
            $33.8$ &
            $34.2$ &
            $34.0$ &
            $35.2$ &
            \cellcolor[rgb]{0.85, 0.85, 0.85} $\underline{35.9}$ &
            $34.8$ &
            $34.3$ &
            $34.6$ &
            $36.3$ &
            $35.7$ &
            $\mathbf{36.5}$ \\
            
            \textbf{bg}&
            $34.4$ &
            $36.1$ &
            $36.2$ &
            $37.1$ &
            $37.2$ &
            $37.4$ &
            $37.9$ &
            $37.6$ &
            $38.1$ &
            $38.2$ &
            \cellcolor[rgb]{0.85, 0.85, 0.85} $\underline{39.5}$ &
            $38.0$ &
            $38.3$ &
            $39.7$ &
            $39.5$ &
            $\mathbf{39.9}$ \\
            
            \textbf{pl}&
            $21.1$ &
            $22.6$ &
            $22.8$ &
            $23.4$ &
            $23.5$ &
            $23.8$ &
            $24.1$ &
            $23.9$ &
            $24.0$ &
            $24.1$ &
            $24.5$ &
            \cellcolor[rgb]{0.85, 0.85, 0.85} $\underline{25.0}$ &
            $24.3$ &
            $\mathbf{25.4}$ &
            $\mathbf{25.4}$ &
            $\mathbf{25.4}$ \\
            
            \textbf{ru}&
            $21.6$ &
            $23.3$ &
            $23.3$ &
            $23.8$ &
            $24.1$ &
            $24.3$ &
            $24.7$ &
            $24.9$ &
            $24.7$ &
            $24.8$ &
            $25.0$ &
            $24.9$ &
            \cellcolor[rgb]{0.85, 0.85, 0.85} $\underline{25.8}$ &
            $\mathbf{26.0}$ &
            $\mathbf{26.0}$ &
            $25.4$ \\\hline 
            
            \textbf{avg} &
            $27.1$ &
            $28.9$ &
            $29.0$ &
            $29.7$ &
            $29.7$ &
            $30.2$ &
            $30.6$ &
            $30.5$ &
            $30.9$ &
            $30.9$ &
            $30.9$ &
            $30.6$ &
            $30.8$ &
            $\mathbf{32.1}$ &
            $31.8$ &
            $\mathbf{32.1}$ \\   
        \end{tabular}
    }
    \caption{X$\rightarrow$en BLEU scores for Slavic languages.
    We display results for all combinations of Slavic translation directions and datastores.
    Languages with $|\mathcal{D}|<1\textrm{M}$ are above the dashed line.
    Bilingual datastores on the diagonal are highlighted in \hl{gray}.
    The three rightmost columns show multilingual datastores built from Slavic languages ($\mathcal{D}_{\textrm{LG}}$), bridge languages ($\mathcal{D}_{\textrm{BR}}$), or all languages ($\mathcal{D}_{\textrm{ALL}}$).
    BLEU scores without \textit{k}NN-MT are in the column labeled base.
    We \underline{underline} the best cross-lingual or bilingual results and mark overall best scores in \textbf{bold}. See Table~\ref{ch4:tabml_knn_germanic_greek} for results on Germanic and Greek-family languages.}
    \label{ch4:tabml_knn_slavic}
\end{table}

\begin{table}[!htb]
    \centering
    \setlength\dashlinedash{2.0pt}
    \setlength\dashlinegap{1.5pt}
    \setlength\arrayrulewidth{0.3pt}
    \definecolor{gray}{RGB}{217,217,217}
    \sethlcolor{gray}
    \footnotesize
    
    % Germanic Languages
    \begin{subtable}{\textwidth}
        \centering
        \begin{tabular}{
            >{\centering\arraybackslash}m{0.05\textwidth}|
            >{\centering\arraybackslash}m{0.05\textwidth}|
            >{\centering\arraybackslash}m{0.05\textwidth}
            >{\centering\arraybackslash}m{0.05\textwidth}
            >{\centering\arraybackslash}m{0.05\textwidth}
            >{\centering\arraybackslash}m{0.05\textwidth}
            >{\centering\arraybackslash}m{0.05\textwidth}|
            >{\centering\arraybackslash}m{0.05\textwidth}
            >{\centering\arraybackslash}m{0.05\textwidth}
            >{\centering\arraybackslash}m{0.05\textwidth}}
        
            $\mathcal{D}$ &
            base &
            $\mathcal{D}_{\textrm{no}}$ &
            $\mathcal{D}_{\textrm{da}}$ &
            $\mathcal{D}_{\textrm{sv}}$ &
            $\mathcal{D}_{\textrm{de}}$ &
            $\mathcal{D}_{\textrm{nl}}$ &
            $\mathcal{D}_{\textrm{LG}}$ &
            $\mathcal{D}_{\textrm{BR}}$ &
            $\mathcal{D}_{\textrm{ALL}}$ \\
                    
            $|\mathcal{D}|$ &
            $0$ &
            $411\mathrm{K}$ &
            $1.2\mathrm{M}$ &
            $1.4\mathrm{M}$ &
            $4.5\mathrm{M}$ &
            $4.9\mathrm{M}$ &
            $12\mathrm{M}$ &
            $86\mathrm{M}$ &
            $125\mathrm{M}$\\\hline 
        
            \textbf{no} &
            $42.8$ &
            \cellcolor[rgb]{0.85, 0.85, 0.85} $45.6$ &
            $\underline{46.7}$ &
            $45.9$ &
            $46.4$ &
            $46.3$ &
            $47.7$ &
            $47.4$ &
            $\mathbf{47.8}$ \\\hdashline
            
            \textbf{da} &
            $40.0$ &
            $43.1$ &
            \cellcolor[rgb]{0.85, 0.85, 0.85} $\underline{44.5}$ &
            $43.3$ &
            $44.0$ &
            $44.0$ &
            $45.5$ &
            $45.0$ &
            $\mathbf{45.7}$ \\
            
            \textbf{sv} &
            $37.3$ &
            $39.6$ &
            $40.4$ &
            \cellcolor[rgb]{0.85, 0.85, 0.85} $\underline{41.0}$ &
            $40.8$ &
            $40.8$ &
            $41.8$ &
            $\mathbf{42.1}$ &
            $42.0$ \\
            
            \textbf{de} &
            $31.7$ &
            $34.3$ &
            $34.9$ &
            $35.0$ &
            \cellcolor[rgb]{0.85, 0.85, 0.85} $\underline{36.9}$ &
            $36.0$ &
            $37.1$ &
            $37.2$ &
            $\mathbf{37.3}$ \\
            
            \textbf{nl} &
            $31.9$ &
            $33.9$ &
            $34.4$ &
            $34.6$ &
            $35.2$ &
            \cellcolor[rgb]{0.85, 0.85, 0.85} $\underline{36.2}$ &
            $36.1$ &
            $36.0$ &
            $\mathbf{36.3}$ \\\hline
        
            \textbf{avg} &
            $36.7$ &
            $39.3$ &
            $40.2$ &
            $40.0$ &
            $40.7$ &
            $40.7$ &
            $41.7$ &
            $41.5$ &
            $\mathbf{41.8}$ \\
        \end{tabular}
        \caption*{(a) Germanic languages}
    \end{subtable}
    
    \vspace{1em}
    
    % Greek Languages
    \begin{subtable}{\textwidth}
        \centering
        \begin{tabular}{
            >{\centering\arraybackslash}m{0.05\textwidth}|
            >{\centering\arraybackslash}m{0.05\textwidth}|
            >{\centering\arraybackslash}m{0.05\textwidth}
            >{\centering\arraybackslash}m{0.05\textwidth}
            >{\centering\arraybackslash}m{0.05\textwidth}
            >{\centering\arraybackslash}m{0.05\textwidth}|
            >{\centering\arraybackslash}m{0.05\textwidth}
            >{\centering\arraybackslash}m{0.05\textwidth}
            >{\centering\arraybackslash}m{0.05\textwidth}}
        
            $\mathcal{D}$ &
            base &
            $\mathcal{D}_{\textrm{ka}}$ &
            $\mathcal{D}_{\textrm{hy}}$ &
            $\mathcal{D}_{\textrm{sq}}$ &
            $\mathcal{D}_{\textrm{el}}$ &
            $\mathcal{D}_{\textrm{LG}}$ &
            $\mathcal{D}_{\textrm{BR}}$ &
            $\mathcal{D}_{\textrm{ALL}}$ \\
        
            $|\mathcal{D}|$ &
            $0$ &
            $332\mathrm{K}$ &
            $544\mathrm{K}$ &
            $1.2\mathrm{M}$ &
            $3.5\mathrm{M}$ &
            $5.6\mathrm{M}$ &
            $86\mathrm{M}$ &
            $125\mathrm{M}$ \\\hline 
        
            \textbf{ka} &
            $10.8$ &
            \cellcolor[rgb]{0.85, 0.85, 0.85} $\underline{14.7}$ &
            $12.7$ &
            $12.8$ &
            $12.9$ &
            $\mathbf{15.2}$ &
            $14.4$ &
            $\mathbf{15.2}$ \\
            
            \textbf{hy} &
            $16.8$ &
            $18.4$ &
            \cellcolor[rgb]{0.85, 0.85, 0.85} $\underline{20.1}$ &
            $19.0$ &
            $19.4$ &
            $\mathbf{20.8}$ &
            $20.3$ &
            $20.7$ \\\hdashline
        
            \textbf{sq} &
            $31.9$ &
            $33.2$ &
            $33.4$ &
            \cellcolor[rgb]{0.85, 0.85, 0.85} $\underline{35.8}$ &
            $34.6$ &
            $36.0$ &
            $35.5$ &
            $\mathbf{36.2}$ \\
        
            \textbf{el} &
            $32.6$ &
            $34.8$ &
            $35.3$ &
            $35.8$ &
            \cellcolor[rgb]{0.85, 0.85, 0.85} $\underline{38.3}$ &
            $38.3$ &
            $38.7$ &
            $\mathbf{38.8}$ \\\hline
    
            \textbf{avg} &
            $23.0$ &
            $25.3$ &
            $25.4$ &
            $25.9$ &
            $26.3$ &
            $27.6$ &
            $27.2$ &
            $\mathbf{27.7}$ \\
        \end{tabular}
        \caption*{(b) Greek-family languages}
    \end{subtable}
    
    \caption{X$\rightarrow$en BLEU scores for Germanic and Greek language families.
    We display results for all combinations of translation directions and datastores within each language family.
    For brevity, we write, e.g., the Norwegian-English datastores as $\mathcal{D}_{\textrm{no}}$ instead of $\mathcal{D}_{(\textrm{no}, \textrm{en})}$.
    Languages with $|\mathcal{D}|<1\textrm{M}$ are above the dashed line in each subtable.
    Bilingual datastores on the diagonal are highlighted in \hl{gray}.
    The rightmost columns show multilingual datastores built from languages in the same family ($\mathcal{D}_{\textrm{LG}}$), bridge languages ($\mathcal{D}_{\textrm{BR}}$), or all languages ($\mathcal{D}_{\textrm{ALL}}$).
    BLEU scores without \textit{k}NN-MT are in the column labeled base.
    We \underline{underline} the best cross-lingual or bilingual results and mark overall best scores in \textbf{bold}.
    See Table~\ref{ch4:tabml_knn_slavic} for results on Slavic languages.}
    \label{ch4:tabml_knn_germanic_greek}
\end{table}

\subsection{Results}
Translation results for bilingual, cross-lingual, and multilingual datastores are shown in Tables~\ref{ch4:tabml_knn_slavic} (Slavic languages) and \ref{ch4:tabml_knn_germanic_greek} (Germanic and Greek-family languages).
Below, we analyze the performance of different datastore configurations, paying particular attention to their impact on low-resource languages.

\subsubsection{Bilingual datastores work to a limited extent}
We observe that bilingual datastores provide consistent improvements across all languages, though the magnitude varies significantly by resource level.

For example, low-resource Slavic languages (Belarusian, Bosnian, Slovenian, and Macedonian) using their respective bilingual datastores achieve modest gains of $+2.3$ BLEU on average, with individual improvements ranging from $+1.7$ to $+3.5$ BLEU.
In contrast, high-resource Slavic languages (Slovak, Czech, Ukrainian, Croatian, Serbian, Bulgarian, Polish, and Russian) achieve more substantial improvements, with an average gain of $+4.5$ BLEU when using their respective bilingual datastores.
The most dramatic improvement is observed for Bulgarian-English (bg-en), which gains $+5.1$ BLEU over the baseline.

This performance gap highlights the inherent limitations of bilingual datastores for low-resource languages, where the smaller datastore size ($<1$M entries) constrains the potential benefits.

\subsubsection{Cross-lingual outperforms bilingual for low-resource languages}
Our results demonstrate that low-resource languages benefit significantly from cross-lingual datastores.
For example, Belarusian-English (be-en) translation with its own bilingual datastore ($116\mathrm{K}$ instances) improves by $+1.7$ BLEU over the baseline, while using the Ukrainian-English (uk-en) cross-lingual datastore ($2.9\mathrm{M}$ instances) yields a further improvement of $+1.5$ BLEU, reaching $22.4$ BLEU in total.

While datastore size and performance are strongly correlated ($\rho=0.88$ with $p<0.001$ for the Slavic language grouping), the additional quality improvements can \textit{not} be fully explained by the size increase.
For instance, Bosnian-English (bs-en) augmented with a cross-lingual Croatian-English (hr-en) datastore, which is only 60\% of the size of Russian-English (ru-en), leads to $+1.0$ BLEU compared to using Russian-English. 
We conclude that it is difficult to predict which cross-lingual datastore will perform best.

In contrast, for high-resource languages it is \textit{always} a better choice to use the bilingual datastore, even when larger cross-lingual datastores are available.
For instance, Serbian-English (sr-en) with Serbian-English datastore of $3.6\textrm{M}$ key-value pairs performs better ($35.9$ BLEU) than Serbian-English with the substantially larger cross-lingual datastore Russian-English ($5.6\textrm{M}$ entries, $34.6$ BLEU). 

When considering cross-lingual datastores that come from a more distant language family, using a bilingual datastore leads to better results, even for low-resource languages.
Considering Georgian-English (ka-en), the bilingual datastore improvement ($+3.9$ BLEU) is larger than the improvement for its best cross-lingual datastore Greek-English (el-en, $+2.1$ BLEU).
This is likely because the Greek language grouping combines different families: Greek (el) is a Hellenic language, whereas Georgian (ka) is from the Kartvelian language family.
Therefore, their representations likely have larger differences than these of Belarusian-English (be-en) and Ukrainian-English (uk-en), which come from the same family.

\subsubsection{Multilingual datastores perform best}
Since it is unclear which cross-lingual datastore performs best, we use as many languages as possible as a first attempt.
This results in our largest datastore $\mathcal{D}_{(\textrm{ALL}, \textrm{en})}$, which has $125\textrm{M}$ entries.
For almost all languages, except Russian-English (ru-en), this leads to better results than bilingual datastores.
Low-resource languages show the largest improvements, where Bosnian-English (bs-en) has the largest improvement of $+3.6$ BLEU compared to Bosnian-English datastore, or $+0.7$ BLEU compared to the best cross-lingual datastore.
A problem for $\mathcal{D}_{(\textrm{ALL}, \textrm{en})}$ is slow inference speed, because \textit{k}NN lookup in a large datastore is expensive.

When decreasing the datastore size by focusing on bridge languages, we can construct a smaller datastore of size $86\textrm{M}$, but its results are worse in almost all cases which is clearly reflected in the average scores.

Finally, we consider a datastore that is constructed using linguistic similarity.
It consists of languages from the same language grouping.
We observe that this multilingual datastore is on par with the largest one, while significantly smaller, which is clearly reflected in the averages.
For some languages, such as Belarusian-English (be-en) and Bosnian-English (bs-en), this even brings improvements of $+0.6$ BLEU and $+0.5$ BLEU compared to using $\mathcal{D}_{(\textrm{ALL}, \textrm{en})}$.
We emphasize that multilingual datastores lead to best results for \textit{all} languages we tested, including higher-resource directions such as Polish-English, (pl-en, $+0.4$ BLEU compared to bilingual) and Ukrainian-English (uk-en, $+0.8$ BLEU).

\subsubsection{Effectiveness of cross-lingual mapping}
We created a cross-lingual mapping from Belarusian (be) to other languages in the Slavic language grouping.
We also created the inverse mapping, and constructed a Slavic language grouping datastore mapped to Belarusian representations.

Results are presented in Table \ref{ch4:tabxlingmap}. We observe that generally, Belarusian-English (be-en) performance is improved, especially for larger cross-lingual datastores such as Polish-English (pl-en, $+0.8$ BLEU).
For Bosnian-English (bs-en) and Ukrainian-English (uk-en) the mapping results in a slight quality decrease.
This can possibly be explained by the small data size (for Bosnian-English), and because Ukrainian-English and Belarusian-English are already relatively well aligned, since Ukrainian-English is the best cross-lingual datastore for Belarusian-English (see Table \ref{ch4:tabknn_all}).

\begin{table}[!htb]
\small
    \begin{center}
        \begin{tabular}{llrrr}
        \toprule
        $\mathcal{D}$ &
        $|\mathbb{T}|$ &
        $\mathbb{T}_{\textrm{be}}$ &
        $A\mathbb{T}_{\textrm{be}}$ &
        $\Delta$ BLEU \\\midrule
        
        $\mathcal{D}_{(\textrm{bs}, \textrm{en})}$ &
        $23\mathrm{K}$ &
        $\mathbf{20.5}$ &
        $20.4$ & $-0.1$ \\
        
        $\mathcal{D}_{(\textrm{sl}, \textrm{en})}$ &
        $95\mathrm{K}$ &
        $20.9$ &
        $\mathbf{21.3}$ &
        $0.4$ \\
        
        $\mathcal{D}_{(\textrm{mk}, \textrm{en})}$ &
        $73\mathrm{K}$ &
        $20.8$ &
        $\mathbf{21.2}$ &
        $0.4$ \\
        
        $\mathcal{D}_{(\textrm{sk}, \textrm{en})}$ &
        $202\mathrm{K}$ &
        $21.9$ &
        $\mathbf{21.3}$ &
        $0.3$ \\

        $\mathcal{D}_{(\textrm{cs}, \textrm{en})}$ &
        $305\mathrm{K}$ &
        $21.5$ &
        $\mathbf{21.7}$ &
        $0.2$ \\
        
        $\mathcal{D}_{(\textrm{uk}, \textrm{en})}$ &
        $347\mathrm{K}$ &
        $\mathbf{22.4}$ &
        $22.2$ &
        $-0.2$ \\
        
        $\mathcal{D}_{(\textrm{hr}, \textrm{en})}$ &
        $359\mathrm{K}$ &
        $21.4$ &
        $\mathbf{21.7}$ &
        $0.3$ \\
        
        $\mathcal{D}_{(\textrm{sr}, \textrm{en})}$ &
        $417\mathrm{K}$&
        $21.3$ &
        $\mathbf{21.8}$ &
        $0.5$ \\
        
        $\mathcal{D}_{(\textrm{bg}, \textrm{en})}$ &
        $431\mathrm{K}$ &
        $21.3$ &
        $\mathbf{21.8}$ &
        $0.5$ \\
        
        $\mathcal{D}_{(\textrm{pl}, \textrm{en})}$ &
        $421\mathrm{K}$ &
        $21.3$ &
        $\mathbf{22.1}$ &
        $0.8$ \\
        
        $\mathcal{D}_{(\textrm{ru}, \textrm{en})}$ &
        $533\mathrm{K}$ &
        $21.7$ &
        $\mathbf{22.3}$ &
        $0.6$ \\\midrule
        
        avg &
        $291\mathrm{K}$ &
        $21.3$ &
        $\mathbf{21.6}$ &
        $0.3$ \\\midrule
        
        $\mathcal{D}_{(\textrm{LG}, \textrm{en})}$ &
        $-$ &
        $23.1$ &
        $\mathbf{23.4}$ &
        $0.3$ \\\bottomrule
                
        \end{tabular}
        \end{center}
    \caption{be$\rightarrow$en BLEU scores for Slavic grouping, with cross-lingual mapping ($A\mathbb{T}_{\textrm{be}}$) and without ($\mathbb{T}_{\textrm{be}}$). Training data size for mapping is shown as $|\mathbb{T}|$. Best results shown in \textbf{bold}.}
    \label{ch4:tabxlingmap}
\end{table}

\subsection{Analysis}
Having established the effectiveness of multilingual datastores for improving translation quality, we now analyze two key aspects of our approach: the distribution of language contributions in multilingual retrieval and the computational efficiency implications of different datastore configurations.

\subsubsection{Which languages are used?}

We explore the language origin of \textit{k}NN-MT target token suggestions when using $\mathcal{D}_{(\textrm{ALL}, \textrm{en})}$ to augment the Norwegian-English (no-en) translation direction.
Table \ref{ch4:tabprob_multilingual_all} presents the complete breakdown of individual language origins.

We observe that the distribution of observed language origins generally follows a uniform distribution based on bilingual datastore size.
The largest outlier is the Norwegian-English datastore, which is used for $6.05\%$ of the generations, as opposed to the $0.34\%$ that would be expected from a uniform distribution. 

Surprisingly, despite consisting of only 5 out of 51 languages, the Germanic language group accounts for $23.2\%$ of the suggestions.
This helps to explain why $\mathcal{D}_{(\textrm{LG}, \textrm{en})}$ performs on par with $\mathcal{D}_{(\textrm{ALL}, \textrm{en})}$, even though it is more than ten times smaller. 

Furthermore, we observe that datastores from several bridge languages, including Arabic-English (ar-en), Polish-English (pl-en), Hungarian-English (hu-en), Korean-English (ko-en) and Japanese-English (ja-en), are undersampled compared to the uniform expectation. 
This discrepancy can likely be attributed to the fact that these languages are relatively distant from Norwegian, resulting in dissimilar representations.

\begin{table}[H]
\centering
\footnotesize
\begin{tabular}{llcccc}\toprule
    $\mathcal{D}$ & group & $|\mathcal{D}|$ & $P_{\textrm{obs}}$ & $P_{\textrm{uni}}$ & $P_{\textrm{obs}} / P_{\textrm{uni}}$ \\\midrule
    \textbf{no-en} & Germanic & $411\textrm{K}$ &\cellcolor[HTML]{999999}6.05\% &\cellcolor[HTML]{fafafa}0.34\% &\cellcolor[HTML]{999999}17.79 \\
    \textbf{da-en} & Germanic & $1.2\textrm{M}$ &\cellcolor[HTML]{c2c2c2}3.64\% &\cellcolor[HTML]{efefef}0.99\% &\cellcolor[HTML]{a9a9a9}3.68 \\
    \textbf{\underline{sv-en}} & Germanic & $1.4\textrm{M}$ &\cellcolor[HTML]{cccccc}3.08\% &\cellcolor[HTML]{ececec}1.18\% &\cellcolor[HTML]{b7b7b7}2.61 \\
    \underline{pt-en} & Romance & $1.2\textrm{M}$ &\cellcolor[HTML]{e1e1e1}1.81\% &\cellcolor[HTML]{eeeeee}1.03\% &\cellcolor[HTML]{c5c5c5}1.76 \\
    gl-en & Romance & $254\textrm{K}$ &\cellcolor[HTML]{fbfbfb}0.31\% &\cellcolor[HTML]{fcfcfc}0.21\% &\cellcolor[HTML]{cacaca}1.48 \\
    mk-en & Slavic & $683\textrm{K}$ &\cellcolor[HTML]{f2f2f2}0.83\% &\cellcolor[HTML]{f6f6f6}0.57\% &\cellcolor[HTML]{cbcbcb}1.46 \\
    \textbf{\underline{nl-en}} & Germanic & $4.9\textrm{M}$ &\cellcolor[HTML]{a3a3a3}5.47\% &\cellcolor[HTML]{bbbbbb}4.06\% &\cellcolor[HTML]{cecece}1.35 \\
    \textbf{\underline{de-en}} & Germanic & $4.7\textrm{M}$ &\cellcolor[HTML]{adadad}4.92\% &\cellcolor[HTML]{c1c1c1}3.73\% &\cellcolor[HTML]{cfcfcf}1.32 \\
    \underline{fr-en} & Romance & $5.1\textrm{M}$ &\cellcolor[HTML]{a1a1a1}5.62\% &\cellcolor[HTML]{b7b7b7}4.29\% &\cellcolor[HTML]{d0d0d0}1.31 \\
    sl-en & Slavic & $520\textrm{K}$ &\cellcolor[HTML]{f7f7f7}0.55\% &\cellcolor[HTML]{f8f8f8}0.44\% &\cellcolor[HTML]{d1d1d1}1.25 \\
    it-en & Romance & $5.5\textrm{M}$ &\cellcolor[HTML]{9f9f9f}5.74\% &\cellcolor[HTML]{b2b2b2}4.62\% &\cellcolor[HTML]{d1d1d1}1.24 \\
    \underline{es-en} & Romance & $5.2\textrm{M}$ &\cellcolor[HTML]{a6a6a6}5.31\% &\cellcolor[HTML]{b6b6b6}4.38\% &\cellcolor[HTML]{d2d2d2}1.21 \\
    bg-en & Slavic & $4.7\textrm{M}$ &\cellcolor[HTML]{afafaf}4.77\% &\cellcolor[HTML]{bdbdbd}3.95\% &\cellcolor[HTML]{d2d2d2}1.21 \\
    \underline{el-en} & Greek & $3.5\textrm{M}$ &\cellcolor[HTML]{c5c5c5}3.48\% &\cellcolor[HTML]{cecece}2.96\% &\cellcolor[HTML]{d3d3d3}1.18 \\
    bs-en & Slavic & $146\textrm{K}$ &\cellcolor[HTML]{fdfdfd}0.14\% &\cellcolor[HTML]{fefefe}0.12\% &\cellcolor[HTML]{d3d3d3}1.17 \\
    hr-en & Slavic & $3.3\textrm{M}$ &\cellcolor[HTML]{cbcbcb}3.15\% &\cellcolor[HTML]{d2d2d2}2.73\% &\cellcolor[HTML]{d3d3d3}1.15 \\
    \underline{id-en} & Malayo & $2.3\textrm{M}$ &\cellcolor[HTML]{dcdcdc}2.11\% &\cellcolor[HTML]{e0e0e0}1.90\% &\cellcolor[HTML]{d5d5d5}1.11 \\
    ro-en & Romance & $4.8\textrm{M}$ &\cellcolor[HTML]{b5b5b5}4.40\% &\cellcolor[HTML]{bbbbbb}4.05\% &\cellcolor[HTML]{d6d6d6}1.09 \\
    sk-en & Slavic & $1.6\textrm{M}$ &\cellcolor[HTML]{e7e7e7}1.48\% &\cellcolor[HTML]{e9e9e9}1.36\% &\cellcolor[HTML]{d6d6d6}1.09 \\
    sq-en & Greek & $1.2\textrm{M}$ &\cellcolor[HTML]{efefef}1.00\% &\cellcolor[HTML]{efefef}0.97\% &\cellcolor[HTML]{d8d8d8}1.03 \\
    sr-en & Slavic & $3.6\textrm{M}$ &\cellcolor[HTML]{cbcbcb}3.11\% &\cellcolor[HTML]{cdcdcd}3.02\% &\cellcolor[HTML]{d8d8d8}1.03 \\
    cs-en & Slavic & $2.7\textrm{M}$ &\cellcolor[HTML]{d8d8d8}2.35\% &\cellcolor[HTML]{d9d9d9}2.27\% &\cellcolor[HTML]{d8d8d8}1.04 \\
    \underline{he-en} & Arabic & $5.7\textrm{M}$ &\cellcolor[HTML]{adadad}4.88\% &\cellcolor[HTML]{afafaf}4.80\% &\cellcolor[HTML]{d8d8d8}1.02 \\
    \underline{fi-en} & Uralic & $623\textrm{K}$ &\cellcolor[HTML]{f8f8f8}0.48\% &\cellcolor[HTML]{f7f7f7}0.52\% &\cellcolor[HTML]{dbdbdb}0.92 \\
    \underline{lt-en} & Uralic & $1.1\textrm{M}$ &\cellcolor[HTML]{f3f3f3}0.78\% &\cellcolor[HTML]{f0f0f0}0.91\% &\cellcolor[HTML]{dcdcdc}0.86 \\
    et-en & Uralic & $280\textrm{K}$ &\cellcolor[HTML]{fcfcfc}0.20\% &\cellcolor[HTML]{fcfcfc}0.23\% &\cellcolor[HTML]{dcdcdc}0.87 \\
    \underline{vi-en} & Chinese & $4.6\textrm{M}$ &\cellcolor[HTML]{cccccc}3.06\% &\cellcolor[HTML]{bfbfbf}3.84\% &\cellcolor[HTML]{e0e0e0}0.80 \\
    ms-en & Malayo & $132\textrm{K}$ &\cellcolor[HTML]{fefefe}0.08\% &\cellcolor[HTML]{fefefe}0.11\% &\cellcolor[HTML]{e3e3e3}0.73 \\
    \underline{hu-en} & Uralic & $3.9\textrm{M}$ &\cellcolor[HTML]{d8d8d8}2.34\% &\cellcolor[HTML]{c8c8c8}3.28\% &\cellcolor[HTML]{e4e4e4}0.71 \\
    be-en & Slavic & $116\textrm{K}$ &0.07\% &\cellcolor[HTML]{fefefe}0.10\% &\cellcolor[HTML]{e4e4e4}0.70 \\
    \underline{ru-en} & Slavic & $5.6\textrm{M}$ &\cellcolor[HTML]{cbcbcb}3.15\% &\cellcolor[HTML]{b0b0b0}4.71\% &\cellcolor[HTML]{e4e4e4}0.67 \\
    \underline{pl-en} & Slavic & $4.7\textrm{M}$ &\cellcolor[HTML]{d4d4d4}2.60\% &\cellcolor[HTML]{bdbdbd}3.95\% &\cellcolor[HTML]{e5e5e5}0.66 \\
    \underline{ar-en} & Arabic & $5.8\textrm{M}$ &\cellcolor[HTML]{cccccc}3.05\% &\cellcolor[HTML]{aeaeae}4.87\% &\cellcolor[HTML]{e6e6e6}0.63 \\
    \underline{fa-en} & Arabic & $4.1\textrm{M}$ &\cellcolor[HTML]{dfdfdf}1.93\% &\cellcolor[HTML]{c6c6c6}3.39\% &\cellcolor[HTML]{e8e8e8}0.57 \\
    \underline{tr-en} & Turkic & $4.9\textrm{M}$ &\cellcolor[HTML]{dadada}2.21\% &\cellcolor[HTML]{bbbbbb}4.07\% &\cellcolor[HTML]{e9e9e9}0.54 \\
    hy-en & Greek & $544\textrm{K}$ &\cellcolor[HTML]{fbfbfb}0.26\% &\cellcolor[HTML]{f8f8f8}0.45\% &\cellcolor[HTML]{e8e8e8}0.58 \\
    ka-en & Greek & $332\textrm{K}$ &\cellcolor[HTML]{fdfdfd}0.14\% &\cellcolor[HTML]{fbfbfb}0.28\% &\cellcolor[HTML]{ececec}0.50 \\
    \underline{hi-en} & Indo & $481\textrm{K}$ &\cellcolor[HTML]{fdfdfd}0.19\% &\cellcolor[HTML]{f9f9f9}0.40\% &\cellcolor[HTML]{ececec}0.48 \\
    ur-en & Indo & $158\textrm{K}$ &0.05\% &\cellcolor[HTML]{fefefe}0.13\% &\cellcolor[HTML]{efefef}0.38 \\
    \underline{zh-en} & Chinese & $5.4\textrm{M}$ &\cellcolor[HTML]{e4e4e4}1.67\% &\cellcolor[HTML]{b4b4b4}4.50\% &\cellcolor[HTML]{efefef}0.37 \\
    \underline{bn-en} & Indo & $127\textrm{K}$ &0.04\% &\cellcolor[HTML]{fefefe}0.11\% &\cellcolor[HTML]{f0f0f0}0.36 \\
    \underline{ko-en} & Chinese & $5.5\textrm{M}$ &\cellcolor[HTML]{e4e4e4}1.66\% &\cellcolor[HTML]{b2b2b2}4.63\% &\cellcolor[HTML]{f0f0f0}0.36 \\
    mr-en & Indo & $241\textrm{K}$ &0.07\% &\cellcolor[HTML]{fcfcfc}0.20\% &\cellcolor[HTML]{f0f0f0}0.35 \\
    kk-en & Turkic & $84\textrm{K}$ &0.02\% &0.07\% &\cellcolor[HTML]{f2f2f2}0.29 \\
    mn-en & Mongolic & $181\textrm{K}$ &0.04\% &\cellcolor[HTML]{fdfdfd}0.15\% &\cellcolor[HTML]{f3f3f3}0.27 \\
    my-en & Mongolic & $558\textrm{K}$ &\cellcolor[HTML]{fefefe}0.12\% &\cellcolor[HTML]{f8f8f8}0.47\% &\cellcolor[HTML]{f3f3f3}0.26 \\
    \underline{ja-en} & Chinese & $5.5\textrm{M}$ &\cellcolor[HTML]{ececec}1.17\% &\cellcolor[HTML]{b2b2b2}4.60\% &\cellcolor[HTML]{f5f5f5}0.25 \\
    \underline{ta-en} & Indo & $156\textrm{K}$ &0.03\% &\cellcolor[HTML]{fefefe}0.13\% &\cellcolor[HTML]{f5f5f5}0.23 \\
    az-en & Turkic & $153\textrm{K}$ &0.03\% &\cellcolor[HTML]{fefefe}0.13\% &\cellcolor[HTML]{f5f5f5}0.23 \\
    th-en & Mongolic & $2.6\textrm{M}$ &\cellcolor[HTML]{fafafa}0.36\% &\cellcolor[HTML]{dbdbdb}2.20\% &\cellcolor[HTML]{f9f9f9}0.16 \\
    \bottomrule
\end{tabular}
\caption{\small{
Origins when augmenting no-en translation with multilingual datastore $\mathcal{D}_{(\textrm{ALL}, \textrm{en})}$.
Languages are sorted by oversampling ratio.
$P_{\textrm{obs}}$ shows observed percentages on the no-en test set; $P_{\textrm{uni}}$ shows uniform percentages by datastore size.
$P_{\textrm{obs}}/P_{\textrm{uni}}$ indicates over/undersampling. 
\textbf{Bold}: Germanic languages (same as no-en); \underline{underlined}: bridge languages.
Darker colors: higher probability in each column.}}
\label{ch4:tabprob_multilingual_all}
\end{table}

It should be noted that not all target token suggestions from \textit{k}NN-MT are included in the generated target sentence.
During beam search decoding, the system makes token selection decisions based on the interpolated probability distribution that combines both the \textit{k}NN-MT suggestions and the base NMT model's predictions, weighted by the parameter $\lambda$.
While the \textit{k}NN retrieval mechanism influences these decisions by providing alternative token probabilities, many retrieved tokens may be outweighed by the base model's predictions or pruned during beam search, particularly when retrieved neighbors have limited relevance to the current context or when the NMT model has high confidence in its own predictions.

\begin{figure}[!htb]
    \centering
    \includegraphics[width=0.75\linewidth]{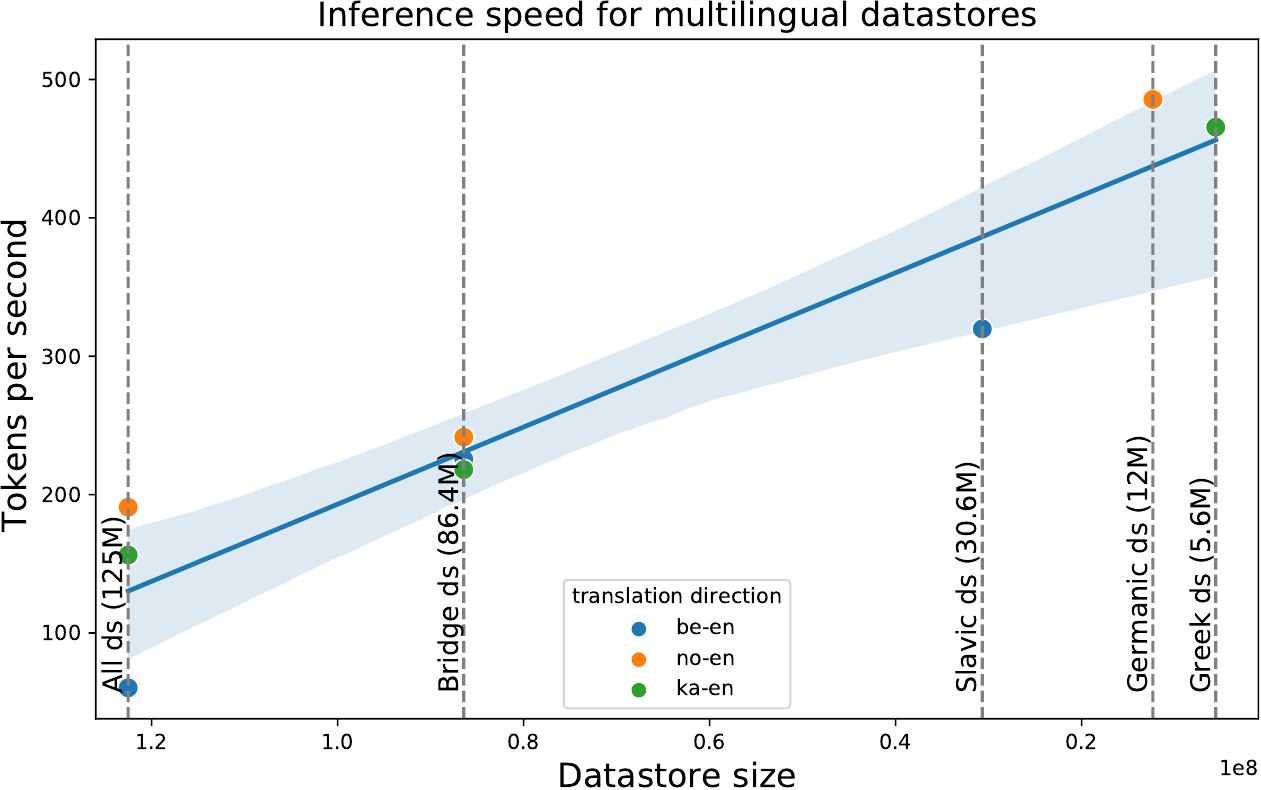}
    \caption{Inference speed for Greek, Germanic, Slavic, Bridge, and All multilingual datastores. The x-axis displays the datastore size (large to small), and the y-axis shows the corresponding tokens per second. A clear linear trend can be observed: smaller datastores result in substantially faster decoding times.}
    \label{ch4:fig:inf_speed}
\end{figure}

\subsubsection{Multilingual datastore speed}
We present multilingual datastore inference speeds in Figure \ref{ch4:fig:inf_speed}. We set $k$ to $64$, and present results for all multilingual datastores, using a single source language into English for each language grouping.
We average results over 3 runs.

We observe a clear trend: smaller datastores result in substantially faster decoding times.
For Belarusian-English (be-en), using the Language Grouping datastore $\mathcal{D}_{(\text{LG},\text{en})}$ with 31M entries instead of the All multilingual datastore $\mathcal{D}_{(\text{ALL},\text{en})}$ consisting of 125M key-values pairs results in significantly faster decoding speeds of $5.3$x.
For Norwegian-English (no-en) and Georgian-English (ka-en), the improvements are $3.0$x and $2.6$x.
In terms of quality, be-en improves when using the Language Grouping datastore ($+0.6$ BLEU), while Norwegian-English (no-en) and Georgian-English (ka-en) have similar performance ($-0.1$ BLEU and $+0.0$ BLEU).

\section{Conclusion}
\label{ch4:sec:conclusion}
In this chapter, we investigated how cross-lingual knowledge transfer can improve \textit{k}NN-MT performance for low-resource languages through three interconnected research questions:\\

\RQsub{2}{1}
\begin{myquote}
We examined whether cross-lingual datastores can improve translation quality for low-resource languages.
Our experiments with various language pairs demonstrated that low-resource languages can indeed benefit substantially from cross-lingual datastores.
As shown in Tables~\ref{ch4:tabml_knn_slavic} and \ref{ch4:tabml_knn_germanic_greek}, using datastores from related high-resource languages consistently improved translation quality for low-resource languages, often surpassing the improvements achieved with smaller bilingual datastores.
Importantly, we found that these improvements cannot be fully explained by datastore size alone, suggesting that linguistic similarity plays a crucial role in effective cross-lingual retrieval.
\end{myquote}

\RQsub{2}{2}
\begin{myquote}
Having established that cross-lingual datastores can benefit low-resource languages, we investigated methods for combining information from multiple languages into a single datastore to benefit both low and high-resource languages.
Our results demonstrate that multilingual datastores constructed from multiple source languages consistently outperform both bilingual and cross-lingual datastores.
For low-resource languages, the improvements are particularly substantial, with gains of up to +3.6 BLEU compared to bilingual datastores.
Surprisingly, even high-resource languages show noticeable improvements of up to +0.5 BLEU when using multilingual datastores.
Our analysis in Table~\ref{ch4:tabprob_multilingual_all} reveals that linguistically similar languages contribute disproportionately to the retrieved examples, highlighting the importance of linguistic similarity in effective multilingual retrieval.
\end{myquote}

\RQsub{2}{3}
\begin{myquote}
Based on our findings from RQ2.1 and RQ2.2, we then examined strategies for optimizing multilingual datastores to balance translation quality and computational efficiency.
As shown in Figure~\ref{ch4:fig:inf_speed}, creating datastores based on linguistic similarity allows for significant reductions in datastore size while maintaining translation quality.
By focusing on languages from the same language grouping, we can create datastores that were up to four times smaller while achieving comparable or even better translation quality than the comprehensive multilingual datastore.
This approach results in up to 5.3x faster inference speeds without sacrificing translation quality.
Additionally, our experiments with cross-lingual representation alignment in Table~\ref{ch4:tabxlingmap} show that linear transformation matrices can further improve cross-lingual retrieval effectiveness, enhancing translation quality by up to +0.8 BLEU.
\end{myquote}

Taken together, these steps answered our main research question:

\RQ{2}
\begin{myquote}
Our research demonstrates that cross-lingual knowledge transfer can significantly enhance \textit{k}NN-MT performance for low-resource languages.
This builds on our findings from Chapter \ref{ch3}, where we established that representational similarities between languages are strongly correlated with effective knowledge transfer.

By leveraging multilingual neural machine translation representations and carefully constructing cross-lingual and multilingual datastores, we enable effective retrieval across language boundaries.
Our proposed multilingual \textit{k}NN-MT approach not only substantially improves translation quality for low-resource languages but also enhances performance for high-resource languages.
Furthermore, we show that linguistically informed datastore construction maintains these quality improvements while significantly reducing computational costs, making \textit{k}NN-MT more accessible for a wider range of languages and applications.
\end{myquote}

\subsubsection{Limitations}
We used an mNMT model that is non-English-centric, which may present limitations in the generalizability of our multilingual \textit{k}NN-MT results to other multilingual settings such as English-centric.

A limitation of \textit{k}NN-MT, which has become the central focus of most subsequent \textit{k}NN-MT research, is the steep increase in decoding time introduced by \textit{k}NN-MT, as each decoding step requires a computationally expensive nearest neighbor search \citep{zheng_adaptive_2021,martins_chunk-based_2022,martins_efficient_2022,yang_nearest_2022,meng_fast_2022,jiang-etal-2022-towards}.

While we built multilingual datastores using up to 51 languages, we did not investigate to what extent even larger multilingual datastores can further improve performance.

\chapter{Fine-Tuning and Capability Preservation in LLM Translation}
\label{ch5}

\renewcommand{\thefootnote}{}
\footnotetext{This chapter was published as: 
David Stap, Eva Hasler, Bill Byrne, Christof Monz, and Ke Tran.
The Fine-Tuning Paradox: Boosting Translation Quality Without Sacrificing LLM Abilities.
In \textit{Proceedings of the 62nd Annual Meeting of the Association for Computational Linguistics (Volume 1: Long Papers)}, pages 6189–6206, Bangkok, Thailand. Association for Computational Linguistics, 2024.}
\renewcommand{\thefootnote}{\arabic{footnote}}

\section{Introduction and research questions}
While Chapters \ref{ch3} and \ref{ch4} focused on positive cross-lingual knowledge transfer in multilingual NMT models to improve low-resource translation performance, this chapter shifts our attention to large language models (LLMs) for machine translation, examining the challenge of negative transfer or interference that occurs during fine-tuning—where improving translation quality through parallel data can diminish the beneficial capabilities that emerge from pre-training.

Recent work has highlighted a range of qualitative advantages that LLMs hold over Neural Machine Translation (NMT) models.
One significant advantage is the controllability of style and language variety which can be achieved through prompting and in-context learning \citep{brown_language_2020,garcia_unreasonable_2023,agrawal_-context_2023}.
LLMs also exhibit inherent document-level translation abilities \citep{wang_document-level_2023, karpinska_large_2023}.
Another advantage is their ability to produce less literal translations \citep{raunak_gpts_2023}.
Finally, LLMs have been shown to have better performance in handling difficult linguistic phenomena such as idioms and ambiguous expressions \citep{neubig_zeno_2023}.
Taken together, LLMs are surpassing NMT models in terms of versatility.

Recent studies have demonstrated that fine-tuning LLMs on parallel data further improves their translations as measured by metrics that reflect overall quality (such as COMET) \citep{li_eliciting_2023,yang_bigtranslate_2023,zeng_teaching_2024}.
However, relying on general translation quality metrics and generic test sets does not fully capture the nuanced abilities of LLMs in machine translation.
This oversight raises questions about the retention of LLM-specific advantages---such as controllability, document-level translation proficiency, and the production of less literal translations---after fine-tuning on parallel data. 

In this chapter, we investigate how qualitative advantages of LLMs change when fine-tuning on parallel data.
To this end, we ask:

\RQ{3}
\begin{myquote}
While it is clear that general machine translation quality improves through fine-tuning, there is a risk that LLMs lose their unique strengths due to catastrophic forgetting \citep{mccloskey_catastrophic_1989,ratcliff_connectionist_1990,luo_empirical_2023}.
We investigate this trade-off through a comprehensive analysis of different fine-tuning approaches, evaluating both translation quality improvements and capability preservation across multiple model scales and data regimes. Our experiments span different model architectures and sizes to develop a thorough understanding of how to effectively balance these competing objectives.\\\\
This overarching question is addressed through three interconnected steps.
First, we identify which specific LLM capabilities are affected by fine-tuning (RQ3.1).
Then, we investigate how different model scales and data quantities influence these effects (RQ3.2).
Finally, we develop and evaluate strategies to maintain capabilities while still improving translation quality (RQ3.3).
\end{myquote}

\RQsub{3}{1}
\begin{myquote}
The impact of fine-tuning LLMs using parallel data on properties that are relevant for translation remains poorly understood.
The LLM properties we investigate are general translation quality, formality steerability, non-literalness in idiom translations (as measured by a novel dataset, IdiomsInCtx-MT, consisting of human-translated idioms), performance on specialized domains, and performance on document-level input which requires contextualization of ambiguous tokens.
\end{myquote}

\RQsub{3}{2}
\begin{myquote}
We examine how the effects on these capabilities vary across different model and data configurations.\\\\
We perform a comprehensive analysis of LLaMA and Falcon models, with parameter counts ranging from 7 billion up to 65 billion.
We systematically evaluate how different amounts of training data (up to 89K of human-written translations and up to 1.4M parallel examples from web data) affect model capabilities across six diverse language pairs, enabling us to understand the relationship between model scale, data quantity, and capability preservation.
\end{myquote}

\RQsub{3}{3}
\begin{myquote}
Based on our findings from RQ3.1 and RQ3.2, we then investigate solutions to mitigate capability loss.
To this end, we investigate an approach that combines monolingual and parallel data during fine-tuning.
Through extensive experimentation, we evaluate whether this mixed-data strategy can improve translation performance while better preserving the model's inherent capabilities compared to traditional parallel-only fine-tuning.
Our analysis spans multiple language pairs and evaluates various capability metrics to comprehensively assess this approach's effectiveness.
\end{myquote}

\paragraph{Organization.}
This chapter is organized as follows: After reviewing the previous work in Section \ref{ch5:sec:related_work}, we present our setup and analysis for fine-tuning on parallel data in Section \ref{ch5:sec:method_parallel_ft}. Next, we propose a method for capability preservation in Section \ref{ch5:sec:method_parallel_mono_ft}. Finally, we discuss the conclusions and implications of this work in Section \ref{ch5:sec:conclusion}

\section{Related work}\label{ch5:sec:related_work}
In this section, we discuss previous work that investigate advantages that LLMs have over NMT systems for machine translation, as well as related work in the area of LLM fine-tuning strategies for machine translation.

\subsection{Advantages of LLMs for machine translation}
Several studies have investigated the use of LLMs for translation.
Generally, current LLMs show strong performance for most language pairs, but lag behind NMT systems when translating into low-resource languages \citep{zhu-etal-2024-multilingual,stap_chatgpt_2023,robinson_chatgpt_2023,kocmi_findings_2023}.

In addition to strong performance, LLMs exhibit certain abilities that are relevant for translation.
NMT systems show a bias towards generating text that is over-represented in the data, such as language varieties \citep{riley_frmt_2023} and formality \citep{rippeth_controlling_2022}, whereas LLMs can easily be controlled for this bias using examples \citep{garcia_unreasonable_2023}.
Furthermore, examples can be supplied to improve general LLM translation quality via in-context learning \citep{agrawal_-context_2023,moslem_adaptive_2023}.
NMT models are often unable to translate idioms accurately and generate literal translations \citep{dankers_can_2022}.
LLMs produce less literal outputs compared to NMT models, particularly for sentences that contain idiomatic expressions \citep{vilar_prompting_2023,raunak_gpts_2023}.
NMT models are trained on sentence level, and thus do not take into account document context.
LLMs outperform NMT models for document translation in general domains such as news and social media \citep{wang_document-level_2023}, as well as in more specialized domains such as literature \citep{karpinska_large_2023}.

\subsection{Fine-tuning LLMs for machine translation}
There are multiple strategies for fine-tuning LLMs for machine translation.
One approach makes use of either a small set of high-quality human-written translations or a set of translation instructions for fine-tuning \citep{li_eliciting_2023,zeng_teaching_2024,jiao_parrot_2023}.
Another line of work makes use of more traditional machine translation data: parallel data from the web, which is orders of magnitude larger compared to what is used in fine-tuning \citep{yang_bigtranslate_2023,alves_steering_2023,zhang_machine_2023,zhu_extrapolating_2023}.
These strategies are focused on improving general machine translation quality, but it remains unclear what happens to other abilities that are relevant for translation.

Unlike the positive cross-lingual transfer investigated in Chapters \ref{ch3} and \ref{ch4}, LLMs present a different challenge where interference or negative transfer can occur during fine-tuning.
In previous chapters, we sought to maximize knowledge sharing between languages; here, we balance the benefits of task-specific fine-tuning against the risk of catastrophic forgetting of pre-trained knowledge.

We investigate the effect of fine-tuning on relevant abilities for translation using publicly available models.

\section{Evaluating fine-tuning impact on LLM translation capabilities}
\label{ch5:sec:method_parallel_ft}
This section details our experimental methodology for evaluating how fine-tuning with parallel data influences both general translation performance and the distinctive capabilities that make LLMs valuable for translation tasks.

\subsection{Experimental setup}

\subsubsection*{Models}
We use the LLaMA \citep{touvron_llama_2023} 7B, 13B, 30B, 65B, and Falcon \citep{almazrouei_falcon_2023} 7B and 40B language models.

\subsubsection*{Optimization}
We perform full fine-tuning on models up to 40B, and store intermediate checkpoints during fine-tuning to track how abilities evaluate over time.
For LLaMA-65B, we fine-tune with QLoRA \citep{dettmers_qlora_2023}, using 8-bit quantization.
We use the AdamW \citep{loshchilov2018decoupled} optimizer with a cosine learning rate scheduler and 3\% warm-up percentage.
Following \citet{zhang_machine_2023} we set the low-rank approximation to 64 and the scaling factor for low-rank adaptation to 32.
We empirically set the batch size to 128, learning rate to $2e-5$, and train for one epoch.
During inference we use beam search with a beam size of 4.
We run our fine-tuning experiments with the Hugging Face transformers library \citep{wolf_transformers_2020} and make use of DeepSpeed \citep{rasley_deepspeed_2020}.

\subsubsection*{Inference}
Depending on the evaluation set, we use a 0-shot or few-shot approach.
The prompt used for fine-tuning and 0-shot is shown in Table~\ref{ch5:tab:0_shot_prompt}.
For our few-shot setup, we find the 5 most similar source sentences and their corresponding target sentences from a corresponding train set (if available) or validation set.
The resulting prompt is displayed in Table~\ref{ch5:tab:5_shot_prompt}.
We use Sentence-BERT \citep{reimers_sentence-bert_2019} for encoding\footnote{We use \texttt{all-MiniLM-L6-v2} for English sentences and \texttt{paraphrase-multilingual-MiniLM-L12-v2} for non-English sentences.} and FAISS \citep{douze_faiss_2025,johnson_billion-scale_2019} for searching similar sentences.

\begin{table*}[!htb]
    \begin{center}
        \small
        \begin{tabular}{l}
        \toprule
        \texttt{Translate this from \{source\_language\} to \{target\_language\}:} \\
        \texttt{\{source\_language\}: \{source\_sentence\}} \\
        \texttt{\{target\_language\}: \{target\_sentence\}} \\
        \bottomrule
        \end{tabular}
    \end{center}
    \caption{Prompting template for fine-tuning and 0-shot inference. For fine-tuning \texttt{\{target\_sentence\}} is filled with the corresponding target sentence, and for 0-shot inference it is the empty string.}
    \label{ch5:tab:0_shot_prompt}
\end{table*}

\begin{table*}[!htb]
    \begin{center}
        \small
        \begin{tabular}{l}
        \toprule
        \texttt{Translate this from \{source\_language\} to \{target\_language\}:} \\
        \texttt{\{source\_language\}: \{source\_sentence$_1$\}} \\
        \texttt{\{target\_language\}: \{target\_sentence$_1$\}} \\
        \texttt{...} \\
        \texttt{\{source\_language\}: \{source\_sentence$_n$\}} \\
        \texttt{\{target\_language\}: } \\
        \bottomrule
        \end{tabular}
    \end{center}
    \caption{Prompting template for few-shot inference.}
    \label{ch5:tab:5_shot_prompt}
\end{table*}

\subsubsection*{Language directions}
We consider the language directions German (de), Russian (ru), and Chinese (zh) into and out of English (en).

\subsubsection*{Human-written training data}
Following \citet{xu_paradigm_2024}, we use human-written translations from WMT17 to WMT20, resulting in 89K training examples that are evenly distributed across the language directions we consider. On this dataset, we perform full fine-tuning for models up to 40B and QLoRA fine-tuning for the LLaMA 65B model.

\subsubsection*{Web-scraped training data}
Additionally we train models on general domain OPUS \citep{tiedemann_parallel_2012} data from the News-Commentary, WikiTitles, and ParaCrawl \citep{banon_paracrawl_2020} corpora.
To ensure that the resulting data is above an acceptable quality threshold we perform data filtering using an Amazon-internal Quality Estimation (QE) model, which has a similar architecture as COMETKiwi \citep{rei_cometkiwi_2022} and is based on the InfoXLM-Large pretrained multilingual encoder \citep{chi_infoxlm_2021}. We train our sentence-level QE model on a large Amazon-internal dataset of human annotations for more than 12 languages, where each translation is rated between 1 (completely random) and 6 (perfect translation).

We use a subset of 1.4M sentence pairs of this filtered data that is evenly distributed across language directions for training. We fine-tune  LLaMA models up to 40B parameters on this dataset but leave out larger models because of the high computational cost.

\subsubsection*{Evaluation data and metrics}
For evaluation, we consider the following test sets:
\begin{itemize}
    \item \textbf{WMT22} To evaluate general machine translation quality, we use WMT22 \citep{kocmi_findings_2022} test sets consisting of news, e-commerce, social, and conversational domains.
    We evaluate all language directions on this test set in a 0-shot setting.
    Following the recommendation of \citet{kocmi-etal-2021-ship}, we report COMET scores\footnote{model: \texttt{unbabel/wmt22-comet-da}} \citep{rei_comet_2020}, which aligns with our transition to neural metrics for evaluating high-quality translations as described in Section~\ref{ch2:sec:comet}.
    While BLEU \citep{papineni_bleu_2002} served as a reliable metric in our earlier chapters focused on multilingual NMT, recent research has shown that neural metrics like COMET provide better correlation with human judgments for evaluating the nuanced capabilities of LLMs \citep{mathur_tangled_2020,freitag-etal-2021-experts,kocmi-etal-2021-ship,freitag-etal-2022-results}.
        
    \item \textbf{CoCoA-MT} To evaluate formality steering ability of LLMs, we make use of the CoCoA-MT \citep{nadejde_cocoa-mt_2022} dataset.
    It consists of 600 test sentences with English source and contrastive target sentences consisting of a formal and informal translation.
    We use German as the target language and report both COMET and accuracy based on the ratio of correctly predicted formality forms.
    We use 5-shot examples to bias the formality of outputs.\footnote{We also experimented with steering formality through prompting, but the results were inferior to using 5-shot examples.}

    \item \textbf{Law}, \textbf{Medical}, and \textbf{TICO-19} To evaluate the in-context learning ability on technical domains, we consider the Law and Medical test sets from \citet{aharoni_unsupervised_2020}.
    We consider 5-shot inputs.
    We evaluate on German $\leftrightarrow$ English and report COMET scores.
    In addition we evaluate technical abilities on  TICO-19  \citep{anastasopoulos_tico-19_2020},
    which consists of translations in the COVID-19 domain.
    We evaluate on Russian $\leftrightarrow$ English and Chinese $\leftrightarrow$ English.

    \item \textbf{ctxpro} We evaluate 0-shot performance on longer inputs that includes sentences that require context to be disambiguated correctly by including ctxpro \citep{wicks_identifying_2023}.
    We consider the animacy ambiguity type in German $\rightarrow$ English and Russian $\rightarrow$ English.\footnote{We also experimented with different ambiguity types (auxiliary and gender in out-of-English directions).
    However, the resulting translations were often incomplete, containing only a subset of the total number of sentences, making it impossible to effectively evaluate decontextualization abilities.}
    While English makes no gender distinction for inanimate objects, some other languages such as Russian do.
    The ambiguous animacy examples in the ctxpro dataset require corresponding document-level context for correct disambiguation.    
    We subsample the test set to 2K examples per language direction.
    The average number of sentences per input is $10.16\pm1.27$.
    We report the generative accuracy score, which measures the accuracy of the contextualization.

    \item \textbf{IdiomsInCtx-MT} We introduce a novel dataset consisting of idiomatic expressions in context and their human-written translations.\footnote{\url{https://github.com/amazon-science/idioms-incontext-mt}}
    To our knowledge, it is the first publicly available dataset that consists of idiomatic expressions in context and their human-written translations.
    The dataset comprises 2 language pairs: German and Russian paired with English.
    For German, the opposite translation directions are also included.
    Current idiom datasets stem from potentially noisy, web-extracted sources \citep{fadaee-etal-2018-examining}, machine-generated translations \citep{tang_petci_2022}, or are monolingual \citep{haagsma-etal-2020-magpie}.
    In contrast, we use professional translators to create a high-quality evaluation benchmark. 
    Idiomatic expressions and their context sentences were sourced in the respective source language and translated to the target language by professional translators.
    In addition, the dataset contains annotations of the source and target idiomatic expressions for each segment.
    This enables running targeted evaluation on the 0-shot translations of the idiomatic expressions in addition to general quality metrics.
    We evaluate on English$\rightarrow$German, German$\rightarrow$English and Russian$\rightarrow$English using test splits of 1000 segments.
    We report COMET, LitTER \cite{baziotis_automatic_2023} and MWEScore, an Amazon-internal multi-reference version of Score$\_$mwe \citep{zaninello-birch-2020-multiword}.
\end{itemize}

Additionally, we considered style transfer and constrained generation, such as prompting a language model to use custom terminology, as additional translation capabilities. However, our final choices were influenced by the availability of suitable test sets and the applicability of targeted automatic metrics, leading us to not pursue these options.

\begin{figure*}[!htb]
    \centering
    \includegraphics[width=\linewidth]{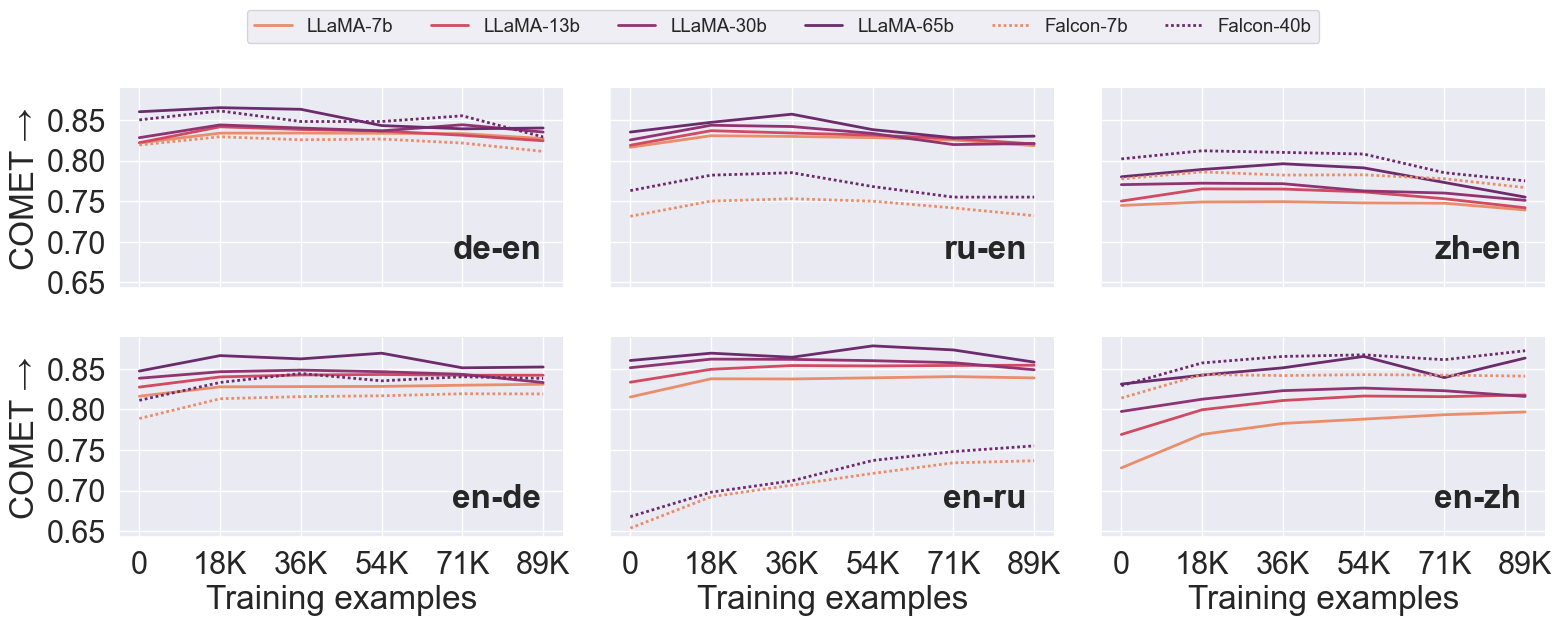}
    \caption{X$\rightarrow$English (top) and English$\rightarrow$X (bottom) COMET scores on WMT22 for models trained on human-written translations with different amounts of training data.
    }
    \label{ch5:fig:wmt_general_quality_wmt}
\end{figure*}

\begin{figure*}[!htb]
    \centering
    \includegraphics[width=\linewidth]{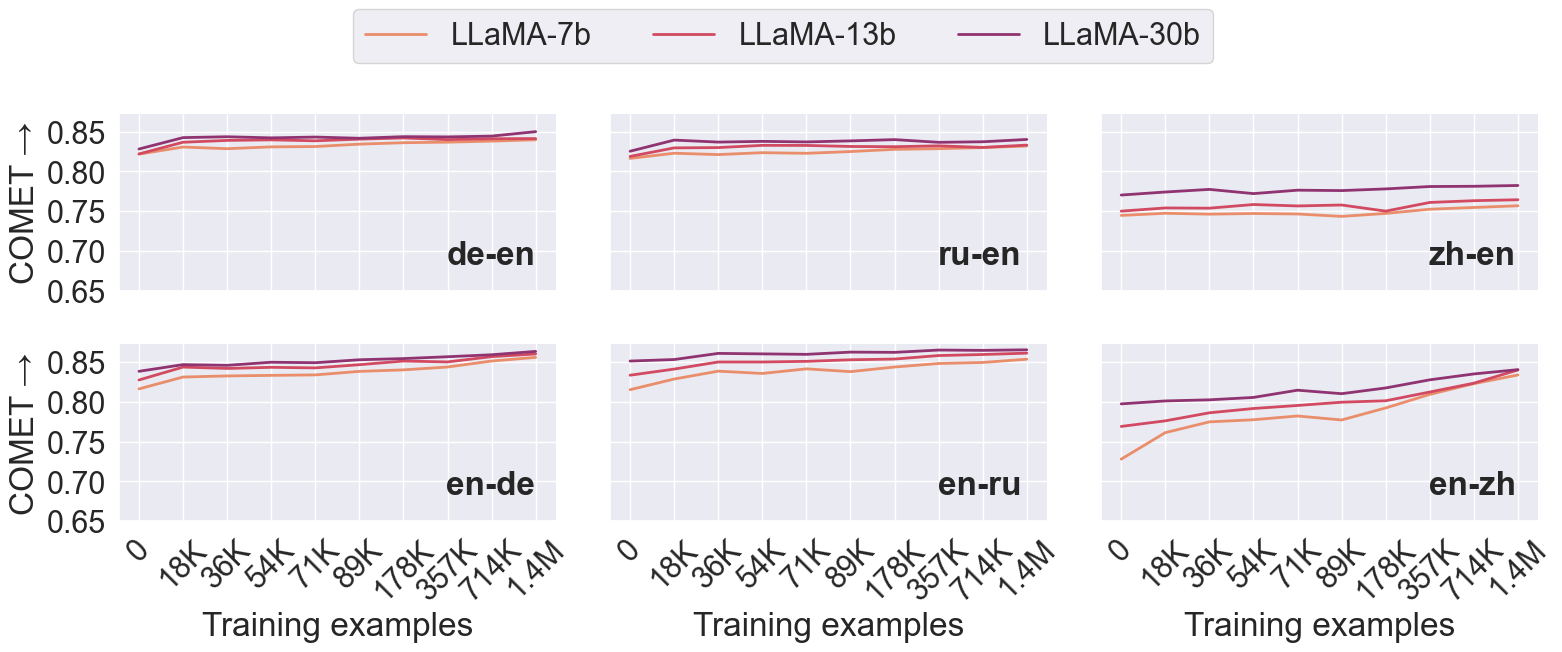}
    \caption{X$\rightarrow$English (top) and English$\rightarrow$X (bottom) COMET scores on WMT22 for different models trained on OPUS parallel data with different amounts of training data.
    }
    \label{ch5:fig:opus_general_quality_wmt}
\end{figure*}

\begin{table*}[!htb]
    \centering
    \small
    \rotatebox{90}{
    \begin{tabular}{lllllll}
        \toprule
        & \multicolumn{2}{c}{de-en}
        & \multicolumn{2}{c}{ru-en}
        & \multicolumn{2}{c}{zh-en} \\
        \cmidrule(lr){2-3} 
        \cmidrule(lr){4-5}
        \cmidrule(lr){6-7}
        & \multicolumn{1}{l}{WMT} 
        & \multicolumn{1}{l}{OPUS}
        & \multicolumn{1}{l}{WMT} 
        & \multicolumn{1}{l}{OPUS}
        & \multicolumn{1}{l}{WMT}
        & \multicolumn{1}{l}{OPUS} \\
        \midrule
        LLaMA-7b         
            & $0.834$ (36K)
            & $\underline{0.840}$ (1.4M)
            & $0.831$ (18K)
            & $\underline{0.832}$ (1.4M)
            & $0.749$ (36K)
            & $\underline{0.757}$ (1.4M) \\                 

        LLaMA-13b
            & $0.842$ (18K)
            & $\underline{0.842}$ (178K)
            & $\underline{0.837}$ (18K)
            & $0.833$ (1.4M)
            & $\underline{0.765}$ (18K)
            & $0.764$ (1.4M) \\

        LLaMA-30b
            & $0.844$ (71K)
            & $\underline{0.850}$ (1.4M)
            & $\underline{0.843}$ (18K)
            & $0.840$ (1.4M)
            & $0.772$ (18K)
            & $\underline{0.782}$ (1.4M) \\
        \midrule \\
        & \multicolumn{2}{c}{en-de}
        & \multicolumn{2}{c}{en-ru}
        & \multicolumn{2}{c}{en-zh} \\
        \cmidrule(lr){2-3} 
        \cmidrule(lr){4-5}
        \cmidrule(lr){6-7}
        & \multicolumn{1}{l}{WMT} 
        & \multicolumn{1}{l}{OPUS}
        & \multicolumn{1}{l}{WMT} 
        & \multicolumn{1}{l}{OPUS}
        & \multicolumn{1}{l}{WMT}
        & \multicolumn{1}{l}{OPUS} \\
        \midrule
        LLaMA-7b         
            & $0.831$ (89K)
            & $\underline{0.856}$ (1.4M)
            & $0.840$ (71K)
            & $\underline{0.853}$ (1.4M)
            & $0.797$ (89K)
            & $\underline{0.834}$ (1.4M) \\                 

        LLaMA-13b
            & $0.843$ (54K)
            & $\underline{0.860}$ (1.4M)
            & $0.854$ (89K)
            & $\underline{0.861}$ (1.4M)
            & $0.818$ (89K)
            & $\underline{0.840}$ (1.4M) \\

        LLaMA-30b
            & $0.848$ (36K)
            & $\underline{0.863}$ (1.4M)
            & $0.862$ (18K)
            & $\underline{0.865}$ (1.4M)
            & $0.826$ (54K)
            & $\underline{0.840}$ (1.4M) \\
        \bottomrule
        \end{tabular}
        }
        \caption{WMT22 COMET scores comparing models fine-tuned on human-written data (WMT) and filtered web-crawled data (OPUS).
        Parentheses indicate the number of fine-tuning examples seen by the best-performing checkpoints.
        Best scores are \underline{underlined}.}
        \label{ch5:tab:wmt_vs_opus_general_quality}
\end{table*}

\subsection{Results}
\label{sec:parallel_ft_results}

\subsubsection*{General translation quality improves}
Results on WMT22 for models trained on human-written translations are summarized in Figure~\ref{ch5:fig:wmt_general_quality_wmt}.
Consistent with expectations, we observe that fine-tuning generally improves the translation quality, and larger models generally perform better.
Using 89K parallel examples does not always yield better results compared to smaller datasets. 
For most language directions, we notice an initial increase in translation quality, followed by a slight decline.                        The most notable increases are found for the out-of-English directions.

In contrast, when models are fine-tuned on a more extensive dataset (up to 1.4 million examples) sourced from the web translation quality continues to improve with the addition of more data, as shown in Figure \ref{ch5:fig:opus_general_quality_wmt}).
Note that the data size up until 89K has relatively small increments, and later data sizes are doubled compared to the data point before it.
Generally, in contrast to training on human-written data, translation quality continues to increase when adding more data.
We compare the best checkpoints of models trained on human-written and OPUS data in Table \ref{ch5:tab:wmt_vs_opus_general_quality}.
In 15 out of 18 cases, models fine-tuned on the larger OPUS dataset result in better scores on WMT22.
We hypothesize that this difference stems from the domain-specific composition of the human-written WMT training data, which is exclusively news content.
In contrast, the OPUS dataset is more diverse and includes multiple domains.
This diversity better reflects domain composition of the WMT22 test set, which has content from news, e-commerce, social media and conversational domains.
The improvements are significantly more marked in translations from other languages into English.

\begin{figure}[!htb]
    \centering
    \includegraphics[width=\linewidth]{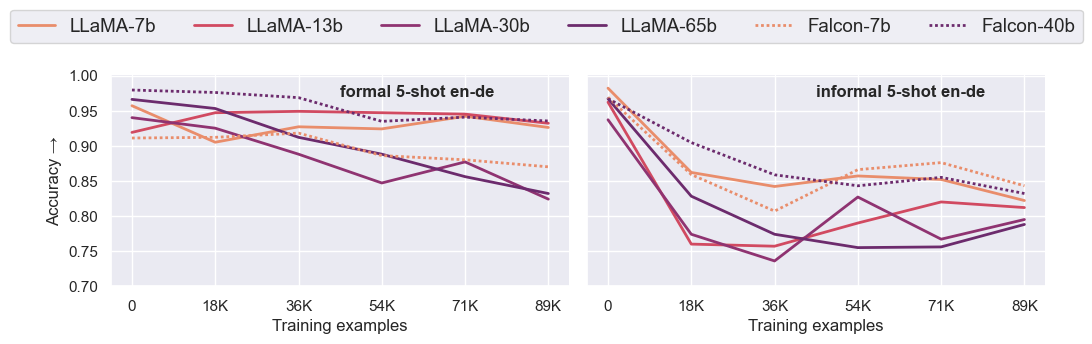}
    \caption{Accuracy of formality markers for models trained on human-written translations. 
    }
    \label{ch5:fig:wmt_formality_steering}
\end{figure}

\subsubsection*{Formality steering ability degrades}
\begin{figure}[!htb]
    \centering
    \includegraphics[width=\linewidth]{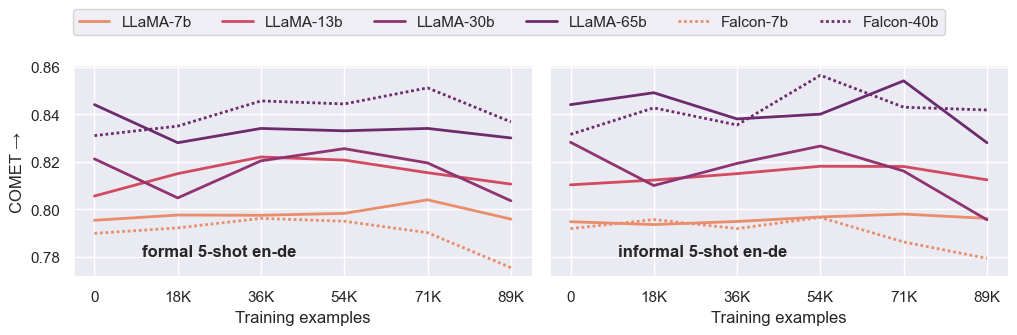}
    \caption{COMET scores on CoCoA-MT for models trained on WMT data.
    }
    \label{ch5:fig:wmt_formality_steering_comet}
\end{figure}

We show results for formality steering using 5-shot examples for models trained on human-written data in Figure \ref{ch5:fig:wmt_formality_steering}.
Notably, the base models exhibit strong performance for this task; for instance, the LLaMA-7b model achieves an accuracy of 0.982 in identifying informal markers.
However, fine-tuning on only 18K examples results in a decline of this ability to 0.862, even though German-English COMET on WMT22 continues to improve up to 36K examples.
The degradation is more pronounced with informal markers, which is likely attributable to the formal bias inherent in the WMT22 training data.
Fine-tuning on more data further degrades formality steering capabilities: there is a significant n egative correlation (Spearman's $\rho=-0.46$, $p<0.001$) between formal and informal marker prediction and dataset size.
In contrast to the accuracy of formality forms, we find that COMET stays relatively constant (see Figure \ref{ch5:fig:wmt_formality_steering_comet}), despite some fluctuations, indicating that it does not adequately capture formality markers.
As a result, the accuracy scores correlate very weakly ($\rho=0.16$, $p<0.001$) with COMET scores, which suggests that comprehensive evaluation of LLMs for machine translation benefits from task-specific metrics.

\begin{figure}[!htb]
    \centering
    \includegraphics[width=\linewidth]{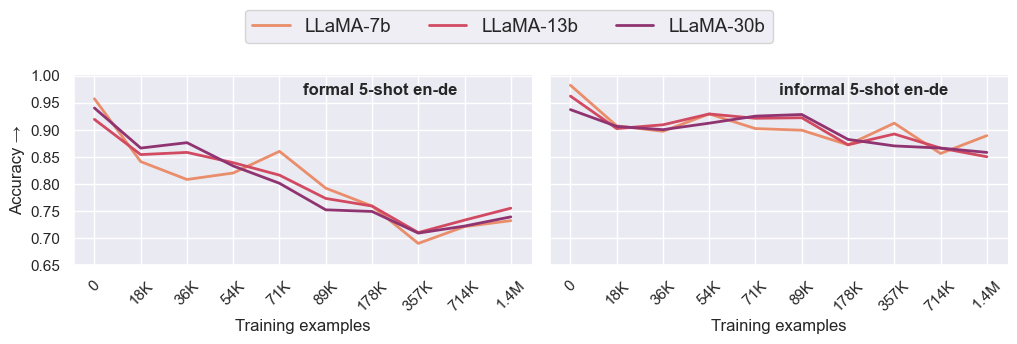}
    \caption{Accuracy of formality markers for models trained on OPUS data.
    }
    \label{ch5:fig:opus_formality_steering}
\end{figure}

Models trained on OPUS data show even more severe degradation of formality control as the fine-tuning data increases to 1.4M parallel sentences (Figure \ref{ch5:fig:opus_formality_steering}).
Notably, while general translation quality continues to improve with more data (Figure \ref{ch5:fig:opus_general_quality_wmt}), the ability to generate correct formality markers steadily declines, highlighting the tension between improving overall performance and maintaining specific capabilities.
Again, we observe a significant negative correlation ($\rho=-0.58$, $p<0.001$) between accuracy and dataset size for OPUS-trained models, reinforcing the conclusion that larger parallel datasets during fine-tuning adversely affect formality steering capabilities.

\subsubsection*{Performance on technical domains degrades}

\begin{figure*}[!htb]
    \centering
    \includegraphics[width=\linewidth]{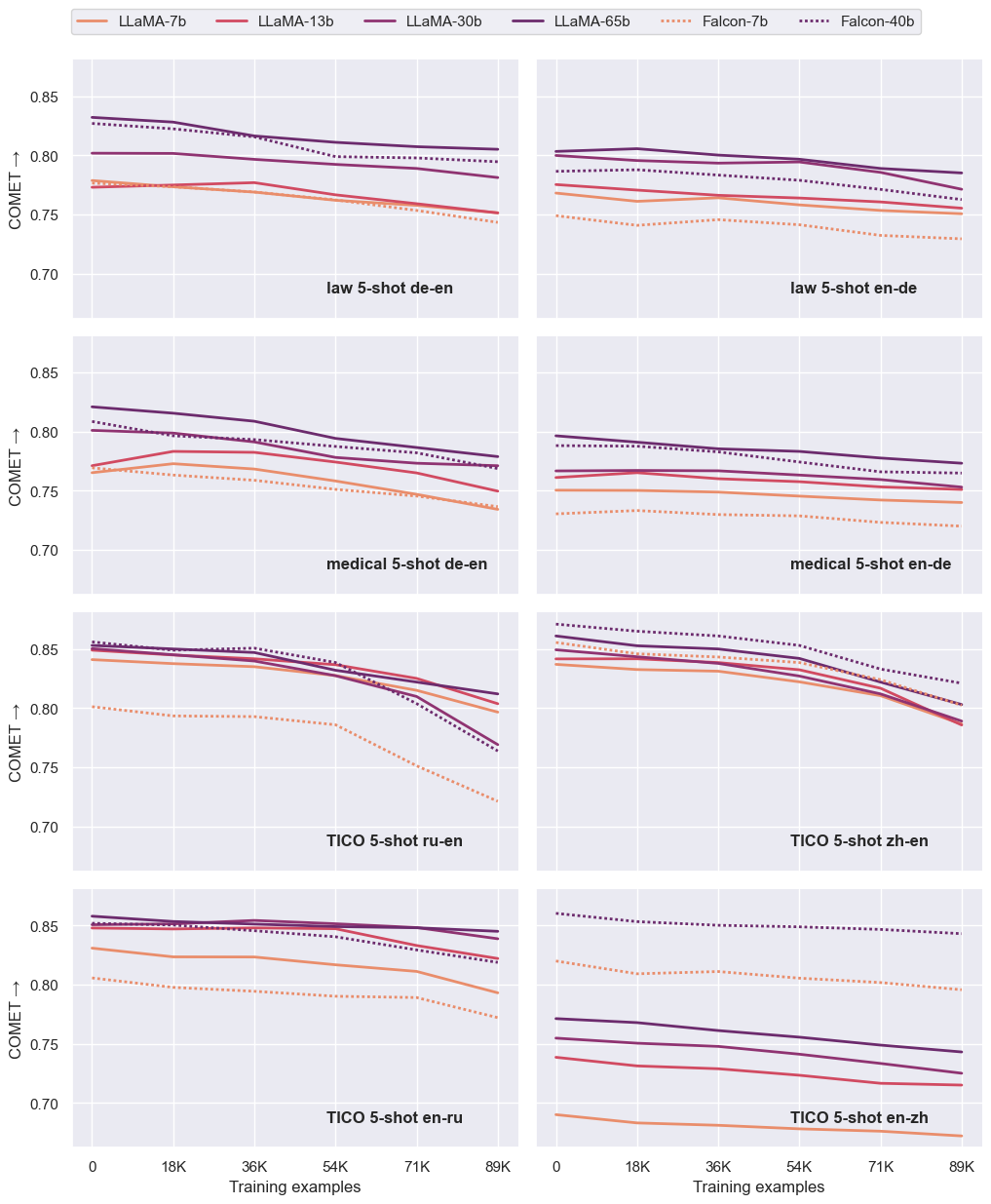}
    \caption{COMET on technical domains using 5-shot examples for models trained on human-written translations. 
    }
    \label{ch5:fig:wmt_technical_domains_full}
\end{figure*}

Next, we evaluate the model capabilities of doing technical translations given 5-shot examples.
Results on human-written training data across all evaluated domains and language directions are shown in Figure \ref{ch5:fig:wmt_technical_domains_full}.
We observe a consistent trend: fine-tuning impairs the few-shot technical translation capabilities, and generally further fine-tuning results in more degradations.
In most cases, performance starts to degrade  after only 18K examples.
For example, LLaMA-7b scores 0.8308 COMET on TICO English-Russian, whereas after fine-tuning on 18K examples COMET is 0.8234, a degradation of 0.0074.
The COMET scores correlate negatively with data size ($\rho=-0.27$, $p<0.001$), indicating that fine-tuning on more data results in larger degradations.

\begin{figure*}[!htb]
    \centering
    \includegraphics[width=\linewidth]{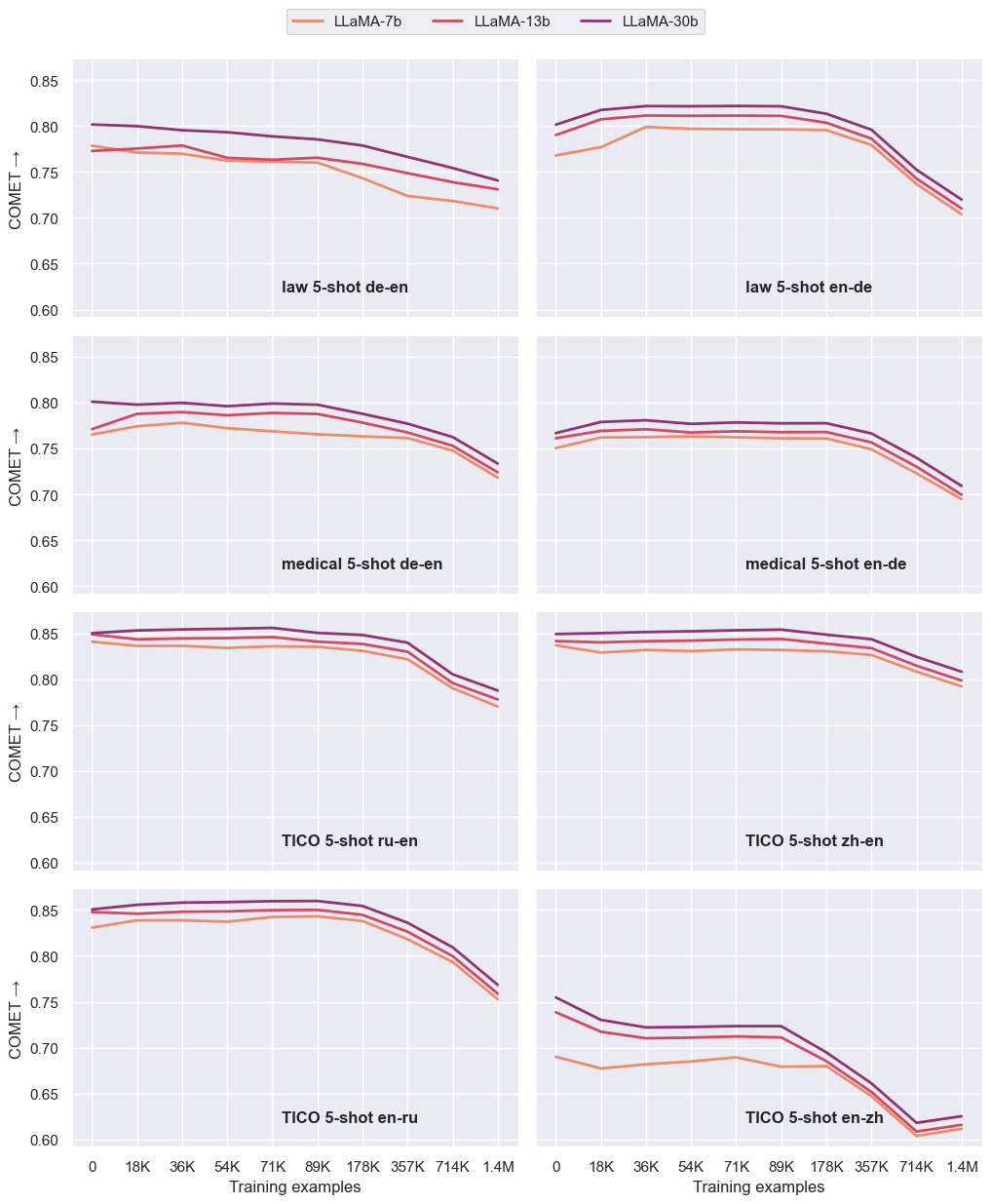}
    \caption{COMET on technical domains using 5-shot examples for models trained on OPUS data. 
    }
    \label{ch5:fig:opus_technical_domains_full}
\end{figure*}

The effects of fine-tuning are also analyzed using filtered web-scraped data from OPUS, as shown in Figure \ref{ch5:fig:opus_technical_domains_full}.
Similar to the previous findings, an increase in data volume for fine-tuning corresponds to performance degradations, evidenced by a negative correlation between COMET scores and datastore size ($\rho=-0.33$, $p<0.001$).
However, OPUS data reveals that these degradations manifest more gradually compared to the WMT dataset.
This discrepancy is likely due to OPUS's broader domain coverage, in contrast to the specialized news content of the human-curated WMT dataset.
Notably, while fine-tuning on OPUS data leads to deterioration in technical domain translation accuracy when leveraging few-shot examples, it concurrently continues to enhance overall translation quality (Figure \ref{ch5:fig:opus_general_quality_wmt}), underscoring a nuanced impact of fine-tuning across different data types and translation tasks.

\begin{figure}[!htb]
    \centering
    \includegraphics[width=\linewidth]{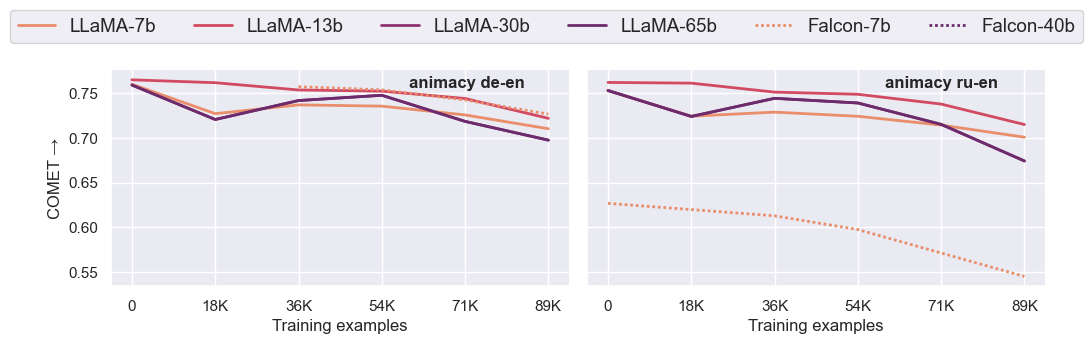}
    \caption{Accuracy of animacy contextualization for German$\rightarrow$English and Russian$\rightarrow$English for models fine-tuned on human-written translations.
    }
    \label{ch5:fig:wmt_animacy}
\end{figure}

\subsubsection*{Contextualization ability on document-level input degrades}
Our analysis extends to the animacy contextualization accuracy of document-level input.
We show results for models trained on human-written data in Figure \ref{ch5:fig:wmt_animacy}.
Mirroring the trend observed in formality steering, a clear degradation in contextualization accuracy is noted when fine-tuning the models on parallel data.
Again, we observe that fine-tuning on only 18K examples results in a decline of this ability, even though general translation quality on WMT continues to improve.
For example, Falcon-40b scores 0.91 before fine-tuning, which degrades to 0.85 after 18K examples.
The decline can be summarized by a negative correlation between accuracy and the size of the dataset used for fine-tuning ($\rho=-0.55$, $p<0.001$), indicating that contextualization abilities further degrade when fine-tuning on more data.
This trend is not exclusive to WMT data.

\begin{figure}[!htb]
    \centering
    \includegraphics[width=\linewidth]{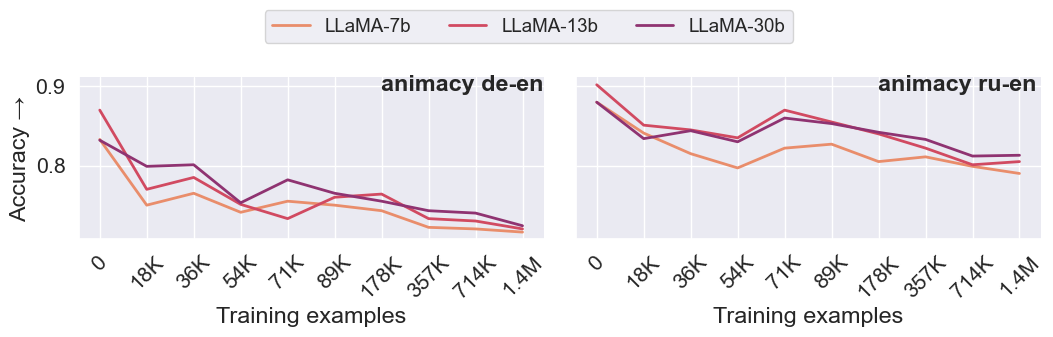}
    \caption{Accuracy of animacy contextualization for German$\rightarrow$English and Russian$\rightarrow$English for models fine-tuned on OPUS data.
    }
    \label{ch5:fig:opus_animacy}
\end{figure}

A similar pattern emerges when analyzing models trained on larger datasets.
Figure \ref{ch5:fig:opus_animacy} shows that the animacy contextualization accuracy of document-level input degrades for models fine-tuned on filtered web-crawled OPUS data.
We observe a negative correlation between accuracy and fine-tuning dataset size ($\rho=-0.49$, $p<0.001$).

\subsubsection*{Performance on idioms remains stable or improves}
\begin{figure}[!htb]
    \centering
    \includegraphics[width=0.75\linewidth]{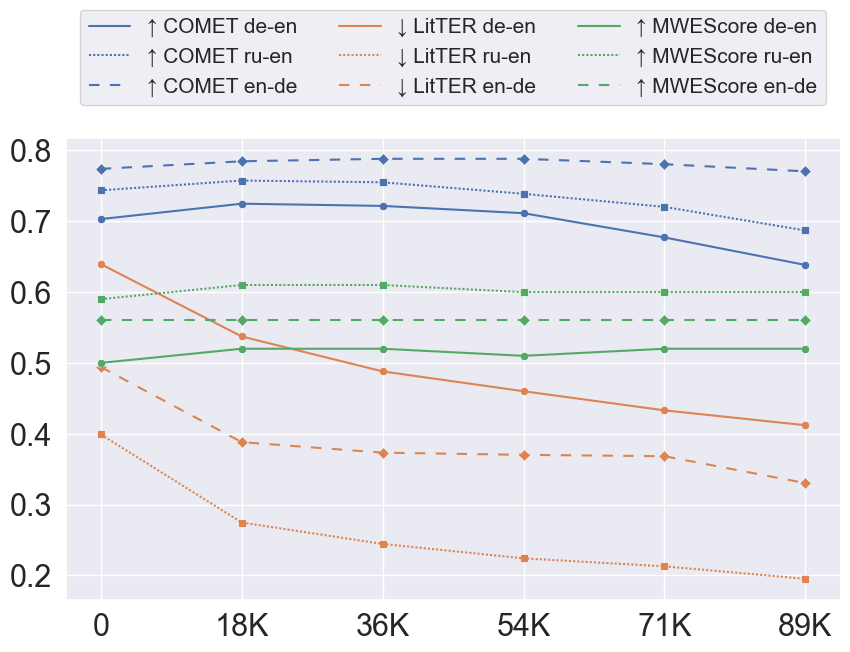}
    \caption{COMET, LitTER and MWEScore on IdiomsInCtx-MT test sets for LLaMA-65b fine-tuned on human-quality parallel data.}
    \label{ch5:fig:idioms}
\end{figure}

We evaluate the quality of idiomatic translations on the IdiomsInCtx-MT test sets.
Since idiomatic translation is inherently difficult, we focus our analysis on the strongest model, LLaMA-65b.
Figure \ref{ch5:fig:idioms} shows COMET, LitTER (lower is better) and MWEScore (higher is better) across training checkpoints for 1 epoch.

While LitTER uses input-specific block lists to assess the literalness of translation outputs based on annotations of idiomatic expressions in the input, MWEScore additionally relies on one or more gold translations of those idiomatic expressions and computes a score based on edit distance of output vs reference idiom tokens. 

Similar to the WMT22 test set, we see that COMET scores improve until the first 1-3 checkpoints before stabilizing or starting to decrease.
However, for idiomatic expressions even the targeted metrics LitTER  and MWEScore improve during fine-tuning or least remain stable.
This indicates that even a large open-source model like LLaMA-65b still has room for improvement when it comes to idiomatic translations.

Future work could investigate translation literalness and idiomaticity of LLM translations on stronger models such as GPT-3.5 during fine-tuning.\footnote{https://openai.com/blog/gpt-3-5-turbo-fine-tuning-and-api-updates}

\section{Fine-tuning on a mix of monolingual and parallel data}\label{ch5:sec:method_parallel_mono_ft}
Having established that fine-tuning on parallel data leads to a decline in various advantages of LLMs for machine translation, this section explores strategies to mitigate this degradation.

A potential approach to counteract degradation involves incorporating examples of desired behaviors during fine-tuning.
For instance, the degradation of few-shot capabilities for domain adaptation can be partially mitigated by including few-shot examples during fine-tuning \citep{alves_steering_2023, moslem_fine-tuning_2023}.
However, our aim is to establish a more general solution that prevents degradation across a broader spectrum of behaviors, without the need to specifically include data for each behavior during fine-tuning.
To achieve this, our experiments involve a blend of monolingual and parallel data in the fine-tuning phase.
This strategy stems from the understanding that the pre-training on monolingual data contributed to these beneficial phenomena, and our goal is to retain these qualities during fine-tuning.

Our mixed-data approach addresses the interference problem in a way that complements our findings in Section \ref{ch3:sec:transfer_m2m_models}, where we introduced the Representational Transfer Potential (RTP) metric to quantify cross-lingual transfer.
While Section \ref{ch3:sec:transfer_m2m_models} focused on enhancing positive transfer between languages through multi-parallel training, here we mitigate negative transfer during fine-tuning.
These investigations demonstrate different sides of the same challenge: how to selectively share knowledge in neural translation systems while preventing unwanted interference.

\subsection{Experimental setup}
To construct our monolingual dataset, we use the News-Commentary data from WMT22.
This dataset includes document-level information, which we preserve by concatenating sentences within each paragraph to form a single data entry.
The resulting processed data closely resembles the type of data used for LLM pre-training.
We then merged this monolingual dataset with parallel data sourced from OPUS (we sample 89K parallel examples), maintaining a 1:1 ratio and resulting in a total of 178K examples.
We refer to this arrangement as the FT 1:1 setup.
We compare this setup to several baselines: 1) the base model prior to any fine-tuning; 2) the FT 1:0 setup, which involves fine-tuning exclusively on parallel OPUS data (89K); and 3) the FT 2:0 setup, where fine-tuning is conducted on parallel OPUS data equal in size to our mixed monolingual and parallel dataset, totaling 178K examples.
We use LLaMA-7B with a context window size of 2048 tokens for this experiment.

\FloatBarrier
\begin{table*}[!ht]
    \begin{center}
        \small
        \begin{tabular}{lrrrrrrr}
        model       & de-en     & ru-en     & zh-en     & en-de     & en-ru     & en-zh     & avg       \\\toprule
        base        & $0.8217$  & $0.8163$  & $0.7477$  & $0.8161$  & $0.8151$  & $0.7279$  & $0.7903$  \\
        FT 1:0      & $0.8342$  & $0.8249$  & $0.7435$  & $0.8381$  & $0.8378$  & $0.7771$  & $0.8093$ \\
        FT 2:0      & $0.8360$  & $0.8277$  & $0.7472$  & $\mathbf{0.8400}$  & $0.8436$  & $0.7922$ & $0.8145$ \\\midrule
        FT 1:1      & $\mathbf{0.8380}$  & $\mathbf{0.8280}$  & $\mathbf{0.7519}$  & $0.8375$  & $\mathbf{0.8484}$  & $\mathbf{0.8053}$ & $\mathbf{0.8182}$ \\
        \bottomrule
        \end{tabular}
    \end{center}
    \caption{COMET scores on WMT22. Best scores in \textbf{bold}. Including both parallel and monolingual data during fine-tuning (FT 1:1) results in better translation performance compared to parallel-only fine-tuning (FT 1:0, FT 2:0).}
    \label{ch5:tab:mono_results_general}
\end{table*}

\subsection{Results}

\subsubsection*{General translation quality  improves further}
The results, as detailed in Table~\ref{ch5:tab:mono_results_general}, show the comparative performance of the baselines and the integration of monolingual data during the fine-tuning phase on general translation quality (WMT22) as measured by COMET.
Including monolingual data (FT 1:1) leads to translations that generally surpass those produced by parallel-only fine-tuning approaches (FT 1:0 and FT 2:0).
Notably, the most significant improvement is observed in the en-zh direction, where the FT 1:1 setup yields an increase of 0.0131 COMET over the best baseline (FT 2:0).
This can be attributed to the pre-training of LLaMA on an English-centric corpus, which contains only minimal amounts of (accidental) Chinese data.
As suggested by \citet{xu_paradigm_2024}, out-of-English capabilities of the model can be substantially augmented through an additional monolingual fine-tuning step, a methodology akin to our approach.

\begin{figure*}[!htb]
    \centering
    \begin{subfigure}[b]{0.49\linewidth}
        \includegraphics[width=\linewidth]{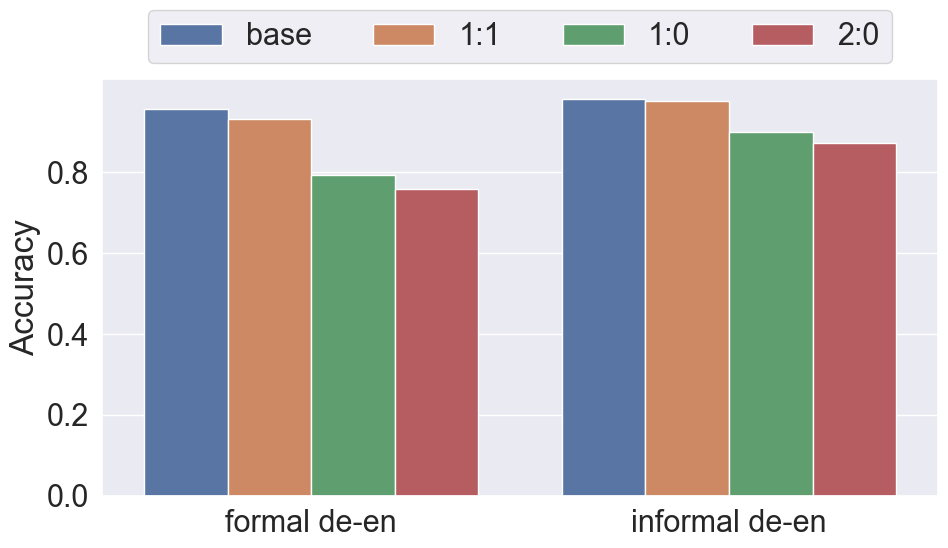}
        \caption{Formality steering}
        \label{ch5:fig:sub1}
    \end{subfigure}
    \hfill
    \begin{subfigure}[b]{0.49\linewidth}
        \includegraphics[width=\linewidth]{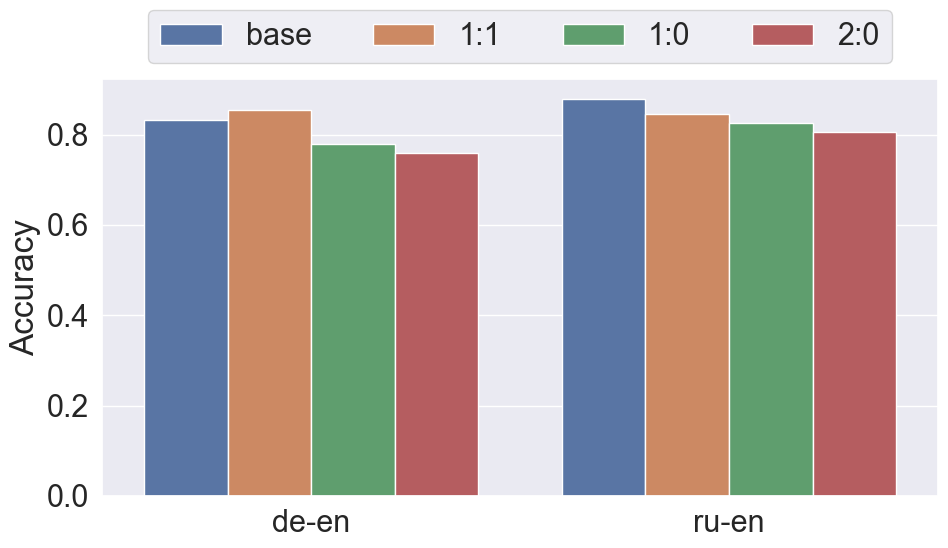}
        \caption{Contextualization}
        \label{ch5:fig:sub2}
    \end{subfigure}
    \par\bigskip
    \begin{subfigure}[b]{0.49\linewidth} 
        \includegraphics[width=\linewidth]{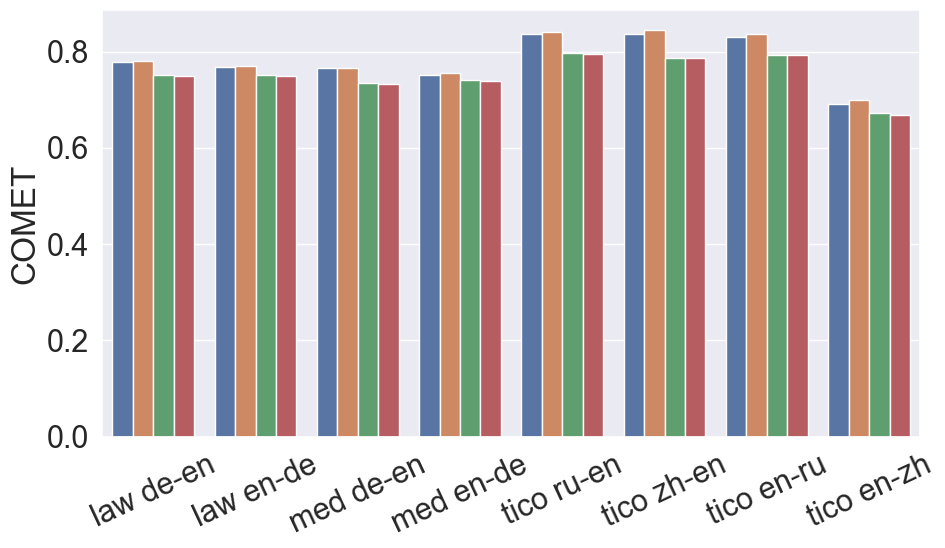}
        \caption{Technical translations}
        \label{ch5:fig:sub3}
    \end{subfigure}
    \hfill
    \begin{subfigure}[b]{0.49\linewidth}
        \includegraphics[width=\linewidth]{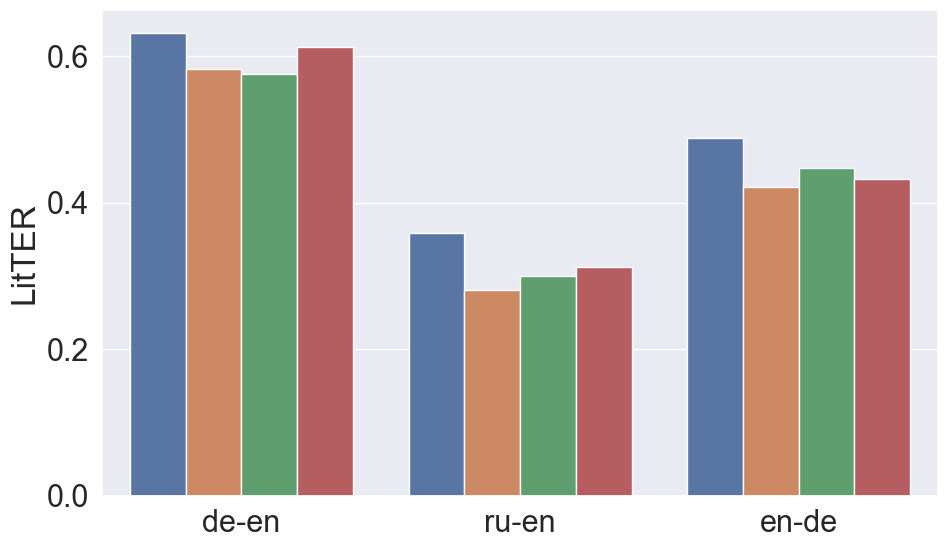}
        \caption{Literalness}
        \label{ch5:fig:sub4}
    \end{subfigure}
    \caption{Formality steering accuracy scores (a), Animacy contextualization accuracy of document-level input (b), COMET for few-shot technical translations (c), and LitTER scores for idioms. Base is LLaMA-7b before fine-tuning, 1:1 is fine-tuned on 89K parallel and 89K monolingual data, 1:0 on 89K parallel data, and 2:0 on 178K parallel data. Integrating monolingual data generally results in more preservation of capabilities.}
    \label{ch5:fig:mono_results_qualitative}
\end{figure*}

While the enhancement of general translation quality is a beneficial outcome, our primary interest lies in evaluating the ability of our method to preserve and possibly enhance the qualitative behaviors inherent in Large Language Models. This aspect forms the next focus of our investigation.
Figure~\ref{ch5:fig:mono_results_qualitative} shows a comparison between our fine-tuning method on formality steering, document-level contextualization, technical translation, and idiom translation tasks.
A consistent trend is observed: the integration of monolingual data with parallel data during fine-tuning generally results in more effective preservation of various translation capabilities.

\subsubsection*{Minimal formality steering degradation}
Baselines using only parallel data show accuracy drops as great as 0.198 for formal and 0.11 for informal steering.
The inclusion of monolingual data mitigates this degradation, reducing it to just 0.025 for formal and 0.007 for informal steering, albeit some degradation persists when compared to the baseline.

\subsubsection*{Degradation of contextualization abilities are lessened}
The impact of combining parallel and monolingual data during fine-tuning is also evident here.
For translations from German to English and Russian to English, the loss in accuracy is up to 0.073 and 0.075, respectively, when using only parallel data.
Incorporating monolingual data diminishes this degradation for Russian to English translations (-0.035), and even shows a notable improvement of +0.021 over the base model.

\subsubsection*{Technical domain performance is enhanced}
The inclusion of monolingual data during fine-tuning enhances performance for English into Russian and Chinese, and vice versa.
For instance, in the TICO English to Chinese translation task, the blended approach of monolingual and parallel data in fine-tuning yields a +0.0089 COMET score improvement over the baseline.
Conversely, relying solely on parallel data results in a 0.022 COMET score decrease (FT 2:0), indicating a substantial differential of 0.03.
For English in and out of German in both the Law and Medical domain, the differences between fine-tuning with monolingual data plus parallel data and the base model are minimal.
In contrast to using parallel data only, we observe no decline.
Notably, we use parallel data up to 178K (FT 2:0), where degradation is relatively modest in the case of few-shot technical domain translation.
When doing extended fine-tuning, this capability will further degrade, as we show in Section \ref{sec:parallel_ft_results}.

\subsubsection*{Performance on idioms improves}
Including monolingual data (1:1) improves the literalness of translations as measured by LitTER for ru-en and en-de.
However, LLaMA-7b does not demonstrate good performance on idiomatic translations, making it an easy baseline to improve upon.

These findings underscore the nuanced benefits of incorporating monolingual data when fine-tuning English-centric LLMs for machine translation tasks, specifically in preserving task-relevant capabilities.
However, for long-term progress, we advocate for the development of LLMs with multilingual data in mind.
In this approach, parallel data would be combined with monolingual data during the pre-training phase \citep{anil_palm_2023,wei_polylm_2023}.
Nevertheless, even when beginning with a more robust multilingual LLM for translation purposes, the exploration of fine-tuning strategies that preserve the emerged capabilities of LLMs remains critical, especially when adapting these models for various use cases and objectives.

\section{Conclusion}\label{ch5:sec:conclusion}
In this chapter, we have investigated how fine-tuning on parallel data affects the qualitative advantages of LLMs for machine translation.
We asked:\\

\RQsub{3}{1}
\begin{myquote}
While previous research predominantly focused on summary quality metrics like COMET, our findings in Section \ref{sec:parallel_ft_results} reveal a more complex interplay between fine-tuning and LLM capabilities.
Consistent with prior work, we find that fine-tuning enhances the general translation quality of LLMs, as shown in Figures \ref{ch5:fig:wmt_general_quality_wmt} and \ref{ch5:fig:opus_general_quality_wmt}.
However, we show that fine-tuning adversely impacts several important qualitative advantages of LLMs that are important for machine translation.
We observe declines in the abilities of LLMs to 1) perform formality steering (Figure \ref{ch5:fig:wmt_formality_steering}), 2) perform technical translation through few-shot examples (Figure \ref{ch5:fig:wmt_technical_domains_full}), as well as 3) a decrease in their document-level translation capabilities (Figure \ref{ch5:fig:wmt_animacy}).
We introduce a novel evaluation dataset, IdiomsInCtx-MT, to measure non-literalness performance.
The ability to produce non-literal translations shows improvement post fine-tuning, likely because the publicly available LLMs we investigate do not perform strongly on this task to begin with.
\end{myquote}

\RQsub{3}{2}
\begin{myquote}
Having established which capabilities are affected by fine-tuning, we addressed the role of data and model size.
Our results demonstrate that these degradations manifest even when using only 18k fine-tuning samples.
Furthermore, the degradations become more pronounced with larger fine-tuning datasets, even as generic translation quality continues to improve.
These trends are consistent across all tested model scales (7b up to 65b), underscoring the generalizability of our findings.
\end{myquote}

\RQsub{3}{3}
\begin{myquote}
Based on our findings from RQ3.1 and RQ3.2, we developed a solution to mitigate capability loss.
We develop a fine-tuning method tailored for machine translation, that combines monolingual and parallel data.
As demonstrated in Table \ref{ch5:tab:mono_results_general} and Figure \ref{ch5:fig:mono_results_qualitative}, this approach mitigates the degradation of LLMs' qualitative advantages, thereby preserving their capabilities while improving general translation quality.
Moreover, overall translation quality is enhanced to a greater extent compared to fine-tuning on parallel data alone, highlighting the dual benefits of our approach.
\end{myquote}

Taken together, these questions answered our main research question:

\RQ{3}
\begin{myquote}
Our findings clearly demonstrate that fine-tuning \textit{exclusively} on parallel data leads to a significant loss of capabilities that are valuable for machine translation.
We demonstrate that incorporating monolingual data during fine-tuning allows us to maintain these capabilities while simultaneously enhancing overall translation quality.
Most importantly, our work highlights the critical importance of developing fine-tuning approaches that not only improve general translation quality, but also preserve the distinctive capabilities that make LLMs particularly valuable for machine translation.
\end{myquote}

\subsubsection{Limitations}
Due to the high cost of repeatedly fine-tuning LLMs of different sizes, we limited ourselves to six translation directions (German, Chinese, and Russian in and out of English).

The impact of finetuning on emergent abilities of LLMs when translating in and out of low-resource languages is not studied in our work.
It is therefore possible that our findings do not generalize to low-resource languages.

\chapter{The Effect of Language Diversity in LLM Translation}
\label{ch6}

\renewcommand{\thefootnote}{}
\footnotetext{This chapter is accepted for publication and will be published in the \textit{Findings of the Association for Computational Linguistics: EMNLP 2025} as:
David Stap and Christof Monz.
The Effect of Language Diversity When Fine-Tuning Large Language Models for Translation.}

\renewcommand{\thefootnote}{\arabic{footnote}}

\section{Introduction and research questions}
\label{ch6:sec:introduction}

In previous chapters, we examined cross-lingual knowledge transfer in multilingual NMT models (Chapters \ref{ch3} and \ref{ch4}).
While in Chapter \ref{ch5} we examined the trade-offs involved in fine-tuning LLMs for machine translation and preserving their capabilities. Here, we focus specifically on how language diversity during LLM fine-tuning affects both seen and unseen languages.

General-purpose LLMs like \textsc{Llama~3} \citep{llama_team_llama_nodate} show promise for machine translation but require targeted fine-tuning beyond their incidental bilingualism \citep{briakou_searching_2023} to match specialized translation systems.
Through fine-tuning approaches ranging from two-stage methods \citep{li_eliciting_2023,zeng_teaching_2024} including our work in Chapter \ref{ch5} \citep{stap_finetuning_2024} to more sophisticated optimization techniques \citep{xu_contrastive_2025,zhu-etal-2024-preference}, LLMs such as \textsc{Tower} \citep{alves_tower_2024} now outperform traditional NMT systems \citep{kocmi-etal-2024-findings,deutsch_wmt24_2025}.

Current research presents conflicting evidence regarding the optimal multilingual fine-tuning strategy.
In this chapter, we set out to resolve these disparities.
We ask the following research question:

\RQ{4}
\begin{myquote}
While some studies show that scaling the number of tasks or languages during instruction tuning improves (cross-lingual) generalization \citep{wang_super-naturalinstructions_2022, muennighoff-etal-2023-crosslingual, dang-etal-2024-rlhf}, others report that just 1--3 fine-tuning languages effectively trigger cross-lingual transfer \citep{kew_turning_2024,zhu_finetuning_2024}.\\

To resolve these disparities and address our overarching question, we start by systematically increasing the number of language pairs during fine-tuning:
\end{myquote}

\RQsub{4}{1}
\begin{myquote}
Recent \textit{inference-only} experiments by \citet{richburg_how_2024} across 132 translation directions show variance in translation quality with off-target generations for non-English sources and inconsistent performance across languages.
While non-English over-tokenization and typological distance provide partial explanations, controlled fine-tuning experiments on the effects of language diversity \textit{during fine-tuning} remain unexplored.\\

In contrast to these inference-only studies, we conduct controlled fine-tuning experiments across 132 translation directions, systematically increasing language diversity during the fine-tuning process to examine its effects on translation quality, cross-lingual transfer, and model representations.\\

The effects of language diversity may vary across different model architectures and training conditions.
This leads to our second sub-question:
\end{myquote}

\RQsub{4}{2}
\begin{myquote}
To address this question, we examine whether the patterns observed with language diversity are consistent across different model sizes and under various fine-tuning conditions.
We explore whether these effects persist with different types of training data, regularization methods, and model scales, providing a more comprehensive understanding of the relationships between diversity, model capacity, and training approaches.\\

Finally, we design experiments to understand representational changes that are related to language diversity, by asking:
\end{myquote}

\RQsub{4}{3}
\begin{myquote}
Through representational analysis, we examine how the internal model representations adapt when exposed to different levels of language diversity during fine-tuning.
This analysis provides insights into the nature of cross-lingual transfer and helps explain the observed performance patterns across different language combinations.
\end{myquote}
By resolving contradictions in prior work and providing empirical evidence on the effects of language diversity during fine-tuning, this research contributes recommendations for developing more effective multilingual translation systems while advancing our theoretical understanding of cross-lingual knowledge transfer in LLMs.

\paragraph{Organization.}
This chapter is organized as follows: In Section~\ref{ch6:sec:related_work} we review prior work. Section~\ref{ch6:sec:language_diversity} presents our experimental design and results for investigating the effect of language diversity across 132 translation directions. We then conduct representational analysis to understand the underlying mechanisms of language transfer in these models in Section \ref{ch6:sec:representational_analysis}. Finally, we summarize our findings in relation to our research questions in Section \ref{ch6:sec:conclusion}.

\section{Related work}
\label{ch6:sec:related_work}

In this section, we review existing work on multilingual LLMs for translation and the evaluation of cross-lingual generalization, providing context for our investigation into the effects of language diversity during fine-tuning.        

\subsection{The \textsc{TowerLLM} architecture}
The \textsc{Tower} family of models \citep{alves_tower_2024} consists of several variants built upon \textsc{LLaMA 2} \citep{touvron_llama2_2023} with different parameter sizes and training stages.
\textsc{TowerBase} (available in 7B and 13B parameter sizes) extends \textsc{LLaMA 2}'s multilingual capabilities through continued pre-training on a mixture of monolingual and parallel data across 10 languages: English, German, French, Dutch, Italian, Spanish, Portuguese, Korean, Russian, and Chinese.
This continued pre-training uses a two-thirds monolingual to one-third parallel data ratio.
\textsc{TowerInstruct} models further fine-tune \textsc{TowerBase} on a specialized dataset (\textsc{TowerBlocks}) encompassing multiple translation-related tasks, including sentence-level translation, document-level translation, and automatic post-editing.

The \textsc{Tower} models demonstrate that incorporating parallel data during continued pre-training delivers superior performance compared to monolingual-only approaches, particularly for out-of-English translation directions.

While \textsc{TowerBase} includes monolingual data from multiple languages, its parallel data and \textsc{TowerBlocks} fine-tuning are English-centric, focusing on English-paired translation directions.
Our work extends beyond this English-centric approach by systematically incorporating non-English language pairs during \textsc{TowerBase} fine-tuning.
This broader language coverage allows us to investigate how diverse multilingual exposure affects translation quality, not only for English-centric directions but also for translation between non-English languages.

\subsection{Evaluating cross-lingual generalization}
\citet{richburg_how_2024} conducted an extensive empirical study examining the translation capabilities of LLMs across a diverse set of languages and translation directions.
Their work evaluated multiple models, including the \textsc{Tower} family and \textsc{NLLB}, across 132 translation directions involving 12 typologically diverse languages.

Their analysis revealed several important findings: While fine-tuned LLMs showed strong performance on language pairs seen during fine-tuning, they exhibited inconsistent behavior on unseen language pairs.
The authors identified persistent challenges with off-target generations (producing output in the wrong language) and significantly lower translation quality for certain languages like Korean and Icelandic, regardless of their inclusion in fine-tuning data.
Their study highlighted several factors that might contribute to these disparities, including over-tokenization of non-English languages by the LLM tokenizer and varying levels of language representation in the pre-training data.
The work by \citet{richburg_how_2024} is limited to observing the behavior of existing models rather than manipulating the conditions under which models were trained or fine-tuned, constraining the conclusions that can be drawn about the effects of language diversity during fine-tuning.

Our research directly addresses this gap by systematically investigating how different levels of language diversity during fine-tuning affect translation performance across both seen and unseen language pairs.
Unlike inference-only evaluations, our approach combines controlled fine-tuning experiments with comprehensive evaluation, allowing us to establish direct relationships between language diversity in training data and translation performance.
By systematically varying the language pairs used during fine-tuning while keeping other factors constant, we derive concrete, actionable insights about multilingual fine-tuning strategies.

\section{Investigating the effect of language diversity in LLM translation}
\label{ch6:sec:language_diversity}

\begin{table*}[!ht]
    \begin{center}
        \small
        \begin{tabular}{lllll}
        \toprule
        \textbf{Language} & \textbf{ISO} & \textbf{Script} & \textbf{LLaMA 2} & \textbf{Similarity} \\
         & \textbf{639-1} & & \textbf{support} & \textbf{groups} \\
        \midrule
        Czech            & \texttt{cs} & Latin    & 0.03\%          & West Slavic \\
        Polish           & \texttt{pl} & Latin    & 0.09\%          & West Slavic \\
        \textbf{Russian} & \texttt{ru} & Cyrillic & 0.13\%          & East Slavic \\
        Ukrainian        & \texttt{uk} & Cyrillic & 0.07\%          & East Slavic \\\midrule
        \textbf{German}  & \texttt{de} & Latin    & 0.17\%          & West Germanic \\
        \textbf{English} & \texttt{en} & Latin    & 89.70\%         & West Germanic \\
        Icelandic        & \texttt{is} & Latin    & possibly unseen & North Germanic \\
        \textbf{Dutch}   & \texttt{nl} & Latin    & 0.12\%          & West Germanic \\
        Swedish          & \texttt{sv} & Latin    & 0.15\%          & North Germanic \\\midrule
        Japanese         & \texttt{ja} & Kana     & 0.10\%          & Kanji from Hanzi \\
                         &             &          &                 & SOV order \\
        \textbf{Korean}  & \texttt{ko} & Hangul   & 0.06\%          & SOV order \\
        \textbf{Chinese} & \texttt{zh} & Hanzi    & 0.13\%          & Hanzi to Kanji\\
                         &             &          &                 & loanwords to \texttt{ja} and \texttt{ko} \\
        \bottomrule
        \end{tabular}
    \end{center}
    \caption{Evaluated languages with rationales for similarity grouping, following the language selection from \citet{richburg_how_2024}.
    Languages marked in \textbf{bold} belong to the supervised set used in the original \textsc{Tower} model continual pre-training.}
    \label{ch6:tab:lang_details}
\end{table*}

In this section, we present our methodology for investigating how language diversity during fine-tuning affects translation performance across various language pairs, describing our experimental design, data, and evaluation metrics.

Following \citet{richburg_how_2024}, we categorize our language pairs into three groups based on their presence in the fine-tuning data of the \textsc{Tower} model we build upon: 
\begin{itemize}
    \item \textbf{Fully supervised translation directions}: pairs between Chinese, Dutch, English, German, Korean, and Russian.
    These are language pairs where both languages were present in the original \textsc{Tower} model's fine-tuning data, meaning the model has been explicitly trained on these translation directions.
    This results in 30 pairs.
    \item \textbf{Zero-shot translation directions}: pairs involving Czech, Icelandic, Japanese, Polish, Swedish, and Ukrainian.
    These are language pairs where neither language was present in the original \textsc{Tower} model's fine-tuning data, requiring the model to perform translation without having seen these specific directions during training.
    This results in 30 pairs.
    \item \textbf{Partially supervised translation directions}: pairs combining one supervised language (from the first group) with one zero-shot language (from the second group).
    These represent an intermediate case where the model has training experience with one language but not the other in the translation pair.
    This results in 72 pairs.
\end{itemize}

The selection of languages is shown in Table \ref{ch6:tab:lang_details}.
It enables evaluation across varied typological properties and scripts while providing a systematic comparison between supervised languages (seen during fine-tuning) and zero-shot languages that share linguistic features with the supervised set.
The languages in the zero-shot set were chosen to represent both varying degrees of resource support in the pre-training data and to have relationships to languages in the supervised set through language family, typological properties, or orthography.

This yields 132 translation directions across 12 typologically diverse languages with varying pre-training representation, enabling comprehensive assessment across different data conditions.

\subsection{Experimental setup}
We now describe the experimental framework used to systematically evaluate the impact of language diversity during fine-tuning, including model configurations, data preparation, optimization strategies, and evaluation metrics.

\subsubsection{Fine-tuning setups}
We compare the following incremental fine-tuning approaches using the \textsc{Tower} family of models, which are built on \textsc{Llama 2} and underwent continued pre-training with a mixture of monolingual and parallel data:
\begin{itemize}
    \item \textsc{base}: \textsc{TowerBase-7B} model without task-specific fine-tuning, serving as our baseline.
    \item \textsc{fsec}: \textsc{base} fine-tuned only on fully supervised (\textsc{fs}) English-centric (\textsc{ec}) translation directions (10 directions), representing minimal supervision.
    \item \textsc{fs}: \textsc{base} fine-tuned on all fully supervised language directions (30 directions), extending beyond English-centric pairs to investigate transfer learning between diverse language combinations.
    \item \textsc{fs+ps+un}: \textsc{base} fine-tuned on fully supervised, partially supervised (\textsc{ps}), and unsupervised (\textsc{un}) directions (132 directions), maximizing language diversity to investigate cross-lingual transfer effects.
\end{itemize}

This controlled experimental design allows us to systematically evaluate how increasing language diversity during fine-tuning affects both supervised and unsupervised translation directions, moving beyond aggregate scores to understand performance patterns across specific language groups.

\subsubsection{Data}
We fine-tune on \textsc{NTREX-128} \citep{federmann_ntrex-128_2022}, a high-quality dataset of 1,997 multi-parallel professionally translated sentences designed for machine translation evaluation.\footnote{Preliminary experiments with additional \textsc{FLORES-200} (\texttt{dev}) data showed no significant improvements, so we exclude it for experimental clarity.}
For evaluation, we use the \textsc{FLORES-200} \citep{nllb_team_no_2022} \texttt{devtest} set, which provides multi-parallel data for controlled cross-language comparison.

\subsubsection{Optimization}
We conducted hyperparameter tuning on our development set (\textsc{FLORES-200} \texttt{dev}), exploring learning rate scheduler $\in\{\text{cosine, inverse square root}\}$, batch size $\in\{128, 256\}$, and learning rate $\in\{2 \times 10^{-5}, 2 \times 10^{-6}\}$.

For all experiments, we performed full fine-tuning using the AdamW optimizer \citep{loshchilov2018decoupled} with 5\% warm-up percentage and trained for one epoch.
Based on development set performance, we selected the optimal configuration: a cosine learning rate scheduler with batch size of 256 and learning rate of $2 \times 10^{-5}$.
We implemented our fine-tuning experiments using the Hugging Face transformers library \citep{wolf_transformers_2020} with DeepSpeed \citep{rasley_deepspeed_2020}.

\subsubsection{Metrics}
Our primary metric is \textsc{COMET-strict}, a modified version of \textsc{COMET} \citep{rei_comet_2020} that assigns zero scores to off-target translations, following recommendations by \citet{zouhar_pitfalls_2024}.\footnote{We use version \texttt{wmt22-comet-da}.}
We also report off-target rates, measured using \textsc{fastText} \citep{joulin-etal-2017-bag,joulin_fasttextzip_2016} language identification.\footnote{We use the \texttt{lid.176.bin} model.}

\begin{table*}[!ht]
    \begin{center}
        \small
        \begin{tabular}{l}
        \toprule
        \texttt{Translate this from \{source\_language\} to \{target\_language\}:} \\
        \texttt{\{source\_language\}: \{source\_sentence\}} \\
        \texttt{\{target\_language\}: \{target\_sentence\}} \\
        \bottomrule
        \end{tabular}
    \end{center}
    \caption{Prompting template for fine-tuning and 0-shot inference.
    For fine-tuning \texttt{\{target\_sentence\}} is filled with the corresponding target sentence, and for 0-shot inference it is the empty string.}
    \label{ch6:tab:prompt}
\end{table*}

\subsubsection{Inference}
For both fine-tuning and zero-shot inference, we used the prompt template shown in Table~\ref{ch6:tab:prompt}.
We mask out the prompt during fine-tuning.
We employed greedy decoding (beam size 1) to balance computational efficiency with comprehensive evaluation across all translation directions.

\begin{figure*}[!htb]
    \centering
    \includegraphics[width=.7\linewidth]{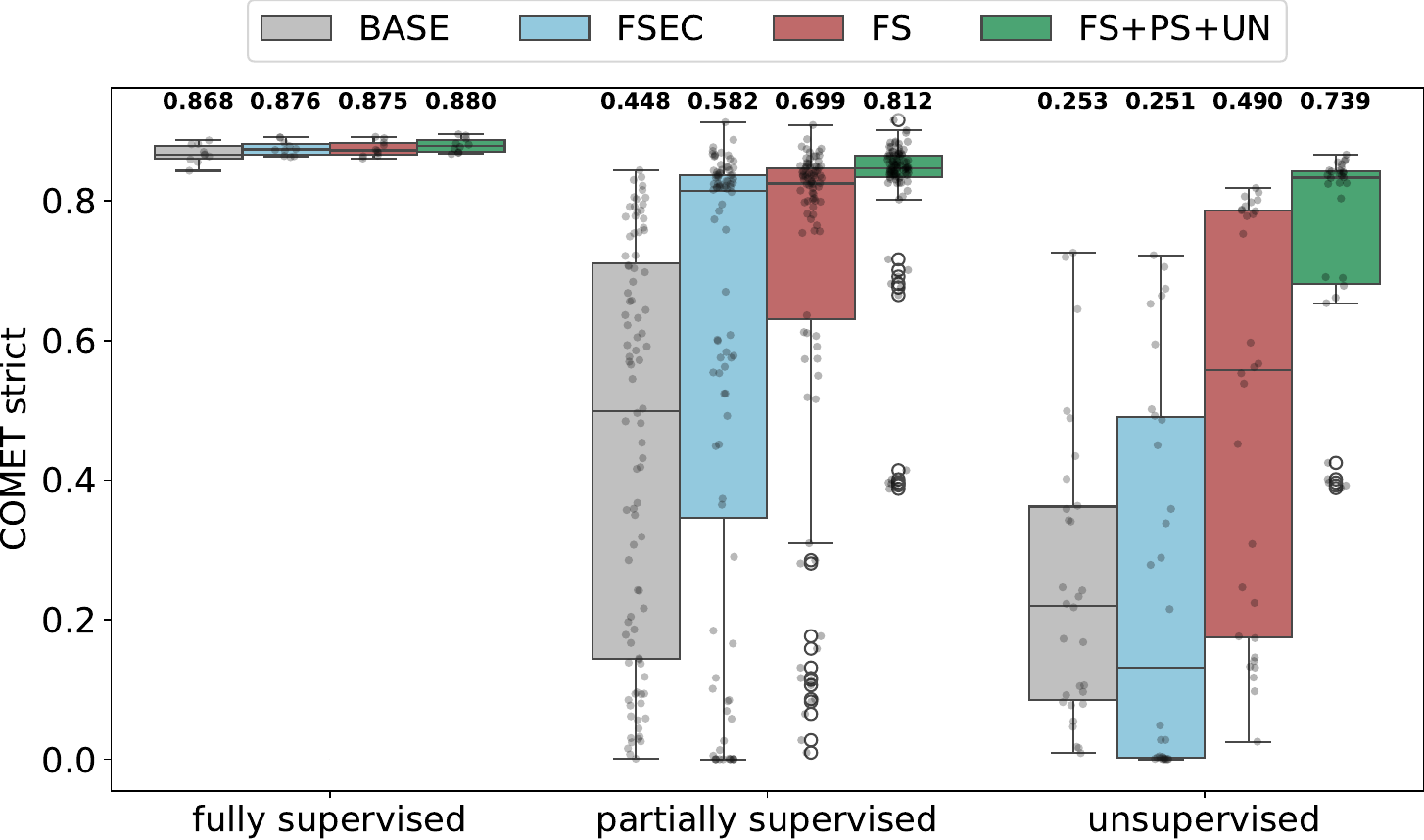}
    \includegraphics[width=.7\linewidth]{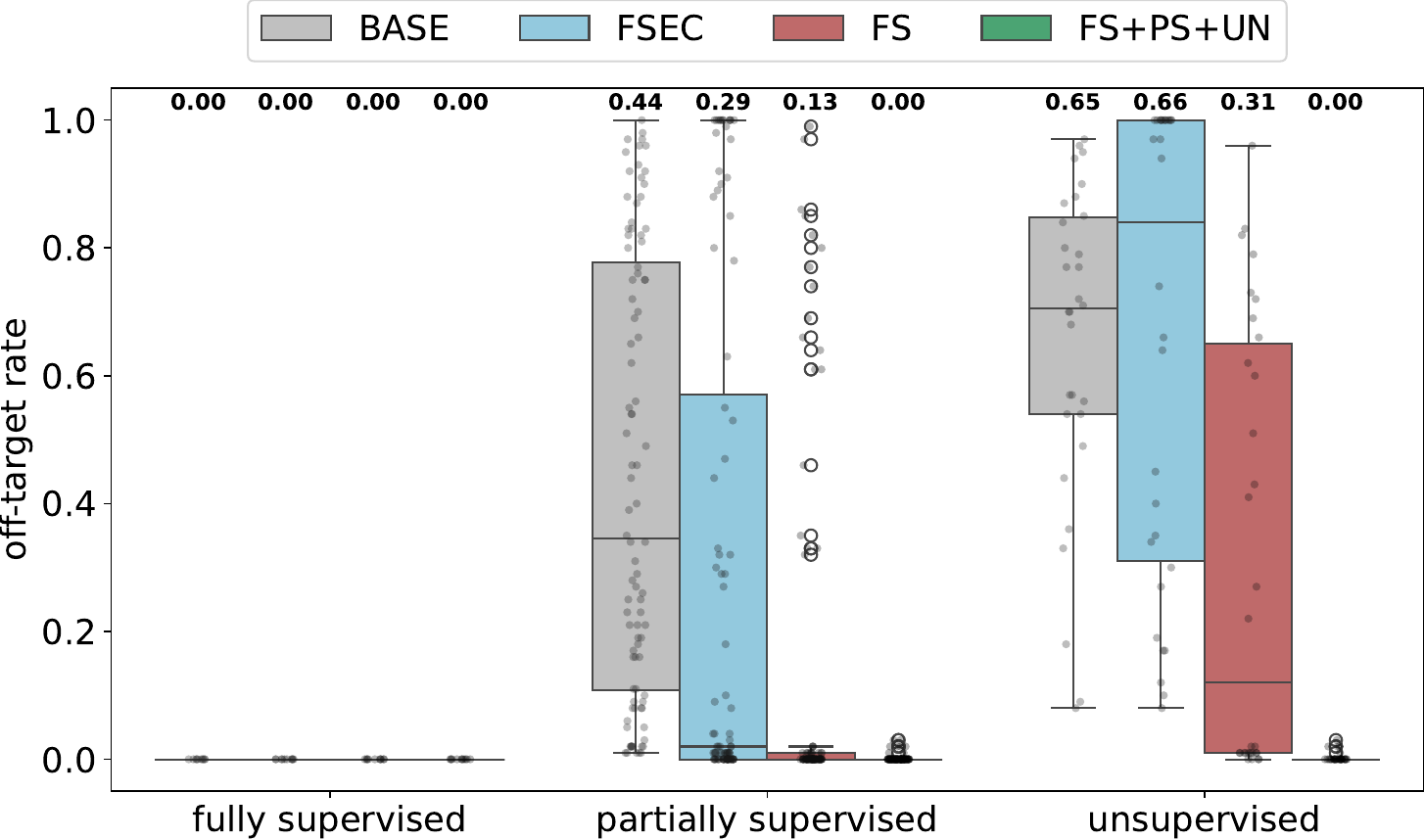}
    \caption{\textsc{COMET-strict} scores (top) and off-target rates (bottom) for \textsc{base} (no fine-tuning), \textsc{fsec} (English-centric), \textsc{fs} (seen directions), \textsc{fs+ps+un} (all directions), evaluated on \textit{fully supervised} (Chinese, Dutch, English, German, Korean, and Russian pairs), \textit{unsupervised} (Czech, Icelandic, Japanese, Polish, Swedish, and Ukrainian pairs), and \textit{partially supervised} (combining supervised and unsupervised) language pairs.
    Numbers above bars show mean scores.
    Training on more diverse sets improves \textit{all} categories, with \textsc{fs+ps+un} achieving best \textsc{COMET-strict} scores even for fully supervised pairs.
    \textsc{fs} substantially reduces off-target rates for unsupervised directions compared to \textsc{base} and \textsc{fsec}, despite these pairs being \textit{absent} from its fine-tuning data.}
    \label{ch6:fig:comet-strict-off-target-7b}
\end{figure*}

\subsection{Results}
In this section, we present the results of our fine-tuning experiments with varying degrees of language diversity, analyzing the impact on translation quality across different language combinations and model configurations.

\subsubsection{Increased diversity leads to better performance}
Figure~\ref{ch6:fig:comet-strict-off-target-7b} (top) demonstrates that expanding language diversity during fine-tuning yields consistent performance improvements across all language pair categories.
The \textsc{COMET-strict} scores show a clear progression from \textsc{base} to \textsc{fs+ps+un} models, with the most diverse model achieving highest scores in every category.
Surprisingly, the \textsc{fs+ps+un} model (fine-tuned on \textit{all} 132 directions) outperforms specialized models even on fully supervised language pairs (0.880 vs.\ 0.876 for \textsc{fsec}), despite the latter being specifically optimized for these directions.
The benefits become more pronounced for partially supervised (0.812 vs.\ 0.448 for \textsc{base}) and unsupervised (0.739 vs.\ 0.253 for \textsc{base}), though this improvement is expected as \textsc{fs+ps+un} is explicitly fine-tuned on these directions.

These results clarify conflicting evidence on language diversity (see Section \ref{ch6:sec:introduction}) and align with \citet{wang_super-naturalinstructions_2022} and \citet{dang-etal-2024-rlhf}, confirming that \textit{broad language diversity} (132 directions vs. 10--30), rather than minimal exposure, substantially enhances cross-lingual transfer, even for language pairs already well supported in more specialized models.

\subsubsection{Increased diversity reduces off-target problem}
Off-target translations, where models generate content in incorrect languages, represent a critical failure mode in LLM-based MT \citep{zhang_prompting_2023,guerreiro_hallucinations_2023,sennrich-etal-2024-mitigating}.

Figure~\ref{ch6:fig:comet-strict-off-target-7b} (bottom) shows that while all models maintain target language fidelity for fully supervised pairs, the \textsc{base} model produces incorrect target languages at alarming rates for partially supervised (44\%) and unsupervised pairs (65\%).
Fine-tuning progressively mitigates this problem, with \textsc{fs} showing substantial improvements (13\% and 31\% respectively) despite not being explicitly trained on these language combinations.
Significantly, the \textsc{fs+ps+un} model completely eliminates off-target translations across all categories.

\subsubsection{Diversity benefits plateau}
Expanding from \textsc{fs+ps+un} (132 directions) to 272 directions reveals nuances in the diversity-performance relationship.
To investigate whether further increasing language diversity yields additional benefits, we compared our most diverse model from the main experiments to an even more diverse setup including 272 translation directions. While maintaining a similar distribution of language families, we added five additional languages:

\begin{itemize}
    \item \textbf{Germanic family}: Danish (\texttt{da}, North Germanic) and Afrikaans (\texttt{af}, West Germanic)
    \item \textbf{Slavic family}: Slovak (\texttt{sk}, West Slavic) and Bulgarian (\texttt{bg}, South Slavic)
    \item \textbf{East Asian languages}: Vietnamese (\texttt{vi}, different writing system but shares vocabulary with Chinese)
\end{itemize}
This selection maintains balanced representation across language families while introducing controlled diversity within each family.
All additional languages are represented in both \textsc{NTREX} and \textsc{FLORES-200}.

Importantly, we evaluate both models on the same set of languages and directions as used throughout the chapter.
The additional languages are only used during fine-tuning to increase diversity, allowing us to measure their impact on the original set of translation directions.

\begin{figure*}[!htb]
    \centering
    \includegraphics[width=0.7\linewidth]{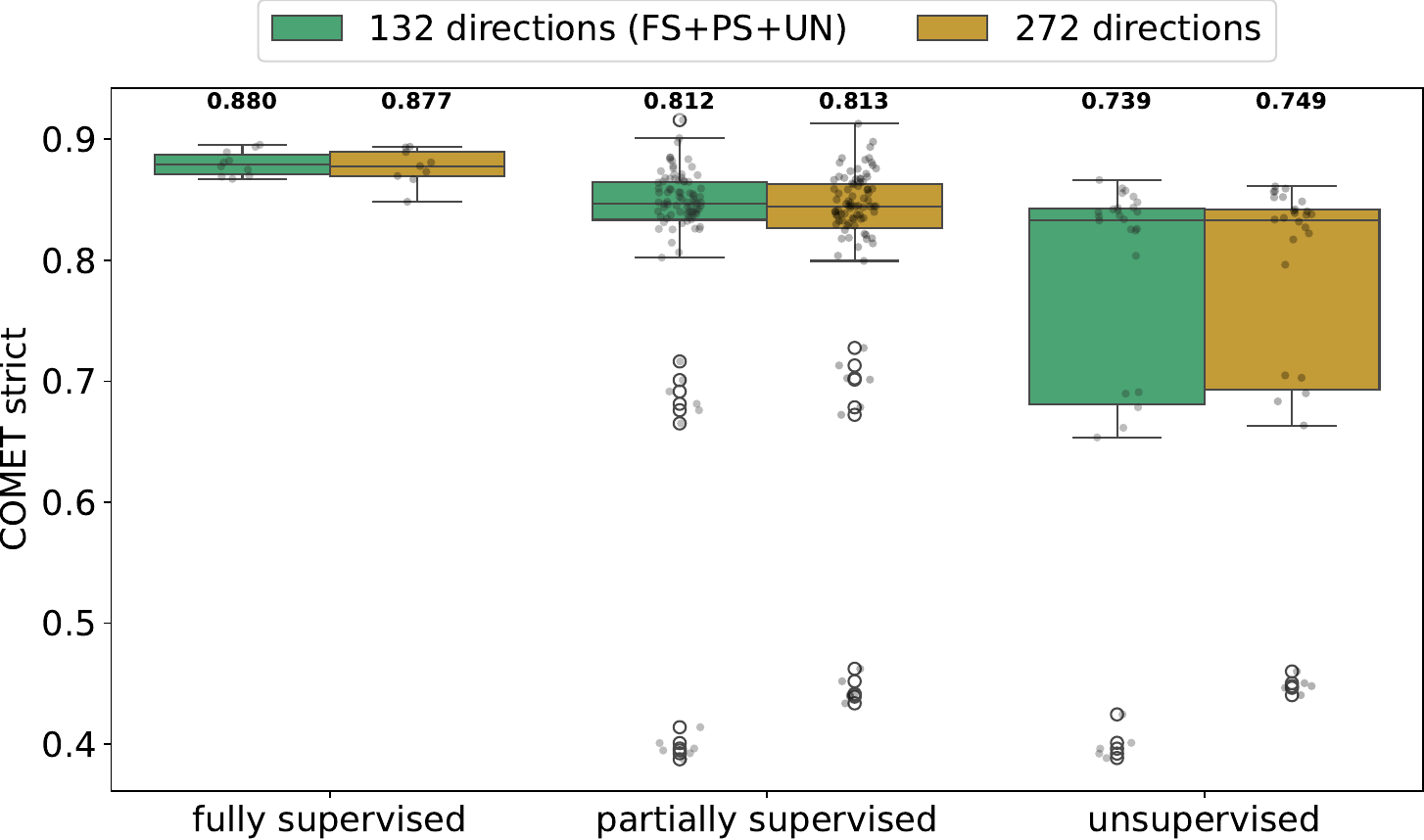}
    \caption{\textsc{COMET-strict} scores comparing models trained on 132 directions and 272 directions.
    Both are evaluated on the original test set with the same language pairs as used throughout the chapter.
    While the numerical differences are small, unsupervised directions show clearest benefits from increased diversity (+0.01), while fully supervised directions show a slight decrease (-0.003), suggesting potential diversity trade-offs.}    
    \label{ch6:fig:comet-strict-7b-272}
\end{figure*}

Figure \ref{ch6:fig:comet-strict-7b-272} shows the performance comparison between our 132-direction model (\textsc{fs+ps+un}) and the expanded 272-direction model.
Unsupervised directions benefit from increased diversity (+0.01), while fully supervised directions show slight performance decline (-0.003 \textsc{COMET-strict}), suggesting benefits plateau beyond a certain threshold.
Partially supervised directions show almost identical performance (+0.001).

These results suggest that language diversity benefits may plateau or even slightly decline for already well-represented language pairs.
The slight reduction in fully supervised performance may indicate a trade-off between focused optimization and broader generalization, where extremely high diversity can dilute the model's effectiveness for specific well-represented languages.
Nonetheless, the continued improvements for unsupervised directions demonstrate that higher diversity provides additional benefits for previously unseen language combinations.

This contradicts prior work that found monotonic improvements with diversity \citep{wang_super-naturalinstructions_2022,dang-etal-2024-rlhf}, but aligns with \citet{muennighoff-etal-2023-crosslingual}'s observation of diminishing returns when scaling multilingual pretraining beyond certain language counts.

\subsubsection{Regularization alone insufficient}
Regularization benefits models by enhancing generalization and calibration, with strong effects when using distant languages \citep{meng-monz-2024-disentangling}.
We investigate whether these benefits can be achieved through explicit regularization techniques rather than language diversity.

To test this hypothesis, we conduct additional experiments using stronger regularization on models with limited language diversity.
If increased language diversity primarily functions as a form of regularization, we hypothesize that similar improvements could be obtained by directly increasing regularization strength in less diverse models.

All our previous experiments use the AdamW optimizer with weight decay set to 0.01 and gradient clipping at 1.0, aligning with common LLM fine-tuning practices where dropout \citep{srivastava_dropout_2014} is rarely employed.
We tested the \textsc{fs} setup with increased weight decay values of 0.05 and 0.10 (compared to our standard 0.01) to examine whether stronger regularization would induce better cross-lingual transfer to partially supervised and unsupervised directions, potentially mimicking the benefits observed in the more diverse \textsc{fs+ps+un} model.

\begin{figure*}[!htb]
    \centering
    \includegraphics[width=0.7\linewidth]{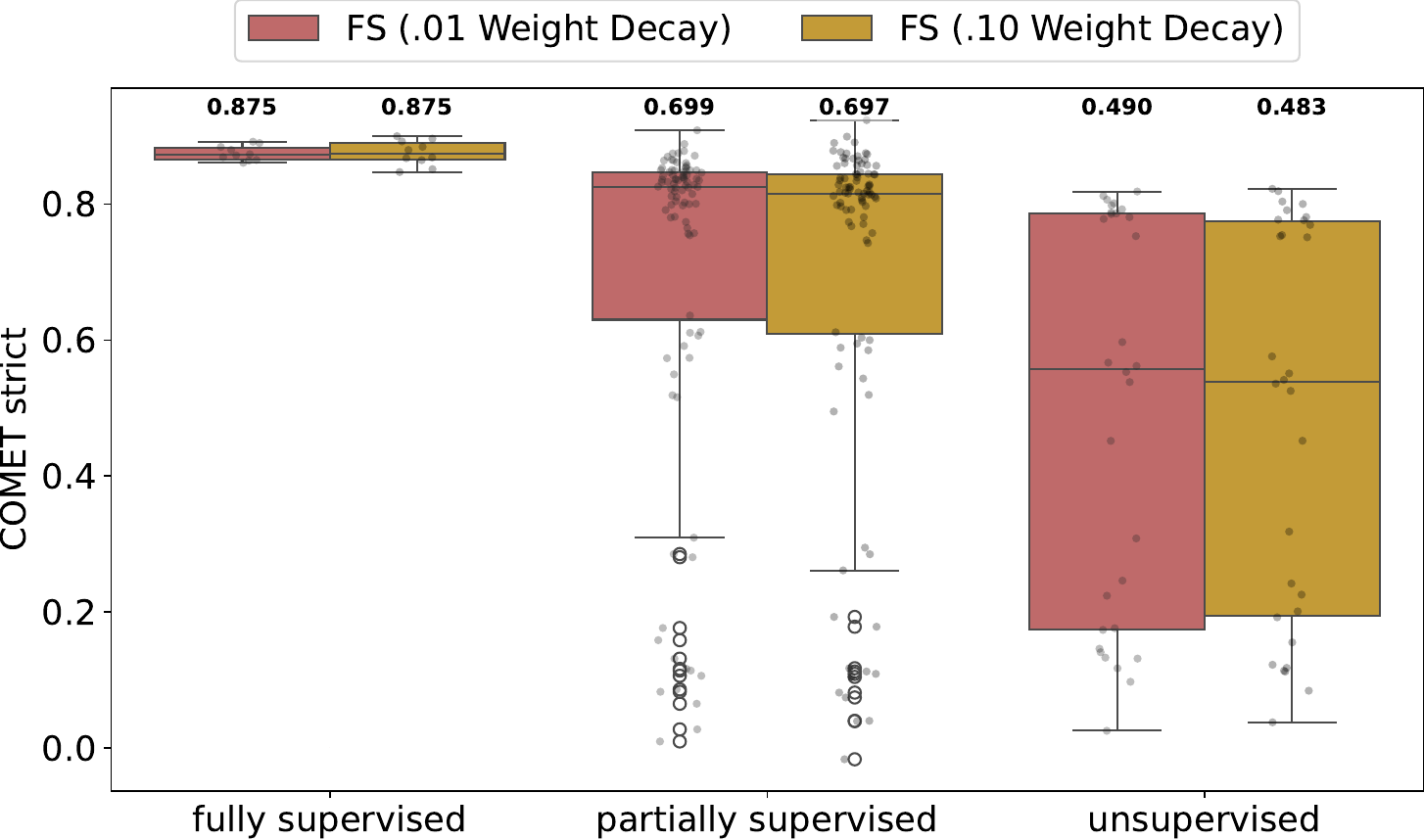}
    \caption{\textsc{COMET-strict} scores comparing \textsc{fs} models with weight decay values of 0.01 (standard) and 0.10.
    Increasing regularization strength shows minimal impact on fully supervised and partially supervised directions, while actually harming performance on unsupervised directions, suggesting that regularization alone cannot replicate the benefits of increased language diversity.}
    \label{ch6:fig:regularization}
\end{figure*}

Figure \ref{ch6:fig:regularization} shows the \textsc{COMET-strict} scores comparing \textsc{fs} models with different weight decay values.
Increasing regularization strength has minimal impact on translation performance across all language categories.
For fully supervised directions, both models achieved identical mean scores (0.875).
For partially supervised directions, the difference was negligible (0.699 vs.\ 0.697).
For unsupervised directions, the model with stronger regularization actually performed slightly worse (0.483 vs.\ 0.490).
The results for weight decay at 0.05 were very similar to 0.10 and are omitted for clarity.

We further explored alternative regularization approaches by implementing LoRA \citep{hu2022lora} with rank 64, which constrains fine-tuning to a low-dimensional subspace.
Results from LoRA experiments align with our weight decay findings: performance for fully and partially supervised directions remained comparable to full fine-tuning with standard regularization, while unsupervised directions showed slight degradation.

These experiments demonstrate that our initial weight decay value already provides an appropriate balance between overfitting prevention and model flexibility.
More importantly, they confirm that the cross-lingual transfer benefits observed in more diverse models cannot be replicated merely by increasing explicit regularization in less diverse models.
The language diversity benefits we observe go beyond simple regularization effects, providing specialized cross-lingual knowledge transfer.
This aligns with \citet{aharoni_massively_2019}, who suggest that multilingualism provides benefits beyond explicit regularization methods.

\subsubsection{Results not due to multi-parallel data}
While recent work by \citet{caswell_smol_2025} found that fine-tuning on multi-parallel data causes catastrophic forgetting in LLMs when trained on X$\rightarrow$\texttt{en} directions, our findings persist beyond multi-parallel settings.

To verify our findings are not artifacts of using multi-parallel data, we constructed a non-multi-parallel dataset from the \textsc{NLLB} corpus \citep{nllb_team_no_2022}.
We maintained the same 132 language directions as in our main experiments but eliminated the multi-parallel property.
Following \citet{koehn-2024-neural}, we extract examples with \textsc{LASER} \citep{artetxe_massively_2019} scores above 1.05.
We then removed sentences that appeared in multiple language pairs and sampled the remaining data to ensure exactly 2,000 examples per direction, creating a completely non-multi-parallel dataset of equivalent size to our NTREX experiments.

\begin{figure*}[!htb]
    \centering
    \includegraphics[width=0.7\linewidth]{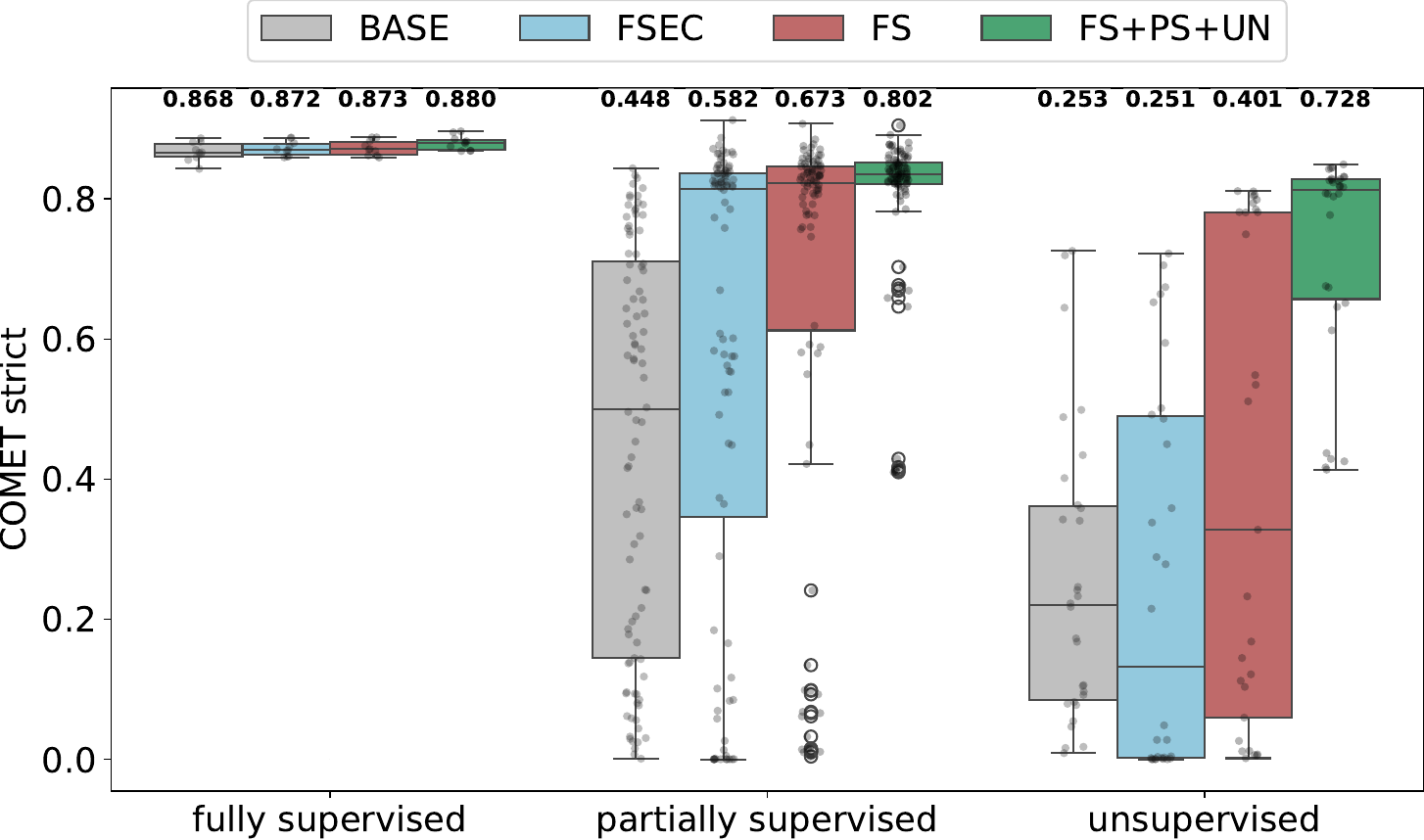}
    \includegraphics[width=0.7\linewidth]{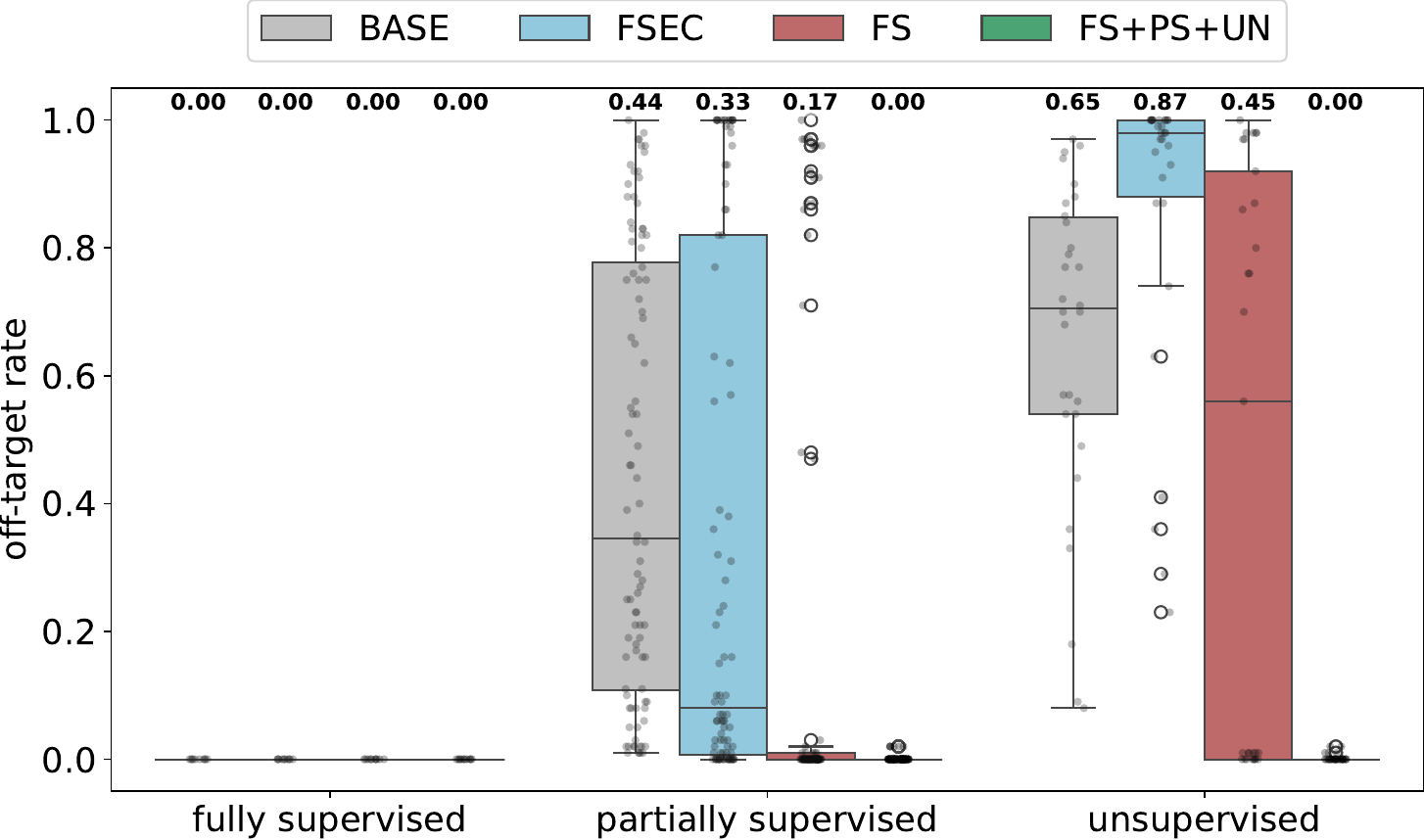}
    \caption{\textsc{COMET-strict} scores (top) and off-target rates (bottom) for 7B models fine-tuned on filtered non-multi-parallel NLLB dataset:  \textsc{base} (no fine-tuning), \textsc{fsec} (English-centric), \textsc{fs} (seen directions), \textsc{fs+ps+un} (all directions), evaluated on \textit{fully supervised} (Chinese, Dutch, English, German, Korean, and Russian pairs), \textit{unsupervised} (Czech, Icelandic, Japanese, Polish, Swedish, and Ukrainian pairs), and \textit{partially supervised} (combining supervised and unsupervised) language pairs.
    Numbers above bars show mean scores.
    The same observations from Figure \ref{ch6:fig:comet-strict-off-target-7b} (trained on multi-parallel data) remain true:
    training on more diverse sets improves \textit{all} categories, with \textsc{fs+ps+un} achieving best results even for fully supervised pairs;
    \textsc{fs} substantially reduces off-target rates for unsupervised directions compared to \textsc{base} and \textsc{fsec}, despite these pairs being \textit{absent} from its fine-tuning data.}
    \label{ch6:fig:comet-strict-off-target-7b-nllb}
\end{figure*}

Figure \ref{ch6:fig:comet-strict-off-target-7b-nllb} shows results from experiments conducted using the filtered \textsc{NLLB} dataset rather than \textsc{NTREX}.
The results demonstrate that our core finding—increased language diversity during fine-tuning leads to better performance—holds when using non-multi-parallel data as well.
The \textsc{fs+ps+un} model still achieves the highest \textsc{COMET-strict} scores across all language categories, including for fully supervised language pairs.

When comparing performance between models fine-tuned on \textsc{NLLB} versus \textsc{NTREX} data, we observe identical ranking patterns across different fine-tuning setups, though the \textsc{NTREX}-trained models show slightly better overall performance.
This marginal improvement is likely attributable to \textsc{NTREX}'s higher data quality, as it consists of professionally translated content specifically designed for machine translation evaluation, which aligns with our findings in Chapter \ref{ch5} where we observed that high-quality human-written translations often yielded better performance than web-scraped parallel data for the same data budget.

Unlike the overfitting issues reported for LLMs in prior work, our models maintain performance, consistent with studies showing multi-parallel data benefits in NMT, including our findings in Chapter \ref{ch3} \citep{stap_viewing_2023} and work by \citet{wu-etal-2024-far}.

\begin{figure*}[!htb]
    \centering
    \includegraphics[width=0.7\linewidth]{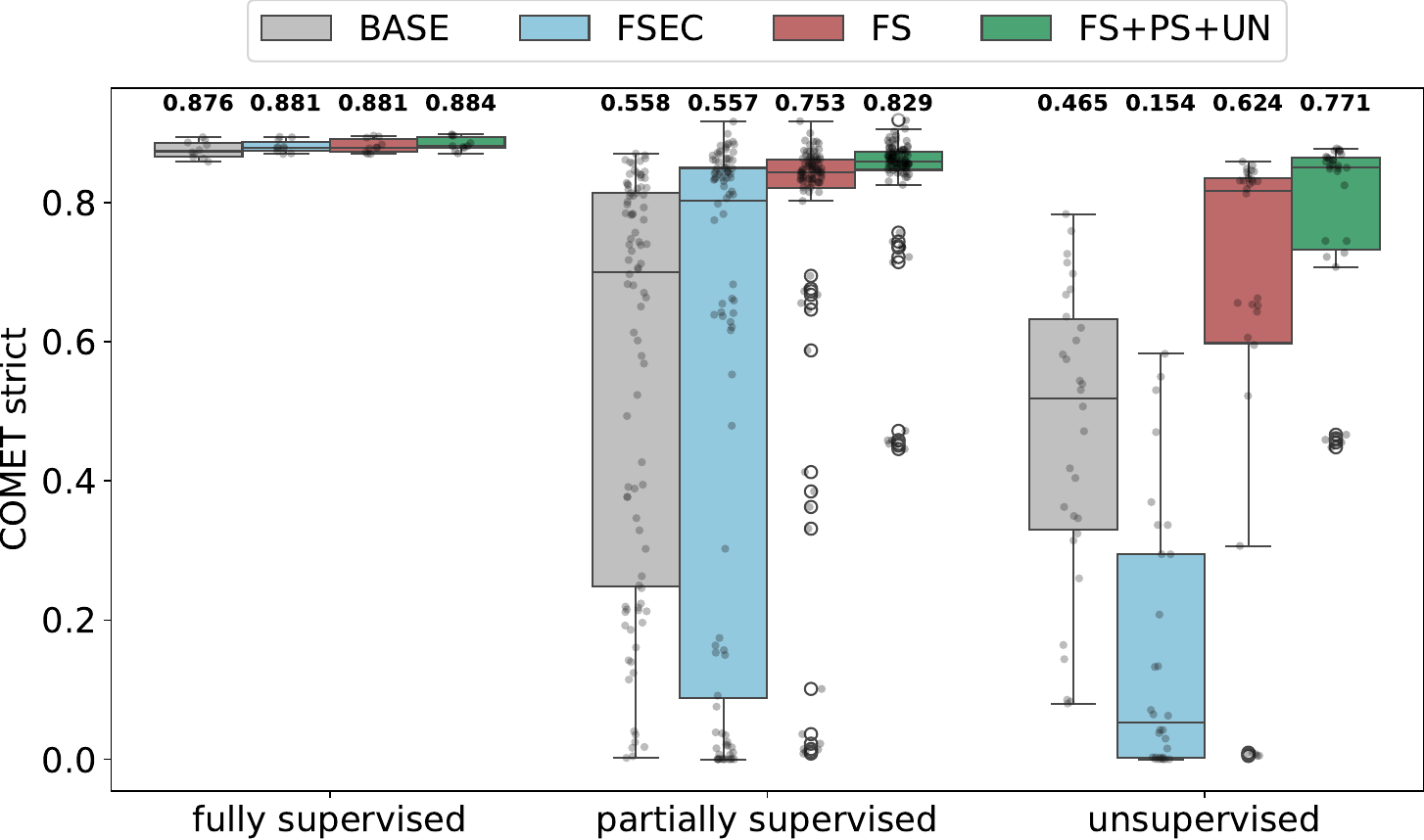}
    \includegraphics[width=0.7\linewidth]{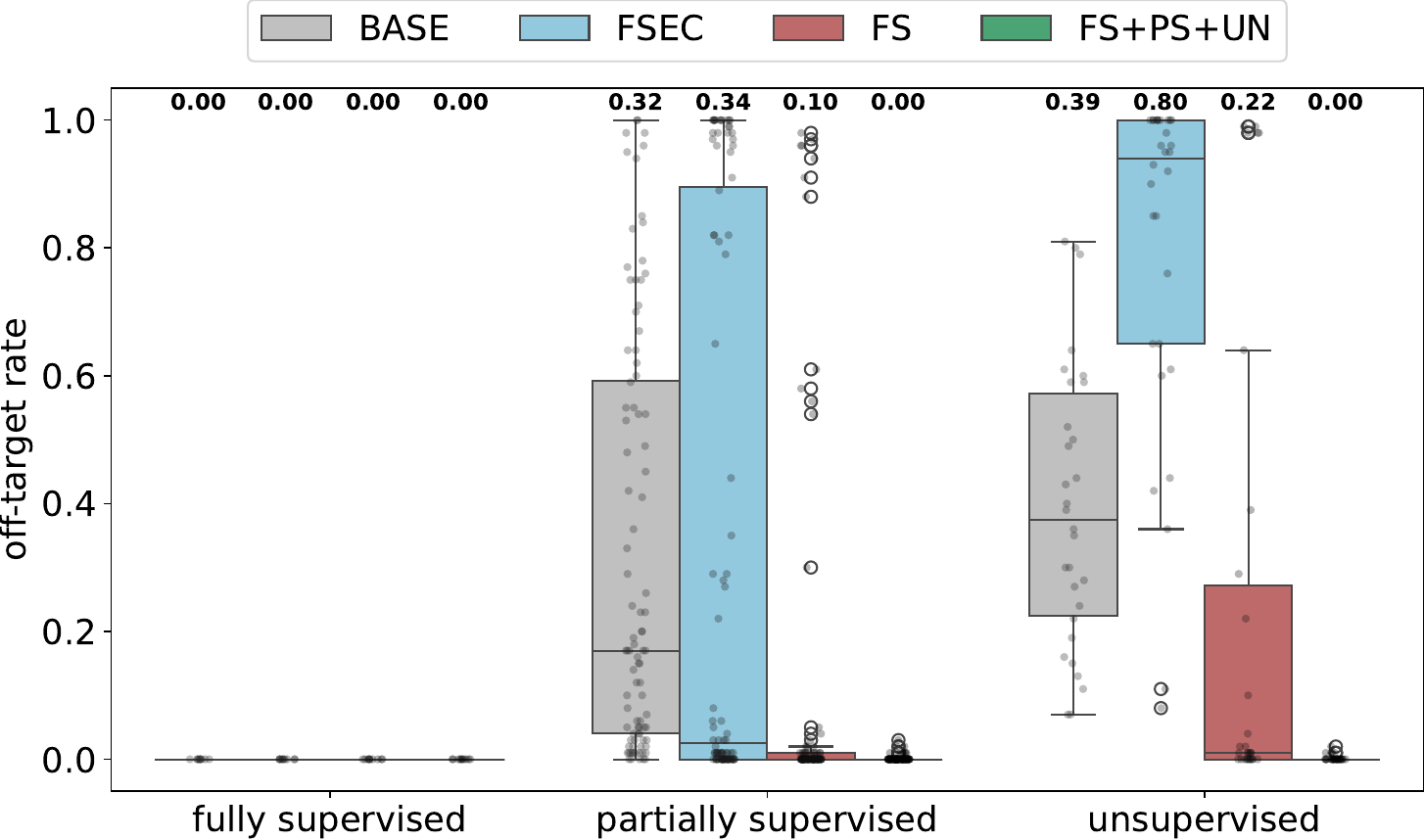}
    \caption{\textsc{COMET-strict} scores (top) and off-target rates (bottom) for 13B models: \textsc{base} (no fine-tuning), \textsc{fsec} (English-centric), \textsc{fs} (seen directions), \textsc{fs+ps+un} (all directions), evaluated on \textit{fully supervised} (Chinese, Dutch, English, German, Korean, and Russian pairs), \textit{unsupervised} (Czech, Icelandic, Japanese, Polish, Swedish, and Ukrainian pairs), and \textit{partially supervised} (combining supervised and unsupervised) language pairs.
    Numbers above bars show mean scores.
    The same observations from Figure \ref{ch6:fig:comet-strict-off-target-7b} (7B instead of 13B) remain true:
    training on more diverse sets improves \textit{all} categories, with \textsc{fs+ps+un} achieving best results even for fully supervised pairs;
    \textsc{fs} substantially reduces off-target rates for unsupervised directions compared to \textsc{base} and \textsc{fsec}, despite these pairs being \textit{absent} from its fine-tuning data.}
    \label{ch6:fig:comet-strict-off-target-13b}
\end{figure*}

\subsubsection{Findings persist at larger scale}
Larger models (13B) exhibit the same trends as their 7B counterparts: increased language diversity leads to reduced off-target rates and improved cross-lingual transfer.
This confirms our findings are robust across model scales.

Figure \ref{ch6:fig:comet-strict-off-target-13b} demonstrates that our findings about language diversity benefits persist when scaling to 13B parameters.
For translation quality (Figure \ref{ch6:fig:comet-strict-off-target-13b}, top), the most diverse setup (\textsc{fs+ps+un}) consistently achieves the best results across all language categories, including fully supervised pairs.
While most 13B models show higher scores than their 7B counterparts, the \textsc{fsec} model unexpectedly performs worse than \textsc{base} in partially supervised and unsupervised settings (0.557 vs.\ 0.558 and 0.154 vs.\ 0.465), unlike in the 7B configuration where \textsc{fsec} outperformed \textsc{base}.

For off-target rates (Figure \ref{ch6:fig:comet-strict-off-target-13b}, bottom), the most diverse setup again eliminates off-target translations completely.
No model produces off-target translations for fully supervised pairs.
The \textsc{fsec} 13B model shows substantially worse performance for partially supervised (0.34) and unsupervised (0.80) pairs compared to its 7B version.
Though \textsc{base} and \textsc{fs} 13B models show improved off-target rates compared to 7B, the problem remains significant (\textsc{base}: 39\% for unsupervised, \textsc{fs}: 22\%).

The decrease in performance for the \textsc{fsec} 13B model can likely be attributed to overfitting to the limited English-centric training data, since larger models are more prone to overfitting on limited data.
These results confirm that language diversity benefits during fine-tuning are robust across model scales.
The consistency of these patterns across model scales parallels our observations in Chapter \ref{ch5}, where we found that capability changes during fine-tuning were consistent across different model sizes.

\subsection{Representational analysis}
\label{ch6:sec:representational_analysis}
Building on our representational analysis approach from Chapter \ref{ch3}, we now examine how the internal model representations adapt when exposed to different levels of language diversity during fine-tuning.

\subsubsection{Middle layers adapt most}

\begin{figure*}[!htb]
    \centering
    \includegraphics[width=0.7\linewidth]{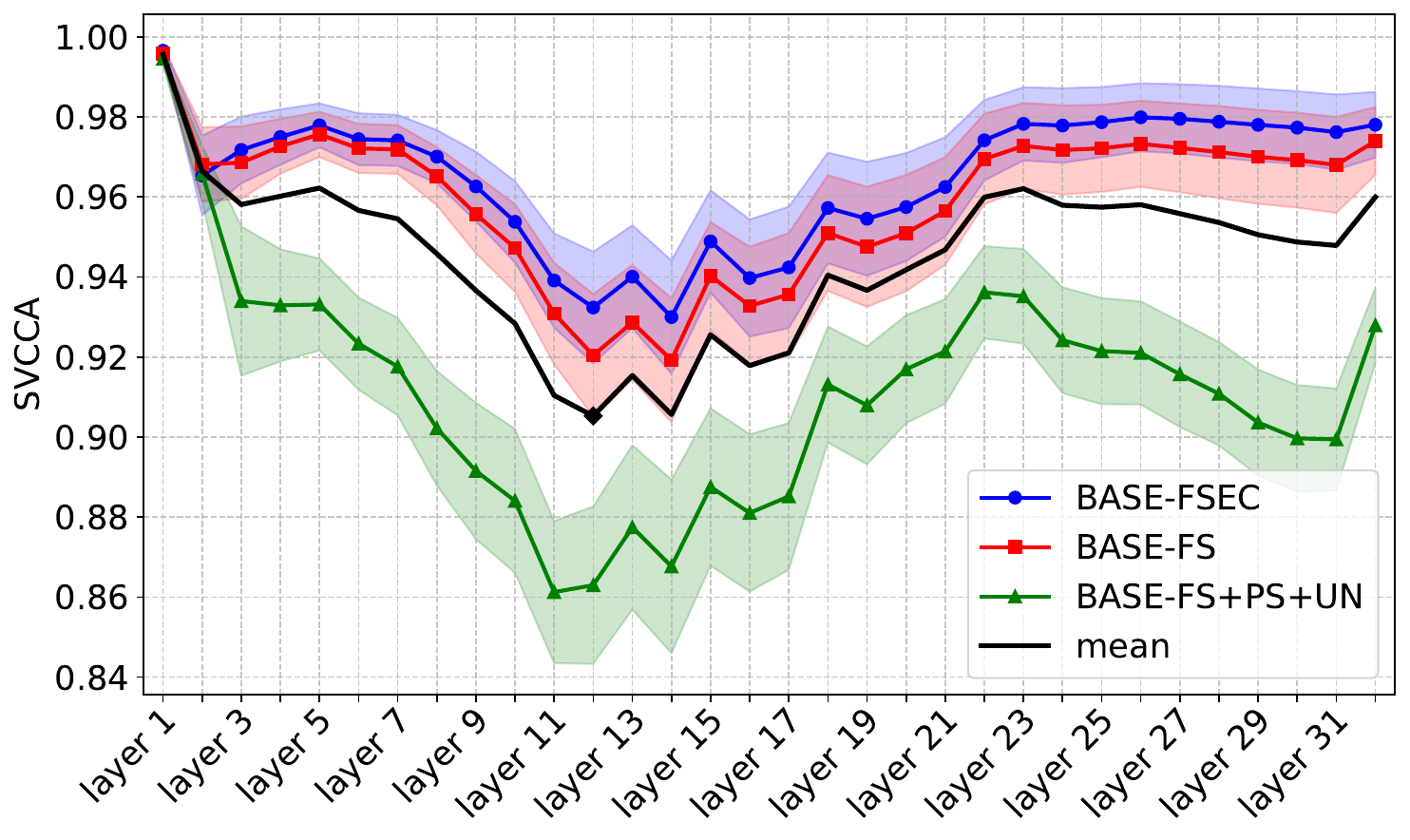}
    \caption{SVCCA similarity scores between fine-tuned and \textsc{base} models across layers. 
    Lower values indicate greater adaptation during fine-tuning. 
    \textsc{base-fsec} (blue), \textsc{base-fs} (red), and \textsc{base-fs+ps+un} (green) are compared, with their mean shown in black. 
    Shaded regions represent confidence intervals.
    Middle layers show most significant adaptation, with lowest mean similarity (0.91) at layer 12. 
    \textsc{fp+ps+un} exhibits greater adaptation throughout the network.}
    \label{ch6:fig:svcca_per_layer}
\end{figure*}

We analyze activation patterns across models by comparing them with the base model using Singular Vector Canonical Correlation Analysis (SVCCA; \citealp{raghu_svcca_2017}).
This analysis identifies \textit{where} and \textit{to what extent} adaptations occur during fine-tuning.
We aggregate activations across all source-target language pairs and present the layer-specific results in Figure \ref{ch6:fig:svcca_per_layer}.

Our analysis reveals that middle layers consistently undergo the most substantial adaptation across all fine-tuned models, with the lowest mean similarity (0.91) occurring at layer 12.
Furthermore, models fine-tuned on more languages exhibit greater divergence from the base model, with \textsc{fs+ps+un} showing most substantial adaptations.

Middle layers encode semantic information and show the strongest cross-lingual transfer capabilities \citep{liu_middle-layer_2025,liu-etal-2025-selected}.
Our findings support that larger degrees of cross-lingual transfer within middle layers explain the performance improvements observed in models fine-tuned on a larger linguistic diversity.

\subsubsection{Diversity increases cross-lingual overlap}

\begin{figure*}[!htb]
    \centering
    \includegraphics[width=0.85\linewidth]{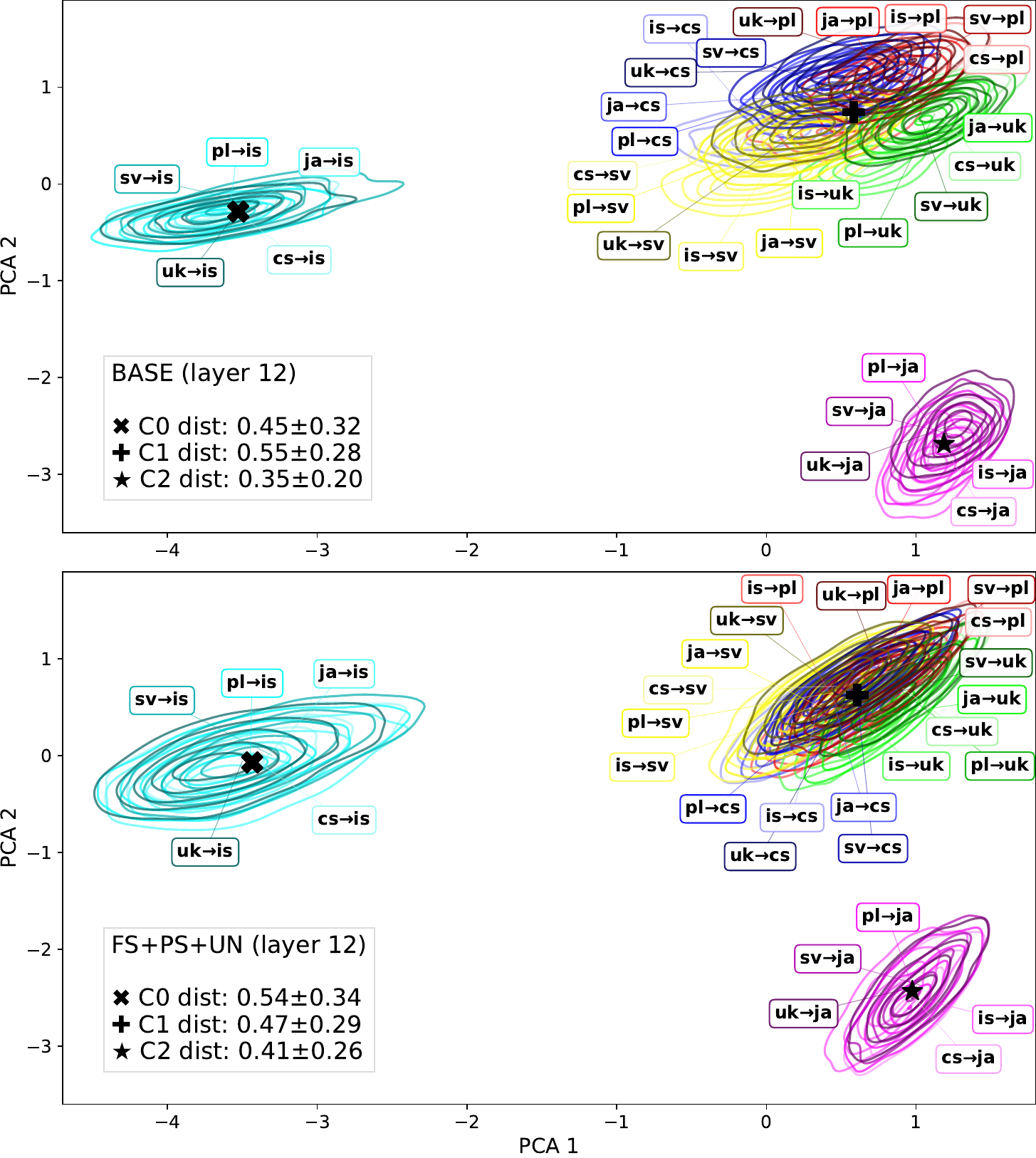}
    \caption{Kernel density estimation of layer 12 activations for \textsc{base} (top) and \textsc{fs+ps+un} (bottom).
    Colors represent translation directions.
    Three distinct clusters emerge (C0, C1, C2), consistently grouped across both models.
    C0 (left, blue) and C2 (bottom, purple) each contain translations into a single target language (Icelandic; \texttt{is}, and Japanese; \texttt{ja}, respectively), while C1 (upper right, multi-colored) contains translations into multiple target languages (Czech; \texttt{cs}, Polish; \texttt{pl}, Swedish; \texttt{sv}, Ukrainian; \texttt{uk}).
    Intra-cluster distances show increased specialization for single-target clusters in \textsc{fs+ps+un}, while multi-target cluster C1 demonstrates increased overlap.}
    \label{ch6:fig:kde_layer_12}
\end{figure*}

We analyze layer 12 (the most significantly modified layer) to understand \textit{which adaptations} occur during fine-tuning.
Following  \citet{gao_towards_2024} and \citet{wang_bridging_2024}, we apply t-SNE dimension reduction \citep{vandermaaten2008visualizing} to layer activations and visualize the bivariate kernel density (KDE) estimation.
Next, we employ \textit{k}-means clustering to identify language groups within these representations, using silhouette score maximization \citep{ROUSSEEUW198753} for optimal cluster determination without requiring manual inspection.
Finally, we calculate the intra-cluster distances.
We compare the \textsc{base} and \textsc{fs+ps+un} models, visualizing unsupervised directions where we expect the most significant adaptations.

Figure \ref{ch6:fig:kde_layer_12} presents the resulting visualization.
Notably, for the single-target language clusters C0 and C2, the \textsc{fs+ps+un} model exhibits \textit{greater intra-cluster distances} (0.54±0.34 and 0.41±0.26) compared to the \textsc{base} model (0.45±0.32 and 0.35±0.20), suggesting \textit{increased specialization} per source-target direction after fine-tuning on diverse data.
Conversely, for the multi-target language cluster (C1), the \textsc{fs+ps+un} model shows \textit{reduced} intra-cluster distances (0.47±0.29) relative to the \textsc{base} model (0.55±0.28), indicating \textit{greater representational overlap} between these linguistically related languages.
This increased overlap provides evidence for enhanced cross-lingual transfer, which contributes to the superior performance of models fine-tuned on greater linguistic diversity.
These results align with our findings in Chapter \ref{ch3}, where we showed that greater representational similarity between languages correlates with more effective cross-lingual transfer.

Table \ref{ch6:tab:cluster_distances} presents intra-cluster distances for all models.
Note that clusters contain the same languages for all setups.
As diversity increases, single-target clusters (C0, C2) show greater specialization while multi-language cluster C1 exhibits enhanced representational overlap, suggesting improved cross-lingual transfer.

While previous work has \textit{explicitly} aligned representations \citep{liu_middle-layer_2025,kargaran_mexa_2024}, including our approach in Chapter \ref{ch3} \citep{stap_viewing_2023}, our findings show \textit{implicit} alignment occurs through multilingual fine-tuning.

\begin{table*}[!ht]
    \centering
    \small
    \begin{tabular}{lccc}
    \toprule
    &  $\boldsymbol{\times}$ C0 & $\boldsymbol{+}$ C1 & $\boldsymbol{\star}$ C2 \\
    \midrule
    \textsc{base}        & $0.45\pm0.32$ & $0.55\pm0.28$ & $0.35\pm0.20$\\
    \textsc{fsec}        & $0.49\pm0.33$ & $0.53\pm0.26$ & $0.34\pm0.20$\\
    \textsc{fs}          & $0.52\pm0.36$ & $0.51\pm0.28$ & $0.39\pm0.24$\\
    \textsc{fs+ps+un}    & $0.54\pm0.34$ & $0.47\pm0.29$ & $0.41\pm0.26$\\
    \bottomrule
    \end{tabular}
    \caption{Intra-cluster distances.
    C0 (Icelandic; \texttt{is} target) and C2 (Japanese; \texttt{ja} target) show increased distances in models fine-tuned on more diverse data, indicating increased specialization, while C1 (Czech; \texttt{cs}, Polish; \texttt{pl}, Swedish; \texttt{sv}, Ukrainian; \texttt{uk} targets) shows decreased distances, indicating enhanced cross-lingual transfer.}
    \label{ch6:tab:cluster_distances}
\end{table*}

\section{Conclusion}
\label{ch6:sec:conclusion}
In this chapter, we investigated how language diversity during fine-tuning affects LLM translation performance across both seen and unseen language pairs through the following research questions:\\

\RQsub{4}{1}
\begin{myquote}
Through controlled experiments across 132 translation directions, we find that increasing language diversity consistently improves performance across all categories of language pairs.
As shown in Figure~\ref{ch6:fig:comet-strict-off-target-7b} (top), our most diverse model (\textsc{fs+ps+un}) outperformed specialized models even on fully supervised language pairs, despite the latter being specifically optimized for these directions.
The benefits become more pronounced for partially supervised and unsupervised directions.
The bottom panel of Figure~\ref{ch6:fig:comet-strict-off-target-7b} demonstrates that increasing language diversity dramatically reduces off-target translations, with our most diverse model completely eliminating this common failure mode across all language categories.
However, as illustrated in Figure~\ref{ch6:fig:comet-strict-7b-272}, when expanding beyond 132 to 272 directions, we observe that benefits plateau or slightly decrease for fully supervised pairs, suggesting an optimal diversity threshold.
\end{myquote}

\RQsub{4}{2}
\begin{myquote}
Our experiments with 13B models, presented in Figure~\ref{ch6:fig:comet-strict-off-target-13b}, demonstrate that the benefits of language diversity are robust across model scales, with 13B models showing the same performance trends as their 7B counterparts.
Both model sizes showed consistent improvements in translation quality and reductions in off-target generations with increased language diversity.
As shown in Figure~\ref{ch6:fig:comet-strict-off-target-7b-nllb}, these benefits persist across different data conditions, including when using non-multi-parallel data from the \textsc{NLLB} corpus rather than multi-parallel \textsc{NTREX} data.
Furthermore, our experiments with different regularization techniques illustrated in Figure~\ref{ch6:fig:regularization} confirmed that the improvements from language diversity cannot be explained by explicit regularization alone, suggesting that diverse language exposure provides unique cross-lingual knowledge transfer beyond simple regularization effects.
\end{myquote}

\RQsub{4}{3}
\begin{myquote}
Our SVCCA analysis in Figure~\ref{ch6:fig:svcca_per_layer} shows that middle layers undergo the most substantial adaptation across all fine-tuned models, with the lowest mean similarity to the base model (0.91) occurring at layer 12.
Cluster analysis of layer 12 activations presented in Figure~\ref{ch6:fig:kde_layer_12} and Table~\ref{ch6:tab:cluster_distances} demonstrate that for single-target language clusters, fine-tuning on diverse data led to increased specialization with greater intra-cluster distances.
Conversely, for multi-target language clusters containing linguistically related languages, our most diverse model shows reduced intra-cluster distances, indicating greater representational overlap and enhanced cross-lingual transfer.
These findings explain the superior performance of models fine-tuned on greater linguistic diversity and demonstrate how multilingual fine-tuning implicitly aligns representations across related languages.\\

These findings extend our work in Chapter \ref{ch3}, where we established metrics for representational similarity between languages, by demonstrating how fine-tuning on diverse language sets naturally increases this similarity for linguistically related languages.
\end{myquote}

Taken together, these steps answered our main research question:

\RQ{4}
\begin{myquote}
Our research resolves conflicting evidence regarding the optimal multilingual fine-tuning strategy for LLMs in machine translation.
As demonstrated through our comprehensive experiments across 132 translation directions, scaling language diversity during fine-tuning consistently improves translation performance across both seen and unseen language pairs, with benefits extending beyond what can be achieved through minimal language exposure or explicit regularization techniques.
This confirms that broad language diversity (132 directions vs. 10--30) substantially enhances cross-lingual transfer, aligning with studies showing benefits from scaling the number of tasks or languages during instruction tuning.\\

However, we also identified limitations to these benefits, with performance improvements plateauing or slightly decreasing beyond a certain diversity threshold, particularly for already well-represented language pairs.
Our representational analysis provides a mechanistic explanation for these findings, showing that increased language diversity creates more language-agnostic representations in middle layers, facilitating better cross-lingual transfer.\\

Our results complement the findings from Chapters \ref{ch3} and \ref{ch4} by demonstrating that diverse language exposure not only improves translation quality across seen and unseen languages but also creates more language-agnostic representations that facilitate better cross-lingual transfer  .
\end{myquote}

\subsubsection{Limitations}
We evaluate on the \textsc{FLORES-200} \citep{nllb_team_no_2022} \texttt{devtest} set, a multi-parallel benchmark consisting of documents originally written in English and professionally translated into multiple languages.
While this may introduce some translationese effects, the multi-parallel nature enables controlled comparison across language pairs.

Our findings are based on the \textsc{Tower} model family \citep{alves_tower_2024} (7B and 13B), built on \textsc{LLaMA 2} \citep{touvron_llama2_2023}.
Further research should verify whether these patterns generalize to other model architectures and even larger model sizes.

\chapter{Conclusions}
\label{ch7}
This dissertation has focused on \textit{analyzing} and \textit{improving} cross-lingual knowledge transfer for machine translation, addressing the challenge of leveraging information across languages to enhance translation quality, particularly for low-resource languages.
Cross-lingual knowledge transfer (the ability of multilingual models to apply what they have learned from one language to another) represents one of the most promising approaches for extending high-quality machine translation to the world's thousands of languages, most of which lack sufficient parallel data for traditional training methods.
Our research has approached this challenge through multiple complementary perspectives.

First, we developed a theoretical framework for \textit{analyzing} cross-lingual transfer through the lens of representational similarities (Chapter~\ref{ch3}), providing quantitative metrics to measure and predict knowledge transfer between languages.
Building on these analytical insights, we then proposed novel methods for \textit{improving} cross-lingual transfer in multilingual machine translation through explicit alignment (Chapter~\ref{ch3}) and retrieval-augmented approaches (Chapter~\ref{ch4}).
These contributions advance our understanding of when and why cross-lingual transfer occurs while offering practical techniques to enhance it.

We also investigated an equally important but often overlooked aspect of knowledge transfer: how to prevent negative transfer or interference that can occur during fine-tuning of large language models (Chapter~\ref{ch5}).
While most prior work has focused exclusively on maximizing positive transfer, our research demonstrates the importance of balanced approaches that preserve beneficial capabilities while improving task-specific performance.

Finally, we explored how scaling language diversity during fine-tuning implicitly affects cross-lingual generalization (Chapter~\ref{ch6}), contrasting with the explicit alignment methods in Chapters~\ref{ch3} and \ref{ch4}.

Throughout the dissertation, our work spans multiple model architectures and scales, from multilingual neural machine translation systems with 93M parameters to large language models with up to 65B parameters, allowing us to identify both scale-invariant principles and size-dependent effects in cross-lingual knowledge transfer.

In the following section, we revisit our research questions and summarize the main findings of this thesis.
We then propose a number of questions that remain open for further exploration in Section \ref{ch7:sec:future_work}.

\section{Main findings}

\RQ{1}
\begin{myquote}
We have established that representational similarities between languages are strongly correlated with effective cross-lingual knowledge transfer in multilingual neural machine translation.
By introducing the Representational Transfer Potential (RTP) metric in Chapter \ref{ch3}, we formalize this relationship, providing a quantitative measure that predicts which languages will benefit most from multilingual training.
Our experiments across diverse language pairs demonstrate that RTP scores correlate significantly with improvements in translation quality, confirming that the performance gains in multilingual models stem from genuine knowledge transfer rather than merely from increased exposure to target language data.
Our analysis of the factors influencing representational similarities reveals that multi-parallel overlap, source subword overlap, and vocabulary occupancy are the most predictive dataset features, while genetic distance emerged as the most important linguistic feature.
Building on these insights, we develop a novel training approach that exploits multi-parallel data through an auxiliary similarity loss.
This method increases the degree of language invariance across language representations, leading to substantial improvements in translation quality for low-resource languages (up to +1.8 BLEU points for the lowest resource languages in our experiments).
Importantly, this approach demonstrates that intentionally optimizing for representational similarity is an effective strategy for enhancing cross-lingual knowledge transfer.
\end{myquote}

\RQ{2}
\begin{myquote}
Our investigation into extending \textit{k}-nearest neighbor machine translation (\textit{k}NN-MT) to multilingual approaches in Chapter \ref{ch4} demonstrates that cross-lingual knowledge transfer plays a crucial role in improving low-resource translation.
We found that cross-lingual datastores, i.e., datastores constructed from high-resource language pairs and used for retrieving translation examples during inference for low-resource language pairs, significantly outperform traditional bilingual datastores, particularly for languages with limited parallel data.
Our experiments with various language combinations show that while datastore size correlates with translation quality, linguistic similarity between source languages is also important.
We find that low-resource languages benefit substantially more from cross-lingual datastores built from linguistically similar high-resource languages than from larger datastores of more distant languages.
We further developed multilingual datastores that combine information from multiple languages, achieving even greater improvements for both low-resource and high-resource language pairs.
Our language-group-specific multilingual datastores achieved comparable performance to comprehensive datastores while being significantly smaller, resulting in up to 5.3× faster inference speeds.
Additionally, our cross-lingual linear mapping approach further enhanced retrieval effectiveness by aligning representations between languages, demonstrating that explicit representation alignment further boosts cross-lingual transfer in \textit{k}NN-MT.
These findings establish multilingual \textit{k}NN-MT as an effective approach for leveraging cross-lingual knowledge transfer to improve low-resource translation, offering both quality improvements and computational efficiency.
\end{myquote}

\RQ{3}
\begin{myquote}
Our research in Chapter \ref{ch5} reveals a significant trade-off when fine-tuning LLMs for machine translation: while general translation quality improves, several capabilities that are relevant for the translation task are compromised.
Through comprehensive experiments across multiple model scales (7B to 65B parameters) and diverse language pairs, we identify that fine-tuning on parallel data (even with as few as 18K examples) leads to notable degradation in formality steerability, few-shot adaptation to specialized domains, and document-level contextualization abilities.
We find that these capability losses become more pronounced as fine-tuning data size increases, even as general translation quality continues to improve.
This pattern is consistent across model scales, demonstrating that this challenge affects LLMs of all sizes we study.
Our novel evaluation dataset for idiom translation, IdiomsInCtx-MT, shows that non-literal translation abilities remain stable or improve after fine-tuning, indicating that not all capabilities are equally vulnerable to negative transfer.
To address this capability preservation challenge, we develop a mixed-data fine-tuning approach that combines parallel translation data with monolingual data.
This method not only better preserves the model's formality control, document-level translation, and few-shot adaptation capabilities compared to parallel-only fine-tuning, but also achieved better overall translation quality.
These findings highlight the importance of developing fine-tuning strategies that maintain the beneficial emergent properties of LLMs while improving task-specific performance.
\end{myquote}

\RQ{4}
\begin{myquote}
Our systematic investigation in Chapter \ref{ch6} into the effects of scaling language diversity during fine-tuning resolves conflicting evidence in prior work by demonstrating that increased diversity consistently improves translation performance across both seen and unseen language combinations.
Through controlled experiments across 132 translation directions, we find that models fine-tuned on the most diverse set of language pairs outperform models trained on limited language combinations, even for language pairs that were explicitly included in the more specialized models' training data.
Our experiments with even greater diversity (272 language directions) reveal that these benefits eventually plateau, suggesting an optimal level of language diversity for balancing generalization and specialization.
These findings are robust across model scales (7B and 13B parameters) and different data conditions, including both multi-parallel and non-multi-parallel training data.
Importantly, we demonstrate that explicit regularization techniques alone can not replicate the benefits of language diversity, indicating that diverse language exposure provides unique cross-lingual knowledge transfer benefits beyond simple regularization effects.
Our representational analysis reveals that multilingual fine-tuning leads to significant adaptations in middle layers of the model, with increased specialization for single-target language clusters and enhanced representational overlap for linguistically related languages.
This improved cross-lingual alignment explains the superior performance of models fine-tuned on greater linguistic diversity and demonstrates how multilingual fine-tuning naturally aligns representations across related languages.
\end{myquote}

\section{Future work}
\label{ch7:sec:future_work}
This thesis has contributed to advancing our understanding of cross-lingual knowledge transfer in machine translation, yet several promising directions for future research remain:

\subsubsection{Standardized cross-lingual transfer metrics}
While our RTP metric provides valuable insights for multilingual MT, future work should develop standardized, task-agnostic metrics designed for measuring cross-lingual transfer in large language models.

Just as BLEU and COMET have become the de facto standards for comparing translation quality across systems, there is a critical need for an analogous widely-adopted metric for cross-lingual transfer.
Current approaches rely primarily on summary statistics and visualization of representations, which offer limited explanatory power and make comparisons across different studies challenging.

A standardized cross-lingual transfer metric would enable direct comparisons between different models and fine-tuning approaches, accelerating progress in the field through more rigorous benchmarking.
Such a metric should be model and task agnostic, computationally efficient, and correlate strongly with downstream performance improvements.
It should also provide nuanced insights beyond aggregate scores, identifying which (linguistic) phenomena transfer successfully and which remain challenging.
Like COMET revolutionized translation evaluation by providing more reliable automatic assessments aligned with human judgment, a unified cross-lingual transfer metric would transform how researchers optimize for and compare transfer effectiveness across language pairs and model architectures.

\subsubsection{Enhanced representational alignment techniques}
While we demonstrated that representational similarities strongly correlate with cross-lingual knowledge transfer, future work could focus on exploring, comparing, and contrasting techniques for explicit representational alignment.

Our auxiliary similarity loss and similar methods like it provide a valuable foundation, but it would be interesting to explore alignment during pre-training rather than fine-tuning.
This approach remains largely unexplored, as most current research focuses on aligning representations after pre-training is complete, when it might be too late to perfectly align representation spaces. 
Incorporating representational alignment objectives directly into the pre-training stage could lead to more fundamentally multilingual models with better transfer capabilities, and this could be feasibly investigated even with smaller model sizes (around 1B parameters) before scaling to larger models.
Additionally, investigating how different model architectures impact representational alignment could yield valuable insights for designing more transfer-efficient multilingual models.

\bibliographystyle{plainnat}
\bibliography{references}

@article{jumper2021highly,
	title = {Highly accurate protein structure prediction with {AlphaFold}},
	volume = {596},
	number = {7873},
	journal = {nature},
	author = {Jumper, John and Evans, Richard and Pritzel, Alexander and Green, Tim and Figurnov, Michael and Ronneberger, Olaf and Tunyasuvunakool, Kathryn and Bates, Russ and Zidek, Augustin and Potapenko, Anna and {others}},
	year = {2021},
	note = {Publisher: Nature Publishing Group},
	pages = {583--589},
}

@misc{llama_team_llama_nodate,
	title = {The {Llama} 3 {Herd} of {Models}},
	url = {http://arxiv.org/abs/2407.21783},
	doi = {10.48550/arXiv.2407.21783},
	urldate = {2025-04-01},
	publisher = {arXiv},
	author = {Grattafiori, Aaron and Dubey, Abhimanyu and Jauhri, Abhinav and Pandey, Abhinav and Kadian, Abhishek and Al-Dahle, Ahmad and Letman, Aiesha and Mathur, Akhil and Schelten, Alan and Vaughan, Alex and Yang, Amy and Fan, Angela and Goyal, Anirudh and Hartshorn, Anthony and Yang, Aobo and Mitra, Archi and Sravankumar, Archie and Korenev, Artem and Hinsvark, Arthur and Rao, Arun and Zhang, Aston and Rodriguez, Aurelien and Gregerson, Austen and Spataru, Ava and Roziere, Baptiste and Biron, Bethany and Tang, Binh and Chern, Bobbie and Caucheteux, Charlotte and Nayak, Chaya and Bi, Chloe and Marra, Chris and McConnell, Chris and Keller, Christian and Touret, Christophe and Wu, Chunyang and Wong, Corinne and Ferrer, Cristian Canton and Nikolaidis, Cyrus and Allonsius, Damien and Song, Daniel and Pintz, Danielle and Livshits, Danny and Wyatt, Danny and Esiobu, David and Choudhary, Dhruv and Mahajan, Dhruv and Garcia-Olano, Diego and Perino, Diego and Hupkes, Dieuwke and Lakomkin, Egor and AlBadawy, Ehab and Lobanova, Elina and Dinan, Emily and Smith, Eric Michael and Radenovic, Filip and Guzmán, Francisco and Zhang, Frank and Synnaeve, Gabriel and Lee, Gabrielle and Anderson, Georgia Lewis and Thattai, Govind and Nail, Graeme and Mialon, Gregoire and Pang, Guan and Cucurell, Guillem and Nguyen, Hailey and Korevaar, Hannah and Xu, Hu and Touvron, Hugo and Zarov, Iliyan and Ibarra, Imanol Arrieta and Kloumann, Isabel and Misra, Ishan and Evtimov, Ivan and Zhang, Jack and Copet, Jade and Lee, Jaewon and Geffert, Jan and Vranes, Jana and Park, Jason and Mahadeokar, Jay and Shah, Jeet and Linde, Jelmer van der and Billock, Jennifer and Hong, Jenny and Lee, Jenya and Fu, Jeremy and Chi, Jianfeng and Huang, Jianyu and Liu, Jiawen and Wang, Jie and Yu, Jiecao and Bitton, Joanna and Spisak, Joe and Park, Jongsoo and Rocca, Joseph and Johnstun, Joshua and Saxe, Joshua and Jia, Junteng and Alwala, Kalyan Vasuden and Prasad, Karthik and Upasani, Kartikeya and Plawiak, Kate and Li, Ke and Heafield, Kenneth and Stone, Kevin and El-Arini, Khalid and Iyer, Krithika and Malik, Kshitiz and Chiu, Kuenley and Bhalla, Kunal and Lakhotia, Kushal and Rantala-Yeary, Lauren and Maaten, Laurens van der and Chen, Lawrence and Tan, Liang and Jenkins, Liz and Martin, Louis and Madaan, Lovish and Malo, Lubo and Blecher, Lukas and Landzaat, Lukas and Oliveira, Luke de and Muzzi, Madeline and Pasupuleti, Mahesh and Singh, Mannat and Paluri, Manohar and Kardas, Marcin and Tsimpoukelli, Maria and Oldham, Mathew and Rita, Mathieu and Pavlova, Maya and Kambadur, Melanie and Lewis, Mike and Si, Min and Singh, Mitesh Kumar and Hassan, Mona and Goyal, Naman and Torabi, Narjes and Bashlykov, Nikolay and Bogoychev, Nikolay and Chatterji, Niladri and Zhang, Ning and Duchenne, Olivier and Çelebi, Onur and Alrassy, Patrick and Zhang, Pengchuan and Li, Pengwei and Vasic, Petar and Weng, Peter and Bhargava, Prajjwal and Dubal, Pratik and Krishnan, Praveen and Koura, Punit Singh and Xu, Puxin and He, Qing and Dong, Qingxiao and Srinivasan, Ragavan and Ganapathy, Raj and Calderer, Ramon and Cabral, Ricardo Silveira and Stojnic, Robert and Raileanu, Roberta and Maheswari, Rohan and Girdhar, Rohit and Patel, Rohit and Sauvestre, Romain and Polidoro, Ronnie and Sumbaly, Roshan and Taylor, Ross and Silva, Ruan and Hou, Rui and Wang, Rui and Hosseini, Saghar and Chennabasappa, Sahana and Singh, Sanjay and Bell, Sean and Kim, Seohyun Sonia and Edunov, Sergey and Nie, Shaoliang and Narang, Sharan and Raparthy, Sharath and Shen, Sheng and Wan, Shengye and Bhosale, Shruti and Zhang, Shun and Vandenhende, Simon and Batra, Soumya and Whitman, Spencer and Sootla, Sten and Collot, Stephane and Gururangan, Suchin and Borodinsky, Sydney and Herman, Tamar and Fowler, Tara and Sheasha, Tarek and Georgiou, Thomas and Scialom, Thomas and Speckbacher, Tobias and Mihaylov, Todor and Xiao, Tong and Karn, Ujjwal and Goswami, Vedanuj and Gupta, Vibhor and Ramanathan, Vignesh and Kerkez, Viktor and Gonguet, Vincent and Do, Virginie and Vogeti, Vish and Albiero, Vítor and Petrovic, Vladan and Chu, Weiwei and Xiong, Wenhan and Fu, Wenyin and Meers, Whitney and Martinet, Xavier and Wang, Xiaodong and Wang, Xiaofang and Tan, Xiaoqing Ellen and Xia, Xide and Xie, Xinfeng and Jia, Xuchao and Wang, Xuewei and Goldschlag, Yaelle and Gaur, Yashesh and Babaei, Yasmine and Wen, Yi and Song, Yiwen and Zhang, Yuchen and Li, Yue and Mao, Yuning and Coudert, Zacharie Delpierre and Yan, Zheng and Chen, Zhengxing and Papakipos, Zoe and Singh, Aaditya and Srivastava, Aayushi and Jain, Abha and Kelsey, Adam and Shajnfeld, Adam and Gangidi, Adithya and Victoria, Adolfo and Goldstand, Ahuva and Menon, Ajay and Sharma, Ajay and Boesenberg, Alex and Baevski, Alexei and Feinstein, Allie and Kallet, Amanda and Sangani, Amit and Teo, Amos and Yunus, Anam and Lupu, Andrei and Alvarado, Andres and Caples, Andrew and Gu, Andrew and Ho, Andrew and Poulton, Andrew and Ryan, Andrew and Ramchandani, Ankit and Dong, Annie and Franco, Annie and Goyal, Anuj and Saraf, Aparajita and Chowdhury, Arkabandhu and Gabriel, Ashley and Bharambe, Ashwin and Eisenman, Assaf and Yazdan, Azadeh and James, Beau and Maurer, Ben and Leonhardi, Benjamin and Huang, Bernie and Loyd, Beth and Paola, Beto De and Paranjape, Bhargavi and Liu, Bing and Wu, Bo and Ni, Boyu and Hancock, Braden and Wasti, Bram and Spence, Brandon and Stojkovic, Brani and Gamido, Brian and Montalvo, Britt and Parker, Carl and Burton, Carly and Mejia, Catalina and Liu, Ce and Wang, Changhan and Kim, Changkyu and Zhou, Chao and Hu, Chester and Chu, Ching-Hsiang and Cai, Chris and Tindal, Chris and Feichtenhofer, Christoph and Gao, Cynthia and Civin, Damon and Beaty, Dana and Kreymer, Daniel and Li, Daniel and Adkins, David and Xu, David and Testuggine, Davide and David, Delia and Parikh, Devi and Liskovich, Diana and Foss, Didem and Wang, Dingkang and Le, Duc and Holland, Dustin and Dowling, Edward and Jamil, Eissa and Montgomery, Elaine and Presani, Eleonora and Hahn, Emily and Wood, Emily and Le, Eric-Tuan and Brinkman, Erik and Arcaute, Esteban and Dunbar, Evan and Smothers, Evan and Sun, Fei and Kreuk, Felix and Tian, Feng and Kokkinos, Filippos and Ozgenel, Firat and Caggioni, Francesco and Kanayet, Frank and Seide, Frank and Florez, Gabriela Medina and Schwarz, Gabriella and Badeer, Gada and Swee, Georgia and Halpern, Gil and Herman, Grant and Sizov, Grigory and Guangyi and Zhang and Lakshminarayanan, Guna and Inan, Hakan and Shojanazeri, Hamid and Zou, Han and Wang, Hannah and Zha, Hanwen and Habeeb, Haroun and Rudolph, Harrison and Suk, Helen and Aspegren, Henry and Goldman, Hunter and Zhan, Hongyuan and Damlaj, Ibrahim and Molybog, Igor and Tufanov, Igor and Leontiadis, Ilias and Veliche, Irina-Elena and Gat, Itai and Weissman, Jake and Geboski, James and Kohli, James and Lam, Janice and Asher, Japhet and Gaya, Jean-Baptiste and Marcus, Jeff and Tang, Jeff and Chan, Jennifer and Zhen, Jenny and Reizenstein, Jeremy and Teboul, Jeremy and Zhong, Jessica and Jin, Jian and Yang, Jingyi and Cummings, Joe and Carvill, Jon and Shepard, Jon and McPhie, Jonathan and Torres, Jonathan and Ginsburg, Josh and Wang, Junjie and Wu, Kai and U, Kam Hou and Saxena, Karan and Khandelwal, Kartikay and Zand, Katayoun and Matosich, Kathy and Veeraraghavan, Kaushik and Michelena, Kelly and Li, Keqian and Jagadeesh, Kiran and Huang, Kun and Chawla, Kunal and Huang, Kyle and Chen, Lailin and Garg, Lakshya and A, Lavender and Silva, Leandro and Bell, Lee and Zhang, Lei and Guo, Liangpeng and Yu, Licheng and Moshkovich, Liron and Wehrstedt, Luca and Khabsa, Madian and Avalani, Manav and Bhatt, Manish and Mankus, Martynas and Hasson, Matan and Lennie, Matthew and Reso, Matthias and Groshev, Maxim and Naumov, Maxim and Lathi, Maya and Keneally, Meghan and Liu, Miao and Seltzer, Michael L. and Valko, Michal and Restrepo, Michelle and Patel, Mihir and Vyatskov, Mik and Samvelyan, Mikayel and Clark, Mike and Macey, Mike and Wang, Mike and Hermoso, Miquel Jubert and Metanat, Mo and Rastegari, Mohammad and Bansal, Munish and Santhanam, Nandhini and Parks, Natascha and White, Natasha and Bawa, Navyata and Singhal, Nayan and Egebo, Nick and Usunier, Nicolas and Mehta, Nikhil and Laptev, Nikolay Pavlovich and Dong, Ning and Cheng, Norman and Chernoguz, Oleg and Hart, Olivia and Salpekar, Omkar and Kalinli, Ozlem and Kent, Parkin and Parekh, Parth and Saab, Paul and Balaji, Pavan and Rittner, Pedro and Bontrager, Philip and Roux, Pierre and Dollar, Piotr and Zvyagina, Polina and Ratanchandani, Prashant and Yuvraj, Pritish and Liang, Qian and Alao, Rachad and Rodriguez, Rachel and Ayub, Rafi and Murthy, Raghotham and Nayani, Raghu and Mitra, Rahul and Parthasarathy, Rangaprabhu and Li, Raymond and Hogan, Rebekkah and Battey, Robin and Wang, Rocky and Howes, Russ and Rinott, Ruty and Mehta, Sachin and Siby, Sachin and Bondu, Sai Jayesh and Datta, Samyak and Chugh, Sara and Hunt, Sara and Dhillon, Sargun and Sidorov, Sasha and Pan, Satadru and Mahajan, Saurabh and Verma, Saurabh and Yamamoto, Seiji and Ramaswamy, Sharadh and Lindsay, Shaun and Lindsay, Shaun and Feng, Sheng and Lin, Shenghao and Zha, Shengxin Cindy and Patil, Shishir and Shankar, Shiva and Zhang, Shuqiang and Zhang, Shuqiang and Wang, Sinong and Agarwal, Sneha and Sajuyigbe, Soji and Chintala, Soumith and Max, Stephanie and Chen, Stephen and Kehoe, Steve and Satterfield, Steve and Govindaprasad, Sudarshan and Gupta, Sumit and Deng, Summer and Cho, Sungmin and Virk, Sunny and Subramanian, Suraj and Choudhury, Sy and Goldman, Sydney and Remez, Tal and Glaser, Tamar and Best, Tamara and Koehler, Thilo and Robinson, Thomas and Li, Tianhe and Zhang, Tianjun and Matthews, Tim and Chou, Timothy and Shaked, Tzook and Vontimitta, Varun and Ajayi, Victoria and Montanez, Victoria and Mohan, Vijai and Kumar, Vinay Satish and Mangla, Vishal and Ionescu, Vlad and Poenaru, Vlad and Mihailescu, Vlad Tiberiu and Ivanov, Vladimir and Li, Wei and Wang, Wenchen and Jiang, Wenwen and Bouaziz, Wes and Constable, Will and Tang, Xiaocheng and Wu, Xiaojian and Wang, Xiaolan and Wu, Xilun and Gao, Xinbo and Kleinman, Yaniv and Chen, Yanjun and Hu, Ye and Jia, Ye and Qi, Ye and Li, Yenda and Zhang, Yilin and Zhang, Ying and Adi, Yossi and Nam, Youngjin and Yu and Wang and Zhao, Yu and Hao, Yuchen and Qian, Yundi and Li, Yunlu and He, Yuzi and Rait, Zach and DeVito, Zachary and Rosnbrick, Zef and Wen, Zhaoduo and Yang, Zhenyu and Zhao, Zhiwei and Ma, Zhiyu},
	month = nov,
	year = {2024},
	note = {arXiv:2407.21783 [cs]},
}

@inproceedings{zhu_knn-box_2024,
	address = {St. Julians, Malta},
	title = {{kNN}-{BOX}: {A} {Unified} {Framework} for {Nearest} {Neighbor} {Generation}},
	url = {https://aclanthology.org/2024.eacl-demo.2/},
	doi = {10.18653/v1/2024.eacl-demo.2},
	booktitle = {Proceedings of the 18th {Conference} of the {European} {Chapter} of the {Association} for {Computational} {Linguistics}: {System} {Demonstrations}},
	publisher = {Association for Computational Linguistics},
	author = {Zhu, Wenhao and Zhao, Qianfeng and Lv, Yunzhe and Huang, Shujian and Zhao, Siheng and Liu, Sizhe and Chen, Jiajun},
	editor = {Aletras, Nikolaos and De Clercq, Orphee},
	month = mar,
	year = {2024},
	pages = {10--17},
}

@inproceedings{zhang_machine_2023,
	address = {Singapore},
	title = {Machine {Translation} with {Large} {Language} {Models}: {Prompting}, {Few}-shot {Learning}, and {Fine}-tuning with {QLoRA}},
	url = {https://aclanthology.org/2023.wmt-1.43/},
	doi = {10.18653/v1/2023.wmt-1.43},
	booktitle = {Proceedings of the {Eighth} {Conference} on {Machine} {Translation}},
	publisher = {Association for Computational Linguistics},
	author = {Zhang, Xuan and Rajabi, Navid and Duh, Kevin and Koehn, Philipp},
	editor = {Koehn, Philipp and Haddow, Barry and Kocmi, Tom and Monz, Christof},
	month = dec,
	year = {2023},
	pages = {468--481},
}

@inproceedings{yang_nearest_2022,
	address = {Seattle, United States},
	title = {Nearest {Neighbor} {Knowledge} {Distillation} for {Neural} {Machine} {Translation}},
	url = {https://aclanthology.org/2022.naacl-main.406/},
	doi = {10.18653/v1/2022.naacl-main.406},
	booktitle = {Proceedings of the 2022 {Conference} of the {North} {American} {Chapter} of the {Association} for {Computational} {Linguistics}: {Human} {Language} {Technologies}},
	publisher = {Association for Computational Linguistics},
	author = {Yang, Zhixian and Sun, Renliang and Wan, Xiaojun},
	editor = {Carpuat, Marine and de Marneffe, Marie-Catherine and Meza Ruiz, Ivan Vladimir},
	month = jul,
	year = {2022},
	pages = {5546--5556},
}

@inproceedings{wu_beyond_2023,
	address = {Singapore},
	title = {Beyond {Shared} {Vocabulary}: {Increasing} {Representational} {Word} {Similarities} across {Languages} for {Multilingual} {Machine} {Translation}},
	url = {https://aclanthology.org/2023.emnlp-main.605/},
	doi = {10.18653/v1/2023.emnlp-main.605},
	booktitle = {Proceedings of the 2023 {Conference} on {Empirical} {Methods} in {Natural} {Language} {Processing}},
	publisher = {Association for Computational Linguistics},
	author = {Wu, Di and Monz, Christof},
	editor = {Bouamor, Houda and Pino, Juan and Bali, Kalika},
	month = dec,
	year = {2023},
	pages = {9749--9764},
}

@inproceedings{wicks_identifying_2023,
	address = {Singapore},
	title = {Identifying {Context}-{Dependent} {Translations} for {Evaluation} {Set} {Production}},
	url = {https://aclanthology.org/2023.wmt-1.42/},
	doi = {10.18653/v1/2023.wmt-1.42},
	booktitle = {Proceedings of the {Eighth} {Conference} on {Machine} {Translation}},
	publisher = {Association for Computational Linguistics},
	author = {Wicks, Rachel and Post, Matt},
	editor = {Koehn, Philipp and Haddow, Barry and Kocmi, Tom and Monz, Christof},
	month = dec,
	year = {2023},
	pages = {452--467},
}

@inproceedings{wang_document-level_2023,
	address = {Singapore},
	title = {Document-{Level} {Machine} {Translation} with {Large} {Language} {Models}},
	url = {https://aclanthology.org/2023.emnlp-main.1036/},
	doi = {10.18653/v1/2023.emnlp-main.1036},
	booktitle = {Proceedings of the 2023 {Conference} on {Empirical} {Methods} in {Natural} {Language} {Processing}},
	publisher = {Association for Computational Linguistics},
	author = {Wang, Longyue and Lyu, Chenyang and Ji, Tianbo and Zhang, Zhirui and Yu, Dian and Shi, Shuming and Tu, Zhaopeng},
	editor = {Bouamor, Houda and Pino, Juan and Bali, Kalika},
	month = dec,
	year = {2023},
	pages = {16646--16661},
}

@article{vijayakumar_diverse_2018,
	title = {Diverse beam search for improved description of complex scenes},
	volume = {32},
	url = {https://ojs.aaai.org/index.php/AAAI/article/view/12340},
	doi = {10.1609/aaai.v32i1.12340},
	number = {1},
	journal = {Proceedings of the AAAI Conference on Artificial Intelligence},
	author = {Vijayakumar, Ashwin and Cogswell, Michael and Selvaraju, Ramprasaath and Sun, Qing and Lee, Stefan and Crandall, David and Batra, Dhruv},
	month = apr,
	year = {2018},
}

@inproceedings{vaswani_attention_2017,
	address = {Long Beach, California},
	title = {Attention is all you need},
	url = {https://proceedings.neurips.cc/paper_files/paper/2017/file/3f5ee243547dee91fbd053c1c4a845aa-Paper.pdf},
	urldate = {2019-12-09},
	booktitle = {Advances in {Neural} {Information} {Processing} {Systems}},
	author = {Vaswani, Ashish and Shazeer, Noam and Parmar, Niki and Uszkoreit, Jakob and Jones, Llion and Gomez, Aidan N. and Kaiser, Łukasz and Polosukhin, Illia},
	year = {2017},
}

@inproceedings{schwenk_ccmatrix_2021,
	address = {Online},
	title = {{CCMatrix}: {Mining} {Billions} of {High}-{Quality} {Parallel} {Sentences} on the {Web}},
	url = {https://aclanthology.org/2021.acl-long.507/},
	doi = {10.18653/v1/2021.acl-long.507},
	booktitle = {Proceedings of the 59th {Annual} {Meeting} of the {Association} for {Computational} {Linguistics} and the 11th {International} {Joint} {Conference} on {Natural} {Language} {Processing} ({Volume} 1: {Long} {Papers})},
	publisher = {Association for Computational Linguistics},
	author = {Schwenk, Holger and Wenzek, Guillaume and Edunov, Sergey and Grave, Edouard and Joulin, Armand and Fan, Angela},
	editor = {Zong, Chengqing and Xia, Fei and Li, Wenjie and Navigli, Roberto},
	month = aug,
	year = {2021},
	pages = {6490--6500},
}

@article{radford_language_2019,
	title = {Language {Models} are {Unsupervised} {Multitask} {Learners}},
	volume = {1},
	number = {8},
	journal = {OpenAI blog},
	author = {Radford, Alec and Wu, Jeffrey and Child, Rewon and Luan, David and Amodei, Dario and Sutskever, Ilya},
	year = {2019},
	pages = {9},
}

@inproceedings{penedo_refinedweb_2023,
	address = {Red Hook, NY, USA},
	series = {{NIPS} '23},
	title = {The {RefinedWeb} dataset for falcon {LLM}: outperforming curated corpora with web data only},
	booktitle = {Proceedings of the 37th {International} {Conference} on {Neural} {Information} {Processing} {Systems}},
	publisher = {Curran Associates Inc.},
	author = {Penedo, Guilherme and Malartic, Quentin and Hesslow, Daniel and Cojocaru, Ruxandra and Alobeidli, Hamza and Cappelli, Alessandro and Pannier, Baptiste and Almazrouei, Ebtesam and Launay, Julien},
	year = {2023},
	note = {event-place: New Orleans, LA, USA},
}

@article{pelicon_zero-shot_2020,
	title = {Zero-shot learning for cross-lingual news sentiment classification},
	volume = {10},
	issn = {2076-3417},
	url = {https://www.mdpi.com/2076-3417/10/17/5993},
	doi = {10.3390/app10175993},
	number = {17},
	journal = {Applied Sciences},
	author = {Pelicon, Andraž and Pranjić, Marko and Miljković, Dragana and Škrlj, Blaž and Pollak, Senja},
	year = {2020},
}

@inproceedings{moslem_adaptive_2023,
	address = {Tampere, Finland},
	title = {Adaptive {Machine} {Translation} with {Large} {Language} {Models}},
	url = {https://aclanthology.org/2023.eamt-1.22/},
	booktitle = {Proceedings of the 24th {Annual} {Conference} of the {European} {Association} for {Machine} {Translation}},
	publisher = {European Association for Machine Translation},
	author = {Moslem, Yasmin and Haque, Rejwanul and Kelleher, John D. and Way, Andy},
	editor = {Nurminen, Mary and Brenner, Judith and Koponen, Maarit and Latomaa, Sirkku and Mikhailov, Mikhail and Schierl, Frederike and Ranasinghe, Tharindu and Vanmassenhove, Eva and Vidal, Sergi Alvarez and Aranberri, Nora and Nunziatini, Mara and Escartín, Carla Parra and Forcada, Mikel and Popovic, Maja and Scarton, Carolina and Moniz, Helena},
	month = jun,
	year = {2023},
	pages = {227--237},
}

@inproceedings{mathur_tangled_2020,
	address = {Online},
	title = {Tangled up in {BLEU}: {Reevaluating} the {Evaluation} of {Automatic} {Machine} {Translation} {Evaluation} {Metrics}},
	url = {https://aclanthology.org/2020.acl-main.448/},
	doi = {10.18653/v1/2020.acl-main.448},
	booktitle = {Proceedings of the 58th {Annual} {Meeting} of the {Association} for {Computational} {Linguistics}},
	publisher = {Association for Computational Linguistics},
	author = {Mathur, Nitika and Baldwin, Timothy and Cohn, Trevor},
	editor = {Jurafsky, Dan and Chai, Joyce and Schluter, Natalie and Tetreault, Joel},
	month = jul,
	year = {2020},
	pages = {4984--4997},
}

@article{maimaiti_multi-round_2019,
	title = {Multi-round transfer learning for low-resource {NMT} using multiple high-resource languages},
	volume = {18},
	issn = {2375-4699},
	url = {https://doi.org/10.1145/3314945},
	doi = {10.1145/3314945},
	number = {4},
	journal = {ACM Transactions on Asian and Low-Resource Language Information Processing},
	author = {Maimaiti, Mieradilijiang and Liu, Yang and Luan, Huanbo and Sun, Maosong},
	month = may,
	year = {2019},
}

@article{ma_knowledge-driven_2021,
	title = {Knowledge-driven data construction for zero-shot evaluation in commonsense question answering},
	volume = {35},
	url = {https://ojs.aaai.org/index.php/AAAI/article/view/17593},
	doi = {10.1609/aaai.v35i15.17593},
	number = {15},
	journal = {Proceedings of the AAAI Conference on Artificial Intelligence},
	author = {Ma, Kaixin and Ilievski, Filip and Francis, Jonathan and Bisk, Yonatan and Nyberg, Eric and Oltramari, Alessandro},
	month = may,
	year = {2021},
	pages = {13507--13515},
}

@phdthesis{10.5555/907741,
	address = {USA},
	type = {{PhD} {Thesis}},
	title = {The harpy speech recognition system.},
	school = {Carnegie Mellon University},
	author = {Lowerre, Bruce T.},
	year = {1976},
}

@inproceedings{kitaev_reformer_2020,
	address = {Online},
	title = {Reformer: the efficient {Transformer}},
	url = {https://openreview.net/forum?id=rkgNKkHtvB},
	booktitle = {International conference on learning representations},
	author = {Kitaev, Nikita and Kaiser, Łukasz and Levskaya, Anselm},
	year = {2020},
}

@inproceedings{karpinska_large_2023,
	address = {Singapore},
	title = {Large {Language} {Models} {Effectively} {Leverage} {Document}-level {Context} for {Literary} {Translation}, but {Critical} {Errors} {Persist}},
	url = {https://aclanthology.org/2023.wmt-1.41/},
	doi = {10.18653/v1/2023.wmt-1.41},
	booktitle = {Proceedings of the {Eighth} {Conference} on {Machine} {Translation}},
	publisher = {Association for Computational Linguistics},
	author = {Karpinska, Marzena and Iyyer, Mohit},
	editor = {Koehn, Philipp and Haddow, Barry and Kocmi, Tom and Monz, Christof},
	month = dec,
	year = {2023},
	pages = {419--451},
}

@inproceedings{kalchbrenner_recurrent_2013,
	address = {Seattle, Washington},
	title = {Recurrent continuous translation models},
	url = {https://aclanthology.org/D13-1176/},
	booktitle = {Proceedings of the 2013 {Conference} on {Empirical} {Methods} in {Natural} {Language} {Processing}},
	publisher = {Association for Computational Linguistics},
	author = {Kalchbrenner, Nal and Blunsom, Phil},
	year = {2013},
	pages = {1700--1709},
}

@inproceedings{jiao_parrot_2023,
	address = {Singapore},
	title = {{ParroT}: {Translating} during {Chat} using {Large} {Language} {Models} tuned with {Human} {Translation} and {Feedback}},
	url = {https://aclanthology.org/2023.findings-emnlp.1001/},
	doi = {10.18653/v1/2023.findings-emnlp.1001},
	booktitle = {Findings of the {Association} for {Computational} {Linguistics}: {EMNLP} 2023},
	publisher = {Association for Computational Linguistics},
	author = {Jiao, Wenxiang and Huang, Jen-tse and Wang, Wenxuan and He, Zhiwei and Liang, Tian and Wang, Xing and Shi, Shuming and Tu, Zhaopeng},
	editor = {Bouamor, Houda and Pino, Juan and Bali, Kalika},
	month = dec,
	year = {2023},
	pages = {15009--15020},
}

@article{hupkes_taxonomy_2023,
	title = {A taxonomy and review of generalization research in {NLP}},
	volume = {5},
	number = {10},
	journal = {Nature Machine Intelligence},
	author = {Hupkes, Dieuwke and Giulianelli, Mario and Dankers, Verna and Artetxe, Mikel and Elazar, Yanai and Pimentel, Tiago and Christodoulopoulos, Christos and Lasri, Karim and Saphra, Naomi and Sinclair, Arabella and {others}},
	year = {2023},
	note = {Publisher: Nature Publishing Group UK London},
	pages = {1161--1174},
}

@article{goyal_flores-101_2022,
	title = {The {Flores}-101 {Evaluation} {Benchmark} for {Low}-{Resource} and {Multilingual} {Machine} {Translation}},
	volume = {10},
	url = {https://aclanthology.org/2022.tacl-1.30/},
	doi = {10.1162/tacl_a_00474},
	journal = {Transactions of the Association for Computational Linguistics},
	author = {Goyal, Naman and Gao, Cynthia and Chaudhary, Vishrav and Chen, Peng-Jen and Wenzek, Guillaume and Ju, Da and Krishnan, Sanjana and Ranzato, Marc'Aurelio and Guzmán, Francisco and Fan, Angela},
	editor = {Roark, Brian and Nenkova, Ani},
	year = {2022},
	note = {Place: Cambridge, MA
Publisher: MIT Press},
	pages = {522--538},
}

@inproceedings{gehring_convolutional_2017,
	address = {Sydney, Australia},
	title = {Convolutional sequence to sequence learning},
	url = {https://proceedings.mlr.press/v70/gehring17a.html},
	booktitle = {Proceedings of the {International} {Conference} on {Machine} {Learning} ({ICML})},
	author = {Gehring, Jonas and Auli, Michael and Grangier, David and Yarats, Denis and Dauphin, Yann N.},
	year = {2017},
	pages = {1243--1252},
}

@inproceedings{garcia_unreasonable_2023,
	address = {Honolulu, Hawaii, USA},
	series = {{ICML}'23},
	title = {The unreasonable effectiveness of few-shot learning for machine translation},
	booktitle = {Proceedings of the 40th {International} {Conference} on {Machine} {Learning}},
	publisher = {JMLR.org},
	author = {Garcia, Xavier and Bansal, Yamini and Cherry, Colin and Foster, George and Krikun, Maxim and Johnson, Melvin and Firat, Orhan},
	year = {2023},
}

@article{eronen_transfer_2022,
	title = {Transfer {Language} {Selection} for {Zero}-{Shot} {Cross}-{Lingual} {Abusive} {Language} {Detection}},
	volume = {59},
	issn = {03064573},
	url = {https://doi.org/10.1016/j.ipm.2022.102981},
	doi = {10.1016/j.ipm.2022.102981},
	number = {4},
	urldate = {2025-04-09},
	journal = {Information Processing \& Management},
	author = {Eronen, Juuso and Ptaszynski, Michal and Masui, Fumito and Arata, Masaki and Leliwa, Gniewosz and Wroczynski, Michal},
	month = jul,
	year = {2022},
	pages = {102981},
}

@inproceedings{dettmers_qlora_2023,
	address = {Red Hook, NY, USA},
	series = {{NIPS} '23},
	title = {{QLORA}: efficient finetuning of quantized {LLMs}},
	booktitle = {Proceedings of the 37th {International} {Conference} on {Neural} {Information} {Processing} {Systems}},
	publisher = {Curran Associates Inc.},
	author = {Dettmers, Tim and Pagnoni, Artidoro and Holtzman, Ari and Zettlemoyer, Luke},
	year = {2023},
	note = {event-place: New Orleans, LA, USA},
}

@article{chowdhery_palm_2023,
	title = {{PaLM}: scaling language modeling with pathways},
	volume = {24},
	issn = {1532-4435},
	number = {1},
	journal = {J. Mach. Learn. Res.},
	author = {Chowdhery, Aakanksha and Narang, Sharan and Devlin, Jacob and Bosma, Maarten and Mishra, Gaurav and Roberts, Adam and Barham, Paul and Chung, Hyung Won and Sutton, Charles and Gehrmann, Sebastian and Schuh, Parker and Shi, Kensen and Tsvyashchenko, Sashank and Maynez, Joshua and Rao, Abhishek and Barnes, Parker and Tay, Yi and Shazeer, Noam and Prabhakaran, Vinodkumar and Reif, Emily and Du, Nan and Hutchinson, Ben and Pope, Reiner and Bradbury, James and Austin, Jacob and Isard, Michael and Gur-Ari, Guy and Yin, Pengcheng and Duke, Toju and Levskaya, Anselm and Ghemawat, Sanjay and Dev, Sunipa and Michalewski, Henryk and Garcia, Xavier and Misra, Vedant and Robinson, Kevin and Fedus, Liam and Zhou, Denny and Ippolito, Daphne and Luan, David and Lim, Hyeontaek and Zoph, Barret and Spiridonov, Alexander and Sepassi, Ryan and Dohan, David and Agrawal, Shivani and Omernick, Mark and Dai, Andrew M. and Pillai, Thanumalayan Sankaranarayana and Pellat, Marie and Lewkowycz, Aitor and Moreira, Erica and Child, Rewon and Polozov, Oleksandr and Lee, Katherine and Zhou, Zongwei and Wang, Xuezhi and Saeta, Brennan and Diaz, Mark and Firat, Orhan and Catasta, Michele and Wei, Jason and Meier-Hellstern, Kathy and Eck, Douglas and Dean, Jeff and Petrov, Slav and Fiedel, Noah},
	month = jan,
	year = {2023},
	note = {Publisher: JMLR.org},
}

@inproceedings{alves_steering_2023,
	address = {Singapore},
	title = {Steering {Large} {Language} {Models} for {Machine} {Translation} with {Finetuning} and {In}-{Context} {Learning}},
	url = {https://aclanthology.org/2023.findings-emnlp.744/},
	doi = {10.18653/v1/2023.findings-emnlp.744},
	booktitle = {Findings of the {Association} for {Computational} {Linguistics}: {EMNLP} 2023},
	publisher = {Association for Computational Linguistics},
	author = {Alves, Duarte and Guerreiro, Nuno and Alves, João and Pombal, José and Rei, Ricardo and de Souza, José and Colombo, Pierre and Martins, Andre},
	editor = {Bouamor, Houda and Pino, Juan and Bali, Kalika},
	month = dec,
	year = {2023},
	pages = {11127--11148},
}

@inproceedings{agrawal_-context_2023,
	address = {Toronto, Canada},
	title = {In-context {Examples} {Selection} for {Machine} {Translation}},
	url = {https://aclanthology.org/2023.findings-acl.564/},
	doi = {10.18653/v1/2023.findings-acl.564},
	booktitle = {Findings of the {Association} for {Computational} {Linguistics}: {ACL} 2023},
	publisher = {Association for Computational Linguistics},
	author = {Agrawal, Sweta and Zhou, Chunting and Lewis, Mike and Zettlemoyer, Luke and Ghazvininejad, Marjan},
	editor = {Rogers, Anna and Boyd-Graber, Jordan and Okazaki, Naoaki},
	month = jul,
	year = {2023},
	pages = {8857--8873},
}

@misc{stap_effect_2025,
	title = {The {Effect} of {Language} {Diversity} {When} {Fine}-{Tuning} {Large} {Language} {Models} for {Translation}},
	url = {http://arxiv.org/abs/2505.13090},
	doi = {10.48550/arXiv.2505.13090},
	urldate = {2025-05-21},
	publisher = {arXiv},
	author = {Stap, David and Monz, Christof},
	month = may,
	year = {2025},
	note = {arXiv:2505.13090 [cs]},
}

@inproceedings{zhang_prompting_2023,
	series = {{ICML}'23},
	title = {Prompting large language model for machine translation: a case study},
	booktitle = {Proceedings of the 40th international conference on machine learning},
	publisher = {JMLR.org},
	author = {Zhang, Biao and Haddow, Barry and Birch, Alexandra},
	year = {2023},
}

@article{zeng_teaching_2024,
	title = {Teaching large language models to translate with comparison},
	volume = {38},
	url = {https://ojs.aaai.org/index.php/AAAI/article/view/29920},
	doi = {10.1609/aaai.v38i17.29920},
	number = {17},
	journal = {Proceedings of the AAAI Conference on Artificial Intelligence},
	author = {Zeng, Jiali and Meng, Fandong and Yin, Yongjing and Zhou, Jie},
	month = mar,
	year = {2024},
	pages = {19488--19496},
}

@article{guerreiro_hallucinations_2023,
	title = {Hallucinations in large multilingual translation models},
	volume = {11},
	url = {https://aclanthology.org/2023.tacl-1.85/},
	doi = {10.1162/tacl_a_00615},
	journal = {Transactions of the Association for Computational Linguistics},
	author = {Guerreiro, Nuno M. and Alves, Duarte M. and Waldendorf, Jonas and Haddow, Barry and Birch, Alexandra and Colombo, Pierre and Martins, André F. T.},
	year = {2023},
	note = {Place: Cambridge, MA
Publisher: MIT Press},
	pages = {1500--1517},
}

@article{li_eliciting_2023,
	title = {Eliciting the translation ability of large language models via multilingual finetuning with translation instructions},
	volume = {12},
	url = {https://aclanthology.org/2024.tacl-1.32/},
	doi = {10.1162/tacl_a_00655},
	journal = {Transactions of the Association for Computational Linguistics},
	author = {Li, Jiahuan and Zhou, Hao and Huang, Shujian and Cheng, Shanbo and Chen, Jiajun},
	year = {2024},
	note = {Place: Cambridge, MA
Publisher: MIT Press},
	pages = {576--592},
}

@inproceedings{richburg_how_2024,
	title = {How multilingual are large language models fine-tuned for translation?},
	url = {https://openreview.net/forum?id=bnscREWUuc},
	booktitle = {First conference on language modeling},
	author = {Richburg, Aquia and Carpuat, Marine},
	year = {2024},
}

@inproceedings{stap_finetuning_2024,
	address = {Bangkok, Thailand},
	title = {The fine-tuning paradox: {Boosting} translation quality without sacrificing {LLM} abilities},
	url = {https://aclanthology.org/2024.acl-long.336/},
	doi = {10.18653/v1/2024.acl-long.336},
	booktitle = {Proceedings of the 62nd annual meeting of the association for computational linguistics (volume 1: {Long} papers)},
	publisher = {Association for Computational Linguistics},
	author = {Stap, David and Hasler, Eva and Byrne, Bill and Monz, Christof and Tran, Ke},
	editor = {Ku, Lun-Wei and Martins, Andre and Srikumar, Vivek},
	month = aug,
	year = {2024},
	pages = {6189--6206},
}

@inproceedings{zouhar_pitfalls_2024,
	address = {Miami, Florida, USA},
	title = {Pitfalls and outlooks in using {COMET}},
	url = {https://aclanthology.org/2024.wmt-1.121/},
	doi = {10.18653/v1/2024.wmt-1.121},
	booktitle = {Proceedings of the ninth conference on machine translation},
	publisher = {Association for Computational Linguistics},
	author = {Zouhar, Vilém and Chen, Pinzhen and Lam, Tsz Kin and Moghe, Nikita and Haddow, Barry},
	editor = {Haddow, Barry and Kocmi, Tom and Koehn, Philipp and Monz, Christof},
	month = nov,
	year = {2024},
	pages = {1272--1288},
}

@inproceedings{muennighoff-etal-2023-crosslingual,
	address = {Toronto, Canada},
	title = {Crosslingual generalization through multitask finetuning},
	url = {https://aclanthology.org/2023.acl-long.891/},
	doi = {10.18653/v1/2023.acl-long.891},
	booktitle = {Proceedings of the 61st annual meeting of the association for computational linguistics (volume 1: {Long} papers)},
	publisher = {Association for Computational Linguistics},
	author = {Muennighoff, Niklas and Wang, Thomas and Sutawika, Lintang and Roberts, Adam and Biderman, Stella and Le Scao, Teven and Bari, M Saiful and Shen, Sheng and Yong, Zheng Xin and Schoelkopf, Hailey and Tang, Xiangru and Radev, Dragomir and Aji, Alham Fikri and Almubarak, Khalid and Albanie, Samuel and Alyafeai, Zaid and Webson, Albert and Raff, Edward and Raffel, Colin},
	editor = {Rogers, Anna and Boyd-Graber, Jordan and Okazaki, Naoaki},
	month = jul,
	year = {2023},
	pages = {15991--16111},
}

@inproceedings{zhu_finetuning_2024,
	address = {Miami, Florida, USA},
	title = {Fine-tuning large language models to translate: {Will} a touch of noisy data in misaligned languages suffice?},
	url = {https://aclanthology.org/2024.emnlp-main.24/},
	doi = {10.18653/v1/2024.emnlp-main.24},
	booktitle = {Proceedings of the 2024 conference on empirical methods in natural language processing},
	publisher = {Association for Computational Linguistics},
	author = {Zhu, Dawei and Chen, Pinzhen and Zhang, Miaoran and Haddow, Barry and Shen, Xiaoyu and Klakow, Dietrich},
	editor = {Al-Onaizan, Yaser and Bansal, Mohit and Chen, Yun-Nung},
	month = nov,
	year = {2024},
	pages = {388--409},
}

@inproceedings{briakou_searching_2023,
	address = {Toronto, Canada},
	title = {Searching for {Needles} in a {Haystack}: {On} the {Role} of {Incidental} {Bilingualism} in {PaLM}'s {Translation} {Capability}},
	url = {https://aclanthology.org/2023.acl-long.524},
	urldate = {2023-07-11},
	booktitle = {Proceedings of the 61st {Annual} {Meeting} of the {Association} for {Computational} {Linguistics} ({Volume} 1: {Long} {Papers})},
	publisher = {Association for Computational Linguistics},
	author = {Briakou, Eleftheria and Cherry, Colin and Foster, George},
	year = {2023},
	pages = {9432--9452},
}

@inproceedings{alves_tower_2024,
	title = {Tower: {An} open multilingual large language model for translation-related tasks},
	url = {https://openreview.net/forum?id=EHPns3hVkj},
	booktitle = {First conference on language modeling},
	author = {Alves, Duarte Miguel and Pombal, José and Guerreiro, Nuno M and Martins, Pedro Henrique and Alves, João and Farajian, Amin and Peters, Ben and Rei, Ricardo and Fernandes, Patrick and Agrawal, Sweta and Colombo, Pierre and de Souza, José G. C. and Martins, Andre},
	year = {2024},
}

@inproceedings{meng-monz-2024-disentangling,
	address = {St. Julian's, Malta},
	title = {Disentangling the roles of target-side transfer and regularization in multilingual machine translation},
	url = {https://aclanthology.org/2024.eacl-long.110/},
	booktitle = {Proceedings of the 18th conference of the european chapter of the association for computational linguistics (volume 1: {Long} papers)},
	publisher = {Association for Computational Linguistics},
	author = {Meng, Yan and Monz, Christof},
	editor = {Graham, Yvette and Purver, Matthew},
	month = mar,
	year = {2024},
	pages = {1828--1840},
}

@inproceedings{koehn-2024-neural,
	address = {Miami, Florida, USA},
	title = {Neural methods for aligning large-scale parallel corpora from the web for south and {East} {Asian} languages},
	url = {https://aclanthology.org/2024.wmt-1.132/},
	doi = {10.18653/v1/2024.wmt-1.132},
	booktitle = {Proceedings of the ninth conference on machine translation},
	publisher = {Association for Computational Linguistics},
	author = {Koehn, Philipp},
	editor = {Haddow, Barry and Kocmi, Tom and Koehn, Philipp and Monz, Christof},
	month = nov,
	year = {2024},
	pages = {1454--1466},
}

@article{ROUSSEEUW198753,
	title = {Silhouettes: {A} graphical aid to the interpretation and validation of cluster analysis},
	volume = {20},
	issn = {0377-0427},
	url = {https://www.sciencedirect.com/science/article/pii/0377042787901257},
	doi = {https://doi.org/10.1016/0377-0427(87)90125-7},
	journal = {Journal of Computational and Applied Mathematics},
	author = {Rousseeuw, Peter J.},
	year = {1987},
	pages = {53--65},
}

@inproceedings{sennrich-etal-2024-mitigating,
	address = {St. Julian's, Malta},
	title = {Mitigating hallucinations and off-target machine translation with source-contrastive and language-contrastive decoding},
	url = {https://aclanthology.org/2024.eacl-short.4/},
	booktitle = {Proceedings of the 18th conference of the european chapter of the association for computational linguistics (volume 2: {Short} papers)},
	publisher = {Association for Computational Linguistics},
	author = {Sennrich, Rico and Vamvas, Jannis and Mohammadshahi, Alireza},
	editor = {Graham, Yvette and Purver, Matthew},
	month = mar,
	year = {2024},
	pages = {21--33},
}

@misc{caswell_smol_2025,
	title = {{SMOL}: {Professionally} translated parallel data for 115 under-represented languages},
	shorttitle = {{SMOL}},
	url = {http://arxiv.org/abs/2502.12301},
	doi = {10.48550/arXiv.2502.12301},
	urldate = {2025-05-08},
	publisher = {arXiv},
	author = {Caswell, Isaac and Nielsen, Elizabeth and Luo, Jiaming and Cherry, Colin and Kovacs, Geza and Shemtov, Hadar and Talukdar, Partha and Tewari, Dinesh and Diane, Baba Mamadi and Doumbouya, Koulako Moussa and Diane, Djibrila and Cissé, Solo Farabado},
	month = feb,
	year = {2025},
	note = {arXiv:2502.12301 [cs]},
}

@inproceedings{dang-etal-2024-rlhf,
	address = {Miami, Florida, USA},
	title = {{RLHF} can speak many languages: {Unlocking} multilingual preference optimization for {LLMs}},
	url = {https://aclanthology.org/2024.emnlp-main.729/},
	doi = {10.18653/v1/2024.emnlp-main.729},
	booktitle = {Proceedings of the 2024 conference on empirical methods in natural language processing},
	publisher = {Association for Computational Linguistics},
	author = {Dang, John and Ahmadian, Arash and Marchisio, Kelly and Kreutzer, Julia and Üstün, Ahmet and Hooker, Sara},
	editor = {Al-Onaizan, Yaser and Bansal, Mohit and Chen, Yun-Nung},
	month = nov,
	year = {2024},
	pages = {13134--13156},
}

@misc{kargaran_mexa_2024,
	title = {{MEXA}: {Multilingual} {Evaluation} of {English}-{Centric} {LLMs} via {Cross}-{Lingual} {Alignment}},
	shorttitle = {{MEXA}},
	url = {http://arxiv.org/abs/2410.05873},
	doi = {10.48550/arXiv.2410.05873},
	urldate = {2025-05-05},
	publisher = {arXiv},
	author = {Kargaran, Amir Hossein and Modarressi, Ali and Nikeghbal, Nafiseh and Diesner, Jana and Yvon, François and Schütze, Hinrich},
	month = oct,
	year = {2024},
	note = {arXiv:2410.05873 [cs]},
}

@misc{touvron_llama2_2023,
	title = {Llama 2: {Open} {Foundation} and {Fine}-{Tuned} {Chat} {Models}},
	shorttitle = {Llama 2},
	url = {http://arxiv.org/abs/2307.09288},
	doi = {10.48550/arXiv.2307.09288},
	urldate = {2025-01-15},
	publisher = {arXiv},
	author = {Touvron, Hugo and Martin, Louis and Stone, Kevin and Albert, Peter and Almahairi, Amjad and Babaei, Yasmine and Bashlykov, Nikolay and Batra, Soumya and Bhargava, Prajjwal and Bhosale, Shruti and Bikel, Dan and Blecher, Lukas and Ferrer, Cristian Canton and Chen, Moya and Cucurull, Guillem and Esiobu, David and Fernandes, Jude and Fu, Jeremy and Fu, Wenyin and Fuller, Brian and Gao, Cynthia and Goswami, Vedanuj and Goyal, Naman and Hartshorn, Anthony and Hosseini, Saghar and Hou, Rui and Inan, Hakan and Kardas, Marcin and Kerkez, Viktor and Khabsa, Madian and Kloumann, Isabel and Korenev, Artem and Koura, Punit Singh and Lachaux, Marie-Anne and Lavril, Thibaut and Lee, Jenya and Liskovich, Diana and Lu, Yinghai and Mao, Yuning and Martinet, Xavier and Mihaylov, Todor and Mishra, Pushkar and Molybog, Igor and Nie, Yixin and Poulton, Andrew and Reizenstein, Jeremy and Rungta, Rashi and Saladi, Kalyan and Schelten, Alan and Silva, Ruan and Smith, Eric Michael and Subramanian, Ranjan and Tan, Xiaoqing Ellen and Tang, Binh and Taylor, Ross and Williams, Adina and Kuan, Jian Xiang and Xu, Puxin and Yan, Zheng and Zarov, Iliyan and Zhang, Yuchen and Fan, Angela and Kambadur, Melanie and Narang, Sharan and Rodriguez, Aurelien and Stojnic, Robert and Edunov, Sergey and Scialom, Thomas},
	month = jul,
	year = {2023},
	note = {arXiv:2307.09288 [cs]},
}

@inproceedings{tu-etal-2016-modeling,
	address = {Berlin, Germany},
	title = {Modeling coverage for neural machine translation},
	url = {https://aclanthology.org/P16-1008/},
	doi = {10.18653/v1/P16-1008},
	booktitle = {Proceedings of the 54th annual meeting of the association for computational linguistics (volume 1: {Long} papers)},
	publisher = {Association for Computational Linguistics},
	author = {Tu, Zhaopeng and Lu, Zhengdong and Liu, Yang and Liu, Xiaohua and Li, Hang},
	editor = {Erk, Katrin and Smith, Noah A.},
	month = aug,
	year = {2016},
	pages = {76--85},
}

@misc{vries_bertje_2019,
	title = {{BERTje}: {A} {Dutch} {BERT} {Model}},
	shorttitle = {{BERTje}},
	url = {http://arxiv.org/abs/1912.09582},
	doi = {10.48550/arXiv.1912.09582},
	urldate = {2025-04-21},
	publisher = {arXiv},
	author = {Vries, Wietse de and Cranenburgh, Andreas van and Bisazza, Arianna and Caselli, Tommaso and Noord, Gertjan van and Nissim, Malvina},
	month = dec,
	year = {2019},
	note = {arXiv:1912.09582 [cs]},
}

@inproceedings{burchardt-2013-multidimensional,
	address = {London, UK},
	title = {Multidimensional quality metrics: a flexible system for assessing translation quality},
	url = {https://aclanthology.org/2013.tc-1.6/},
	booktitle = {Proceedings of translating and the computer 35},
	publisher = {Aslib},
	author = {Lommel, Arle Richard and Burchardt, Aljoscha and Uszkoreit, Hans},
	year = {2013},
}

@article{GRAHAM_BALDWIN_MOFFAT_ZOBEL_2017,
	title = {Can machine translation systems be evaluated by the crowd alone},
	volume = {23},
	doi = {10.1017/S1351324915000339},
	number = {1},
	journal = {Natural Language Engineering},
	author = {Graham, Yvette and Baldwin, Timothy and Moffat, Alistair and Zobel, Justin},
	year = {2017},
	pages = {3--30},
}

@article{doi:10.1021/acscentsci.9b00576,
	title = {Molecular transformer: a model for uncertainty-calibrated chemical reaction prediction},
	volume = {5},
	url = {https://doi.org/10.1021/acscentsci.9b00576},
	doi = {10.1021/acscentsci.9b00576},
	number = {9},
	journal = {ACS Central Science},
	author = {Schwaller, Philippe and Laino, Teodoro and Gaudin, Théophile and Bolgar, Peter and Hunter, Christopher A. and Bekas, Costas and Lee, Alpha A.},
	year = {2019},
	note = {tex.eprint: https://doi.org/10.1021/acscentsci.9b00576},
	pages = {1572--1583},
}

@inproceedings{ramesh2021zeroshot,
	series = {Proceedings of machine learning research},
	title = {Zero-shot text-to-image generation},
	volume = {139},
	url = {https://proceedings.mlr.press/v139/ramesh21a.html},
	booktitle = {Proceedings of the 38th international conference on machine learning},
	publisher = {PMLR},
	author = {Ramesh, Aditya and Pavlov, Mikhail and Goh, Gabriel and Gray, Scott and Voss, Chelsea and Radford, Alec and Chen, Mark and Sutskever, Ilya},
	editor = {Meila, Marina and Zhang, Tong},
	month = jul,
	year = {2021},
	pages = {8821--8831},
}

@inproceedings{pmlr-v139-radford21a,
	series = {Proceedings of machine learning research},
	title = {Learning transferable visual models from natural language supervision},
	volume = {139},
	url = {https://proceedings.mlr.press/v139/radford21a.html},
	booktitle = {Proceedings of the 38th international conference on machine learning},
	publisher = {PMLR},
	author = {Radford, Alec and Kim, Jong Wook and Hallacy, Chris and Ramesh, Aditya and Goh, Gabriel and Agarwal, Sandhini and Sastry, Girish and Askell, Amanda and Mishkin, Pamela and Clark, Jack and Krueger, Gretchen and Sutskever, Ilya},
	editor = {Meila, Marina and Zhang, Tong},
	month = jul,
	year = {2021},
	pages = {8748--8763},
}

@article{CHEN2024103280,
	title = {{TransUNet}: {Rethinking} the {U}-{Net} architecture design for medical image segmentation through the lens of transformers},
	volume = {97},
	issn = {1361-8415},
	url = {https://www.sciencedirect.com/science/article/pii/S1361841524002056},
	doi = {https://doi.org/10.1016/j.media.2024.103280},
	journal = {Medical Image Analysis},
	author = {Chen, Jieneng and Mei, Jieru and Li, Xianhang and Lu, Yongyi and Yu, Qihang and Wei, Qingyue and Luo, Xiangde and Xie, Yutong and Adeli, Ehsan and Wang, Yan and Lungren, Matthew P. and Zhang, Shaoting and Xing, Lei and Lu, Le and Yuille, Alan and Zhou, Yuyin},
	year = {2024},
	pages = {103280},
}

@inproceedings{Zhao_2021_ICCV,
	title = {Point transformer},
	booktitle = {Proceedings of the {IEEE}/{CVF} international conference on computer vision ({ICCV})},
	author = {Zhao, Hengshuang and Jiang, Li and Jia, Jiaya and Torr, Philip H.S. and Koltun, Vladlen},
	month = oct,
	year = {2021},
	pages = {16259--16268},
}

@inproceedings{9710415,
	title = {{ViViT}: a video vision transformer},
	doi = {10.1109/ICCV48922.2021.00676},
	booktitle = {2021 {IEEE}/{CVF} international conference on computer vision ({ICCV})},
	author = {Arnab, Anurag and Dehghani, Mostafa and Heigold, Georg and Sun, Chen and Lučić, Mario and Schmid, Cordelia},
	year = {2021},
	pages = {6816--6826},
}

@inproceedings{dosovitskiy2021an,
	title = {An image is worth 16x16 words: {Transformers} for image recognition at scale},
	url = {https://openreview.net/forum?id=YicbFdNTTy},
	booktitle = {International conference on learning representations},
	author = {Dosovitskiy, Alexey and Beyer, Lucas and Kolesnikov, Alexander and Weissenborn, Dirk and Zhai, Xiaohua and Unterthiner, Thomas and Dehghani, Mostafa and Minderer, Matthias and Heigold, Georg and Gelly, Sylvain and Uszkoreit, Jakob and Houlsby, Neil},
	year = {2021},
}

@inproceedings{gu_non-autoregressive_2018,
	address = {Vancouver, Canada},
	title = {Non-autoregressive neural machine translation},
	url = {https://openreview.net/forum?id=B1l8BtlCb},
	booktitle = {International conference on learning representations},
	author = {Gu, Jiatao and Bradbury, James and Xiong, Caiming and Li, Victor O.K. and Socher, Richard},
	year = {2018},
}

@inproceedings{yee-etal-2019-simple,
	address = {Hong Kong, China},
	title = {Simple and effective noisy channel modeling for neural machine translation},
	url = {https://aclanthology.org/D19-1571/},
	doi = {10.18653/v1/D19-1571},
	booktitle = {Proceedings of the 2019 conference on empirical methods in natural language processing and the 9th international joint conference on natural language processing ({EMNLP}-{IJCNLP})},
	publisher = {Association for Computational Linguistics},
	author = {Yee, Kyra and Dauphin, Yann and Auli, Michael},
	editor = {Inui, Kentaro and Jiang, Jing and Ng, Vincent and Wan, Xiaojun},
	month = nov,
	year = {2019},
	pages = {5696--5701},
}

@inproceedings{eikema-aziz-2022-sampling,
	address = {Abu Dhabi, United Arab Emirates},
	title = {Sampling-based approximations to minimum {Bayes} risk decoding for neural machine translation},
	url = {https://aclanthology.org/2022.emnlp-main.754/},
	doi = {10.18653/v1/2022.emnlp-main.754},
	booktitle = {Proceedings of the 2022 conference on empirical methods in natural language processing},
	publisher = {Association for Computational Linguistics},
	author = {Eikema, Bryan and Aziz, Wilker},
	editor = {Goldberg, Yoav and Kozareva, Zornitsa and Zhang, Yue},
	month = dec,
	year = {2022},
	pages = {10978--10993},
}

@inproceedings{stahlberg-byrne-2019-nmt,
	address = {Hong Kong, China},
	title = {On {NMT} search errors and model errors: {Cat} got your tongue?},
	url = {https://aclanthology.org/D19-1331/},
	doi = {10.18653/v1/D19-1331},
	booktitle = {Proceedings of the 2019 conference on empirical methods in natural language processing and the 9th international joint conference on natural language processing ({EMNLP}-{IJCNLP})},
	publisher = {Association for Computational Linguistics},
	author = {Stahlberg, Felix and Byrne, Bill},
	editor = {Inui, Kentaro and Jiang, Jing and Ng, Vincent and Wan, Xiaojun},
	month = nov,
	year = {2019},
	pages = {3356--3362},
}

@inproceedings{murray-chiang-2018-correcting,
	address = {Brussels, Belgium},
	title = {Correcting length bias in neural machine translation},
	url = {https://aclanthology.org/W18-6322/},
	doi = {10.18653/v1/W18-6322},
	booktitle = {Proceedings of the third conference on machine translation: {Research} papers},
	publisher = {Association for Computational Linguistics},
	author = {Murray, Kenton and Chiang, David},
	editor = {Bojar, Ondřej and Chatterjee, Rajen and Federmann, Christian and Fishel, Mark and Graham, Yvette and Haddow, Barry and Huck, Matthias and Yepes, Antonio Jimeno and Koehn, Philipp and Monz, Christof and Negri, Matteo and Névéol, Aurélie and Neves, Mariana and Post, Matt and Specia, Lucia and Turchi, Marco and Verspoor, Karin},
	month = oct,
	year = {2018},
	pages = {212--223},
}

@inproceedings{xu_paradigm_2024,
	title = {A paradigm shift in machine translation: {Boosting} translation performance of large language models},
	url = {https://openreview.net/forum?id=farT6XXntP},
	booktitle = {The twelfth international conference on learning representations},
	author = {Xu, Haoran and Kim, Young Jin and Sharaf, Amr and Awadalla, Hany Hassan},
	year = {2024},
}

@misc{xu_advancing_2024,
	title = {Advancing {Translation} {Preference} {Modeling} with {RLHF}: {A} {Step} {Towards} {Cost}-{Effective} {Solution}},
	shorttitle = {Advancing {Translation} {Preference} {Modeling} with {RLHF}},
	url = {http://arxiv.org/abs/2402.11525},
	doi = {10.48550/arXiv.2402.11525},
	urldate = {2025-04-09},
	publisher = {arXiv},
	author = {Xu, Nuo and Zhao, Jun and Zu, Can and Li, Sixian and Chen, Lu and Zhang, Zhihao and Zheng, Rui and Dou, Shihan and Qin, Wenjuan and Gui, Tao and Zhang, Qi and Huang, Xuanjing},
	month = feb,
	year = {2024},
	note = {arXiv:2402.11525 [cs]},
}

@inproceedings{hu2022lora,
	title = {{LoRA}: {Low}-rank adaptation of large language models},
	url = {https://openreview.net/forum?id=nZeVKeeFYf9},
	booktitle = {International conference on learning representations},
	author = {Hu, Edward J and shen, yelong and Wallis, Phillip and Allen-Zhu, Zeyuan and Li, Yuanzhi and Wang, Shean and Wang, Lu and Chen, Weizhu},
	year = {2022},
}

@inproceedings{fu-etal-2024-gptscore,
	address = {Mexico City, Mexico},
	title = {{GPTScore}: {Evaluate} as you desire},
	url = {https://aclanthology.org/2024.naacl-long.365/},
	doi = {10.18653/v1/2024.naacl-long.365},
	booktitle = {Proceedings of the 2024 conference of the north american chapter of the association for computational linguistics: {Human} language technologies (volume 1: {Long} papers)},
	publisher = {Association for Computational Linguistics},
	author = {Fu, Jinlan and Ng, See-Kiong and Jiang, Zhengbao and Liu, Pengfei},
	editor = {Duh, Kevin and Gomez, Helena and Bethard, Steven},
	month = jun,
	year = {2024},
	pages = {6556--6576},
}

@inproceedings{ji-etal-2024-submodular,
	address = {Torino, Italia},
	title = {Submodular-based in-context example selection for {LLMs}-based machine translation},
	url = {https://aclanthology.org/2024.lrec-main.1337/},
	booktitle = {Proceedings of the 2024 joint international conference on computational linguistics, language resources and evaluation ({LREC}-{COLING} 2024)},
	publisher = {ELRA and ICCL},
	author = {Ji, Baijun and Duan, Xiangyu and Qiu, Zhenyu and Zhang, Tong and Li, Junhui and Yang, Hao and Zhang, Min},
	editor = {Calzolari, Nicoletta and Kan, Min-Yen and Hoste, Veronique and Lenci, Alessandro and Sakti, Sakriani and Xue, Nianwen},
	month = may,
	year = {2024},
	pages = {15398--15409},
}

@inproceedings{m-etal-2023-ctqscorer,
	address = {Singapore},
	title = {{CTQScorer}: {Combining} multiple features for in-context example selection for machine translation},
	url = {https://aclanthology.org/2023.findings-emnlp.519/},
	doi = {10.18653/v1/2023.findings-emnlp.519},
	booktitle = {Findings of the association for computational linguistics: {EMNLP} 2023},
	publisher = {Association for Computational Linguistics},
	author = {Kumar, Aswanth and Puduppully, Ratish and Dabre, Raj and Kunchukuttan, Anoop},
	editor = {Bouamor, Houda and Pino, Juan and Bali, Kalika},
	month = dec,
	year = {2023},
	pages = {7736--7752},
}

@inproceedings{sia-duh-2023-context,
	address = {Macau SAR, China},
	title = {In-context learning as maintaining coherency: a study of on-the-fly machine translation using large language models},
	url = {https://aclanthology.org/2023.mtsummit-research.15/},
	booktitle = {Proceedings of machine translation summit {XIX}, vol. 1: {Research} track},
	publisher = {Asia-Pacific Association for Machine Translation},
	author = {Sia, Suzanna and Duh, Kevin},
	editor = {Utiyama, Masao and Wang, Rui},
	month = sep,
	year = {2023},
	pages = {173--185},
}

@inproceedings{aycock-bawden-2024-topic,
	address = {St. Julian's, Malta},
	title = {Topic-guided example selection for domain adaptation in {LLM}-based machine translation},
	url = {https://aclanthology.org/2024.eacl-srw.13/},
	booktitle = {Proceedings of the 18th conference of the european chapter of the association for computational linguistics: {Student} research workshop},
	publisher = {Association for Computational Linguistics},
	author = {Aycock, Seth and Bawden, Rachel},
	editor = {Falk, Neele and Papi, Sara and Zhang, Mike},
	month = mar,
	year = {2024},
	pages = {175--195},
}

@inproceedings{bang-etal-2023-multitask,
	address = {Nusa Dua, Bali},
	title = {A multitask, multilingual, multimodal evaluation of {ChatGPT} on reasoning, hallucination, and interactivity},
	url = {https://aclanthology.org/2023.ijcnlp-main.45/},
	doi = {10.18653/v1/2023.ijcnlp-main.45},
	booktitle = {Proceedings of the 13th international joint conference on natural language processing and the 3rd conference of the asia-pacific chapter of the association for computational linguistics (volume 1: {Long} papers)},
	publisher = {Association for Computational Linguistics},
	author = {Bang, Yejin and Cahyawijaya, Samuel and Lee, Nayeon and Dai, Wenliang and Su, Dan and Wilie, Bryan and Lovenia, Holy and Ji, Ziwei and Yu, Tiezheng and Chung, Willy and Do, Quyet V. and Xu, Yan and Fung, Pascale},
	editor = {Park, Jong C. and Arase, Yuki and Hu, Baotian and Lu, Wei and Wijaya, Derry and Purwarianti, Ayu and Krisnadhi, Adila Alfa},
	month = nov,
	year = {2023},
	pages = {675--718},
}

@misc{jiao_is_2023,
	title = {Is {ChatGPT} {A} {Good} {Translator}? {Yes} {With} {GPT}-4 {As} {The} {Engine}},
	shorttitle = {Is {ChatGPT} {A} {Good} {Translator}?},
	url = {http://arxiv.org/abs/2301.08745},
	urldate = {2023-05-15},
	publisher = {arXiv},
	author = {Jiao, Wenxiang and Wang, Wenxuan and Huang, Jen-tse and Wang, Xing and Tu, Zhaopeng},
	month = mar,
	year = {2023},
	note = {arXiv:2301.08745 [cs]},
}

@inproceedings{nguyen-tuan-nguyen-2020-phobert,
	address = {Online},
	title = {{PhoBERT}: {Pre}-trained language models for {Vietnamese}},
	url = {https://aclanthology.org/2020.findings-emnlp.92/},
	doi = {10.18653/v1/2020.findings-emnlp.92},
	booktitle = {Findings of the association for computational linguistics: {EMNLP} 2020},
	publisher = {Association for Computational Linguistics},
	author = {Nguyen, Dat Quoc and Tuan Nguyen, Anh},
	editor = {Cohn, Trevor and He, Yulan and Liu, Yang},
	month = nov,
	year = {2020},
	pages = {1037--1042},
}

@inproceedings{antoun-etal-2020-arabert,
	address = {Marseille, France},
	title = {{AraBERT}: {Transformer}-based model for {Arabic} language understanding},
	isbn = {979-10-95546-51-1},
	url = {https://aclanthology.org/2020.osact-1.2/},
	language = {eng},
	booktitle = {Proceedings of the 4th workshop on open-source arabic corpora and processing tools, with a shared task on offensive language detection},
	publisher = {European Language Resource Association},
	author = {Antoun, Wissam and Baly, Fady and Hajj, Hazem},
	editor = {Al-Khalifa, Hend and Magdy, Walid and Darwish, Kareem and Elsayed, Tamer and Mubarak, Hamdy},
	month = may,
	year = {2020},
	pages = {9--15},
}

@inproceedings{delobelle-etal-2020-robbert,
	address = {Online},
	title = {{RobBERT}: a {Dutch} {RoBERTa}-based {Language} {Model}},
	url = {https://aclanthology.org/2020.findings-emnlp.292/},
	doi = {10.18653/v1/2020.findings-emnlp.292},
	booktitle = {Findings of the association for computational linguistics: {EMNLP} 2020},
	publisher = {Association for Computational Linguistics},
	author = {Delobelle, Pieter and Winters, Thomas and Berendt, Bettina},
	editor = {Cohn, Trevor and He, Yulan and Liu, Yang},
	month = nov,
	year = {2020},
	pages = {3255--3265},
}

@inproceedings{martin-etal-2020-camembert,
	address = {Online},
	title = {{CamemBERT}: a tasty {French} language model},
	url = {https://aclanthology.org/2020.acl-main.645/},
	doi = {10.18653/v1/2020.acl-main.645},
	booktitle = {Proceedings of the 58th annual meeting of the association for computational linguistics},
	publisher = {Association for Computational Linguistics},
	author = {Martin, Louis and Muller, Benjamin and Ortiz Suárez, Pedro Javier and Dupont, Yoann and Romary, Laurent and de la Clergerie, Éric and Seddah, Djamé and Sagot, Benoît},
	editor = {Jurafsky, Dan and Chai, Joyce and Schluter, Natalie and Tetreault, Joel},
	month = jul,
	year = {2020},
	pages = {7203--7219},
}

@inproceedings{wu-etal-2020-single,
	address = {Online},
	title = {Single-/multi-source cross-lingual {NER} via teacher-student learning on unlabeled data in target language},
	url = {https://aclanthology.org/2020.acl-main.581/},
	doi = {10.18653/v1/2020.acl-main.581},
	booktitle = {Proceedings of the 58th annual meeting of the association for computational linguistics},
	publisher = {Association for Computational Linguistics},
	author = {Wu, Qianhui and Lin, Zijia and Karlsson, Börje and Lou, Jian-Guang and Huang, Biqing},
	editor = {Jurafsky, Dan and Chai, Joyce and Schluter, Natalie and Tetreault, Joel},
	month = jul,
	year = {2020},
	pages = {6505--6514},
}

@misc{pushp_train_2017,
	title = {Train {Once}, {Test} {Anywhere}: {Zero}-{Shot} {Learning} for {Text} {Classification}},
	shorttitle = {Train {Once}, {Test} {Anywhere}},
	url = {http://arxiv.org/abs/1712.05972},
	doi = {10.48550/arXiv.1712.05972},
	urldate = {2025-04-09},
	publisher = {arXiv},
	author = {Pushp, Pushpankar Kumar and Srivastava, Muktabh Mayank},
	month = dec,
	year = {2017},
	note = {arXiv:1712.05972 [cs]},
}

@inproceedings{rust-etal-2021-good,
	address = {Online},
	title = {How good is your tokenizer? {On} the monolingual performance of multilingual language models},
	url = {https://aclanthology.org/2021.acl-long.243/},
	doi = {10.18653/v1/2021.acl-long.243},
	booktitle = {Proceedings of the 59th annual meeting of the association for computational linguistics and the 11th international joint conference on natural language processing (volume 1: {Long} papers)},
	publisher = {Association for Computational Linguistics},
	author = {Rust, Phillip and Pfeiffer, Jonas and Vulić, Ivan and Ruder, Sebastian and Gurevych, Iryna},
	editor = {Zong, Chengqing and Xia, Fei and Li, Wenjie and Navigli, Roberto},
	month = aug,
	year = {2021},
	pages = {3118--3135},
}

@misc{liu_study_2020,
	title = {A {Study} of {Cross}-{Lingual} {Ability} and {Language}-specific {Information} in {Multilingual} {BERT}},
	url = {http://arxiv.org/abs/2004.09205},
	doi = {10.48550/arXiv.2004.09205},
	urldate = {2025-04-09},
	publisher = {arXiv},
	author = {Liu, Chi-Liang and Hsu, Tsung-Yuan and Chuang, Yung-Sung and Lee, Hung-Yi},
	month = apr,
	year = {2020},
	note = {arXiv:2004.09205 [cs]},
}

@misc{srinivasan_predicting_2021,
	title = {Predicting the {Performance} of {Multilingual} {NLP} {Models}},
	url = {http://arxiv.org/abs/2110.08875},
	doi = {10.48550/arXiv.2110.08875},
	urldate = {2025-04-09},
	publisher = {arXiv},
	author = {Srinivasan, Anirudh and Sitaram, Sunayana and Ganu, Tanuja and Dandapat, Sandipan and Bali, Kalika and Choudhury, Monojit},
	month = oct,
	year = {2021},
	note = {arXiv:2110.08875 [cs]},
}

@inproceedings{wu-etal-2023-oolong,
	address = {Singapore},
	title = {Oolong: {Investigating} what makes transfer learning hard with controlled studies},
	url = {https://aclanthology.org/2023.emnlp-main.198/},
	doi = {10.18653/v1/2023.emnlp-main.198},
	booktitle = {Proceedings of the 2023 conference on empirical methods in natural language processing},
	publisher = {Association for Computational Linguistics},
	author = {Wu, Zhengxuan and Tamkin, Alex and Papadimitriou, Isabel},
	editor = {Bouamor, Houda and Pino, Juan and Bali, Kalika},
	month = dec,
	year = {2023},
	pages = {3280--3289},
}

@misc{kovacevic_ezglot_2022,
	title = {Ezglot},
	author = {Kovacevic, Lazar and Bradic, Vladimir and de Melo, Gerard and Zdravkovic and Ryzhova, Olga},
	year = {2022},
}

@inproceedings{deshpande-etal-2022-bert,
	address = {Seattle, United States},
	title = {When is {BERT} multilingual? {Isolating} crucial ingredients for cross-lingual transfer},
	url = {https://aclanthology.org/2022.naacl-main.264/},
	doi = {10.18653/v1/2022.naacl-main.264},
	booktitle = {Proceedings of the 2022 conference of the north american chapter of the association for computational linguistics: {Human} language technologies},
	publisher = {Association for Computational Linguistics},
	author = {Deshpande, Ameet and Talukdar, Partha and Narasimhan, Karthik},
	editor = {Carpuat, Marine and de Marneffe, Marie-Catherine and Meza Ruiz, Ivan Vladimir},
	month = jul,
	year = {2022},
	pages = {3610--3623},
}

@inproceedings{de-vries-etal-2022-make,
	address = {Dublin, Ireland},
	title = {Make the best of cross-lingual transfer: {Evidence} from {POS} tagging with over 100 languages},
	url = {https://aclanthology.org/2022.acl-long.529/},
	doi = {10.18653/v1/2022.acl-long.529},
	booktitle = {Proceedings of the 60th annual meeting of the association for computational linguistics (volume 1: {Long} papers)},
	publisher = {Association for Computational Linguistics},
	author = {de Vries, Wietse and Wieling, Martijn and Nissim, Malvina},
	editor = {Muresan, Smaranda and Nakov, Preslav and Villavicencio, Aline},
	month = may,
	year = {2022},
	pages = {7676--7685},
}

@misc{dolicki_analysing_2021,
	title = {Analysing {The} {Impact} {Of} {Linguistic} {Features} {On} {Cross}-{Lingual} {Transfer}},
	url = {http://arxiv.org/abs/2105.05975},
	doi = {10.48550/arXiv.2105.05975},
	urldate = {2025-04-09},
	publisher = {arXiv},
	author = {Dolicki, Błażej and Spanakis, Gerasimos},
	month = may,
	year = {2021},
	note = {arXiv:2105.05975 [cs]},
}

@inproceedings{ahuja-etal-2022-multi,
	address = {Dublin, Ireland},
	title = {Multi task learning for zero shot performance prediction of multilingual models},
	url = {https://aclanthology.org/2022.acl-long.374/},
	doi = {10.18653/v1/2022.acl-long.374},
	booktitle = {Proceedings of the 60th annual meeting of the association for computational linguistics (volume 1: {Long} papers)},
	publisher = {Association for Computational Linguistics},
	author = {Ahuja, Kabir and Kumar, Shanu and Dandapat, Sandipan and Choudhury, Monojit},
	editor = {Muresan, Smaranda and Nakov, Preslav and Villavicencio, Aline},
	month = may,
	year = {2022},
	pages = {5454--5467},
}

@inproceedings{gu_meta-learning_2018,
	address = {Brussels, Belgium},
	title = {Meta-{Learning} for {Low}-{Resource} {Neural} {Machine} {Translation}},
	url = {http://aclweb.org/anthology/D18-1398},
	doi = {10.18653/v1/D18-1398},
	language = {en},
	urldate = {2025-04-08},
	booktitle = {Proceedings of the 2018 {Conference} on {Empirical} {Methods} in {Natural} {Language} {Processing}},
	publisher = {Association for Computational Linguistics},
	author = {Gu, Jiatao and Wang, Yong and Chen, Yun and Li, Victor O. K. and Cho, Kyunghyun},
	year = {2018},
	pages = {3622--3631},
}

@inproceedings{murthy_addressing_2019,
	address = {Minneapolis, Minnesota},
	title = {Addressing word-order {Divergence} in {Multilingual} {Neural} {Machine} {Translation} for extremely {Low} {Resource} {Languages}},
	url = {http://aclweb.org/anthology/N19-1387},
	doi = {10.18653/v1/N19-1387},
	language = {en},
	urldate = {2025-04-08},
	booktitle = {Proceedings of the 2019 {Conference} of the {North}},
	publisher = {Association for Computational Linguistics},
	author = {Murthy, Rudra and Kunchukuttan, Anoop and Bhattacharyya, Pushpak},
	year = {2019},
	pages = {3868--3873},
}

@article{kunchukuttan_leveraging_2018,
	title = {Leveraging {Orthographic} {Similarity} for {Multilingual} {Neural} {Transliteration}},
	volume = {6},
	issn = {2307-387X},
	url = {https://direct.mit.edu/tacl/article/43438},
	doi = {10.1162/tacl_a_00022},
	language = {en},
	urldate = {2025-04-08},
	journal = {Transactions of the Association for Computational Linguistics},
	author = {Kunchukuttan, Anoop and Khapra, Mitesh and Singh, Gurneet and Bhattacharyya, Pushpak},
	month = dec,
	year = {2018},
	pages = {303--316},
}

@inproceedings{dabre-etal-2017-empirical,
	title = {An empirical study of language relatedness for transfer learning in neural machine translation},
	url = {https://aclanthology.org/Y17-1038/},
	booktitle = {Proceedings of the 31st pacific asia conference on language, information and computation},
	publisher = {The National University (Phillippines)},
	author = {Dabre, Raj and Nakagawa, Tetsuji and Kazawa, Hideto},
	editor = {Roxas, Rachel Edita},
	month = nov,
	year = {2017},
	pages = {282--286},
}

@inproceedings{ustun_hyper-x_2022,
	address = {Abu Dhabi, United Arab Emirates},
	title = {Hyper-{X}: {A} {Unified} {Hypernetwork} for {Multi}-{Task} {Multilingual} {Transfer}},
	shorttitle = {Hyper-{X}},
	url = {https://aclanthology.org/2022.emnlp-main.541},
	doi = {10.18653/v1/2022.emnlp-main.541},
	language = {en},
	urldate = {2025-04-08},
	booktitle = {Proceedings of the 2022 {Conference} on {Empirical} {Methods} in {Natural} {Language} {Processing}},
	publisher = {Association for Computational Linguistics},
	author = {Üstün, Ahmet and Bisazza, Arianna and Bouma, Gosse and Van Noord, Gertjan and Ruder, Sebastian},
	year = {2022},
	pages = {7934--7949},
}

@inproceedings{pfeiffer_mad-x_2020,
	address = {Online},
	title = {{MAD}-{X}: {An} {Adapter}-{Based} {Framework} for {Multi}-{Task} {Cross}-{Lingual} {Transfer}},
	shorttitle = {{MAD}-{X}},
	url = {https://www.aclweb.org/anthology/2020.emnlp-main.617},
	doi = {10.18653/v1/2020.emnlp-main.617},
	language = {en},
	urldate = {2025-04-08},
	booktitle = {Proceedings of the 2020 {Conference} on {Empirical} {Methods} in {Natural} {Language} {Processing} ({EMNLP})},
	publisher = {Association for Computational Linguistics},
	author = {Pfeiffer, Jonas and Vulić, Ivan and Gurevych, Iryna and Ruder, Sebastian},
	year = {2020},
	pages = {7654--7673},
}

@misc{touvron_llama_2023,
	title = {{LLaMA}: {Open} and {Efficient} {Foundation} {Language} {Models}},
	shorttitle = {{LLaMA}},
	url = {http://arxiv.org/abs/2302.13971},
	urldate = {2023-12-18},
	publisher = {arXiv},
	author = {Touvron, Hugo and Lavril, Thibaut and Izacard, Gautier and Martinet, Xavier and Lachaux, Marie-Anne and Lacroix, Timothée and Rozière, Baptiste and Goyal, Naman and Hambro, Eric and Azhar, Faisal and Rodriguez, Aurelien and Joulin, Armand and Grave, Edouard and Lample, Guillaume},
	month = feb,
	year = {2023},
	note = {arXiv:2302.13971 [cs]},
}

@article{wei2022emergent,
	title = {Emergent abilities of large language models},
	issn = {2835-8856},
	url = {https://openreview.net/forum?id=yzkSU5zdwD},
	journal = {Transactions on Machine Learning Research},
	author = {Wei, Jason and Tay, Yi and Bommasani, Rishi and Raffel, Colin and Zoph, Barret and Borgeaud, Sebastian and Yogatama, Dani and Bosma, Maarten and Zhou, Denny and Metzler, Donald and Chi, Ed H. and Hashimoto, Tatsunori and Vinyals, Oriol and Liang, Percy and Dean, Jeff and Fedus, William},
	year = {2022},
}

@article{su_roformer_2024,
	title = {{RoFormer}: {Enhanced} transformer with rotary position embedding},
	volume = {568},
	issn = {0925-2312},
	url = {https://doi.org/10.1016/j.neucom.2023.127063},
	doi = {10.1016/j.neucom.2023.127063},
	number = {C},
	journal = {Neurocomput.},
	author = {Su, Jianlin and Ahmed, Murtadha and Lu, Yu and Pan, Shengfeng and Bo, Wen and Liu, Yunfeng},
	month = feb,
	year = {2024},
	note = {Number of pages: 12
Place: NLD
Publisher: Elsevier Science Publishers B. V.
tex.issue\_date: Feb 2024},
}

@incollection{10.5555/3454287.3454289,
	address = {Red Hook, NY, USA},
	title = {{ViLBERT}: pretraining task-agnostic visiolinguistic representations for vision-and-language tasks},
	booktitle = {Proceedings of the 33rd international conference on neural information processing systems},
	publisher = {Curran Associates Inc.},
	author = {Lu, Jiasen and Batra, Dhruv and Parikh, Devi and Lee, Stefan},
	year = {2019},
	note = {Number of pages: 11
tex.articleno: 2},
}

@inproceedings{feng_codebert_2020,
	address = {Online},
	title = {{CodeBERT}: {A} {Pre}-{Trained} {Model} for {Programming} and {Natural} {Languages}},
	shorttitle = {{CodeBERT}},
	url = {https://www.aclweb.org/anthology/2020.findings-emnlp.139},
	doi = {10.18653/v1/2020.findings-emnlp.139},
	language = {en},
	urldate = {2025-04-08},
	booktitle = {Findings of the {Association} for {Computational} {Linguistics}: {EMNLP} 2020},
	publisher = {Association for Computational Linguistics},
	author = {Feng, Zhangyin and Guo, Daya and Tang, Duyu and Duan, Nan and Feng, Xiaocheng and Gong, Ming and Shou, Linjun and Qin, Bing and Liu, Ting and Jiang, Daxin and Zhou, Ming},
	year = {2020},
	pages = {1536--1547},
}

@inproceedings{beltagy_scibert_2019,
	address = {Hong Kong, China},
	title = {{SciBERT}: {A} {Pretrained} {Language} {Model} for {Scientific} {Text}},
	shorttitle = {{SciBERT}},
	url = {https://www.aclweb.org/anthology/D19-1371},
	doi = {10.18653/v1/D19-1371},
	language = {en},
	urldate = {2025-04-08},
	booktitle = {Proceedings of the 2019 {Conference} on {Empirical} {Methods} in {Natural} {Language} {Processing} and the 9th {International} {Joint} {Conference} on {Natural} {Language} {Processing} ({EMNLP}-{IJCNLP})},
	publisher = {Association for Computational Linguistics},
	author = {Beltagy, Iz and Lo, Kyle and Cohan, Arman},
	year = {2019},
	pages = {3613--3618},
}

@inproceedings{zhang_pegasus_2020,
	series = {Proceedings of machine learning research},
	title = {{PEGASUS}: {Pre}-training with extracted gap-sentences for abstractive summarization},
	volume = {119},
	url = {https://proceedings.mlr.press/v119/zhang20ae.html},
	booktitle = {Proceedings of the 37th international conference on machine learning},
	publisher = {PMLR},
	author = {Zhang, Jingqing and Zhao, Yao and Saleh, Mohammad and Liu, Peter},
	editor = {III, Hal Daumé and Singh, Aarti},
	month = jul,
	year = {2020},
	pages = {11328--11339},
}

@inproceedings{NEURIPS2020_6b493230,
	title = {Retrieval-augmented generation for knowledge-intensive {NLP} tasks},
	volume = {33},
	url = {https://proceedings.neurips.cc/paper_files/paper/2020/file/6b493230205f780e1bc26945df7481e5-Paper.pdf},
	booktitle = {Advances in neural information processing systems},
	publisher = {Curran Associates, Inc.},
	author = {Lewis, Patrick and Perez, Ethan and Piktus, Aleksandra and Petroni, Fabio and Karpukhin, Vladimir and Goyal, Naman and Küttler, Heinrich and Lewis, Mike and Yih, Wen-tau and Rocktäschel, Tim and Riedel, Sebastian and Kiela, Douwe},
	editor = {Larochelle, H. and Ranzato, M. and Hadsell, R. and Balcan, M.F. and Lin, H.},
	year = {2020},
	pages = {9459--9474},
}

@misc{beltagy_longformer_2020,
	title = {Longformer: {The} {Long}-{Document} {Transformer}},
	shorttitle = {Longformer},
	url = {http://arxiv.org/abs/2004.05150},
	doi = {10.48550/arXiv.2004.05150},
	urldate = {2025-04-08},
	publisher = {arXiv},
	author = {Beltagy, Iz and Peters, Matthew E. and Cohan, Arman},
	month = dec,
	year = {2020},
	note = {arXiv:2004.05150 [cs]},
}

@inproceedings{choromanski2021rethinking,
	title = {Rethinking attention with performers},
	url = {https://openreview.net/forum?id=Ua6zuk0WRH},
	booktitle = {International conference on learning representations},
	author = {Choromanski, Krzysztof Marcin and Likhosherstov, Valerii and Dohan, David and Song, Xingyou and Gane, Andreea and Sarlos, Tamas and Hawkins, Peter and Davis, Jared Quincy and Mohiuddin, Afroz and Kaiser, Lukasz and Belanger, David Benjamin and Colwell, Lucy J and Weller, Adrian},
	year = {2021},
}

@misc{wang_linformer_2020,
	title = {Linformer: {Self}-{Attention} with {Linear} {Complexity}},
	shorttitle = {Linformer},
	url = {http://arxiv.org/abs/2006.04768},
	doi = {10.48550/arXiv.2006.04768},
	urldate = {2025-04-08},
	publisher = {arXiv},
	author = {Wang, Sinong and Li, Belinda Z. and Khabsa, Madian and Fang, Han and Ma, Hao},
	month = jun,
	year = {2020},
	note = {arXiv:2006.04768 [cs]},
}

@article{raffel_exploring_2020,
	title = {Exploring the limits of transfer learning with a unified text-to-text transformer},
	volume = {21},
	url = {http://jmlr.org/papers/v21/20-074.html},
	number = {140},
	journal = {Journal of Machine Learning Research},
	author = {Raffel, Colin and Shazeer, Noam and Roberts, Adam and Lee, Katherine and Narang, Sharan and Matena, Michael and Zhou, Yanqi and Li, Wei and Liu, Peter J.},
	year = {2020},
	pages = {1--67},
}

@inproceedings{de-gibert-etal-2024-new,
	address = {Torino, Italia},
	title = {A new massive multilingual dataset for high-performance language technologies},
	url = {https://aclanthology.org/2024.lrec-main.100/},
	booktitle = {Proceedings of the 2024 joint international conference on computational linguistics, language resources and evaluation ({LREC}-{COLING} 2024)},
	publisher = {ELRA and ICCL},
	author = {de Gibert, Ona and Nail, Graeme and Arefyev, Nikolay and Bañón, Marta and van der Linde, Jelmer and Ji, Shaoxiong and Zaragoza-Bernabeu, Jaume and Aulamo, Mikko and Ramírez-Sánchez, Gema and Kutuzov, Andrey and Pyysalo, Sampo and Oepen, Stephan and Tiedemann, Jörg},
	editor = {Calzolari, Nicoletta and Kan, Min-Yen and Hoste, Veronique and Lenci, Alessandro and Sakti, Sakriani and Xue, Nianwen},
	month = may,
	year = {2024},
	pages = {1116--1128},
}

@inproceedings{kudugunta2023madlad,
	title = {{MADLAD}-400: a multilingual and document-level large audited dataset},
	url = {https://openreview.net/forum?id=Y45ZCxslFx},
	booktitle = {Thirty-seventh conference on neural information processing systems datasets and benchmarks track},
	author = {Kudugunta, Sneha and Caswell, Isaac Rayburn and Zhang, Biao and Garcia, Xavier and Xin, Derrick and Kusupati, Aditya and Stella, Romi and Bapna, Ankur and Firat, Orhan},
	year = {2023},
}

@misc{abadji_towards_2022,
	title = {Towards a {Cleaner} {Document}-{Oriented} {Multilingual} {Crawled} {Corpus}},
	url = {http://arxiv.org/abs/2201.06642},
	doi = {10.48550/arXiv.2201.06642},
	urldate = {2025-04-07},
	publisher = {arXiv},
	author = {Abadji, Julien and Suarez, Pedro Ortiz and Romary, Laurent and Sagot, Benoît},
	month = jan,
	year = {2022},
	note = {arXiv:2201.06642 [cs]},
}

@inproceedings{xue_mt5_2021,
	address = {Online},
	title = {{mT5}: {A} {Massively} {Multilingual} {Pre}-trained {Text}-to-{Text} {Transformer}},
	shorttitle = {{mT5}},
	url = {https://aclanthology.org/2021.naacl-main.41},
	doi = {10.18653/v1/2021.naacl-main.41},
	language = {en},
	urldate = {2025-04-07},
	booktitle = {Proceedings of the 2021 {Conference} of the {North} {American} {Chapter} of the {Association} for {Computational} {Linguistics}: {Human} {Language} {Technologies}},
	publisher = {Association for Computational Linguistics},
	author = {Xue, Linting and Constant, Noah and Roberts, Adam and Kale, Mihir and Al-Rfou, Rami and Siddhant, Aditya and Barua, Aditya and Raffel, Colin},
	year = {2021},
	pages = {483--498},
}

@inproceedings{soldaini_dolma_2024,
	address = {Bangkok, Thailand},
	title = {Dolma: an {Open} {Corpus} of {Three} {Trillion} {Tokens} for {Language} {Model} {Pretraining} {Research}},
	shorttitle = {Dolma},
	url = {https://aclanthology.org/2024.acl-long.840},
	doi = {10.18653/v1/2024.acl-long.840},
	language = {en},
	urldate = {2025-04-07},
	booktitle = {Proceedings of the 62nd {Annual} {Meeting} of the {Association} for {Computational} {Linguistics} ({Volume} 1: {Long} {Papers})},
	publisher = {Association for Computational Linguistics},
	author = {Soldaini, Luca and Kinney, Rodney and Bhagia, Akshita and Schwenk, Dustin and Atkinson, David and Authur, Russell and Bogin, Ben and Chandu, Khyathi and Dumas, Jennifer and Elazar, Yanai and Hofmann, Valentin and Jha, Ananya and Kumar, Sachin and Lucy, Li and Lyu, Xinxi and Lambert, Nathan and Magnusson, Ian and Morrison, Jacob and Muennighoff, Niklas and Naik, Aakanksha and Nam, Crystal and Peters, Matthew and Ravichander, Abhilasha and Richardson, Kyle and Shen, Zejiang and Strubell, Emma and Subramani, Nishant and Tafjord, Oyvind and Walsh, Evan and Zettlemoyer, Luke and Smith, Noah and Hajishirzi, Hannaneh and Beltagy, Iz and Groeneveld, Dirk and Dodge, Jesse and Lo, Kyle},
	year = {2024},
	pages = {15725--15788},
}

@misc{gao_pile_2020,
	title = {The {Pile}: {An} {800GB} {Dataset} of {Diverse} {Text} for {Language} {Modeling}},
	shorttitle = {The {Pile}},
	url = {http://arxiv.org/abs/2101.00027},
	doi = {10.48550/arXiv.2101.00027},
	urldate = {2025-04-07},
	publisher = {arXiv},
	author = {Gao, Leo and Biderman, Stella and Black, Sid and Golding, Laurence and Hoppe, Travis and Foster, Charles and Phang, Jason and He, Horace and Thite, Anish and Nabeshima, Noa and Presser, Shawn and Leahy, Connor},
	month = dec,
	year = {2020},
	note = {arXiv:2101.00027 [cs]},
}

@inproceedings{domhan_using_2017,
	address = {Copenhagen, Denmark},
	title = {Using {Target}-side {Monolingual} {Data} for {Neural} {Machine} {Translation} through {Multi}-task {Learning}},
	url = {http://aclweb.org/anthology/D17-1158},
	doi = {10.18653/v1/D17-1158},
	language = {en},
	urldate = {2025-04-07},
	booktitle = {Proceedings of the 2017 {Conference} on {Empirical} {Methods} in {Natural}           {Language} {Processing}},
	publisher = {Association for Computational Linguistics},
	author = {Domhan, Tobias and Hieber, Felix},
	year = {2017},
	pages = {1500--1505},
}

@inproceedings{burlot_using_2018,
	address = {Belgium, Brussels},
	title = {Using {Monolingual} {Data} in {Neural} {Machine} {Translation}: a {Systematic} {Study}},
	shorttitle = {Using {Monolingual} {Data} in {Neural} {Machine} {Translation}},
	url = {http://aclweb.org/anthology/W18-6315},
	doi = {10.18653/v1/W18-6315},
	language = {en},
	urldate = {2025-04-07},
	booktitle = {Proceedings of the {Third} {Conference} on {Machine} {Translation}: {Research} {Papers}},
	publisher = {Association for Computational Linguistics},
	author = {Burlot, Franck and Yvon, François},
	year = {2018},
	pages = {144--155},
}

@inproceedings{zhang_exploiting_2016,
	address = {Austin, Texas},
	title = {Exploiting {Source}-side {Monolingual} {Data} in {Neural} {Machine} {Translation}},
	url = {http://aclweb.org/anthology/D16-1160},
	doi = {10.18653/v1/D16-1160},
	language = {en},
	urldate = {2025-04-07},
	booktitle = {Proceedings of the 2016 {Conference} on {Empirical} {Methods} in {Natural}           {Language} {Processing}},
	publisher = {Association for Computational Linguistics},
	author = {Zhang, Jiajun and Zong, Chengqing},
	year = {2016},
	pages = {1535--1545},
}

@inproceedings{kocmi_findings_2023,
	address = {Singapore},
	title = {Findings of the 2023 {Conference} on {Machine} {Translation} ({WMT23}): {LLMs} {Are} {Here} but {Not} {Quite} {There} {Yet}},
	shorttitle = {Findings of the 2023 {Conference} on {Machine} {Translation} ({WMT23})},
	url = {https://aclanthology.org/2023.wmt-1.1},
	doi = {10.18653/v1/2023.wmt-1.1},
	language = {en},
	urldate = {2024-01-30},
	booktitle = {Proceedings of the {Eighth} {Conference} on {Machine} {Translation}},
	publisher = {Association for Computational Linguistics},
	author = {Kocmi, Tom and Avramidis, Eleftherios and Bawden, Rachel and Bojar, Ondřej and Dvorkovich, Anton and Federmann, Christian and Fishel, Mark and Freitag, Markus and Gowda, Thamme and Grundkiewicz, Roman and Haddow, Barry and Koehn, Philipp and Marie, Benjamin and Monz, Christof and Morishita, Makoto and Murray, Kenton and Nagata, Makoto and Nakazawa, Toshiaki and Popel, Martin and Popović, Maja and Shmatova, Mariya},
	year = {2023},
	pages = {1--42},
}

@inproceedings{akhbardeh-etal-2021-findings,
	address = {Online},
	title = {Findings of the 2021 conference on machine translation ({WMT21})},
	url = {https://aclanthology.org/2021.wmt-1.1/},
	booktitle = {Proceedings of the sixth conference on machine translation},
	publisher = {Association for Computational Linguistics},
	author = {Akhbardeh, Farhad and Arkhangorodsky, Arkady and Biesialska, Magdalena and Bojar, Ondřej and Chatterjee, Rajen and Chaudhary, Vishrav and Costa-jussa, Marta R. and España-Bonet, Cristina and Fan, Angela and Federmann, Christian and Freitag, Markus and Graham, Yvette and Grundkiewicz, Roman and Haddow, Barry and Harter, Leonie and Heafield, Kenneth and Homan, Christopher and Huck, Matthias and Amponsah-Kaakyire, Kwabena and Kasai, Jungo and Khashabi, Daniel and Knight, Kevin and Kocmi, Tom and Koehn, Philipp and Lourie, Nicholas and Monz, Christof and Morishita, Makoto and Nagata, Masaaki and Nagesh, Ajay and Nakazawa, Toshiaki and Negri, Matteo and Pal, Santanu and Tapo, Allahsera Auguste and Turchi, Marco and Vydrin, Valentin and Zampieri, Marcos},
	editor = {Barrault, Loic and Bojar, Ondrej and Bougares, Fethi and Chatterjee, Rajen and Costa-jussa, Marta R. and Federmann, Christian and Fishel, Mark and Fraser, Alexander and Freitag, Markus and Graham, Yvette and Grundkiewicz, Roman and Guzman, Paco and Haddow, Barry and Huck, Matthias and Yepes, Antonio Jimeno and Koehn, Philipp and Kocmi, Tom and Martins, Andre and Morishita, Makoto and Monz, Christof},
	month = nov,
	year = {2021},
	pages = {1--88},
}

@inproceedings{banon_paracrawl_2020,
	address = {Online},
	title = {{ParaCrawl}: {Web}-{Scale} {Acquisition} of {Parallel} {Corpora}},
	url = {https://aclanthology.org/2020.acl-main.417},
	doi = {10.18653/v1/2020.acl-main.417},
	booktitle = {Proceedings of the 58th {Annual} {Meeting} of the {Association} for {Computational} {Linguistics}},
	publisher = {Association for Computational Linguistics},
	author = {Bañón, Marta and Chen, Pinzhen and Haddow, Barry and Heafield, Kenneth and Hoang, Hieu and Esplà-Gomis, Miquel and Forcada, Mikel L. and Kamran, Amir and Kirefu, Faheem and Koehn, Philipp and Ortiz Rojas, Sergio and Pla Sempere, Leopoldo and Ramírez-Sánchez, Gema and Sarrías, Elsa and Strelec, Marek and Thompson, Brian and Waites, William and Wiggins, Dion and Zaragoza, Jaume},
	editor = {Jurafsky, Dan and Chai, Joyce and Schluter, Natalie and Tetreault, Joel},
	month = jul,
	year = {2020},
	pages = {4555--4567},
}

@inproceedings{barrault-etal-2020-findings,
	address = {Online},
	title = {Findings of the 2020 conference on machine translation ({WMT20})},
	url = {https://aclanthology.org/2020.wmt-1.1/},
	booktitle = {Proceedings of the fifth conference on machine translation},
	publisher = {Association for Computational Linguistics},
	author = {Barrault, Loïc and Biesialska, Magdalena and Bojar, Ondřej and Costa-jussà, Marta R. and Federmann, Christian and Graham, Yvette and Grundkiewicz, Roman and Haddow, Barry and Huck, Matthias and Joanis, Eric and Kocmi, Tom and Koehn, Philipp and Lo, Chi-kiu and Ljubešić, Nikola and Monz, Christof and Morishita, Makoto and Nagata, Masaaki and Nakazawa, Toshiaki and Pal, Santanu and Post, Matt and Zampieri, Marcos},
	editor = {Barrault, Loïc and Bojar, Ondřej and Bougares, Fethi and Chatterjee, Rajen and Costa-jussà, Marta R. and Federmann, Christian and Fishel, Mark and Fraser, Alexander and Graham, Yvette and Guzman, Paco and Haddow, Barry and Huck, Matthias and Yepes, Antonio Jimeno and Koehn, Philipp and Martins, André and Morishita, Makoto and Monz, Christof and Nagata, Masaaki and Nakazawa, Toshiaki and Negri, Matteo},
	month = nov,
	year = {2020},
	pages = {1--55},
}

@inproceedings{barrault_findings_2019,
	address = {Florence, Italy},
	title = {Findings of the 2019 {Conference} on {Machine} {Translation} ({WMT19})},
	url = {https://www.aclweb.org/anthology/W19-5301},
	doi = {10.18653/v1/W19-5301},
	language = {en},
	urldate = {2025-04-07},
	booktitle = {Proceedings of the {Fourth} {Conference} on {Machine} {Translation} ({Volume} 2: {Shared} {Task} {Papers}, {Day} 1)},
	publisher = {Association for Computational Linguistics},
	author = {Barrault, Loïc and Bojar, Ondřej and Costa-jussà, Marta R. and Federmann, Christian and Fishel, Mark and Graham, Yvette and Haddow, Barry and Huck, Matthias and Koehn, Philipp and Malmasi, Shervin and Monz, Christof and Müller, Mathias and Pal, Santanu and Post, Matt and Zampieri, Marcos},
	year = {2019},
	pages = {1--61},
}

@inproceedings{steinberger-etal-2006-jrc,
	address = {Genoa, Italy},
	title = {The {JRC}-{Acquis}: a multilingual aligned parallel corpus with 20+ languages},
	url = {https://aclanthology.org/L06-1196/},
	booktitle = {Proceedings of the fifth international conference on language resources and evaluation ({LREC}`06)},
	publisher = {European Language Resources Association (ELRA)},
	author = {Steinberger, Ralf and Pouliquen, Bruno and Widiger, Anna and Ignat, Camelia and Erjavec, Tomaž and Tufiş, Dan and Varga, Dániel},
	editor = {Calzolari, Nicoletta and Choukri, Khalid and Gangemi, Aldo and Maegaard, Bente and Mariani, Joseph and Odijk, Jan and Tapias, Daniel},
	month = may,
	year = {2006},
}

@inproceedings{kew_turning_2024,
	address = {Miami, Florida, USA},
	title = {Turning {English}-centric {LLMs} {Into} {Polyglots}: {How} {Much} {Multilinguality} {Is} {Needed}?},
	shorttitle = {Turning {English}-centric {LLMs} {Into} {Polyglots}},
	url = {https://aclanthology.org/2024.findings-emnlp.766},
	doi = {10.18653/v1/2024.findings-emnlp.766},
	language = {en},
	urldate = {2025-04-07},
	booktitle = {Findings of the {Association} for {Computational} {Linguistics}: {EMNLP} 2024},
	publisher = {Association for Computational Linguistics},
	author = {Kew, Tannon and Schottmann, Florian and Sennrich, Rico},
	year = {2024},
	pages = {13097--13124},
}

@article{vandermaaten2008visualizing,
	title = {Visualizing data using t-{SNE}},
	volume = {9},
	url = {http://jmlr.org/papers/v9/vandermaaten08a.html},
	number = {86},
	journal = {Journal of Machine Learning Research},
	author = {van der Maaten, Laurens and Hinton, Geoffrey},
	year = {2008},
	pages = {2579--2605},
}

@misc{wang_bridging_2024,
	title = {Bridging the {Language} {Gaps} in {Large} {Language} {Models} with {Inference}-{Time} {Cross}-{Lingual} {Intervention}},
	url = {http://arxiv.org/abs/2410.12462},
	doi = {10.48550/arXiv.2410.12462},
	urldate = {2025-04-01},
	publisher = {arXiv},
	author = {Wang, Weixuan and Wu, Minghao and Haddow, Barry and Birch, Alexandra},
	month = oct,
	year = {2024},
	note = {arXiv:2410.12462 [cs]},
}

@misc{liu_middle-layer_2025,
	title = {Middle-{Layer} {Representation} {Alignment} for {Cross}-{Lingual} {Transfer} in {Fine}-{Tuned} {LLMs}},
	url = {http://arxiv.org/abs/2502.14830},
	doi = {10.48550/arXiv.2502.14830},
	urldate = {2025-04-01},
	publisher = {arXiv},
	author = {Liu, Danni and Niehues, Jan},
	month = feb,
	year = {2025},
	note = {arXiv:2502.14830 [cs]},
}

@misc{deutsch_wmt24_2025,
	title = {{WMT24}++: {Expanding} the {Language} {Coverage} of {WMT24} to 55 {Languages} \& {Dialects}},
	shorttitle = {{WMT24}++},
	url = {http://arxiv.org/abs/2502.12404},
	doi = {10.48550/arXiv.2502.12404},
	urldate = {2025-04-01},
	publisher = {arXiv},
	author = {Deutsch, Daniel and Briakou, Eleftheria and Caswell, Isaac and Finkelstein, Mara and Galor, Rebecca and Juraska, Juraj and Kovacs, Geza and Lui, Alison and Rei, Ricardo and Riesa, Jason and Rijhwani, Shruti and Riley, Parker and Salesky, Elizabeth and Trabelsi, Firas and Winkler, Stephanie and Zhang, Biao and Freitag, Markus},
	month = feb,
	year = {2025},
	note = {arXiv:2502.12404 [cs]},
}

@misc{phan_humanitys_2025,
	title = {Humanity's {Last} {Exam}},
	url = {http://arxiv.org/abs/2501.14249},
	doi = {10.48550/arXiv.2501.14249},
	urldate = {2025-03-19},
	publisher = {arXiv},
	author = {Phan, Long and Gatti, Alice and Han, Ziwen and Li, Nathaniel and Hu, Josephina and Zhang, Hugh and Zhang, Chen Bo Calvin and Shaaban, Mohamed and Ling, John and Shi, Sean and Choi, Michael and Agrawal, Anish and Chopra, Arnav and Khoja, Adam and Kim, Ryan and Ren, Richard and Hausenloy, Jason and Zhang, Oliver and Mazeika, Mantas and Nguyen, Tung and Anderson, Daron and Shah, Imad Ali and Doroshenko, Mikhail and Stokes, Alun Cennyth and Mahmood, Mobeen and Lee, Jaeho and Pokutnyi, Oleksandr and Iskra, Oleg and Wang, Jessica P. and Gerbicz, Robert and Levin, John-Clark and Popov, Serguei and Feng, Fiona and Feng, Steven Y. and Zhao, Haoran and Yu, Michael and Gangal, Varun and Zou, Chelsea and Wang, Zihan and Kazakov, Mstyslav and Galgon, Geoff and Schmitt, Johannes and Sanchez, Alvaro and Lee, Yongki and Yeadon, Will and Sauers, Scott and Roth, Marc and Agu, Chidozie and Riis, Søren and Giska, Fabian and Utpala, Saiteja and Cheatom, Antrell and Giboney, Zachary and Goshu, Gashaw M. and Crowson, Sarah-Jane and Naiya, Mohinder Maheshbhai and Burns, Noah and Finke, Lennart and Cheng, Zerui and Park, Hyunwoo and Fournier-Facio, Francesco and Zampese, Jennifer and Wydallis, John B. and Hoerr, Ryan G. and Nandor, Mark and Gehrunger, Tim and Cai, Jiaqi and McCarty, Ben and Nam, Jungbae and Taylor, Edwin and Jin, Jun and Loume, Gautier Abou and Cao, Hangrui and Garretson, Alexis C. and Sileo, Damien and Ren, Qiuyu and Cojoc, Doru and Arkhipov, Pavel and Qazi, Usman and Bacho, Aras and Li, Lianghui and Motwani, Sumeet and Witt, Christian Schroeder de and Kopylov, Alexei and Veith, Johannes and Singer, Eric and Rissone, Paolo and Jin, Jaehyeok and Shi, Jack Wei Lun and Willcocks, Chris G. and Prabhu, Ameya and Tang, Longke and Zhou, Kevin and Santos, Emily de Oliveira and Maksimov, Andrey Pupasov and Vendrow, Edward and Zenitani, Kengo and Robinson, Joshua and Mikov, Aleksandar and Guillod, Julien and Li, Yuqi and Pageler, Ben and Vendrow, Joshua and Kuchkin, Vladyslav and Marion, Pierre and Efremov, Denis and Lynch, Jayson and Liang, Kaiqu and Gritsevskiy, Andrew and Martinez, Dakotah and Crispino, Nick and Zvonkine, Dimitri and Fraga, Natanael Wildner and Soori, Saeed and Press, Ori and Tang, Henry and Salazar, Julian and Green, Sean R. and Brüssel, Lina and Twayana, Moon and Dieuleveut, Aymeric and Rogers, T. Ryan and Zhang, Wenjin and Finocchio, Ross and Li, Bikun and Yang, Jinzhou and Rao, Arun and Loiseau, Gabriel and Kalinin, Mikhail and Lukas, Marco and Manolescu, Ciprian and Stambaugh, Nate and Mishra, Subrata and Kamdoum, Ariel Ghislain Kemogne and Hogg, Tad and Jin, Alvin and Bosio, Carlo and Sun, Gongbo and Coppola, Brian P. and Heidinger, Haline and Sayous, Rafael and Ivanov, Stefan and Cavanagh, Joseph M. and Shen, Jiawei and Imperial, Joseph Marvin and Schwaller, Philippe and Senthilkuma, Shaipranesh and Bran, Andres M. and Algaba, Andres and Verbeken, Brecht and Houte, Kelsey Van den and Sypt, Lynn Van Der and Noever, David and Schut, Lisa and Sucholutsky, Ilia and Zheltonozhskii, Evgenii and Yuan, Qiaochu and Lim, Derek and Stanley, Richard and Sivarajan, Shankar and Yang, Tong and Maar, John and Wykowski, Julian and Oller, Martí and Sandlin, Jennifer and Sahu, Anmol and Ardito, Cesare Giulio and Hu, Yuzheng and Dias, Felipe Meneguitti and Kreiman, Tobias and Rawal, Kaivalya and Vilchis, Tobias Garcia and Zu, Yuexuan and Lackner, Martin and Koppel, James and Nguyen, Jeremy and Antonenko, Daniil S. and Chern, Steffi and Zhao, Bingchen and Arsene, Pierrot and Ivanov, Sergey and Poświata, Rafał and Wang, Chenguang and Li, Daofeng and Crisostomi, Donato and Dehghan, Ali and Achilleos, Andrea and Ambay, John Arnold and Myklebust, Benjamin and Sen, Archan and Perrella, David and Kaparov, Nurdin and Inlow, Mark H. and Zang, Allen and Ramakrishnan, Kalyan and Orel, Daniil and Poritski, Vladislav and Ben-David, Shalev and Berger, Zachary and Whitfill, Parker and Foster, Michael and Munro, Daniel and Ho, Linh and Hava, Dan Bar and Kuchkin, Aleksey and Lauff, Robert and Holmes, David and Sommerhage, Frank and Zhang, Anji and Moat, Richard and Schneider, Keith and Pyda, Daniel and Kazibwe, Zakayo and Singh, Mukhwinder and Clarke, Don and Kim, Dae Hyun and Fish, Sara and Elser, Veit and Vilchis, Victor Efren Guadarrama and Klose, Immo and Demian, Christoph and Anantheswaran, Ujjwala and Zweiger, Adam and Albani, Guglielmo and Li, Jeffery and Daans, Nicolas and Radionov, Maksim and Rozhoň, Václav and Ginis, Vincent and Ma, Ziqiao and Stump, Christian and Platnick, Jacob and Nevirkovets, Volodymyr and Basler, Luke and Piccardo, Marco and Cohen, Niv and Singh, Virendra and Tkadlec, Josef and Rosu, Paul and Goldfarb, Alan and Padlewski, Piotr and Barzowski, Stanislaw and Montgomery, Kyle and Menezes, Aline and Patel, Arkil and Wang, Zixuan and Tucker-Foltz, Jamie and Stade, Jack and Grabb, Declan and Goertzen, Tom and Kazemi, Fereshteh and Milbauer, Jeremiah and Shukla, Abhishek and Elgnainy, Hossam and Labrador, Yan Carlos Leyva and He, Hao and Zhang, Ling and Givré, Alan and Wolff, Hew and Demir, Gözdenur and Aziz, Muhammad Fayez and Kaddar, Younesse and Ängquist, Ivar and Chen, Yanxu and Thornley, Elliott and Zhang, Robin and Pan, Jiayi and Terpin, Antonio and Muennighoff, Niklas and Schoelkopf, Hailey and Zheng, Eric and Carmi, Avishy and Shah, Jainam and Brown, Ethan D. L. and Zhu, Kelin and Bartolo, Max and Wheeler, Richard and Ho, Andrew and Barkan, Shaul and Wang, Jiaqi and Stehberger, Martin and Kretov, Egor and Bradshaw, Peter and Heimonen, J. P. and Sridhar, Kaustubh and Hossain, Zaki and Akov, Ido and Makarychev, Yury and Tam, Joanna and Hoang, Hieu and Cunningham, David M. and Goryachev, Vladimir and Patramanis, Demosthenes and Krause, Michael and Redenti, Andrew and Aldous, David and Lai, Jesyin and Coleman, Shannon and Xu, Jiangnan and Lee, Sangwon and Magoulas, Ilias and Zhao, Sandy and Tang, Ning and Cohen, Michael K. and Carroll, Micah and Paradise, Orr and Kirchner, Jan Hendrik and Steinerberger, Stefan and Ovchynnikov, Maksym and Matos, Jason O. and Shenoy, Adithya and Wang, Michael and Nie, Yuzhou and Giordano, Paolo and Petersen, Philipp and Sztyber-Betley, Anna and Faraboschi, Paolo and Riblet, Robin and Crozier, Jonathan and Halasyamani, Shiv and Pinto, Antonella and Verma, Shreyas and Joshi, Prashant and Meril, Eli and Yong, Zheng-Xin and Tee, Allison and Andréoletti, Jérémy and Weller, Orion and Singhal, Raghav and Zhang, Gang and Ivanov, Alexander and Khoury, Seri and Gustafsson, Nils and Mostaghimi, Hamid and Thaman, Kunvar and Chen, Qijia and Khánh, Tran Quoc and Loader, Jacob and Cavalleri, Stefano and Szlyk, Hannah and Brown, Zachary and Narayan, Himanshu and Roberts, Jonathan and Alley, William and Sun, Kunyang and Stendall, Ryan and Lamparth, Max and Reuel, Anka and Wang, Ting and Xu, Hanmeng and Hernández-Cámara, Pablo and Martin, Freddie and Preu, Thomas and Korbak, Tomek and Abramovitch, Marcus and Williamson, Dominic and Bosio, Ida and Chen, Ziye and Bálint, Biró and Lo, Eve J. Y. and Nunes, Maria Inês S. and Jiang, Yibo and Bari, M. Saiful and Kassani, Peyman and Wang, Zihao and Ansarinejad, Behzad and Sun, Yewen and Durand, Stephane and Douville, Guillaume and Tordera, Daniel and Balabanian, George and Anderson, Earth and Kvistad, Lynna and Moyano, Alejandro José and Milliron, Hsiaoyun and Sakor, Ahmad and Eron, Murat and McAlister, Isaac C. and O, Andrew Favre D. and Shah, Shailesh and Zhou, Xiaoxiang and Kamalov, Firuz and Clark, Ronald and Abdoli, Sherwin and Santens, Tim and Wang, Harrison K. and Chen, Evan and Tomasiello, Alessandro and Luca, G. Bruno De and Looi, Shi-Zhuo and Le, Vinh-Kha and Kolt, Noam and Mündler, Niels and Semler, Avi and Rodman, Emma and Drori, Jacob and Fossum, Carl J. and Gloor, Luk and Jagota, Milind and Pradeep, Ronak and Fan, Honglu and Shah, Tej and Eicher, Jonathan and Chen, Michael and Thaman, Kushal and Merrill, William and Firsching, Moritz and Harris, Carter and Ciobâcă, Stefan and Gross, Jason and Pandey, Rohan and Gusev, Ilya and Jones, Adam and Agnihotri, Shashank and Zhelnov, Pavel and Usawasutsakorn, Siranut and Mofayezi, Mohammadreza and Piperski, Alexander and Carauleanu, Marc and Zhang, David K. and Dobarskyi, Kostiantyn and Ler, Dylan and Leventov, Roman and Soroko, Ignat and Jansen, Thorben and Creighton, Scott and Lauer, Pascal and Duersch, Joshua and Taamazyan, Vage and Bezzi, Dario and Morak, Wiktor and Ma, Wenjie and Held, William and Huy, Tran Đuc and Xian, Ruicheng and Zebaze, Armel Randy and Mohamed, Mohanad and Leser, Julian Noah and Yuan, Michelle X. and Yacar, Laila and Lengler, Johannes and Olszewska, Katarzyna and Shahrtash, Hossein and Oliveira, Edson and Jackson, Joseph W. and Gonzalez, Daniel Espinosa and Zou, Andy and Chidambaram, Muthu and Manik, Timothy and Haffenden, Hector and Stander, Dashiell and Dasouqi, Ali and Shen, Alexander and Duc, Emilien and Golshani, Bita and Stap, David and Uzhou, Mikalai and Zhidkovskaya, Alina Borisovna and Lewark, Lukas and Rodriguez, Miguel Orbegozo and Vincze, Mátyás and Wehr, Dustin and Tang, Colin and Phillips, Shaun and Samuele, Fortuna and Muzhen, Jiang and Ekström, Fredrik and Hammon, Angela and Patel, Oam and Farhidi, Faraz and Medley, George and Mohammadzadeh, Forough and Peñaflor, Madellene and Kassahun, Haile and Friedrich, Alena and Sparrow, Claire and Perez, Rayner Hernandez and Sakal, Taom and Dhamane, Omkar and Mirabadi, Ali Khajegili and Hallman, Eric and Okutsu, Kenchi and Battaglia, Mike and Maghsoudimehrabani, Mohammad and Amit, Alon and Hulbert, Dave and Pereira, Roberto and Weber, Simon and Handoko and Peristyy, Anton and Malina, Stephen and Albanie, Samuel and Cai, Will and Mehkary, Mustafa and Aly, Rami and Reidegeld, Frank and Dick, Anna-Katharina and Friday, Cary and Sidhu, Jasdeep and Shapourian, Hassan and Kim, Wanyoung and Costa, Mariana and Gurdogan, Hubeyb and Weber, Brian and Kumar, Harsh and Jiang, Tong and Agarwal, Arunim and Ceconello, Chiara and Vaz, Warren S. and Zhuang, Chao and Park, Haon and Tawfeek, Andrew R. and Aggarwal, Daattavya and Kirchhof, Michael and Dai, Linjie and Kim, Evan and Ferret, Johan and Wang, Yuzhou and Yan, Minghao and Burdzy, Krzysztof and Zhang, Lixin and Franca, Antonio and Pham, Diana T. and Loh, Kang Yong and Robinson, Joshua and Jackson, Abram and Gul, Shreen and Chhablani, Gunjan and Du, Zhehang and Cosma, Adrian and Colino, Jesus and White, Colin and Votava, Jacob and Vinnikov, Vladimir and Delaney, Ethan and Spelda, Petr and Stritecky, Vit and Shahid, Syed M. and Mourrat, Jean-Christophe and Vetoshkin, Lavr and Sponselee, Koen and Bacho, Renas and Rosa, Florencia de la and Li, Xiuyu and Malod, Guillaume and Lang, Leon and Laurendeau, Julien and Kazakov, Dmitry and Adesanya, Fatimah and Portier, Julien and Hollom, Lawrence and Souza, Victor and Zhou, Yuchen Anna and Degorre, Julien and Yalın, Yiğit and Obikoya, Gbenga Daniel and Arnaboldi, Luca and Rai and Bigi, Filippo and Boscá, M. C. and Shumar, Oleg and Bacho, Kaniuar and Clavier, Pierre and Recchia, Gabriel and Popescu, Mara and Shulga, Nikita and Tanwie, Ngefor Mildred and Peskoff, Denis and Lux, Thomas C. H. and Rank, Ben and Ni, Colin and Brooks, Matthew and Yakimchyk, Alesia and Huanxu and Liu and Häggström, Olle and Verkama, Emil and Gundlach, Hans and Brito-Santana, Leonor and Amaro, Brian and Vajipey, Vivek and Grover, Rynaa and Fan, Yiyang and Silva, Gabriel Poesia Reis e and Xin, Linwei and Kratish, Yosi and Łucki, Jakub and Li, Wen-Ding and Gopi, Sivakanth and Caciolai, Andrea and Xu, Justin and Scaria, Kevin Joseph and Vargus, Freddie and Habibi, Farzad and Long and Lian and Rodolà, Emanuele and Robins, Jules and Cheng, Vincent and Fruhauff, Tony and Raynor, Brad and Qi, Hao and Jiang, Xi and Segev, Ben and Fan, Jingxuan and Martinson, Sarah and Wang, Erik Y. and Hausknecht, Kaylie and Brenner, Michael P. and Mao, Mao and Zhang, Xinyu and Avagian, David and Scipio, Eshawn Jessica and Ragoler, Alon and Tan, Justin and Sims, Blake and Plecnik, Rebeka and Kirtland, Aaron and Bodur, Omer Faruk and Shinde, D. P. and Adoul, Zahra and Zekry, Mohamed and Karakoc, Ali and Santos, Tania C. B. and Shamseldeen, Samir and Karim, Loukmane and Liakhovitskaia, Anna and Resman, Nate and Farina, Nicholas and Gonzalez, Juan Carlos and Maayan, Gabe and Hoback, Sarah and Pena, Rodrigo De Oliveira and Sherman, Glen and Kelley, Elizabeth and Mariji, Hodjat and Pouriamanesh, Rasoul and Wu, Wentao and Mendoza, Sandra and Alarab, Ismail and Cole, Joshua and Ferreira, Danyelle and Johnson, Bryan and Safdari, Mohammad and Dai, Liangti and Arthornthurasuk, Siriphan and Pronin, Alexey and Fan, Jing and Ramirez-Trinidad, Angel and Cartwright, Ashley and Pottmaier, Daphiny and Taheri, Omid and Outevsky, David and Stepanic, Stanley and Perry, Samuel and Askew, Luke and Rodríguez, Raúl Adrián Huerta and Minissi, Ali M. R. and Ali, Sam and Lorena, Ricardo and Iyer, Krishnamurthy and Fasiludeen, Arshad Anil and Salauddin, Sk Md and Islam, Murat and Gonzalez, Juan and Ducey, Josh and Somrak, Maja and Mavroudis, Vasilios and Vergo, Eric and Qin, Juehang and Borbás, Benjámin and Chu, Eric and Lindsey, Jack and Radhakrishnan, Anil and Jallon, Antoine and McInnis, I. M. J. and Kumar, Pawan and Goswami, Laxman Prasad and Bugas, Daniel and Heydari, Nasser and Jeanplong, Ferenc and Apronti, Archimedes and Galal, Abdallah and Ze-An, Ng and Singh, Ankit and Xavier, Joan of Arc and Agarwal, Kanu Priya and Berkani, Mohammed and Junior, Benedito Alves de Oliveira and Malishev, Dmitry and Remy, Nicolas and Hartman, Taylor D. and Tarver, Tim and Mensah, Stephen and Gimenez, Javier and Montecillo, Roselynn Grace and Campbell, Russell and Sharma, Asankhaya and Meer, Khalida and Alapont, Xavier and Patil, Deepakkumar and Maheshwari, Rajat and Dendane, Abdelkader and Shukla, Priti and Bogdanov, Sergei and Möller, Sören and Siddiqi, Muhammad Rehan and Saxena, Prajvi and Gupta, Himanshu and Enyekwe, Innocent and V, Ragavendran P. and EL-Wasif, Zienab and Maksapetyan, Aleksandr and Rossbach, Vivien and Harjadi, Chris and Bahaloohoreh, Mohsen and Bian, Song and Lai, John and Uro, Justine Leon and Bateman, Greg and Sayed, Mohamed and Menshawy, Ahmed and Duclosel, Darling and Jain, Yashaswini and Aaron, Ashley and Tiryakioglu, Murat and Siddh, Sheeshram and Krenek, Keith and Hoover, Alex and McGowan, Joseph and Patwardhan, Tejal and Yue, Summer and Wang, Alexandr and Hendrycks, Dan},
	month = feb,
	year = {2025},
	note = {arXiv:2501.14249 [cs]},
}

@book{mrl-2024-1,
	address = {Miami, Florida, USA},
	title = {Proceedings of the fourth workshop on multilingual representation learning ({MRL} 2024)},
	url = {https://aclanthology.org/2024.mrl-1.0/},
	publisher = {Association for Computational Linguistics},
	editor = {Sälevä, Jonne and Owodunni, Abraham},
	month = nov,
	year = {2024},
	doi = {10.18653/v1/2024.mrl-1.0},
}

@inproceedings{aycock2025can,
	title = {Can {LLMs} really learn to translate a low-resource language from one grammar book?},
	url = {https://openreview.net/forum?id=aMBSY2ebPw},
	booktitle = {The thirteenth international conference on learning representations},
	author = {Aycock, Seth and Stap, David and Wu, Di and Monz, Christof and Sima'an, Khalil},
	year = {2025},
}

@inproceedings{enevoldsen2025mmteb,
	title = {{MMTEB}: {Massive} multilingual text embedding benchmark},
	url = {https://openreview.net/forum?id=zl3pfz4VCV},
	booktitle = {The thirteenth international conference on learning representations},
	author = {Enevoldsen, Kenneth and Chung, Isaac and Kerboua, Imene and Kardos, Márton and Mathur, Ashwin and Stap, David and Gala, Jay and Siblini, Wissam and Krzemiński, Dominik and Winata, Genta Indra and Sturua, Saba and Utpala, Saiteja and Ciancone, Mathieu and Schaeffer, Marion and Misra, Diganta and Dhakal, Shreeya and Rystrøm, Jonathan and Solomatin, Roman and Çağatan, Ömer Veysel and Kundu, Akash and Bernstorff, Martin and Xiao, Shitao and Sukhlecha, Akshita and Pahwa, Bhavish and Poświata, Rafał and GV, Kranthi Kiran and Ashraf, Shawon and Auras, Daniel and Plüster, Björn and Harries, Jan Philipp and Magne, Loı̈c and Mohr, Isabelle and Zhu, Dawei and Gisserot-Boukhlef, Hippolyte and Aarsen, Tom and Kostkan, Jan and Wojtasik, Konrad and Lee, Taemin and Suppa, Marek and Zhang, Crystina and Rocca, Roberta and Hamdy, Mohammed and Michail, Andrianos and Yang, John and Faysse, Manuel and Vatolin, Aleksei and Thakur, Nandan and Dey, Manan and Vasani, Dipam and Chitale, Pranjal A and Tedeschi, Simone and Tai, Nguyen and Snegirev, Artem and Hendriksen, Mariya and Günther, Michael and Xia, Mengzhou and Shi, Weijia and Lù, Xing Han and Clive, Jordan and K, Gayatri and Anna, Maksimova and Wehrli, Silvan and Tikhonova, Maria and Panchal, Henil Shalin and Abramov, Aleksandr and Ostendorff, Malte and Liu, Zheng and Clematide, Simon and Miranda, Lester James Validad and Fenogenova, Alena and Song, Guangyu and Safi, Ruqiya Bin and Li, Wen-Ding and Borghini, Alessia and Cassano, Federico and Hansen, Lasse and Hooker, Sara and Xiao, Chenghao and Adlakha, Vaibhav and Weller, Orion and Reddy, Siva and Muennighoff, Niklas},
	year = {2025},
}

@inproceedings{tan-etal-2024-uva,
	address = {Miami, Florida, USA},
	title = {{UvA}-{MT}`s participation in the {WMT24} general translation shared task},
	url = {https://aclanthology.org/2024.wmt-1.11/},
	doi = {10.18653/v1/2024.wmt-1.11},
	booktitle = {Proceedings of the ninth conference on machine translation},
	publisher = {Association for Computational Linguistics},
	author = {Tan, Shaomu and Stap, David and Aycock, Seth and Monz, Christof and Wu, Di},
	editor = {Haddow, Barry and Kocmi, Tom and Koehn, Philipp and Monz, Christof},
	month = nov,
	year = {2024},
	pages = {176--184},
}

@inproceedings{sainz-etal-2024-data,
	address = {Bangkok, Thailand},
	title = {Data contamination report from the 2024 {CONDA} shared task},
	url = {https://aclanthology.org/2024.conda-1.4/},
	doi = {10.18653/v1/2024.conda-1.4},
	booktitle = {Proceedings of the 1st workshop on data contamination ({CONDA})},
	publisher = {Association for Computational Linguistics},
	author = {Sainz, Oscar and García-Ferrero, Iker and Jacovi, Alon and Ander Campos, Jon and Elazar, Yanai and Agirre, Eneko and Goldberg, Yoav and Chen, Wei-Lin and Chim, Jenny and Choshen, Leshem and D'Amico-Wong, Luca and Dell, Melissa and Fan, Run-Ze and Golchin, Shahriar and Li, Yucheng and Liu, Pengfei and Pahwa, Bhavish and Prabhu, Ameya and Sharma, Suryansh and Silcock, Emily and Solonko, Kateryna and Stap, David and Surdeanu, Mihai and Tseng, Yu-Min and Udandarao, Vishaal and Wang, Zengzhi and Xu, Ruijie and Yang, Jinglin},
	editor = {Sainz, Oscar and García Ferrero, Iker and Agirre, Eneko and Ander Campos, Jon and Jacovi, Alon and Elazar, Yanai and Goldberg, Yoav},
	month = aug,
	year = {2024},
	pages = {41--56},
}

@inproceedings{wu-etal-2024-far,
	address = {Bangkok, Thailand},
	title = {How far can 100 samples go? {Unlocking} zero-shot translation with tiny multi-parallel data},
	url = {https://aclanthology.org/2024.findings-acl.896/},
	doi = {10.18653/v1/2024.findings-acl.896},
	booktitle = {Findings of the association for computational linguistics: {ACL} 2024},
	publisher = {Association for Computational Linguistics},
	author = {Wu, Di and Tan, Shaomu and Meng, Yan and Stap, David and Monz, Christof},
	editor = {Ku, Lun-Wei and Martins, Andre and Srikumar, Vivek},
	month = aug,
	year = {2024},
	pages = {15092--15108},
}

@misc{douze_faiss_2025,
	title = {The {Faiss} library},
	url = {http://arxiv.org/abs/2401.08281},
	doi = {10.48550/arXiv.2401.08281},
	urldate = {2025-03-19},
	publisher = {arXiv},
	author = {Douze, Matthijs and Guzhva, Alexandr and Deng, Chengqi and Johnson, Jeff and Szilvasy, Gergely and Mazaré, Pierre-Emmanuel and Lomeli, Maria and Hosseini, Lucas and Jégou, Hervé},
	month = feb,
	year = {2025},
	note = {arXiv:2401.08281 [cs]},
}

@inproceedings{liu-etal-2025-selected,
	address = {Abu Dhabi, UAE},
	title = {Selected languages are all you need for cross-lingual truthfulness transfer},
	url = {https://aclanthology.org/2025.coling-main.601/},
	booktitle = {Proceedings of the 31st international conference on computational linguistics},
	publisher = {Association for Computational Linguistics},
	author = {Liu, Weihao and Wu, Ning and Ding, Wenbiao and Liang, Shining and Gong, Ming and Zhang, Dongmei},
	editor = {Rambow, Owen and Wanner, Leo and Apidianaki, Marianna and Al-Khalifa, Hend and Eugenio, Barbara Di and Schockaert, Steven},
	month = jan,
	year = {2025},
	pages = {8963--8978},
}

@inproceedings{zhu-etal-2024-preference,
	address = {Mexico City, Mexico},
	title = {A preference-driven paradigm for enhanced translation with large language models},
	url = {https://aclanthology.org/2024.naacl-long.186/},
	doi = {10.18653/v1/2024.naacl-long.186},
	booktitle = {Proceedings of the 2024 conference of the north american chapter of the association for computational linguistics: {Human} language technologies (volume 1: {Long} papers)},
	publisher = {Association for Computational Linguistics},
	author = {Zhu, Dawei and Trenous, Sony and Shen, Xiaoyu and Klakow, Dietrich and Byrne, Bill and Hasler, Eva},
	editor = {Duh, Kevin and Gomez, Helena and Bethard, Steven},
	month = jun,
	year = {2024},
	pages = {3385--3403},
}

@inproceedings{xu_contrastive_2025,
	series = {{ICML}'24},
	title = {Contrastive preference optimization: pushing the boundaries of {LLM} performance in machine translation},
	booktitle = {Proceedings of the 41st international conference on machine learning},
	publisher = {JMLR.org},
	author = {Xu, Haoran and Sharaf, Amr and Chen, Yunmo and Tan, Weiting and Shen, Lingfeng and Van Durme, Benjamin and Murray, Kenton and Kim, Young Jin},
	year = {2025},
}

@inproceedings{kocmi-etal-2024-findings,
	address = {Miami, Florida, USA},
	title = {Findings of the {WMT24} general machine translation shared task: {The} {LLM} era is here but {MT} is not solved yet},
	url = {https://aclanthology.org/2024.wmt-1.1/},
	doi = {10.18653/v1/2024.wmt-1.1},
	booktitle = {Proceedings of the ninth conference on machine translation},
	publisher = {Association for Computational Linguistics},
	author = {Kocmi, Tom and Avramidis, Eleftherios and Bawden, Rachel and Bojar, Ondřej and Dvorkovich, Anton and Federmann, Christian and Fishel, Mark and Freitag, Markus and Gowda, Thamme and Grundkiewicz, Roman and Haddow, Barry and Karpinska, Marzena and Koehn, Philipp and Marie, Benjamin and Monz, Christof and Murray, Kenton and Nagata, Masaaki and Popel, Martin and Popović, Maja and Shmatova, Mariya and Steingrímsson, Steinthór and Zouhar, Vilém},
	editor = {Haddow, Barry and Kocmi, Tom and Koehn, Philipp and Monz, Christof},
	month = nov,
	year = {2024},
	pages = {1--46},
}

@inproceedings{joulin-etal-2017-bag,
	address = {Valencia, Spain},
	title = {Bag of tricks for efficient text classification},
	url = {https://aclanthology.org/E17-2068/},
	booktitle = {Proceedings of the 15th conference of the {European} chapter of the association for computational linguistics: {Volume} 2, short papers},
	publisher = {Association for Computational Linguistics},
	author = {Joulin, Armand and Grave, Edouard and Bojanowski, Piotr and Mikolov, Tomas},
	editor = {Lapata, Mirella and Blunsom, Phil and Koller, Alexander},
	month = apr,
	year = {2017},
	pages = {427--431},
}

@misc{joulin_fasttextzip_2016,
	title = {{FastText}.zip: {Compressing} text classification models},
	shorttitle = {{FastText}.zip},
	url = {http://arxiv.org/abs/1612.03651},
	doi = {10.48550/arXiv.1612.03651},
	urldate = {2025-01-15},
	publisher = {arXiv},
	author = {Joulin, Armand and Grave, Edouard and Bojanowski, Piotr and Douze, Matthijs and Jégou, Hérve and Mikolov, Tomas},
	month = dec,
	year = {2016},
	note = {arXiv:1612.03651 [cs]},
}

@inproceedings{loshchilov2018decoupled,
	title = {Decoupled weight decay regularization},
	url = {https://openreview.net/forum?id=Bkg6RiCqY7},
	booktitle = {International conference on learning representations},
	author = {Loshchilov, Ilya and Hutter, Frank},
	year = {2019},
}

@inproceedings{kocmi-etal-2021-ship,
	address = {Online},
	title = {To ship or not to ship: {An} extensive evaluation of automatic metrics for machine translation},
	url = {https://aclanthology.org/2021.wmt-1.57},
	booktitle = {Proceedings of the sixth conference on machine translation},
	publisher = {Association for Computational Linguistics},
	author = {Kocmi, Tom and Federmann, Christian and Grundkiewicz, Roman and Junczys-Dowmunt, Marcin and Matsushita, Hitokazu and Menezes, Arul},
	editor = {Barrault, Loic and Bojar, Ondřej and Bougares, Fethi and Chatterjee, Rajen and Costa-jussa, Marta R. and Federmann, Christian and Fishel, Mark and Fraser, Alexander and Freitag, Markus and Graham, Yvette and Grundkiewicz, Roman and Guzman, Paco and Haddow, Barry and Huck, Matthias and Yepes, Antonio Jimeno and Koehn, Philipp and Kocmi, Tom and Martins, Andre and Morishita, Makoto and Monz, Christof},
	month = nov,
	year = {2021},
	pages = {478--494},
}

@inproceedings{zaninello-birch-2020-multiword,
	address = {Marseille, France},
	title = {Multiword expression aware neural machine translation},
	isbn = {979-10-95546-34-4},
	url = {https://aclanthology.org/2020.lrec-1.471},
	language = {English},
	booktitle = {Proceedings of the twelfth language resources and evaluation conference},
	publisher = {European Language Resources Association},
	author = {Zaninello, Andrea and Birch, Alexandra},
	editor = {Calzolari, Nicoletta and Béchet, Frédéric and Blache, Philippe and Choukri, Khalid and Cieri, Christopher and Declerck, Thierry and Goggi, Sara and Isahara, Hitoshi and Maegaard, Bente and Mariani, Joseph and Mazo, Hélène and Moreno, Asuncion and Odijk, Jan and Piperidis, Stelios},
	month = may,
	year = {2020},
	pages = {3816--3825},
}

@inproceedings{freitag-etal-2022-results,
	address = {Abu Dhabi, United Arab Emirates (Hybrid)},
	title = {Results of {WMT22} metrics shared task: {Stop} using {BLEU} – neural metrics are better and more robust},
	url = {https://aclanthology.org/2022.wmt-1.2},
	booktitle = {Proceedings of the seventh conference on machine translation ({WMT})},
	publisher = {Association for Computational Linguistics},
	author = {Freitag, Markus and Rei, Ricardo and Mathur, Nitika and Lo, Chi-kiu and Stewart, Craig and Avramidis, Eleftherios and Kocmi, Tom and Foster, George and Lavie, Alon and Martins, André F. T.},
	editor = {Koehn, Philipp and Barrault, Loı̈c and Bojar, Ondřej and Bougares, Fethi and Chatterjee, Rajen and Costa-jussà, Marta R. and Federmann, Christian and Fishel, Mark and Fraser, Alexander and Freitag, Markus and Graham, Yvette and Grundkiewicz, Roman and Guzman, Paco and Haddow, Barry and Huck, Matthias and Jimeno Yepes, Antonio and Kocmi, Tom and Martins, André and Morishita, Makoto and Monz, Christof and Nagata, Masaaki and Nakazawa, Toshiaki and Negri, Matteo and Névéol, Aurélie and Neves, Mariana and Popel, Martin and Turchi, Marco and Zampieri, Marcos},
	month = dec,
	year = {2022},
	pages = {46--68},
}

@article{freitag-etal-2021-experts,
	title = {Experts, errors, and context: a large-scale study of human evaluation for machine translation},
	volume = {9},
	url = {https://aclanthology.org/2021.tacl-1.87},
	doi = {10.1162/tacl_a_00437},
	journal = {Transactions of the Association for Computational Linguistics},
	author = {Freitag, Markus and Foster, George and Grangier, David and Ratnakar, Viresh and Tan, Qijun and Macherey, Wolfgang},
	editor = {Roark, Brian and Nenkova, Ani},
	year = {2021},
	note = {Place: Cambridge, MA
Publisher: MIT Press},
	pages = {1460--1474},
}

@inproceedings{stap_multilingual_2023,
	address = {Singapore},
	title = {Multilingual k-{Nearest}-{Neighbor} {Machine} {Translation}-{Nearest}-{Neighbor} {Machine} {Translation}},
	url = {https://aclanthology.org/2023.emnlp-main.571},
	doi = {10.18653/v1/2023.emnlp-main.571},
	language = {en},
	urldate = {2024-06-26},
	booktitle = {Proceedings of the 2023 {Conference} on {Empirical} {Methods} in {Natural} {Language} {Processing}},
	publisher = {Association for Computational Linguistics},
	author = {Stap, David and Monz, Christof},
	year = {2023},
	pages = {9200--9208},
}

@inproceedings{stap_conditional_2020,
	title = {Conditional {Image} {Generation} and {Manipulation} for {User}-{Specified} {Content}},
	url = {http://arxiv.org/abs/2005.04909},
	urldate = {2024-06-26},
	booktitle = {Proceedings of the {IEEE}/{CVF} {Conference} on {Computer} {Vision} and {Pattern} {Recognition} ({CVPR}) {Workshops}},
	author = {Stap, David and Bleeker, Maurits and Ibrahimi, Sarah and ter Hoeve, Maartje},
	year = {2020},
}

@inproceedings{zhu-etal-2024-multilingual,
	address = {Mexico City, Mexico},
	title = {Multilingual machine translation with large language models: {Empirical} results and analysis},
	url = {https://aclanthology.org/2024.findings-naacl.176},
	booktitle = {Findings of the association for computational linguistics: {NAACL} 2024},
	publisher = {Association for Computational Linguistics},
	author = {Zhu, Wenhao and Liu, Hongyi and Dong, Qingxiu and Xu, Jingjing and Huang, Shujian and Kong, Lingpeng and Chen, Jiajun and Li, Lei},
	editor = {Duh, Kevin and Gomez, Helena and Bethard, Steven},
	month = jun,
	year = {2024},
	pages = {2765--2781},
}

@misc{nllb_team_no_2022,
	title = {No {Language} {Left} {Behind}: {Scaling} {Human}-{Centered} {Machine} {Translation}},
	shorttitle = {No {Language} {Left} {Behind}},
	url = {http://arxiv.org/abs/2207.04672},
	urldate = {2024-05-30},
	publisher = {arXiv},
	author = {NLLB Team and Costa-jussà, Marta R. and Cross, James and Çelebi, Onur and Elbayad, Maha and Heafield, Kenneth and Heffernan, Kevin and Kalbassi, Elahe and Lam, Janice and Licht, Daniel and Maillard, Jean and Sun, Anna and Wang, Skyler and Wenzek, Guillaume and Youngblood, Al and Akula, Bapi and Barrault, Loic and Gonzalez, Gabriel Mejia and Hansanti, Prangthip and Hoffman, John and Jarrett, Semarley and Sadagopan, Kaushik Ram and Rowe, Dirk and Spruit, Shannon and Tran, Chau and Andrews, Pierre and Ayan, Necip Fazil and Bhosale, Shruti and Edunov, Sergey and Fan, Angela and Gao, Cynthia and Goswami, Vedanuj and Guzmán, Francisco and Koehn, Philipp and Mourachko, Alexandre and Ropers, Christophe and Saleem, Safiyyah and Schwenk, Holger and Wang, Jeff},
	month = aug,
	year = {2022},
	note = {arXiv:2207.04672 [cs]},
}

@misc{gao_towards_2024,
	title = {Towards {Boosting} {Many}-to-{Many} {Multilingual} {Machine} {Translation} with {Large} {Language} {Models}},
	url = {http://arxiv.org/abs/2401.05861},
	urldate = {2024-05-29},
	publisher = {arXiv},
	author = {Gao, Pengzhi and He, Zhongjun and Wu, Hua and Wang, Haifeng},
	month = feb,
	year = {2024},
	note = {arXiv:2401.05861 [cs]},
}

@inproceedings{stap_viewing_2023,
	address = {Singapore},
	title = {Viewing {Knowledge} {Transfer} in {Multilingual} {Machine} {Translation} {Through} a {Representational} {Lens}},
	url = {https://aclanthology.org/2023.findings-emnlp.998},
	doi = {10.18653/v1/2023.findings-emnlp.998},
	language = {en},
	urldate = {2024-04-08},
	booktitle = {Findings of the {Association} for {Computational} {Linguistics}: {EMNLP} 2023},
	publisher = {Association for Computational Linguistics},
	author = {Stap, David and Niculae, Vlad and Monz, Christof},
	year = {2023},
	pages = {14973--14987},
}

@inproceedings{xezonaki-etal-2023-improving,
	address = {Macau SAR, China},
	title = {Improving domain robustness in neural machine translation with fused topic knowledge embeddings},
	url = {https://aclanthology.org/2023.mtsummit-research.18},
	booktitle = {Proceedings of machine translation summit {XIX}, vol. 1: {Research} track},
	publisher = {Asia-Pacific Association for Machine Translation},
	author = {Xezonaki, Danai and Khalil, Talaat and Stap, David and Denis, Brandon},
	editor = {Utiyama, Masao and Wang, Rui},
	month = sep,
	year = {2023},
	pages = {209--221},
}

@inproceedings{wu_uva-mts_2023,
	address = {Singapore},
	title = {{UvA}-{MT}’s {Participation} in the {WMT} 2023 {General} {Translation} {Shared} {Task}},
	url = {https://aclanthology.org/2023.wmt-1.17},
	doi = {10.18653/v1/2023.wmt-1.17},
	language = {en},
	urldate = {2024-04-08},
	booktitle = {Proceedings of the {Eighth} {Conference} on {Machine} {Translation}},
	publisher = {Association for Computational Linguistics},
	author = {Wu, Di and Tan, Shaomu and Stap, David and Araabi, Ali and Monz, Christof},
	year = {2023},
	pages = {175--180},
}

@inproceedings{zwennicker_towards_2022,
	address = {Abu Dhabi, United Arab Emirates (Hybrid)},
	title = {Towards a general purpose machine translation system for {Sranantongo}},
	url = {https://arxiv.org/abs/2212.06383},
	booktitle = {Proceedings of the {Sixth} {Widening} {NLP} {Workshop} ({WiNLP})},
	publisher = {Association for Computational Linguistics},
	author = {Zwennicker, Just and Stap, David},
	year = {2022},
}

@inproceedings{wang_super-naturalinstructions_2022,
	address = {Abu Dhabi, United Arab Emirates},
	title = {Super-{NaturalInstructions}: {Generalization} via {Declarative} {Instructions} on 1600+ {NLP} {Tasks}},
	shorttitle = {Super-{NaturalInstructions}},
	url = {https://aclanthology.org/2022.emnlp-main.340},
	doi = {10.18653/v1/2022.emnlp-main.340},
	language = {en},
	urldate = {2024-04-08},
	booktitle = {Proceedings of the 2022 {Conference} on {Empirical} {Methods} in {Natural} {Language} {Processing}},
	publisher = {Association for Computational Linguistics},
	author = {Wang, Yizhong and Mishra, Swaroop and Alipoormolabashi, Pegah and Kordi, Yeganeh and Mirzaei, Amirreza and Naik, Atharva and Ashok, Arjun and Dhanasekaran, Arut Selvan and Arunkumar, Anjana and Stap, David and Pathak, Eshaan and Karamanolakis, Giannis and Lai, Haizhi and Purohit, Ishan and Mondal, Ishani and Anderson, Jacob and Kuznia, Kirby and Doshi, Krima and Pal, Kuntal Kumar and Patel, Maitreya and Moradshahi, Mehrad and Parmar, Mihir and Purohit, Mirali and Varshney, Neeraj and Kaza, Phani Rohitha and Verma, Pulkit and Puri, Ravsehaj Singh and Karia, Rushang and Doshi, Savan and Sampat, Shailaja Keyur and Mishra, Siddhartha and Reddy A, Sujan and Patro, Sumanta and Dixit, Tanay and Shen, Xudong},
	year = {2022},
	pages = {5085--5109},
}

@misc{gemini_team_gemini_nodate,
	title = {Gemini 1.5: {Unlocking} multimodal understanding across millions of tokens of context},
	url = {https://storage.googleapis.com/deepmind-media/gemini/gemini_v1_5_report.pdf},
	author = {Gemini Team, Google},
}

@inproceedings{haagsma-etal-2020-magpie,
	address = {Marseille, France},
	title = {{MAGPIE}: {A} large corpus of potentially idiomatic expressions},
	isbn = {979-10-95546-34-4},
	url = {https://aclanthology.org/2020.lrec-1.35},
	language = {English},
	booktitle = {Proceedings of the twelfth language resources and evaluation conference},
	publisher = {European Language Resources Association},
	author = {Haagsma, Hessel and Bos, Johan and Nissim, Malvina},
	editor = {Calzolari, Nicoletta and Béchet, Frédéric and Blache, Philippe and Choukri, Khalid and Cieri, Christopher and Declerck, Thierry and Goggi, Sara and Isahara, Hitoshi and Maegaard, Bente and Mariani, Joseph and Mazo, Hélène and Moreno, Asuncion and Odijk, Jan and Piperidis, Stelios},
	month = may,
	year = {2020},
	pages = {279--287},
}

@misc{tang_petci_2022,
	title = {{PETCI}: {A} {Parallel} {English} {Translation} {Dataset} of {Chinese} {Idioms}},
	shorttitle = {{PETCI}},
	url = {http://arxiv.org/abs/2202.09509},
	urldate = {2024-02-06},
	publisher = {arXiv},
	author = {Tang, Kenan},
	month = feb,
	year = {2022},
	note = {arXiv:2202.09509 [cs]},
}

@inproceedings{fadaee-etal-2018-examining,
	address = {Miyazaki, Japan},
	title = {Examining the tip of the iceberg: {A} data set for idiom translation},
	url = {https://aclanthology.org/L18-1148},
	booktitle = {Proceedings of the eleventh international conference on language resources and evaluation ({LREC} 2018)},
	publisher = {European Language Resources Association (ELRA)},
	author = {Fadaee, Marzieh and Bisazza, Arianna and Monz, Christof},
	editor = {Calzolari, Nicoletta and Choukri, Khalid and Cieri, Christopher and Declerck, Thierry and Goggi, Sara and Hasida, Koiti and Isahara, Hitoshi and Maegaard, Bente and Mariani, Joseph and Mazo, Hélène and Moreno, Asuncion and Odijk, Jan and Piperidis, Stelios and Tokunaga, Takenobu},
	month = may,
	year = {2018},
}

@misc{moslem_fine-tuning_2023,
	title = {Fine-tuning {Large} {Language} {Models} for {Adaptive} {Machine} {Translation}},
	url = {http://arxiv.org/abs/2312.12740},
	urldate = {2024-02-06},
	publisher = {arXiv},
	author = {Moslem, Yasmin and Haque, Rejwanul and Way, Andy},
	month = dec,
	year = {2023},
	note = {arXiv:2312.12740 [cs]},
}

@inproceedings{rippeth_controlling_2022,
	address = {Dublin, Ireland (in-person and online)},
	title = {Controlling {Translation} {Formality} {Using} {Pre}-trained {Multilingual} {Language} {Models}},
	url = {https://aclanthology.org/2022.iwslt-1.30},
	doi = {10.18653/v1/2022.iwslt-1.30},
	language = {en},
	urldate = {2024-01-30},
	booktitle = {Proceedings of the 19th {International} {Conference} on {Spoken} {Language} {Translation} ({IWSLT} 2022)},
	publisher = {Association for Computational Linguistics},
	author = {Rippeth, Elijah and Agrawal, Sweta and Carpuat, Marine},
	year = {2022},
	pages = {327--340},
}

@article{riley_frmt_2023,
	title = {{FRMT}: {A} {Benchmark} for {Few}-{Shot} {Region}-{Aware} {Machine} {Translation}},
	volume = {11},
	issn = {2307-387X},
	shorttitle = {{FRMT}},
	url = {https://direct.mit.edu/tacl/article/doi/10.1162/tacl_a_00568/116617/FRMT-A-Benchmark-for-Few-Shot-Region-Aware-Machine},
	doi = {10.1162/tacl_a_00568},
	language = {en},
	urldate = {2024-01-30},
	journal = {Transactions of the Association for Computational Linguistics},
	author = {Riley, Parker and Dozat, Timothy and Botha, Jan A. and Garcia, Xavier and Garrette, Dan and Riesa, Jason and Firat, Orhan and Constant, Noah},
	month = jun,
	year = {2023},
	pages = {671--685},
}

@inproceedings{dankers_can_2022,
	address = {Dublin, Ireland},
	title = {Can {Transformer} be {Too} {Compositional}? {Analysing} {Idiom} {Processing} in {Neural} {Machine} {Translation}},
	shorttitle = {Can {Transformer} be {Too} {Compositional}?},
	url = {https://aclanthology.org/2022.acl-long.252},
	doi = {10.18653/v1/2022.acl-long.252},
	language = {en},
	urldate = {2024-01-30},
	booktitle = {Proceedings of the 60th {Annual} {Meeting} of the {Association} for {Computational} {Linguistics} ({Volume} 1: {Long} {Papers})},
	publisher = {Association for Computational Linguistics},
	author = {Dankers, Verna and Lucas, Christopher and Titov, Ivan},
	year = {2022},
	pages = {3608--3626},
}

@inproceedings{robinson_chatgpt_2023,
	address = {Singapore},
	title = {{ChatGPT} {MT}: {Competitive} for {High}- (but {Not} {Low}-) {Resource} {Languages}},
	shorttitle = {{ChatGPT} {MT}},
	url = {https://aclanthology.org/2023.wmt-1.40},
	doi = {10.18653/v1/2023.wmt-1.40},
	language = {en},
	urldate = {2024-01-30},
	booktitle = {Proceedings of the {Eighth} {Conference} on {Machine} {Translation}},
	publisher = {Association for Computational Linguistics},
	author = {Robinson, Nathaniel and Ogayo, Perez and Mortensen, David R. and Neubig, Graham},
	year = {2023},
	pages = {392--418},
}

@misc{wei_polylm_2023,
	title = {{PolyLM}: {An} {Open} {Source} {Polyglot} {Large} {Language} {Model}},
	shorttitle = {{PolyLM}},
	url = {http://arxiv.org/abs/2307.06018},
	urldate = {2024-01-30},
	publisher = {arXiv},
	author = {Wei, Xiangpeng and Wei, Haoran and Lin, Huan and Li, Tianhao and Zhang, Pei and Ren, Xingzhang and Li, Mei and Wan, Yu and Cao, Zhiwei and Xie, Binbin and Hu, Tianxiang and Li, Shangjie and Hui, Binyuan and Yu, Bowen and Liu, Dayiheng and Yang, Baosong and Huang, Fei and Xie, Jun},
	month = jul,
	year = {2023},
	note = {arXiv:2307.06018 [cs]},
}

@inproceedings{stap_chatgpt_2023,
	address = {Toronto, Canada},
	title = {{ChatGPT} is not a good indigenous translator},
	url = {https://aclanthology.org/2023.americasnlp-1.17},
	doi = {10.18653/v1/2023.americasnlp-1.17},
	language = {en},
	urldate = {2024-01-30},
	booktitle = {Proceedings of the {Workshop} on {Natural} {Language} {Processing} for {Indigenous} {Languages} of the {Americas} ({AmericasNLP})},
	publisher = {Association for Computational Linguistics},
	author = {Stap, David and Araabi, Ali},
	year = {2023},
	pages = {163--167},
}

@misc{openai_gpt-4_2023,
	title = {{GPT}-4 {Technical} {Report}},
	url = {http://arxiv.org/abs/2303.08774},
	urldate = {2024-01-30},
	publisher = {arXiv},
	author = {OpenAI and Achiam, Josh and Adler, Steven and Agarwal, Sandhini and Ahmad, Lama and Akkaya, Ilge and Aleman, Florencia Leoni and Almeida, Diogo and Altenschmidt, Janko and Altman, Sam and Anadkat, Shyamal and Avila, Red and Babuschkin, Igor and Balaji, Suchir and Balcom, Valerie and Baltescu, Paul and Bao, Haiming and Bavarian, Mo and Belgum, Jeff and Bello, Irwan and Berdine, Jake and Bernadett-Shapiro, Gabriel and Berner, Christopher and Bogdonoff, Lenny and Boiko, Oleg and Boyd, Madelaine and Brakman, Anna-Luisa and Brockman, Greg and Brooks, Tim and Brundage, Miles and Button, Kevin and Cai, Trevor and Campbell, Rosie and Cann, Andrew and Carey, Brittany and Carlson, Chelsea and Carmichael, Rory and Chan, Brooke and Chang, Che and Chantzis, Fotis and Chen, Derek and Chen, Sully and Chen, Ruby and Chen, Jason and Chen, Mark and Chess, Ben and Cho, Chester and Chu, Casey and Chung, Hyung Won and Cummings, Dave and Currier, Jeremiah and Dai, Yunxing and Decareaux, Cory and Degry, Thomas and Deutsch, Noah and Deville, Damien and Dhar, Arka and Dohan, David and Dowling, Steve and Dunning, Sheila and Ecoffet, Adrien and Eleti, Atty and Eloundou, Tyna and Farhi, David and Fedus, Liam and Felix, Niko and Fishman, Simón Posada and Forte, Juston and Fulford, Isabella and Gao, Leo and Georges, Elie and Gibson, Christian and Goel, Vik and Gogineni, Tarun and Goh, Gabriel and Gontijo-Lopes, Rapha and Gordon, Jonathan and Grafstein, Morgan and Gray, Scott and Greene, Ryan and Gross, Joshua and Gu, Shixiang Shane and Guo, Yufei and Hallacy, Chris and Han, Jesse and Harris, Jeff and He, Yuchen and Heaton, Mike and Heidecke, Johannes and Hesse, Chris and Hickey, Alan and Hickey, Wade and Hoeschele, Peter and Houghton, Brandon and Hsu, Kenny and Hu, Shengli and Hu, Xin and Huizinga, Joost and Jain, Shantanu and Jain, Shawn and Jang, Joanne and Jiang, Angela and Jiang, Roger and Jin, Haozhun and Jin, Denny and Jomoto, Shino and Jonn, Billie and Jun, Heewoo and Kaftan, Tomer and Kaiser, Łukasz and Kamali, Ali and Kanitscheider, Ingmar and Keskar, Nitish Shirish and Khan, Tabarak and Kilpatrick, Logan and Kim, Jong Wook and Kim, Christina and Kim, Yongjik and Kirchner, Hendrik and Kiros, Jamie and Knight, Matt and Kokotajlo, Daniel and Kondraciuk, Łukasz and Kondrich, Andrew and Konstantinidis, Aris and Kosic, Kyle and Krueger, Gretchen and Kuo, Vishal and Lampe, Michael and Lan, Ikai and Lee, Teddy and Leike, Jan and Leung, Jade and Levy, Daniel and Li, Chak Ming and Lim, Rachel and Lin, Molly and Lin, Stephanie and Litwin, Mateusz and Lopez, Theresa and Lowe, Ryan and Lue, Patricia and Makanju, Anna and Malfacini, Kim and Manning, Sam and Markov, Todor and Markovski, Yaniv and Martin, Bianca and Mayer, Katie and Mayne, Andrew and McGrew, Bob and McKinney, Scott Mayer and McLeavey, Christine and McMillan, Paul and McNeil, Jake and Medina, David and Mehta, Aalok and Menick, Jacob and Metz, Luke and Mishchenko, Andrey and Mishkin, Pamela and Monaco, Vinnie and Morikawa, Evan and Mossing, Daniel and Mu, Tong and Murati, Mira and Murk, Oleg and Mély, David and Nair, Ashvin and Nakano, Reiichiro and Nayak, Rajeev and Neelakantan, Arvind and Ngo, Richard and Noh, Hyeonwoo and Ouyang, Long and O'Keefe, Cullen and Pachocki, Jakub and Paino, Alex and Palermo, Joe and Pantuliano, Ashley and Parascandolo, Giambattista and Parish, Joel and Parparita, Emy and Passos, Alex and Pavlov, Mikhail and Peng, Andrew and Perelman, Adam and Peres, Filipe de Avila Belbute and Petrov, Michael and Pinto, Henrique Ponde de Oliveira and Michael and Pokorny and Pokrass, Michelle and Pong, Vitchyr and Powell, Tolly and Power, Alethea and Power, Boris and Proehl, Elizabeth and Puri, Raul and Radford, Alec and Rae, Jack and Ramesh, Aditya and Raymond, Cameron and Real, Francis and Rimbach, Kendra and Ross, Carl and Rotsted, Bob and Roussez, Henri and Ryder, Nick and Saltarelli, Mario and Sanders, Ted and Santurkar, Shibani and Sastry, Girish and Schmidt, Heather and Schnurr, David and Schulman, John and Selsam, Daniel and Sheppard, Kyla and Sherbakov, Toki and Shieh, Jessica and Shoker, Sarah and Shyam, Pranav and Sidor, Szymon and Sigler, Eric and Simens, Maddie and Sitkin, Jordan and Slama, Katarina and Sohl, Ian and Sokolowsky, Benjamin and Song, Yang and Staudacher, Natalie and Such, Felipe Petroski and Summers, Natalie and Sutskever, Ilya and Tang, Jie and Tezak, Nikolas and Thompson, Madeleine and Tillet, Phil and Tootoonchian, Amin and Tseng, Elizabeth and Tuggle, Preston and Turley, Nick and Tworek, Jerry and Uribe, Juan Felipe Cerón and Vallone, Andrea and Vijayvergiya, Arun and Voss, Chelsea and Wainwright, Carroll and Wang, Justin Jay and Wang, Alvin and Wang, Ben and Ward, Jonathan and Wei, Jason and Weinmann, C. J. and Welihinda, Akila and Welinder, Peter and Weng, Jiayi and Weng, Lilian and Wiethoff, Matt and Willner, Dave and Winter, Clemens and Wolrich, Samuel and Wong, Hannah and Workman, Lauren and Wu, Sherwin and Wu, Jeff and Wu, Michael and Xiao, Kai and Xu, Tao and Yoo, Sarah and Yu, Kevin and Yuan, Qiming and Zaremba, Wojciech and Zellers, Rowan and Zhang, Chong and Zhang, Marvin and Zhao, Shengjia and Zheng, Tianhao and Zhuang, Juntang and Zhuk, William and Zoph, Barret},
	month = dec,
	year = {2023},
	note = {arXiv:2303.08774 [cs]},
}

@inproceedings{rasley_deepspeed_2020,
	title = {Deepspeed: {System} optimizations enable training deep learning models with over 100 billion parameters},
	booktitle = {Proceedings of the 26th {ACM} {SIGKDD} {International} {Conference} on {Knowledge} {Discovery} \& {Data} {Mining}},
	author = {Rasley, Jeff and Rajbhandari, Samyam and Ruwase, Olatunji and He, Yuxiong},
	year = {2020},
	pages = {3505--3506},
}

@inproceedings{reimers_sentence-bert_2019,
	address = {Hong Kong, China},
	title = {Sentence-{BERT}: {Sentence} {Embeddings} using {Siamese} {BERT}-{Networks}},
	url = {https://aclanthology.org/D19-1410},
	doi = {10.18653/v1/D19-1410},
	booktitle = {Proceedings of the 2019 {Conference} on {Empirical} {Methods} in {Natural} {Language} {Processing} and the 9th {International} {Joint} {Conference} on {Natural} {Language} {Processing} ({EMNLP}-{IJCNLP})},
	publisher = {Association for Computational Linguistics},
	author = {Reimers, Nils and Gurevych, Iryna},
	editor = {Inui, Kentaro and Jiang, Jing and Ng, Vincent and Wan, Xiaojun},
	month = nov,
	year = {2019},
	pages = {3982--3992},
}

@article{johnson_billion-scale_2019,
	title = {Billion-scale similarity search with {GPUs}},
	volume = {7},
	number = {3},
	journal = {IEEE Transactions on Big Data},
	author = {Johnson, Jeff and Douze, Matthijs and Jégou, Hervé},
	year = {2019},
	note = {Publisher: IEEE},
	pages = {535--547},
}

@inproceedings{anastasopoulos_tico-19_2020,
	address = {Online},
	title = {{TICO}-19: the {Translation} {Initiative} for {COvid}-19},
	url = {https://aclanthology.org/2020.nlpcovid19-2.5},
	doi = {10.18653/v1/2020.nlpcovid19-2.5},
	booktitle = {Proceedings of the 1st {Workshop} on {NLP} for {COVID}-19 ({Part} 2) at {EMNLP} 2020},
	publisher = {Association for Computational Linguistics},
	author = {Anastasopoulos, Antonios and Cattelan, Alessandro and Dou, Zi-Yi and Federico, Marcello and Federmann, Christian and Genzel, Dmitriy and Guzmán, Franscisco and Hu, Junjie and Hughes, Macduff and Koehn, Philipp and Lazar, Rosie and Lewis, Will and Neubig, Graham and Niu, Mengmeng and Öktem, Alp and Paquin, Eric and Tang, Grace and Tur, Sylwia},
	editor = {Verspoor, Karin and Cohen, Kevin Bretonnel and Conway, Michael and de Bruijn, Berry and Dredze, Mark and Mihalcea, Rada and Wallace, Byron},
	month = dec,
	year = {2020},
}

@inproceedings{aharoni_unsupervised_2020,
	address = {Online},
	title = {Unsupervised {Domain} {Clusters} in {Pretrained} {Language} {Models}},
	url = {https://aclanthology.org/2020.acl-main.692},
	doi = {10.18653/v1/2020.acl-main.692},
	booktitle = {Proceedings of the 58th {Annual} {Meeting} of the {Association} for {Computational} {Linguistics}},
	publisher = {Association for Computational Linguistics},
	author = {Aharoni, Roee and Goldberg, Yoav},
	editor = {Jurafsky, Dan and Chai, Joyce and Schluter, Natalie and Tetreault, Joel},
	month = jul,
	year = {2020},
	pages = {7747--7763},
}

@inproceedings{nadejde_cocoa-mt_2022,
	address = {Seattle, United States},
	title = {{CoCoA}-{MT}: {A} {Dataset} and {Benchmark} for {Contrastive} {Controlled} {MT} with {Application} to {Formality}},
	url = {https://aclanthology.org/2022.findings-naacl.47},
	doi = {10.18653/v1/2022.findings-naacl.47},
	booktitle = {Findings of the {Association} for {Computational} {Linguistics}: {NAACL} 2022},
	publisher = {Association for Computational Linguistics},
	author = {Nadejde, Maria and Currey, Anna and Hsu, Benjamin and Niu, Xing and Federico, Marcello and Dinu, Georgiana},
	editor = {Carpuat, Marine and de Marneffe, Marie-Catherine and Meza Ruiz, Ivan Vladimir},
	month = jul,
	year = {2022},
	pages = {616--632},
}

@inproceedings{kocmi_findings_2022,
	address = {Abu Dhabi, United Arab Emirates (Hybrid)},
	title = {Findings of the 2022 {Conference} on {Machine} {Translation} ({WMT22})},
	url = {https://aclanthology.org/2022.wmt-1.1},
	booktitle = {Proceedings of the {Seventh} {Conference} on {Machine} {Translation} ({WMT})},
	publisher = {Association for Computational Linguistics},
	author = {Kocmi, Tom and Bawden, Rachel and Bojar, Ondřej and Dvorkovich, Anton and Federmann, Christian and Fishel, Mark and Gowda, Thamme and Graham, Yvette and Grundkiewicz, Roman and Haddow, Barry and Knowles, Rebecca and Koehn, Philipp and Monz, Christof and Morishita, Makoto and Nagata, Masaaki and Nakazawa, Toshiaki and Novák, Michal and Popel, Martin and Popović, Maja},
	editor = {Koehn, Philipp and Barrault, Loïc and Bojar, Ondřej and Bougares, Fethi and Chatterjee, Rajen and Costa-jussà, Marta R. and Federmann, Christian and Fishel, Mark and Fraser, Alexander and Freitag, Markus and Graham, Yvette and Grundkiewicz, Roman and Guzman, Paco and Haddow, Barry and Huck, Matthias and Jimeno Yepes, Antonio and Kocmi, Tom and Martins, André and Morishita, Makoto and Monz, Christof and Nagata, Masaaki and Nakazawa, Toshiaki and Negri, Matteo and Névéol, Aurélie and Neves, Mariana and Popel, Martin and Turchi, Marco and Zampieri, Marcos},
	month = dec,
	year = {2022},
	pages = {1--45},
}

@inproceedings{chi_infoxlm_2021,
	address = {Online},
	title = {{InfoXLM}: {An} {Information}-{Theoretic} {Framework} for {Cross}-{Lingual} {Language} {Model} {Pre}-{Training}},
	url = {https://aclanthology.org/2021.naacl-main.280},
	doi = {10.18653/v1/2021.naacl-main.280},
	booktitle = {Proceedings of the 2021 {Conference} of the {North} {American} {Chapter} of the {Association} for {Computational} {Linguistics}: {Human} {Language} {Technologies}},
	publisher = {Association for Computational Linguistics},
	author = {Chi, Zewen and Dong, Li and Wei, Furu and Yang, Nan and Singhal, Saksham and Wang, Wenhui and Song, Xia and Mao, Xian-Ling and Huang, Heyan and Zhou, Ming},
	editor = {Toutanova, Kristina and Rumshisky, Anna and Zettlemoyer, Luke and Hakkani-Tur, Dilek and Beltagy, Iz and Bethard, Steven and Cotterell, Ryan and Chakraborty, Tanmoy and Zhou, Yichao},
	month = jun,
	year = {2021},
	pages = {3576--3588},
}

@inproceedings{rei_cometkiwi_2022,
	address = {Abu Dhabi, United Arab Emirates (Hybrid)},
	title = {{CometKiwi}: {IST}-{Unbabel} 2022 {Submission} for the {Quality} {Estimation} {Shared} {Task}},
	url = {https://aclanthology.org/2022.wmt-1.60},
	booktitle = {Proceedings of the {Seventh} {Conference} on {Machine} {Translation} ({WMT})},
	publisher = {Association for Computational Linguistics},
	author = {Rei, Ricardo and Treviso, Marcos and Guerreiro, Nuno M. and Zerva, Chrysoula and Farinha, Ana C and Maroti, Christine and C. de Souza, José G. and Glushkova, Taisiya and Alves, Duarte and Coheur, Luisa and Lavie, Alon and Martins, André F. T.},
	editor = {Koehn, Philipp and Barrault, Loïc and Bojar, Ondřej and Bougares, Fethi and Chatterjee, Rajen and Costa-jussà, Marta R. and Federmann, Christian and Fishel, Mark and Fraser, Alexander and Freitag, Markus and Graham, Yvette and Grundkiewicz, Roman and Guzman, Paco and Haddow, Barry and Huck, Matthias and Jimeno Yepes, Antonio and Kocmi, Tom and Martins, André and Morishita, Makoto and Monz, Christof and Nagata, Masaaki and Nakazawa, Toshiaki and Negri, Matteo and Névéol, Aurélie and Neves, Mariana and Popel, Martin and Turchi, Marco and Zampieri, Marcos},
	month = dec,
	year = {2022},
	pages = {634--645},
}

@techreport{neubig_zeno_2023,
	title = {Zeno {GPT}-{MT} {Report}},
	url = {https://github.com/zeno-ml/zeno-build/tree/main/examples/analysis_gpt_mt/report},
	author = {Neubig, Graham},
	year = {2023},
}

@article{ratcliff_connectionist_1990,
	title = {Connectionist models of recognition memory: constraints imposed by learning and forgetting functions.},
	volume = {97},
	number = {2},
	journal = {Psychological review},
	author = {Ratcliff, Roger},
	year = {1990},
	note = {Publisher: American Psychological Association},
	pages = {285},
}

@incollection{mccloskey_catastrophic_1989,
	title = {Catastrophic interference in connectionist networks: {The} sequential learning problem},
	volume = {24},
	booktitle = {Psychology of learning and motivation},
	publisher = {Elsevier},
	author = {McCloskey, Michael and Cohen, Neal J},
	year = {1989},
	pages = {109--165},
}

@misc{almazrouei_falcon_2023,
	title = {The {Falcon} {Series} of {Open} {Language} {Models}},
	url = {http://arxiv.org/abs/2311.16867},
	urldate = {2023-12-18},
	publisher = {arXiv},
	author = {Almazrouei, Ebtesam and Alobeidli, Hamza and Alshamsi, Abdulaziz and Cappelli, Alessandro and Cojocaru, Ruxandra and Debbah, Mérouane and Goffinet, Étienne and Hesslow, Daniel and Launay, Julien and Malartic, Quentin and Mazzotta, Daniele and Noune, Badreddine and Pannier, Baptiste and Penedo, Guilherme},
	month = nov,
	year = {2023},
	note = {arXiv:2311.16867 [cs]},
}

@inproceedings{baziotis_automatic_2023,
	address = {Dubrovnik, Croatia},
	title = {Automatic {Evaluation} and {Analysis} of {Idioms} in {Neural} {Machine} {Translation}},
	url = {https://aclanthology.org/2023.eacl-main.267},
	doi = {10.18653/v1/2023.eacl-main.267},
	language = {en},
	urldate = {2023-11-12},
	booktitle = {Proceedings of the 17th {Conference} of the {European} {Chapter} of the {Association} for {Computational} {Linguistics}},
	publisher = {Association for Computational Linguistics},
	author = {Baziotis, Christos and Mathur, Prashant and Hasler, Eva},
	year = {2023},
	pages = {3682--3700},
}

@misc{luo_empirical_2023,
	title = {An {Empirical} {Study} of {Catastrophic} {Forgetting} in {Large} {Language} {Models} {During} {Continual} {Fine}-tuning},
	url = {http://arxiv.org/abs/2308.08747},
	doi = {10.48550/arXiv.2308.08747},
	urldate = {2023-08-22},
	author = {Luo, Yun and Yang, Zhen and Meng, Fandong and Li, Yafu and Zhou, Jie and Zhang, Yue},
	month = aug,
	year = {2023},
	note = {arXiv:2308.08747 [cs]},
}

@misc{anil_palm_2023,
	title = {{PaLM} 2 {Technical} {Report}},
	url = {http://arxiv.org/abs/2305.10403},
	urldate = {2023-08-10},
	publisher = {arXiv},
	author = {Anil, Rohan and Dai, Andrew M. and Firat, Orhan and Johnson, Melvin and Lepikhin, Dmitry and Passos, Alexandre and Shakeri, Siamak and Taropa, Emanuel and Bailey, Paige and Chen, Zhifeng and Chu, Eric and Clark, Jonathan H. and Shafey, Laurent El and Huang, Yanping and Meier-Hellstern, Kathy and Mishra, Gaurav and Moreira, Erica and Omernick, Mark and Robinson, Kevin and Ruder, Sebastian and Tay, Yi and Xiao, Kefan and Xu, Yuanzhong and Zhang, Yujing and Abrego, Gustavo Hernandez and Ahn, Junwhan and Austin, Jacob and Barham, Paul and Botha, Jan and Bradbury, James and Brahma, Siddhartha and Brooks, Kevin and Catasta, Michele and Cheng, Yong and Cherry, Colin and Choquette-Choo, Christopher A. and Chowdhery, Aakanksha and Crepy, Clément and Dave, Shachi and Dehghani, Mostafa and Dev, Sunipa and Devlin, Jacob and Díaz, Mark and Du, Nan and Dyer, Ethan and Feinberg, Vlad and Feng, Fangxiaoyu and Fienber, Vlad and Freitag, Markus and Garcia, Xavier and Gehrmann, Sebastian and Gonzalez, Lucas and Gur-Ari, Guy and Hand, Steven and Hashemi, Hadi and Hou, Le and Howland, Joshua and Hu, Andrea and Hui, Jeffrey and Hurwitz, Jeremy and Isard, Michael and Ittycheriah, Abe and Jagielski, Matthew and Jia, Wenhao and Kenealy, Kathleen and Krikun, Maxim and Kudugunta, Sneha and Lan, Chang and Lee, Katherine and Lee, Benjamin and Li, Eric and Li, Music and Li, Wei and Li, YaGuang and Li, Jian and Lim, Hyeontaek and Lin, Hanzhao and Liu, Zhongtao and Liu, Frederick and Maggioni, Marcello and Mahendru, Aroma and Maynez, Joshua and Misra, Vedant and Moussalem, Maysam and Nado, Zachary and Nham, John and Ni, Eric and Nystrom, Andrew and Parrish, Alicia and Pellat, Marie and Polacek, Martin and Polozov, Alex and Pope, Reiner and Qiao, Siyuan and Reif, Emily and Richter, Bryan and Riley, Parker and Ros, Alex Castro and Roy, Aurko and Saeta, Brennan and Samuel, Rajkumar and Shelby, Renee and Slone, Ambrose and Smilkov, Daniel and So, David R. and Sohn, Daniel and Tokumine, Simon and Valter, Dasha and Vasudevan, Vijay and Vodrahalli, Kiran and Wang, Xuezhi and Wang, Pidong and Wang, Zirui and Wang, Tao and Wieting, John and Wu, Yuhuai and Xu, Kelvin and Xu, Yunhan and Xue, Linting and Yin, Pengcheng and Yu, Jiahui and Zhang, Qiao and Zheng, Steven and Zheng, Ce and Zhou, Weikang and Zhou, Denny and Petrov, Slav and Wu, Yonghui},
	month = may,
	year = {2023},
	note = {arXiv:2305.10403 [cs]},
}

@misc{zhu_extrapolating_2023,
	title = {Extrapolating {Large} {Language} {Models} to {Non}-{English} by {Aligning} {Languages}},
	url = {http://arxiv.org/abs/2308.04948},
	urldate = {2023-08-10},
	publisher = {arXiv},
	author = {Zhu, Wenhao and Lv, Yunzhe and Dong, Qingxiu and Yuan, Fei and Xu, Jingjing and Huang, Shujian and Kong, Lingpeng and Chen, Jiajun and Li, Lei},
	month = aug,
	year = {2023},
	note = {arXiv:2308.04948 [cs]},
}

@inproceedings{brown_language_2020,
	title = {Language {Models} are {Few}-{Shot} {Learners}},
	volume = {33},
	url = {https://proceedings.neurips.cc/paper_files/paper/2020/file/1457c0d6bfcb4967418bfb8ac142f64a-Paper.pdf},
	booktitle = {Advances in {Neural} {Information} {Processing} {Systems}},
	publisher = {Curran Associates, Inc.},
	author = {Brown, Tom and Mann, Benjamin and Ryder, Nick and Subbiah, Melanie and Kaplan, Jared D and Dhariwal, Prafulla and Neelakantan, Arvind and Shyam, Pranav and Sastry, Girish and Askell, Amanda and Agarwal, Sandhini and Herbert-Voss, Ariel and Krueger, Gretchen and Henighan, Tom and Child, Rewon and Ramesh, Aditya and Ziegler, Daniel and Wu, Jeffrey and Winter, Clemens and Hesse, Chris and Chen, Mark and Sigler, Eric and Litwin, Mateusz and Gray, Scott and Chess, Benjamin and Clark, Jack and Berner, Christopher and McCandlish, Sam and Radford, Alec and Sutskever, Ilya and Amodei, Dario},
	editor = {Larochelle, H. and Ranzato, M. and Hadsell, R. and Balcan, M. F. and Lin, H.},
	year = {2020},
	pages = {1877--1901},
}

@inproceedings{vilar_prompting_2023,
	address = {Toronto, Canada},
	title = {Prompting {PaLM} for {Translation}: {Assessing} {Strategies} and {Performance}},
	url = {https://aclanthology.org/2023.acl-long.859},
	booktitle = {Proceedings of the 61st {Annual} {Meeting} of the {Association} for {Computational} {Linguistics} ({Volume} 1: {Long} {Papers})},
	publisher = {Association for Computational Linguistics},
	author = {Vilar, David and Freitag, Markus and Cherry, Colin and Luo, Jiaming and Ratnakar, Viresh and Foster, George},
	month = jul,
	year = {2023},
	pages = {15406--15427},
}

@inproceedings{raunak_gpts_2023,
	address = {Toronto, Canada},
	title = {Do {GPTs} {Produce} {Less} {Literal} {Translations}?},
	url = {https://aclanthology.org/2023.acl-short.90},
	urldate = {2023-07-11},
	booktitle = {Proceedings of the 61st {Annual} {Meeting} of the {Association} for {Computational} {Linguistics} ({Volume} 2: {Short} {Papers})},
	publisher = {Association for Computational Linguistics},
	author = {Raunak, Vikas and Menezes, Arul and Post, Matt and Awadalla, Hany Hassan},
	year = {2023},
	note = {arXiv:2305.16806 [cs]},
	pages = {1041--1050},
}

@misc{yang_bigtranslate_2023,
	title = {{BigTranslate}: {Augmenting} {Large} {Language} {Models} with {Multilingual} {Translation} {Capability} over 100 {Languages}},
	shorttitle = {{BigTranslate}},
	url = {http://arxiv.org/abs/2305.18098},
	urldate = {2023-07-11},
	publisher = {arXiv},
	author = {Yang, Wen and Li, Chong and Zhang, Jiajun and Zong, Chengqing},
	month = jul,
	year = {2023},
	note = {arXiv:2305.18098 [cs]},
}

@inproceedings{khandelwal_nearest_2021,
	title = {Nearest {Neighbor} {Machine} {Translation}},
	url = {https://openreview.net/forum?id=7wCBOfJ8hJM},
	booktitle = {International {Conference} on {Learning} {Representations}},
	author = {Khandelwal, Urvashi and Fan, Angela and Jurafsky, Dan and Zettlemoyer, Luke and Lewis, Mike},
	year = {2021},
}

@inproceedings{khandelwal_generalization_2020,
	title = {Generalization through {Memorization}: {Nearest} {Neighbor} {Language} {Models}},
	url = {https://openreview.net/forum?id=HklBjCEKvH},
	urldate = {2022-07-12},
	booktitle = {International {Conference} on {Learning} {Representations}},
	author = {Khandelwal, Urvashi and Levy, Omer and Jurafsky, Dan and Zettlemoyer, Luke and Lewis, Mike},
	year = {2020},
}

@inproceedings{drozdov-etal-2022-cant,
	address = {Abu Dhabi, United Arab Emirates},
	title = {You can't pick your neighbors, or can you? {When} and {How} to {Rely} on {Retrieval} in the {kNN}-{LM}},
	url = {https://aclanthology.org/2022.findings-emnlp.218},
	booktitle = {Findings of the association for computational linguistics: {EMNLP} 2022},
	publisher = {Association for Computational Linguistics},
	author = {Drozdov, Andrew and Wang, Shufan and Rahimi, Razieh and McCallum, Andrew and Zamani, Hamed and Iyyer, Mohit},
	month = dec,
	year = {2022},
	pages = {2997--3007},
}

@inproceedings{wang2022non,
	title = {Non-parametric online learning from human feedback for neural machine translation},
	volume = {36},
	booktitle = {Proceedings of the {AAAI} conference on artificial intelligence},
	author = {Wang, Dongqi and Wei, Haoran and Zhang, Zhirui and Huang, Shujian and Xie, Jun and Chen, Jiajun},
	year = {2022},
	note = {Number: 10},
	pages = {11431--11439},
}

@inproceedings{jin-etal-2022-plug,
	address = {Online only},
	title = {Plug and play knowledge distillation for {kNN}-{LM} with external logits},
	url = {https://aclanthology.org/2022.aacl-short.57},
	booktitle = {Proceedings of the 2nd conference of the asia-pacific chapter of the association for computational linguistics and the 12th international joint conference on natural language processing (volume 2: {Short} papers)},
	publisher = {Association for Computational Linguistics},
	author = {Jin, Xuyang and Ge, Tao and Wei, Furu},
	month = nov,
	year = {2022},
	pages = {463--469},
}

@inproceedings{jiang_learning_2021,
	address = {Online and Punta Cana, Dominican Republic},
	title = {Learning {Kernel}-{Smoothed} {Machine} {Translation} with {Retrieved} {Examples}},
	url = {https://aclanthology.org/2021.emnlp-main.579},
	doi = {10.18653/v1/2021.emnlp-main.579},
	language = {en},
	urldate = {2023-06-09},
	booktitle = {Proceedings of the 2021 {Conference} on {Empirical} {Methods} in {Natural} {Language} {Processing}},
	publisher = {Association for Computational Linguistics},
	author = {Jiang, Qingnan and Wang, Mingxuan and Cao, Jun and Cheng, Shanbo and Huang, Shujian and Li, Lei},
	year = {2021},
	pages = {7280--7290},
}

@article{guu_generating_2018,
	title = {Generating {Sentences} by {Editing} {Prototypes}},
	volume = {6},
	issn = {2307-387X},
	url = {https://direct.mit.edu/tacl/article/43446},
	doi = {10.1162/tacl_a_00030},
	language = {en},
	urldate = {2023-06-08},
	journal = {Transactions of the Association for Computational Linguistics},
	author = {Guu, Kelvin and Hashimoto, Tatsunori B. and Oren, Yonatan and Liang, Percy},
	month = dec,
	year = {2018},
	pages = {437--450},
}

@inproceedings{cai_neural_2021,
	address = {Online},
	title = {Neural {Machine} {Translation} with {Monolingual} {Translation} {Memory}},
	url = {https://aclanthology.org/2021.acl-long.567},
	doi = {10.18653/v1/2021.acl-long.567},
	language = {en},
	urldate = {2022-07-08},
	booktitle = {Proceedings of the 59th {Annual} {Meeting} of the {Association} for {Computational} {Linguistics} and the 11th {International} {Joint} {Conference} on {Natural} {Language} {Processing} ({Volume} 1: {Long} {Papers})},
	publisher = {Association for Computational Linguistics},
	author = {Cai, Deng and Wang, Yan and Li, Huayang and Lam, Wai and Liu, Lemao},
	year = {2021},
	pages = {7307--7318},
}

@inproceedings{jiang-etal-2022-towards,
	address = {Abu Dhabi, United Arab Emirates},
	title = {Towards robust k-{Nearest}-{Neighbor} machine translation},
	url = {https://aclanthology.org/2022.emnlp-main.367},
	booktitle = {Proceedings of the 2022 conference on empirical methods in natural language processing},
	publisher = {Association for Computational Linguistics},
	author = {Jiang, Hui and Lu, Ziyao and Meng, Fandong and Zhou, Chulun and Zhou, Jie and Huang, Degen and Su, Jinsong},
	month = dec,
	year = {2022},
	pages = {5468--5477},
}

@inproceedings{zheng_adaptive_2021,
	address = {Online},
	title = {Adaptive {Nearest} {Neighbor} {Machine} {Translation}},
	url = {https://aclanthology.org/2021.acl-short.47},
	doi = {10.18653/v1/2021.acl-short.47},
	language = {en},
	urldate = {2023-05-24},
	booktitle = {Proceedings of the 59th {Annual} {Meeting} of the {Association} for {Computational} {Linguistics} and the 11th {International} {Joint} {Conference} on {Natural} {Language} {Processing} ({Volume} 2: {Short} {Papers})},
	publisher = {Association for Computational Linguistics},
	author = {Zheng, Xin and Zhang, Zhirui and Guo, Junliang and Huang, Shujian and Chen, Boxing and Luo, Weihua and Chen, Jiajun},
	year = {2021},
	pages = {368--374},
}

@inproceedings{martins_chunk-based_2022,
	address = {Abu Dhabi, United Arab Emirates},
	title = {Chunk-based {Nearest} {Neighbor} {Machine} {Translation}},
	url = {https://aclanthology.org/2022.emnlp-main.284/},
	booktitle = {Proceedings of the 2022 {Conference} on {Empirical} {Methods} in {Natural} {Language} {Processing}},
	publisher = {Association for Computational Linguistics},
	author = {Martins, Pedro Henrique and Marinho, Zita and Martins, André F. T.},
	year = {2022},
	pages = {4228--4245},
}

@inproceedings{federmann_ntrex-128_2022,
	address = {Online},
	title = {{NTREX}-128 – {News} {Test} {References} for {MT} {Evaluation} of 128 {Languages}},
	booktitle = {Proceedings of the {First} {Workshop} on {Scaling} {Up} {Multilingual} {Evaluation}},
	publisher = {Association for Computational Linguistics},
	author = {Federmann, Christian and Kocmi, Tom and Xin, Ying},
	year = {2022},
	pages = {21--24},
}

@inproceedings{wolf_transformers_2020,
	address = {Online},
	title = {Transformers: {State}-of-the-{Art} {Natural} {Language} {Processing}},
	shorttitle = {Transformers},
	url = {https://www.aclweb.org/anthology/2020.emnlp-demos.6},
	doi = {10.18653/v1/2020.emnlp-demos.6},
	language = {en},
	urldate = {2023-05-10},
	booktitle = {Proceedings of the 2020 {Conference} on {Empirical} {Methods} in {Natural} {Language} {Processing}: {System} {Demonstrations}},
	publisher = {Association for Computational Linguistics},
	author = {Wolf, Thomas and Debut, Lysandre and Sanh, Victor and Chaumond, Julien and Delangue, Clement and Moi, Anthony and Cistac, Pierric and Rault, Tim and Louf, Remi and Funtowicz, Morgan and Davison, Joe and Shleifer, Sam and Von Platen, Patrick and Ma, Clara and Jernite, Yacine and Plu, Julien and Xu, Canwen and Le Scao, Teven and Gugger, Sylvain and Drame, Mariama and Lhoest, Quentin and Rush, Alexander},
	year = {2020},
	pages = {38--45},
}

@inproceedings{Vardhan2022,
	title = {Low resource retrieval augmented adaptive neural machine translation},
	url = {https://www.amazon.science/publications/low-resource-retrieval-augmented-adaptive-neural-machine-translation},
	booktitle = {{NeurIPS} 2022 workshop on trustworthy and socially responsible machine learning ({TSRML})},
	author = {Vardhan, Harsha and Beniwal, Anurag and Sadagopan, Narayanan and Shah, Swair},
	year = {2022},
}

@inproceedings{kudugunta_investigating_2019,
	address = {Hong Kong, China},
	title = {Investigating multilingual {NMT} representations at scale},
	url = {https://www.aclweb.org/anthology/D19-1167},
	doi = {10.18653/v1/D19-1167},
	language = {en},
	urldate = {2019-12-12},
	booktitle = {Proceedings of the 2019 {Conference} on {Empirical} {Methods} in {Natural} {Language} {Processing} and the 9th {International} {Joint} {Conference} on {Natural} {Language} {Processing}},
	publisher = {Association for Computational Linguistics},
	author = {Kudugunta, Sneha and Bapna, Ankur and Caswell, Isaac and Firat, Orhan},
	year = {2019},
	pages = {1565--1575},
}

@misc{li_better_2022,
	title = {Better {Datastore}, {Better} {Translation}: {Generating} {Datastores} from {Pre}-{Trained} {Models} for {Nearest} {Neural} {Machine} {Translation}},
	shorttitle = {Better {Datastore}, {Better} {Translation}},
	url = {http://arxiv.org/abs/2212.08822},
	urldate = {2023-02-14},
	publisher = {arXiv},
	author = {Li, Jiahuan and Cheng, Shanbo and Sun, Zewei and Wang, Mingxuan and Huang, Shujian},
	month = dec,
	year = {2022},
	note = {arXiv:2212.08822 [cs]},
}

@book{lewis_ethnologue_2009,
	title = {Ethnologue: {Languages} of the world},
	url = {https://www.ethnologue.com},
	publisher = {SIL international},
	author = {Lewis, M. Paul},
	year = {2009},
}

@article{pedregosa_scikit-learn_2011,
	title = {Scikit-learn: {Machine} {Learning} in {Python}},
	volume = {12},
	url = {http://jmlr.org/papers/v12/pedregosa11a.html},
	number = {85},
	journal = {Journal of Machine Learning Research},
	author = {Pedregosa, Fabian and Varoquaux, Gaël and Gramfort, Alexandre and Michel, Vincent and Thirion, Bertrand and Grisel, Olivier and Blondel, Mathieu and Prettenhofer, Peter and Weiss, Ron and Dubourg, Vincent and Vanderplas, Jake and Passos, Alexandre and Cournapeau, David and Brucher, Matthieu and Perrot, Matthieu and Duchesnay, Edouard},
	year = {2011},
	pages = {2825--2830},
}

@inproceedings{timkey_all_2021,
	address = {Online and Punta Cana, Dominican Republic},
	title = {All {Bark} and {No} {Bite}: {Rogue} {Dimensions} in {Transformer} {Language} {Models} {Obscure} {Representational} {Quality}},
	shorttitle = {All {Bark} and {No} {Bite}},
	url = {https://aclanthology.org/2021.emnlp-main.372},
	doi = {10.18653/v1/2021.emnlp-main.372},
	language = {en},
	urldate = {2022-12-22},
	booktitle = {Proceedings of the 2021 {Conference} on {Empirical} {Methods} in {Natural} {Language} {Processing}},
	publisher = {Association for Computational Linguistics},
	author = {Timkey, William and van Schijndel, Marten},
	year = {2021},
	pages = {4527--4546},
}

@inproceedings{zhang_improving_2020,
	address = {Seattle, Washington},
	title = {Improving massively multilingual neural machine translation and zero-shot translation},
	url = {https://aclanthology.org/2020.acl-main.148},
	doi = {10.18653/v1/2020.acl-main.148},
	booktitle = {Proceedings of the 58th {Annual} {Meeting} of the {Association} for {Computational} {Linguistics}},
	author = {Zhang, Biao and Williams, Philip and Titov, Ivan and Sennrich, Rico},
	year = {2020},
}

@inproceedings{xu_eag_2022,
	address = {Dublin, Ireland},
	title = {{EAG}: {Extract} and {Generate} {Multi}-way {Aligned} {Corpus} for {Complete} {Multi}-lingual {Neural} {Machine} {Translation}},
	url = {https://aclanthology.org/2022.acl-long.560},
	doi = {10.18653/v1/2022.acl-long.560},
	booktitle = {Proceedings of the 60th {Annual} {Meeting} of the {Association} for {Computational} {Linguistics} and the 11th {International} {Joint} {Conference} on {Natural} {Language} {Processing}},
	publisher = {Association for Computational Linguistics},
	author = {Xu, Yulin and Yang, Zhen and Meng, Fandong and Jie, Zhou},
	year = {2022},
}

@inproceedings{szegedy_rethinking_2016,
	address = {Las Vegas, Nevada},
	title = {Rethinking the inception architecture for computer vision},
	url = {https://ieeexplore.ieee.org/document/7780677},
	doi = {10.1109/CVPR.2016.308},
	booktitle = {{IEEE} {Conference} on {Computer} {Vision} and {Pattern} {Recognition} ({CVPR})},
	publisher = {IEEE},
	author = {Szegedy, Christian and Vanhoucke, Vincent and Ioffe, Sergey and Shlens, Jon},
	year = {2016},
	pages = {2818--2826},
}

@article{srivastava_dropout_2014,
	title = {Dropout: a simple way to prevent neural networks from overfitting},
	volume = {15},
	url = {https://jmlr.org/papers/v15/srivastava14a.html},
	number = {1},
	journal = {The Journal of Machine Learning Research},
	author = {Srivastava, Nitish and Hinton, Geoffrey and Krizhevsky, Alex and Sutskever, Ilya and Salakhutdinov, Ruslan},
	year = {2014},
	pages = {1929--1958},
}

@inproceedings{press_using_2017,
	address = {Valencia, Spain},
	title = {Using the output embedding to improve language models},
	url = {https://aclanthology.org/E17-2025},
	booktitle = {Proceedings of the 15th {Conference} of the {European} {Chapter} of the {Association} for {Computational} {Linguistics}},
	publisher = {Association for Computational Linguistics},
	author = {Press, Ofir and Wolf, Lior},
	year = {2017},
	pages = {157--163},
}

@inproceedings{patil_overlap-based_2022,
	address = {Dublin, Ireland},
	title = {Overlap-based {Vocabulary} {Generation} {Improves} {Cross}-lingual {Transfer} {Among} {Related} {Languages}},
	url = {https://aclanthology.org/2022.acl-long.18},
	doi = {10.18653/v1/2022.acl-long.18},
	booktitle = {Proceedings of the 60th {Annual} {Meeting} of the {Association} for {Computational} {Linguistics} and the 11th {International} {Joint} {Conference} on {Natural} {Language} {Processing}},
	publisher = {Association for Computational Linguistics},
	author = {Patil, Vaidehi and Talukdar, Partha and Sarawagi, Sunita},
	year = {2022},
}

@inproceedings{kingma_adam_2015,
	address = {San Diego, California},
	title = {Adam: a method for stochastic optimization},
	url = {https://arxiv.org/abs/1412.6980},
	booktitle = {International conference on learning representations},
	author = {Kingma, Diederik P. and Ba, Jimmy L.},
	year = {2015},
}

@inproceedings{ha_toward_2016,
	address = {Seattle, Washington},
	title = {Toward multilingual neural machine translation with universal encoder and decoder},
	url = {https://aclanthology.org/2016.iwslt-1.6},
	booktitle = {Proceedings of the {International} {Workshop} on {Spoken} {Language} {Translation} ({IWSLT})},
	author = {Ha, Thanh-Le and Niehues, Jan and Waibel, Alexander},
	year = {2016},
}

@article{fisher_all_2019,
	title = {All {Models} are {Wrong}, but {Many} are {Useful}: {Learning} a {Variable}'s {Importance} by {Studying} an {Entire} {Class} of {Prediction} {Models} {Simultaneously}},
	volume = {20},
	url = {http://jmlr.org/papers/v20/18-760.html},
	number = {177},
	journal = {Journal of Machine Learning Research},
	author = {Fisher, Aaron and Rudin, Cynthia and Dominici, Francesca},
	year = {2019},
	pages = {1--81},
}

@article{breiman_random_2001,
	title = {Random {Forests}},
	volume = {45},
	url = {https://doi.org/10.1023/A:1010933404324},
	doi = {https://doi.org/10.1023/A:1010933404324},
	journal = {Springer},
	author = {Breiman, Leo},
	year = {2001},
	pages = {5--32},
}

@inproceedings{sun_alternative_2022,
	address = {Dublin, Ireland},
	title = {Alternative {Input} {Signals} {Ease} {Transfer} in {Multilingual} {Machine} {Translation}},
	url = {https://aclanthology.org/2022.acl-long.363},
	doi = {10.18653/v1/2022.acl-long.363},
	language = {en},
	urldate = {2022-12-19},
	booktitle = {Proceedings of the 60th {Annual} {Meeting} of the {Association} for {Computational} {Linguistics} ({Volume} 1: {Long} {Papers})},
	publisher = {Association for Computational Linguistics},
	author = {Sun, Simeng and Fan, Angela and Cross, James and Chaudhary, Vishrav and Tran, Chau and Koehn, Philipp and Guzmán, Francisco},
	year = {2022},
	pages = {5291--5305},
}

@inproceedings{liao_back-translation_2021,
	address = {Online},
	title = {Back-translation for {Large}-{Scale} {Multilingual} {Machine} {Translation}},
	url = {https://aclanthology.org/2021.wmt-1.50},
	booktitle = {Proceedings of the {Sixth} {Conference} on {Machine} {Translation}},
	publisher = {Association for Computational Linguistics},
	author = {Liao, Baohao and Khadivi, Shahram and Hewavitharana, Sanjika},
	year = {2021},
	pages = {418--424},
}

@inproceedings{kim_multilingual_2021,
	address = {Online},
	title = {Do {Multilingual} {Neural} {Machine} {Translation} {Models} {Contain} {Language} {Pair} {Specific} {Attention} {Heads}?},
	url = {https://aclanthology.org/2021.findings-acl.250},
	doi = {10.18653/v1/2021.findings-acl.250},
	language = {en},
	urldate = {2022-11-15},
	booktitle = {Findings of the {Association} for {Computational} {Linguistics}: {ACL}-{IJCNLP} 2021},
	publisher = {Association for Computational Linguistics},
	author = {Kim, Zae Myung and Besacier, Laurent and Nikoulina, Vassilina and Schwab, Didier},
	year = {2021},
	pages = {2832--2841},
}

@article{fan_beyond_2021,
	title = {Beyond {English}-centric multilingual machine translation},
	volume = {22},
	url = {http://jmlr.org/papers/v22/20-1307.html},
	number = {107},
	journal = {Journal of Machine Learning Research},
	author = {Fan, Angela and Bhosale, Shruti and Schwenk, Holger and Ma, Zhiyi and El-Kishky, Ahmed and Goyal, Siddharth and Baines, Mandeep and Celebi, Onur and Wenzek, Guillaume and Chaudhary, Vishrav and Goyal, Naman and Birch, Tom and Liptchinsky, Vitaliy and Edunov, Sergey and Grave, Edouard and Auli, Michael and Joulin, Armand},
	year = {2021},
	pages = {1--48},
}

@inproceedings{meng_fast_2022,
	address = {Dublin, Ireland},
	title = {Fast {Nearest} {Neighbor} {Machine} {Translation}},
	url = {https://aclanthology.org/2022.findings-acl.47},
	doi = {10.18653/v1/2022.findings-acl.47},
	language = {en},
	urldate = {2022-07-11},
	booktitle = {Findings of the {Association} for {Computational} {Linguistics}: {ACL} 2022},
	publisher = {Association for Computational Linguistics},
	author = {Meng, Yuxian and Li, Xiaoya and Zheng, Xiayu and Wu, Fei and Sun, Xiaofei and Zhang, Tianwei and Li, Jiwei},
	year = {2022},
	pages = {555--565},
}

@inproceedings{martins_efficient_2022,
	address = {Dublin, Ireland and Online},
	title = {Efficient {Machine} {Translation} {Domain} {Adaptation}},
	url = {https://aclanthology.org/2022.spanlp-1.3},
	doi = {10.18653/v1/2022.spanlp-1.3},
	booktitle = {Proceedings of the 1st {Workshop} on {Semiparametric} {Methods} in {NLP}: {Decoupling} {Logic} from {Knowledge}},
	publisher = {Association for Computational Linguistics},
	author = {Martins, Pedro Henrique and Marinho, Zita and Martins, André F. T.},
	year = {2022},
	pages = {23--29},
}

@inproceedings{he_efficient_2021,
	address = {Online and Punta Cana, Dominican Republic},
	title = {Efficient {Nearest} {Neighbor} {Language} {Models}},
	url = {https://aclanthology.org/2021.emnlp-main.461},
	doi = {10.18653/v1/2021.emnlp-main.461},
	language = {en},
	urldate = {2022-07-12},
	booktitle = {Proceedings of the 2021 {Conference} on {Empirical} {Methods} in {Natural} {Language} {Processing}},
	publisher = {Association for Computational Linguistics},
	author = {He, Junxian and Neubig, Graham and Berg-Kirkpatrick, Taylor},
	year = {2021},
	pages = {5703--5714},
}

@article{ba_layer_2016,
	title = {Layer normalization},
	url = {http://arxiv.org/abs/1607.06450},
	urldate = {2019-12-31},
	journal = {arXiv:1607.06450 [cs, stat]},
	author = {Ba, Jimmy Lei and Kiros, Jamie Ryan and Hinton, Geoffrey E.},
	month = jul,
	year = {2016},
	note = {arXiv: 1607.06450},
}

@article{escolano_multilingual_2022,
	title = {Multilingual {Machine} {Translation}: {Deep} {Analysis} of {Language}-{Specific} {Encoder}-{Decoders}},
	volume = {73},
	issn = {1076-9757},
	shorttitle = {Multilingual {Machine} {Translation}},
	url = {http://jair.org/index.php/jair/article/view/12699},
	doi = {10.1613/jair.1.12699},
	urldate = {2022-05-11},
	journal = {Journal of Artificial Intelligence Research},
	author = {Escolano, Carlos and Ruiz Costa-jussà, Marta and R. Fonollosa, José A.},
	month = apr,
	year = {2022},
	pages = {1535--1552},
}

@misc{dryer_wals_2013,
	title = {{WALS} {Online}},
	url = {https://wals.info/},
	publisher = {Max Planck Institute for Evolutionary Anthropology},
	author = {Dryer, Matthew S. and Haspelmath, Martin},
	year = {2013},
}

@misc{moran_cldf-datasetsphoible_2019,
	title = {cldf-datasets/phoible: {PHOIBLE} 2.0 as {CLDF} dataset},
	copyright = {Creative Commons Attribution 3.0 Unported, Open Access},
	shorttitle = {cldf-datasets/phoible},
	url = {https://zenodo.org/record/2593234},
	doi = {10.5281/ZENODO.2593234},
	urldate = {2021-12-13},
	publisher = {Zenodo},
	author = {Moran, Steven and McCloy, Daniel},
	month = mar,
	year = {2019},
}

@misc{hammarstrom_glottologglottolog_2021,
	title = {glottolog/glottolog: {Glottolog} database 4.5},
	copyright = {Creative Commons Attribution 4.0 International, Open Access},
	shorttitle = {glottolog/glottolog},
	url = {https://zenodo.org/record/5772642},
	doi = {10.5281/ZENODO.5772642},
	urldate = {2021-12-13},
	publisher = {Zenodo},
	author = {Hammarström, Harald and Forkel, Robert and Haspelmath, Martin and Bank, Sebastian},
	month = dec,
	year = {2021},
}

@inproceedings{lin_choosing_2019,
	address = {Florence, Italy},
	title = {Choosing {Transfer} {Languages} for {Cross}-{Lingual} {Learning}},
	url = {https://www.aclweb.org/anthology/P19-1301},
	doi = {10.18653/v1/P19-1301},
	language = {en},
	urldate = {2021-10-18},
	booktitle = {Proceedings of the 57th {Annual} {Meeting} of the {Association} for {Computational} {Linguistics}},
	publisher = {Association for Computational Linguistics},
	author = {Lin, Yu-Hsiang and Chen, Chian-Yu and Lee, Jean and Li, Zirui and Zhang, Yuyan and Xia, Mengzhou and Rijhwani, Shruti and He, Junxian and Zhang, Zhisong and Ma, Xuezhe and Anastasopoulos, Antonios and Littell, Patrick and Neubig, Graham},
	year = {2019},
	pages = {3125--3135},
}

@inproceedings{rei_comet_2020,
	address = {Online},
	title = {{COMET}: {A} {Neural} {Framework} for {MT} {Evaluation}},
	shorttitle = {{COMET}},
	url = {https://www.aclweb.org/anthology/2020.emnlp-main.213},
	doi = {10.18653/v1/2020.emnlp-main.213},
	language = {en},
	urldate = {2021-08-18},
	booktitle = {Proceedings of the 2020 {Conference} on {Empirical} {Methods} in {Natural} {Language} {Processing} ({EMNLP})},
	publisher = {Association for Computational Linguistics},
	author = {Rei, Ricardo and Stewart, Craig and Farinha, Ana C and Lavie, Alon},
	year = {2020},
	pages = {2685--2702},
}

@inproceedings{papineni_bleu_2002,
	address = {Philadelphia, Pennsylvania},
	title = {{BLEU}: a method for automatic evaluation of machine translation},
	shorttitle = {{BLEU}},
	url = {http://portal.acm.org/citation.cfm?doid=1073083.1073135},
	doi = {10.3115/1073083.1073135},
	language = {en},
	urldate = {2019-12-13},
	booktitle = {Proceedings of the 40th {Annual} {Meeting} of the {Association} for {Computational} {Linguistics}},
	publisher = {Association for Computational Linguistics},
	author = {Papineni, Kishore and Roukos, Salim and Ward, Todd and Zhu, Wei-Jing},
	year = {2002},
	pages = {311},
}

@inproceedings{wu_beto_2019,
	address = {Hong Kong, China},
	title = {Beto, bentz, becas: the surprising cross-lingual effectiveness of {BERT}},
	shorttitle = {Beto, {Bentz}, {Becas}},
	url = {https://www.aclweb.org/anthology/D19-1077},
	doi = {10.18653/v1/D19-1077},
	language = {en},
	urldate = {2021-01-28},
	booktitle = {Proceedings of the 2019 {Conference} on {Empirical} {Methods} in {Natural} {Language} {Processing} and the 9th {International} {Joint} {Conference} on {Natural} {Language} {Processing} ({EMNLP}-{IJCNLP})},
	publisher = {Association for Computational Linguistics},
	author = {Wu, Shijie and Dredze, Mark},
	year = {2019},
	pages = {833--844},
}

@inproceedings{sennrich_neural_2016,
	address = {Berlin, Germany},
	title = {Neural machine translation of rare words with subword units},
	url = {http://aclweb.org/anthology/P16-1162},
	doi = {10.18653/v1/P16-1162},
	language = {en},
	urldate = {2019-12-13},
	booktitle = {Proceedings of the 54th {Annual} {Meeting} of the {Association} for {Computational} {Linguistics}},
	publisher = {Association for Computational Linguistics},
	author = {Sennrich, Rico and Haddow, Barry and Birch, Alexandra},
	year = {2016},
	pages = {1715--1725},
}

@article{artetxe_massively_2019,
	title = {Massively multilingual sentence embeddings for zero-shot cross-lingual transfer and beyond},
	volume = {7},
	issn = {2307-387X},
	url = {https://www.mitpressjournals.org/doi/abs/10.1162/tacl_a_00288},
	doi = {10.1162/tacl_a_00288},
	language = {en},
	urldate = {2020-12-01},
	journal = {Transactions of the Association for Computational Linguistics},
	author = {Artetxe, Mikel and Schwenk, Holger},
	month = nov,
	year = {2019},
	pages = {597--610},
}

@inproceedings{conneau_unsupervised_2020,
	address = {Online},
	title = {Unsupervised cross-lingual representation learning at scale},
	url = {https://www.aclweb.org/anthology/2020.acl-main.747},
	doi = {10.18653/v1/2020.acl-main.747},
	language = {en},
	urldate = {2020-11-23},
	booktitle = {Proceedings of the 58th {Annual} {Meeting} of the {Association} for {Computational} {Linguistics}},
	publisher = {Association for Computational Linguistics},
	author = {Conneau, Alexis and Khandelwal, Kartikay and Goyal, Naman and Chaudhary, Vishrav and Wenzek, Guillaume and Guzmán, Francisco and Grave, Edouard and Ott, Myle and Zettlemoyer, Luke and Stoyanov, Veselin},
	year = {2020},
	pages = {8440--8451},
}

@inproceedings{koehn_six_2017,
	address = {Vancouver, Canada},
	title = {Six challenges for neural machine translation},
	url = {http://aclweb.org/anthology/W17-3204},
	doi = {10.18653/v1/W17-3204},
	language = {en},
	urldate = {2019-12-09},
	booktitle = {Proceedings of the {First} {Workshop} on {Neural} {Machine} {Translation}},
	publisher = {Association for Computational Linguistics},
	author = {Koehn, Philipp and Knowles, Rebecca},
	year = {2017},
	pages = {28--39},
}

@inproceedings{freitag_complete_2020,
	address = {Online},
	title = {Complete multilingual neural machine translation},
	url = {https://www.aclweb.org/anthology/2020.wmt-1.66},
	booktitle = {Proceedings of the {Fifth} {Conference} on {Machine} {Translation}},
	publisher = {Association for Computational Linguistics},
	author = {Freitag, Markus and Firat, Orhan},
	year = {2020},
	pages = {548--558},
}

@inproceedings{post_call_2018,
	address = {Belgium, Brussels},
	title = {A call for clarity in reporting {BLEU} scores},
	url = {http://aclweb.org/anthology/W18-6319},
	doi = {10.18653/v1/W18-6319},
	language = {en},
	urldate = {2020-02-12},
	booktitle = {Proceedings of the {Third} {Conference} on {Machine} {Translation}: {Research} {Papers}},
	publisher = {Association for Computational Linguistics},
	author = {Post, Matt},
	year = {2018},
	pages = {186--191},
}

@inproceedings{pires_how_2019,
	address = {Florence, Italy},
	title = {How multilingual is multilingual {BERT}?},
	url = {https://www.aclweb.org/anthology/P19-1493},
	doi = {10.18653/v1/P19-1493},
	language = {en},
	urldate = {2020-11-13},
	booktitle = {Proceedings of the 57th {Annual} {Meeting} of the {Association} for {Computational} {Linguistics}},
	publisher = {Association for Computational Linguistics},
	author = {Pires, Telmo and Schlinger, Eva and Garrette, Dan},
	year = {2019},
	pages = {4996--5001},
}

@inproceedings{zoph_transfer_2016,
	address = {Austin, Texas},
	title = {Transfer learning for low-resource neural machine translation},
	url = {http://aclweb.org/anthology/D16-1163},
	doi = {10.18653/v1/D16-1163},
	language = {en},
	urldate = {2019-12-09},
	booktitle = {Proceedings of the 2016 {Conference} on {Empirical} {Methods} in {Natural}           {Language} {Processing}},
	publisher = {Association for Computational Linguistics},
	author = {Zoph, Barret and Yuret, Deniz and May, Jonathan and Knight, Kevin},
	year = {2016},
	pages = {1568--1575},
}

@inproceedings{artetxe_cross-lingual_2020,
	address = {Online},
	title = {On the cross-lingual transferability of monolingual representations},
	url = {https://www.aclweb.org/anthology/2020.acl-main.421},
	doi = {10.18653/v1/2020.acl-main.421},
	language = {en},
	urldate = {2020-11-17},
	booktitle = {Proceedings of the 58th {Annual} {Meeting} of the {Association} for {Computational} {Linguistics}},
	publisher = {Association for Computational Linguistics},
	author = {Artetxe, Mikel and Ruder, Sebastian and Yogatama, Dani},
	year = {2020},
	pages = {4623--4637},
}

@inproceedings{philip_monolingual_2020,
	address = {Online},
	title = {Monolingual adapters for zero-shot neural machine translation},
	url = {https://www.aclweb.org/anthology/2020.emnlp-main.361},
	booktitle = {Proceedings of the 2020 {Conference} on {Empirical} {Methods} in {Natural} {Language} {Processing} ({EMNLP})},
	publisher = {Association for Computational Linguistics},
	author = {Philip, Jerin and Berard, Alexandre and Gallé, Matthias and Besacier, Laurent},
	year = {2020},
	pages = {4465--4470},
}

@inproceedings{chung_improving_2020,
	address = {Online},
	title = {Improving multilingual models with language-clustered vocabularies},
	url = {https://www.aclweb.org/anthology/2020.emnlp-main.367},
	booktitle = {Proceedings of the 2020 {Conference} on {Empirical} {Methods} in {Natural} {Language} {Processing} ({EMNLP})},
	publisher = {Association for Computational Linguistics},
	author = {Chung, Hyung Won and Garrette, Dan and Tan, Kiat Chuan and Riesa, Jason},
	year = {2020},
	pages = {4536--4546},
}

@inproceedings{lauscher_zero_2020,
	address = {Online},
	title = {From zero to hero: on the limitations of zero-shot language transfer with multilingual {Transformers}},
	url = {https://www.aclweb.org/anthology/2020.emnlp-main.363},
	booktitle = {Proceedings of the 2020 {Conference} on {Empirical} {Methods} in {Natural} {Language} {Processing} ({EMNLP})},
	publisher = {Association for Computational Linguistics},
	author = {Lauscher, Anne and Ravishankar, Vinit and Vulić, Ivan and Glavaš, Goran},
	year = {2020},
	pages = {4483--4499},
}

@inproceedings{conneau_emerging_2020,
	address = {Online},
	title = {Emerging cross-lingual structure in pretrained language models},
	url = {https://www.aclweb.org/anthology/2020.acl-main.536},
	doi = {10.18653/v1/2020.acl-main.536},
	language = {en},
	urldate = {2020-11-17},
	booktitle = {Proceedings of the 58th {Annual} {Meeting} of the {Association} for {Computational} {Linguistics}},
	publisher = {Association for Computational Linguistics},
	author = {Conneau, Alexis and Wu, Shijie and Li, Haoran and Zettlemoyer, Luke and Stoyanov, Veselin},
	year = {2020},
	pages = {6022--6034},
}

@inproceedings{k_cross-lingual_2020,
	address = {Online},
	title = {Cross-lingual ability of multilingual {BERT}: an empirical study},
	url = {https://openreview.net/pdf?id=HJeT3yrtDr},
	booktitle = {International {Conference} on {Learning} {Representations}},
	author = {K, Karthikeyan and Wang, Zihan and Mayhew, Stephen and Roth, Dan},
	year = {2020},
}

@inproceedings{freitag_bleu_2020,
	address = {Online},
	title = {{BLEU} might be guilty but references are not innocent},
	url = {https://www.aclweb.org/anthology/2020.emnlp-main.5},
	booktitle = {Proceedings of the 2020 {Conference} on {Empirical} {Methods} in {Natural} {Language} {Processing} ({EMNLP})},
	publisher = {Association for Computational Linguistics},
	author = {Freitag, Markus and Grangier, David and Caswell, Isaac},
	year = {2020},
	pages = {61--71},
}

@inproceedings{raghu_svcca_2017,
	title = {{SVCCA}: singular vector canonical correlation analysis for deep learning dynamics and interpretability},
	volume = {30},
	url = {https://proceedings.neurips.cc/paper/2017/file/dc6a7e655d7e5840e66733e9ee67cc69-Paper.pdf},
	booktitle = {Advances in {Neural} {Information} {Processing} {Systems}},
	publisher = {Curran Associates, Inc.},
	author = {Raghu, Maithra and Gilmer, Justin and Yosinski, Jason and Sohl-Dickstein, Jascha},
	year = {2017},
	pages = {6076--6085},
}

@inproceedings{dufter_identifying_2020,
	address = {Online},
	title = {Identifying elements essential for {BERT}'s multilinguality},
	url = {https://www.aclweb.org/anthology/2020.emnlp-main.358},
	booktitle = {Proceedings of the 2020 {Conference} on {Empirical} {Methods} in {Natural} {Language} {Processing} ({EMNLP})},
	publisher = {Association for Computational Linguistics},
	author = {Dufter, Philipp and Schütze, Hinrich},
	year = {2020},
	pages = {4423--4437},
}

@inproceedings{oncevay_bridging_2020,
	address = {Online},
	title = {Bridging linguistic topology and multilingual machine translation with multi-view language representations},
	url = {https://www.aclweb.org/anthology/2020.emnlp-main.187},
	booktitle = {Proceedings of the 2020 {Conference} on {Empirical} {Methods} in {Natural} {Language} {Processing} ({EMNLP})},
	publisher = {Association for Computational Linguistics},
	author = {Oncevay, Arturo and Haddow, Barry and Birch, Alexandra},
	year = {2020},
	pages = {2391--2406},
}

@inproceedings{escolano_bilingual_2019,
	address = {Florence, Italy},
	title = {From bilingual to multilingual neural machine translation by incremental training},
	url = {https://www.aclweb.org/anthology/P19-2033},
	doi = {10.18653/v1/P19-2033},
	language = {en},
	urldate = {2020-05-26},
	booktitle = {Proceedings of the 57th {Annual} {Meeting} of the {Association} for {Computational} {Linguistics}: {Student} {Research} {Workshop}},
	publisher = {Association for Computational Linguistics},
	author = {Escolano, Carlos and Costa-jussà, Marta R. and Fonollosa, José A. R.},
	year = {2019},
	pages = {236--242},
}

@inproceedings{shaw_self-attention_2018,
	address = {New Orleans, Louisiana},
	title = {Self-attention with relative position representations},
	url = {http://aclweb.org/anthology/N18-2074},
	doi = {10.18653/v1/N18-2074},
	language = {en},
	urldate = {2020-05-04},
	booktitle = {Proceedings of the 2018 {Conference} of the {North} {American} {Chapter} of           the {Association} for {Computational} {Linguistics}: {Human} {Language}           {Technologies}, {Volume} 2 ({Short} {Papers})},
	publisher = {Association for Computational Linguistics},
	author = {Shaw, Peter and Uszkoreit, Jakob and Vaswani, Ashish},
	year = {2018},
	pages = {464--468},
}

@article{child_generating_2019,
	title = {Generating long sequences with sparse {Transformers}},
	url = {http://arxiv.org/abs/1904.10509},
	urldate = {2020-01-02},
	journal = {arXiv:1904.10509 [cs, stat]},
	author = {Child, Rewon and Gray, Scott and Radford, Alec and Sutskever, Ilya},
	month = apr,
	year = {2019},
	note = {arXiv: 1904.10509},
}

@inproceedings{koehn_europarl_2005,
	address = {Phuket, Thailand},
	title = {Europarl: {A} parallel corpus for statistical machine translation},
	booktitle = {Proceedings of the 10th {Machine} {Translation} {Summit} ({MT}-{Summit})},
	author = {Koehn, Philipp},
	year = {2005},
	pages = {79--86},
}

@inproceedings{ziemski_united_2016,
	address = {Portorož, Slovenia},
	title = {The {United} {Nations} parallel corpus v1.0},
	url = {https://www.aclweb.org/anthology/L16-1561},
	booktitle = {Proceedings of the {Tenth} {International} {Conference} on {Language} {Resources} and {Evaluation}},
	publisher = {European Language Resources Association (ELRA)},
	author = {Ziemski, Michał and Junczys-Dowmunt, Marcin and Pouliquen, Bruno},
	year = {2016},
	pages = {3530--3534},
}

@inproceedings{dabre_exploiting_2019,
	address = {Hong Kong, China},
	title = {Exploiting multilingualism through multistage fine-tuning for low-resource neural machine translation},
	url = {https://www.aclweb.org/anthology/D19-1146},
	doi = {10.18653/v1/D19-1146},
	language = {en},
	urldate = {2019-12-20},
	booktitle = {Proceedings of the 2019 {Conference} on {Empirical} {Methods} in {Natural} {Language} {Processing} and the 9th {International} {Joint} {Conference} on {Natural} {Language} {Processing}},
	publisher = {Association for Computational Linguistics},
	author = {Dabre, Raj and Fujita, Atsushi and Chu, Chenhui},
	year = {2019},
	pages = {1410--1416},
}

@inproceedings{qi_when_2018,
	address = {New Orleans, Louisiana},
	title = {When and why are pre-trained word embeddings useful for neural machine translation?},
	url = {http://aclweb.org/anthology/N18-2084},
	doi = {10.18653/v1/N18-2084},
	language = {en},
	urldate = {2019-12-19},
	booktitle = {Proceedings of the 2018 {Conference} of the {North} {American} {Chapter} of the {Association} for {Computational} {Linguistics}},
	publisher = {Association for Computational Linguistics},
	author = {Qi, Ye and Sachan, Devendra and Felix, Matthieu and Padmanabhan, Sarguna and Neubig, Graham},
	year = {2018},
	pages = {529--535},
}

@inproceedings{tiedemann_parallel_2012,
	address = {Istanbul, Turkey},
	title = {Parallel data, tools and interfaces in {OPUS}},
	booktitle = {Proceedings of the {Eight} {International} {Conference} on {Language} {Resources} and {Evaluation}},
	publisher = {European Language Resources Association (ELRA)},
	author = {Tiedemann, Jörg},
	year = {2012},
	pages = {2214--2218},
}

@inproceedings{bapna_non-parametric_2019,
	address = {Minneapolis, Minnesota},
	title = {Non-parametric adaptation for neural machine translation},
	url = {http://aclweb.org/anthology/N19-1191},
	doi = {10.18653/v1/N19-1191},
	language = {en},
	urldate = {2019-12-12},
	booktitle = {Proceedings of the 2019 {Conference} of the {North} {American} {Chapter} of the {Association} for {Computational} {Linguistics}},
	publisher = {Association for Computational Linguistics},
	author = {Bapna, Ankur and Firat, Orhan},
	year = {2019},
	pages = {1921--1931},
}

@inproceedings{sutskever_sequence_2014,
	address = {Montreal, Canada},
	title = {Sequence to sequence learning with neural networks},
	booktitle = {Advances in {Neural} {Information} {Processing} {Systems} ({NIPS})},
	author = {Sutskever, Ilya and Vinyals, Oriol and Le, Quoc V.},
	year = {2014},
}

@article{hochreiter_long_1997,
	title = {Long short-term memory},
	volume = {9},
	number = {8},
	journal = {Neural Computation},
	author = {Hochreiter, Sepp and Schmidhuber, Jürgen},
	year = {1997},
	pages = {1735--1780},
}

@article{wu_googles_2016,
	title = {Google's neural machine translation system: bridging the gap between human and machine translation},
	shorttitle = {Google's {Neural} {Machine} {Translation} {System}},
	url = {http://arxiv.org/abs/1609.08144},
	urldate = {2019-12-09},
	journal = {arXiv:1609.08144 [cs]},
	author = {Wu, Yonghui and Schuster, Mike and Chen, Zhifeng and Le, Quoc V. and Norouzi, Mohammad and Macherey, Wolfgang and Krikun, Maxim and Cao, Yuan and Gao, Qin and Macherey, Klaus and Klingner, Jeff and Shah, Apurva and Johnson, Melvin and Liu, Xiaobing and Kaiser, Łukasz and Gouws, Stephan and Kato, Yoshikiyo and Kudo, Taku and Kazawa, Hideto and Stevens, Keith and Kurian, George and Patil, Nishant and Wang, Wei and Young, Cliff and Smith, Jason and Riesa, Jason and Rudnick, Alex and Vinyals, Oriol and Corrado, Greg and Hughes, Macduff and Dean, Jeffrey},
	month = oct,
	year = {2016},
	note = {arXiv: 1609.08144},
}

@inproceedings{wang_compact_2019,
	address = {Florence, Italy},
	title = {A compact and language-sensitive multilingual translation method},
	url = {https://www.aclweb.org/anthology/P19-1117},
	doi = {10.18653/v1/P19-1117},
	language = {en},
	urldate = {2019-12-09},
	booktitle = {Proceedings of the 57th {Annual} {Meeting} of the {Association} for {Computational} {Linguistics}},
	publisher = {Association for Computational Linguistics},
	author = {Wang, Yining and Zhou, Long and Zhang, Jiajun and Zhai, Feifei and Xu, Jingfang and Zong, Chengqing},
	year = {2019},
	pages = {1213--1223},
}

@inproceedings{sennrich_improving_2016,
	address = {Berlin, Germany},
	title = {Improving neural machine translation models with monolingual data},
	url = {http://aclweb.org/anthology/P16-1009},
	doi = {10.18653/v1/P16-1009},
	language = {en},
	urldate = {2019-12-09},
	booktitle = {Proceedings of the 54th {Annual} {Meeting} of the {Association} for {Computational} {Linguistics}},
	publisher = {Association for Computational Linguistics},
	author = {Sennrich, Rico and Haddow, Barry and Birch, Alexandra},
	year = {2016},
	pages = {86--96},
}

@inproceedings{sachan_parameter_2018,
	address = {Belgium, Brussels},
	title = {Parameter sharing methods for multilingual self-attentional translation models},
	url = {http://aclweb.org/anthology/W18-6327},
	doi = {10.18653/v1/W18-6327},
	language = {en},
	urldate = {2019-12-09},
	booktitle = {Proceedings of the {Third} {Conference} on {Machine} {Translation}},
	publisher = {Association for Computational Linguistics},
	author = {Sachan, Devendra and Neubig, Graham},
	year = {2018},
	pages = {261--271},
}

@inproceedings{kim_effective_2019,
	address = {Florence, Italy},
	title = {Effective cross-lingual transfer of neural machine translation models without shared vocabularies},
	url = {https://www.aclweb.org/anthology/P19-1120},
	doi = {10.18653/v1/P19-1120},
	language = {en},
	urldate = {2019-12-09},
	booktitle = {Proceedings of the 57th {Annual} {Meeting} of the {Association} for {Computational} {Linguistics}},
	publisher = {Association for Computational Linguistics},
	author = {Kim, Yunsu and Gao, Yingbo and Ney, Hermann},
	year = {2019},
	pages = {1246--1257},
}

@inproceedings{neubig_rapid_2018,
	address = {Brussels, Belgium},
	title = {Rapid adaptation of neural machine translation to new languages},
	url = {http://aclweb.org/anthology/D18-1103},
	doi = {10.18653/v1/D18-1103},
	language = {en},
	urldate = {2019-12-09},
	booktitle = {Proceedings of the 2018 {Conference} on {Empirical} {Methods} in {Natural} {Language} {Processing}},
	publisher = {Association for Computational Linguistics},
	author = {Neubig, Graham and Hu, Junjie},
	year = {2018},
	pages = {875--880},
}

@inproceedings{luong_effective_2015,
	address = {Lisbon, Portugal},
	title = {Effective approaches to attention-based neural machine translation},
	url = {http://aclweb.org/anthology/D15-1166},
	doi = {10.18653/v1/D15-1166},
	language = {en},
	urldate = {2019-12-09},
	booktitle = {Proceedings of the 2015 {Conference} on {Empirical} {Methods} in {Natural} {Language} {Processing}},
	publisher = {Association for Computational Linguistics},
	author = {Luong, Thang and Pham, Hieu and Manning, Christopher D.},
	year = {2015},
	pages = {1412--1421},
}

@inproceedings{kocmi_trivial_2018,
	address = {Belgium, Brussels},
	title = {Trivial transfer learning for low-resource neural machine translation},
	url = {http://aclweb.org/anthology/W18-6325},
	doi = {10.18653/v1/W18-6325},
	language = {en},
	urldate = {2019-12-09},
	booktitle = {Proceedings of the {Third} {Conference} on {Machine} {Translation}},
	publisher = {Association for Computational Linguistics},
	author = {Kocmi, Tom and Bojar, Ondřej},
	year = {2018},
	pages = {244--252},
}

@inproceedings{gu_universal_2018,
	address = {New Orleans, Louisiana},
	title = {Universal neural machine translation for extremely low resource languages},
	url = {http://aclweb.org/anthology/N18-1032},
	doi = {10.18653/v1/N18-1032},
	language = {en},
	urldate = {2019-12-09},
	booktitle = {Proceedings of the 2018 {Conference} of the {North} {American} {Chapter} of the {Association} for {Computational} {Linguistics}},
	publisher = {Association for Computational Linguistics},
	author = {Gu, Jiatao and Hassan, Hany and Devlin, Jacob and Li, Victor O.K.},
	year = {2018},
	pages = {344--354},
}

@article{johnson_googles_2017,
	title = {Google’s multilingual neural machine translation system: enabling zero-shot translation},
	volume = {5},
	issn = {2307-387X},
	shorttitle = {Google’s {Multilingual} {Neural} {Machine} {Translation} {System}},
	url = {https://www.mitpressjournals.org/doi/abs/10.1162/tacl_a_00065},
	doi = {10.1162/tacl_a_00065},
	language = {en},
	urldate = {2019-12-09},
	journal = {Transactions of the Association for Computational Linguistics},
	author = {Johnson, Melvin and Schuster, Mike and Le, Quoc V. and Krikun, Maxim and Wu, Yonghui and Chen, Zhifeng and Thorat, Nikhil and Viégas, Fernanda and Wattenberg, Martin and Corrado, Greg and Hughes, Macduff and Dean, Jeffrey},
	year = {2017},
	pages = {339--351},
}

@inproceedings{dong_multi-task_2015,
	address = {Beijing, China},
	title = {Multi-task learning for multiple language translation},
	url = {http://aclweb.org/anthology/P15-1166},
	doi = {10.3115/v1/P15-1166},
	language = {en},
	urldate = {2019-12-09},
	booktitle = {Proceedings of the 53rd {Annual} {Meeting} of the {Association} for {Computational} {Linguistics} and the 7th {International} {Joint} {Conference} on {Natural} {Language} {Processing}},
	publisher = {Association for Computational Linguistics},
	author = {Dong, Daxiang and Wu, Hua and He, Wei and Yu, Dianhai and Wang, Haifeng},
	year = {2015},
	pages = {1723--1732},
}

@inproceedings{firat_multi-way_2016,
	address = {San Diego, California},
	title = {Multi-way, multilingual neural machine translation with a shared attention mechanism},
	url = {http://aclweb.org/anthology/N16-1101},
	doi = {10.18653/v1/N16-1101},
	language = {en},
	urldate = {2019-12-09},
	booktitle = {Proceedings of the 2016 {Conference} of the {North} {American} {Chapter} of the {Association} for {Computational} {Linguistics}},
	publisher = {Association for Computational Linguistics},
	author = {Firat, Orhan and Cho, Kyunghyun and Bengio, Yoshua},
	year = {2016},
	pages = {866--875},
}

@inproceedings{devlin_bert_2019,
	address = {Minneapolis, Minnesota},
	title = {{BERT}: pre-training of deep bidirectional {Transformers} for language understanding},
	url = {http://aclweb.org/anthology/N19-1423},
	doi = {10.18653/v1/N19-1423},
	language = {en},
	urldate = {2019-12-09},
	booktitle = {Proceedings of the 2019 {Conference} of the {North} {American} {Chapter} of the {Association} for {Computational} {Linguistics}},
	publisher = {Association for Computational Linguistics},
	author = {Devlin, Jacob and Chang, Ming-Wei and Lee, Kenton and Toutanova, Kristina},
	year = {2019},
	pages = {4171--4186},
}

@inproceedings{cho_learning_2014,
	address = {Doha, Qatar},
	title = {Learning phrase representations using {RNN} encoder–decoder for statistical machine translation},
	url = {http://aclweb.org/anthology/D14-1179},
	doi = {10.3115/v1/D14-1179},
	language = {en},
	urldate = {2019-12-09},
	booktitle = {Proceedings of the 2014 {Conference} on {Empirical} {Methods} in {Natural} {Language} {Processing}},
	publisher = {Association for Computational Linguistics},
	author = {Cho, Kyunghyun and van Merrienboer, Bart and Gulcehre, Caglar and Bahdanau, Dzmitry and Bougares, Fethi and Schwenk, Holger and Bengio, Yoshua},
	year = {2014},
	pages = {1724--1734},
}

@inproceedings{bahdanau_neural_2015,
	address = {San Diego, California},
	title = {Neural machine translation by jointly learning to align and translate},
	booktitle = {International {Conference} on {Learning} {Representations}},
	author = {Bahdanau, Dzmitry and Cho, Kyunghyun and Bengio, Yoshua},
	year = {2015},
}

@inproceedings{aharoni_massively_2019,
	address = {Minneapolis, Minnesota},
	title = {Massively multilingual neural machine translation},
	url = {http://aclweb.org/anthology/N19-1388},
	doi = {10.18653/v1/N19-1388},
	language = {en},
	urldate = {2019-12-09},
	booktitle = {Proceedings of the 2019 {Conference} of the {North} {American} {Chapter} of the {Association} for {Computational} {Linguistics}},
	publisher = {Association for Computational Linguistics},
	author = {Aharoni, Roee and Johnson, Melvin and Firat, Orhan},
	year = {2019},
	pages = {3874--3884},
}

@article{arivazhagan_massively_2019,
	title = {Massively multilingual neural machine translation in the wild: findings and challenges},
	url = {http://arxiv.org/abs/1907.05019},
	urldate = {2019-12-09},
	journal = {arXiv:1907.05019 [cs]},
	author = {Arivazhagan, Naveen and Bapna, Ankur and Firat, Orhan and Lepikhin, Dmitry and Johnson, Melvin and Krikun, Maxim and Chen, Mia Xu and Cao, Yuan and Foster, George and Cherry, Colin and Macherey, Wolfgang and Chen, Zhifeng and Wu, Yonghui},
	month = jul,
	year = {2019},
	note = {arXiv: 1907.05019},
}

\chapter*{Summary}

Multilingual machine translation systems learn patterns from parallel text corpora to make knowledge accessible across the world's languages.
Understanding how knowledge flows between languages requires navigating subtle nuances in model representations, data availability, and architecture.
Harnessing these nuances is challenging because low-resource languages lack the wealth of parallel data that makes transfer comparatively straightforward for their high-resource counterparts.

In this thesis, we analyze and enhance cross-lingual knowledge transfer to deliver robust multilingual NLP capabilities, with machine translation as a central testbed.
First, we introduce Representational Transfer Potential to quantify cross-language similarity, uncover predictors like multi-parallel overlap and genetic distance, and design an auxiliary similarity loss that strengthens low- and mid-resource translations.
Second, we extend semi-parametric translation with cross-lingual and multilingual \textit{k}-nearest-neighbor datastores, showing that linguistically structured retrieval boosts quality and enables faster inference without sacrificing accuracy.
Third, we examine the fine-tuning paradox in large language models, revealing how parallel data can erode formality control, domain adaptation, and document-level coherence, and we mitigate these losses by blending monolingual and parallel supervision.
Fourth, we study the role of language diversity during fine-tuning, demonstrating that scaling the number of translation directions improves both seen and unseen pairs, reduces off-target generations, and aligns representations until gains plateau.

Together, these findings show that deliberate modeling and data choices can extend high-quality multilingual NLP beyond well-resourced languages.
Deepening our grasp of cross-lingual transfer not only illuminates current vulnerabilities but also points the way toward inclusive, resilient multilingual technologies that serve the full spectrum of the world's languages.

\chapter*{Samenvatting}

Machinale vertalingssystemen leren patronen uit meertalige tekstcorpora om kennis toegankelijk te maken in alle talen van de wereld.
Inzicht krijgen in hoe kennis tussen talen stroomt vereist het omgaan met subtiele nuances in modelrepresentaties, databeschikbaarheid en architectuur.
Het benutten van deze nuances is uitdagend, aangezien kennisoverdracht tussen talen afhankelijk is van grote hoeveelheden data, die niet beschikbaar zijn voor talen met minder digitale representatie.

In dit proefschrift analyseren en versterken we kennisoverdracht tussen talen om robuuste meertalige taalverwerkingsvaardigheden te leveren, met machinaal vertalen als centraal testbed.
Ten eerste introduceren we Representational Transfer Potential om gelijkenis tussen multilinguale taalrepresentaties te kwantificeren, identificeren we voorspellende factoren zoals multi-parallele overlap en genetische afstand, en ontwerpen we similariteitsregularisatie die vertalingen voor talen met weinig en middelgrote middelen versterkt.
Ten tweede breiden we semi-parametrische vertaling uit met meertalige \textit{k}-nearest-neighbor-datastores, waaruit blijkt dat linguïstisch gestructureerde retrieval de kwaliteit verhoogt en snellere inferentie mogelijk maakt zonder vertaalkwaliteit op te offeren.
Ten derde onderzoeken we het fine-tuning-paradox bij grote taalmodellen, laten we zien hoe parallelle data controle over formaliteit, domeinaanpassing en coherentie op documentniveau kunnen aantasten, en beperken we deze verliezen door monolinguale en parallelle supervisie te combineren.
Ten vierde bestuderen we de rol van taaldiversiteit tijdens fine-tuning en tonen we aan dat het opschalen van het aantal vertaalrichtingen zowel bekende als onbekende taalrichtingen verbetert, generaties in de verkeerde taal vermindert en representaties op elkaar afstemt totdat de voordelen afvlakken.

Gezamenlijk laten deze bevindingen zien dat weloverwogen keuzes in modellering en data tot meertalige natuurlijke taalverwerking-systemen van hoge kwaliteit leiden, ook voor talen met minder data.
Door ons begrip over meertalige kennisoverdracht tussen talen verder te verdiepen, brengen we niet alleen kwetsbaarheden van huidige systemen in kaart, maar leggen we de basis voor inclusieve, veerkrachtige, meertalige technologieën die het volledige spectrum aan talen bedienen.

\end{document}